\documentclass{article}
\usepackage{iclr2027_conference,times}

\usepackage{amsmath,amsfonts,bm}

\def\eqref#1{equation~\ref{#1}}

\def\1{\bm{1}}

\def\vzero{{\bm{0}}}
\def\vone{{\bm{1}}}

\def\vtheta{{\bm{\theta}}}

\def\vc{{\bm{c}}}

\def\ve{{\bm{e}}}

\def\vs{{\bm{s}}}

\def\vu{{\bm{u}}}
\def\vv{{\bm{v}}}

\def\vx{{\bm{x}}}
\def\vy{{\bm{y}}}
\def\vz{{\bm{z}}}

\def\mA{{\bm{A}}}
\def\mB{{\bm{B}}}

\def\mD{{\bm{D}}}
\def\mE{{\bm{E}}}
\def\mF{{\bm{F}}}
\def\mG{{\bm{G}}}
\def\mH{{\bm{H}}}
\def\mI{{\bm{I}}}
\def\mJ{{\bm{J}}}

\def\mL{{\bm{L}}}
\def\mM{{\bm{M}}}
\def\mN{{\bm{N}}}

\def\mP{{\bm{P}}}
\def\mQ{{\bm{Q}}}
\def\mR{{\bm{R}}}
\def\mS{{\bm{S}}}
\def\mT{{\bm{T}}}

\def\mV{{\bm{V}}}

\def\mX{{\bm{X}}}
\def\mY{{\bm{Y}}}
\def\mZ{{\bm{Z}}}

\def\mSigma{{\bm{\Sigma}}}

\DeclareMathAlphabet{\mathsfit}{\encodingdefault}{\sfdefault}{m}{sl}
\SetMathAlphabet{\mathsfit}{bold}{\encodingdefault}{\sfdefault}{bx}{n}

\def\gB{{\mathcal{B}}}

\def\gE{{\mathcal{E}}}
\def\gF{{\mathcal{F}}}

\def\gH{{\mathcal{H}}}
\def\gI{{\mathcal{I}}}

\def\gK{{\mathcal{K}}}

\def\gN{{\mathcal{N}}}

\newcommand{\E}{\mathbb{E}}

\newcommand{\R}{\mathbb{R}}

\newcommand{\Var}{\mathrm{Var}}

\DeclareMathOperator*{\argmax}{arg\,max}

\usepackage{amssymb,mathtools,amsthm,thmtools,thm-restate}
\usepackage{algorithm}
\usepackage{algpseudocode}
\usepackage{booktabs,longtable,pifont}
\usepackage{multirow} %
\usepackage[most]{tcolorbox}
\usepackage{xcolor}
\usepackage[normalem]{ulem}
\usepackage{enumitem}

\usepackage{wrapfig}
\usepackage{threeparttable}
\usepackage{subcaption,tabularx}
\newcommand{\cmark}{\ding{51}}
\newcommand{\xmark}{\ding{55}}

\usepackage{listings}
\usepackage[T1]{fontenc}
\usepackage[varqu,scaled=0.92]{zi4}
 
\newtcblisting[auto counter]{promptbox}[3][]{%
  listing only,
  breakable,
  enhanced jigsaw,
  colback=black!3,
  colframe=black!50,
  boxrule=0.4pt,
  arc=1pt,
  left=4pt, right=4pt, top=4pt, bottom=4pt,
  before skip=10pt, after skip=10pt,
  title={\small\textbf{Prompt \thetcbcounter:} #2},
  coltitle=black,
  colbacktitle=black!8,
  attach boxed title to top left={yshift=-\tcboxedtitleheight/2, xshift=6pt},
  boxed title style={colframe=black!50, boxrule=0.4pt, arc=1pt,
    left=3pt, right=3pt, top=1.5pt, bottom=1.5pt},
  label={#3},
  listing options={
    basicstyle=\footnotesize\ttfamily,
    breaklines=true,
    breakatwhitespace=true,
    breakindent=0pt,
    columns=fullflexible,
    keepspaces=true,
    upquote=true,
    showstringspaces=false,
    escapeinside={(*@}{@*)},
    literate={—}{{\textemdash}}1
             {“}{{``}}1
             {”}{{''}}1
             {‘}{{`}}1
             {’}{{'}}1,
  },
  #1
}
 
\newcommand{\ph}[1]{\textnormal{\itshape$\langle$#1$\rangle$}}

\renewcommand{\eqref}[1]{\textup{(\ref{#1})}}

\newcommand{\PP}{\mathbb{P}}                 %
\newcommand{\lammin}{\lambda_{\min}}
\newcommand{\lammax}{\lambda_{\max}}
\newcommand{\lamperp}{\lambda_{\min,\perp}}  %
\newcommand{\popsup}{\mathrm{pop}}
\newcommand{\vgamma}{{\bm{\gamma}}}
\newcommand{\vbeta}{{\bm{\beta}}}
\newcommand{\vzeta}{{\bm{\zeta}}}
\newcommand{\vomega}{{\bm{\omega}}}
\newcommand{\norm}[1]{\left\lVert#1\right\rVert}
\newcommand{\abs}[1]{\left\lvert#1\right\rvert}
\providecommand{\vertiii}[1]{\ensuremath{\left\vert\kern-0.25ex\left\vert\kern-0.25ex\left\vert #1
    \right\vert\kern-0.25ex\right\vert\kern-0.25ex\right\vert}}
\newcommand{\opnorm}[1]{\vertiii{#1}}
\newcommand{\trans}{^{\top}}
\newcommand{\Fm}{\gF_{-m}}

\newcommand{\metricStatR}{Stat.R.}

\newcommand{\metricLeakage}{Leakage}

\newcommand{\notchlines}{3} %
\DeclareCaptionFormat{notch}{\hangindent=-0.52\linewidth\hangafter=-\notchlines\relax #1#2#3\par}

\definecolor{dialC}{RGB}{42,70,120}    %
\definecolor{coreC}{RGB}{40,100,70}    %
\definecolor{nicheC}{RGB}{140,85,30}   %
\definecolor{nuisC}{RGB}{110,110,110}  %

\newcommand{\tslot}[2]{\textcolor{#1}{$\langle$\textsc{#2}$\rangle$}}
\newcommand{\dialslot}[1]{\tslot{dialC}{#1}}
\newcommand{\coreslot}[1]{\tslot{coreC}{#1}}
\newcommand{\nicheslot}[1]{\tslot{nicheC}{#1}}
\newcommand{\nuisslot}[1]{\tslot{nuisC}{#1}}
\newcommand{\band}[1]{{\scriptsize\color{black!60}#1}}

\newcommand{\blocktag}[1]{%
  \par\vspace{5pt}\nopagebreak
  {\footnotesize\itshape\color{black!50}%
   \hspace{0pt plus 1filll}#1\par}\nopagebreak\vspace{1pt}}

\newcommand{\adtag}[2]{{\footnotesize\textcolor{#1}{$\langle$\textsc{#2}$\rangle$}}}
\newcommand{\dialtag}[1]{\adtag{dialC}{#1}}
\newcommand{\coretag}[1]{\adtag{coreC}{#1}}

\newenvironment{adlist}
  {\begin{list}{\textbullet}{%
     \setlength{\topsep}{1pt}\setlength{\partopsep}{0pt}%
     \setlength{\itemsep}{1pt}\setlength{\parsep}{0pt}%
     \setlength{\leftmargin}{12pt}\setlength{\itemindent}{0pt}%
     \setlength{\labelwidth}{7pt}\setlength{\labelsep}{5pt}}}
  {\end{list}}

\newcommand{\tagright}[1]{\unskip\nobreak\hspace{0pt plus 1filll}\mbox{#1}}

\allowdisplaybreaks

\theoremstyle{plain}
\newtheorem{theorem}{Theorem}
\newtheorem{lemma}{Lemma}
\newtheorem{corollary}{Corollary}
\newtheorem{fact}{Fact}          %
\theoremstyle{definition}
\newtheorem{assumption}{Assumption}
\theoremstyle{remark}
\newtheorem{remark}{Remark}

\usepackage{hyperref}
\usepackage{url}

\title{Interpretable Feature Discovery from Unstructured Data Using LLMs and Statistical Choice Models}

\author{
Baptiste Arnaudo$^{1}$\thanks{Now at the Institute for Computational and Mathematical
Engineering (ICME), Stanford University.} \quad Amine Bennouna$^{2}$ \quad Sébastien Martin$^{2}$ \\[4pt]
$^{1}$École polytechnique, Palaiseau, France \\
$^{2}$Operations Department, Kellogg School of Management, Northwestern University, Evanston, IL, USA \\[4pt]
\texttt{baptiste.arnaudo@stanford.edu} \\
\texttt{\{amine.bennouna, sebastien.martin\}@kellogg.northwestern.edu}
}

\iclrfinalcopy %
\begin{document}

\maketitle

\addtocontents{toc}{\protect\setcounter{tocdepth}{-1}}

\begin{abstract}
Feature extraction from unstructured data typically relies on opaque embeddings unsuited to interpretable decision-making. Interpretable features, on the other hand, typically require costly, and often infeasible, hand-engineering. Large language models (LLMs) offer a natural bridge, as they can read raw data and describe it in interpretable terms. One could simply ask an LLM to extract features from a large sample of the dataset, but context limitations and diluted attention can lead to unsatisfactory results. We introduce a method that automatically discovers interpretable, decision-relevant features from unstructured data. Our key idea is to perform pairwise comparisons between items, asking an LLM to articulate, in interpretable terms, the features distinguishing each pair. We treat these explanations, or \emph{rationales}, as data: we develop a statistical choice model that combines these rationales across the comparisons to identify relevant features and estimate per-item, per-feature scores. We first test our method on a synthetic hiring dataset of cover letters with known ground-truth features and show that it significantly outperforms direct LLM feature extraction in feature recovery. We then apply it to the ChangeMyView (Reddit) debate dataset, where it discovers several interpretable features associated with persuasion while reaching predictive accuracy comparable to established baselines. We also show how the same rationales can be used to audit LLM decisions, yielding a holistic score whose decomposition over interpretable features reveals what the LLM rewards.
\end{abstract}

\section{Introduction}
\label{sec:introduction}

Consider an instructor tasked with grading essays. She typically first inspects a sample of the essays to inform her grading policy: some may show excellent writing while others have an impressive vocabulary, etc. 
These comparable attributes between essays are called \emph{features}. From this small sample, she can draw up a rubric, namely the detailed features she will use to grade with \emph{limited bias}. As she grades, she is likely to realize that the original sample that first calibrated her grading policy was too small to contain every important dimension: as the grading goes on, she realizes that grammar is also distinguishing and important, so she needs to go back to the first essays to revise their grading.

This is clearly inefficient, and the instructor would hope to use AI tools to assist in this process of discriminative feature extraction, namely the extraction of the relevant features of the data on which she can base her decisions. If she had essays from previous years together with the grade they obtained, she could train a machine learning model to predict the essays' grades from their \emph{embeddings}, which would serve as features for grading. This would automate the grading, but the features used would be opaque and uninterpretable, which is incompatible with transparent decision-making: a crucial aspect in such applications. Current techniques cannot automate the \emph{interpretable rubric extraction} step. 
A crucial challenge for decision-makers is therefore: given an unstructured set of items, discover interpretable and discriminative features \emph{for this dataset}, along with per-item per-feature scores, that they can reliably base their decisions on. Additionally, when decision-making is delegated to an LLM, the same features should make the LLM's decisions auditable.

A natural approach is to give an LLM the whole set of items and ask it
directly for the features that distinguish them (a grading rubric, in
the case of essays). However, processing an entire dataset requires an
extremely long context, and LLM performance typically degrades as input
length increases~\citep{liu2024lost,levy2024sametask}. Consistent with this, we
find that direct LLM extraction misses features that are present in the data
and conflates others (Section~\ref{sec:feature_extraction_empirical_results}).
The difficulty is therefore one of \emph{context management}: how to
use LLMs on the set of items so that they can reliably identify the features
that matter. We introduce an end-to-end approach that, given an
unstructured dataset, discovers human-interpretable, dataset-relevant
features using LLMs.

\paragraph{Contributions.}
Our idea is that the relevance of a feature can be revealed by a set of
pairwise comparisons of items, and that LLM judges can articulate these
comparisons in human language. We call such an articulated comparison a
\emph{rationale}: which features the judge cites, and which item it
favors on each. The key is that, with enough comparisons, a statistical
model can reverse-engineer the judge's rationales and recover which
features distinguish the items and how each item scores along each
feature.

We treat the rationales as data. A statistical choice model
(Section~\ref{sec:method}) describes each rationale as a two-stage
outcome: whether the LLM cites a feature, and, when it does, which item
wins on it. The model absorbs the LLM's potential biases, such as a
preference for the first or second position in the prompt. Fitting it
yields per-item, per-feature scores with confidence intervals, together
with diagnostics that filter out features with unreliably estimated scores. Estimation is non-trivial: the likelihood is non-convex and the number of parameters grows with the number of items. We nonetheless prove, under mild assumptions, that the likelihood admits a local maximizer that is consistent for every item score at once in the regime where the number of items grows (Theorems~\ref{thm:consistency} and~\ref{thm:joint}).

We validate our approach extensively in three settings. First, we benchmark it on a synthetic hiring dataset of cover letters with known ground-truth features, where our method correctly recovers all planted features, and most discovered features track a single planted one, whereas methods that prompt an LLM with the whole dataset/a sample of it miss some features and blur several together (Section~\ref{sec:feature_extraction_empirical_results}). 
We then apply it to a real dataset of online debates (the ChangeMyView subreddit) to discover and quantify which argumentation techniques are the most effective. 
We find that our method reaches an accuracy comparable to the state of the art while providing interpretable features (Section~\ref{sec:cmv}) along with insights on argumentation techniques. Finally, we show in Appendix~\ref{app:verdict-extension} how an LLM's rationales can also be used to audit its decisions and represent, in interpretable terms, what actually mattered in its decision when comparing two items, yielding interpretability insights for LLM-based scoring.

Our approach provides a concrete procedure that can be supplied as instructions to an LLM when tasked with feature discovery. It offers a principled way to organize multiple LLM agents for this purpose: a lead agent delegates pairwise comparisons along a graph to parallel sub-agents and aggregates their rationales through a statistical model. This aligns with recent efforts to scale AI performance through the orchestration of multiple agents~\citep{li2024agentsneed}, where how the work is split and recombined is critical to performance~\citep{kim2026sciencescalingagentsystems}.

\section{Related work}
\label{sec:related_work}

\paragraph{The LLM-as-a-Judge paradigm} The \emph{LLM-as-a-Judge} paradigm is an approach to several problems that uses LLMs to evaluate or rank items where one could have solicited a human judge. Examples include model benchmarking: in their foundational work, \citet{zheng2023judgingllmasajudgemtbenchchatbot} asked both LLMs and humans which of two AI-generated answers to a prompt they preferred, and found substantial agreement between the two, suggesting that LLMs could replace humans for comparing items. Nowadays, LLMs are used as judges in a variety of domains, spanning information retrieval~\citep{sun2024chatgptgoodsearchinvestigating},
recommendation systems~\citep{hou2024largelanguagemodelszeroshot}, alignment training~\citep{bai2022constitutionalaiharmlessnessai}, automated essay scoring~\citep{hou2025improve}, etc.

\paragraph{LLM biases.} Nonetheless, several works~\citep{zheng2023judgingllmasajudgemtbenchchatbot, ye2024justiceprejudicequantifyingbiases, gu2025surveyllmasajudge} document biases in LLMs, such as a \emph{positional bias} (when asked to compare two items, the LLM prefers the item that appears first/second in the prompt for no particular reason), a \emph{verbosity bias} (a tendency to prefer more verbose items, such as longer essays if the items being assessed are essays), etc. These phenomena can severely undermine LLM-based decision-making if not accounted for.

\paragraph{Feature evaluation with LLMs.} Closer to our work, general
LLM-based approaches typically work from a predetermined pool of
features: the features are scored, but never \emph{discovered from the
data}. Evaluation features are either fixed in the prompt \citep{zheng2023judgingllmasajudgemtbenchchatbot, liu2023geval}, drawn
from a predefined skill taxonomy \citep{ye2024flaskfinegrainedlanguagemodel}, or supplied by the user as a scoring rubric \citep{kim2024prometheus}. These approaches work by asking the judge to output a few tokens per feature: either a pointwise score \citep{liu2023geval, kim2024prometheus} or the verdict of a pairwise comparison \citep{zheng2023judgingllmasajudgemtbenchchatbot,
liusie2024comparative} (``A'' if item A is better, ``B'' otherwise). The judge's \emph{rationales} are often discarded or used as a qualitative confirmation, while we argue that they could be treated as data useful for feature discovery. Some approaches~\citep{xie2026autorubriclearningimplicitweights} leverage
them to create scoring guidelines to explain existing scores and
preference labels, but to our knowledge no approach uses them for
interpretable feature extraction.

\section{Method}
\label{sec:method}

\begin{figure}[t]
    \centering
    \includegraphics[width=\linewidth]{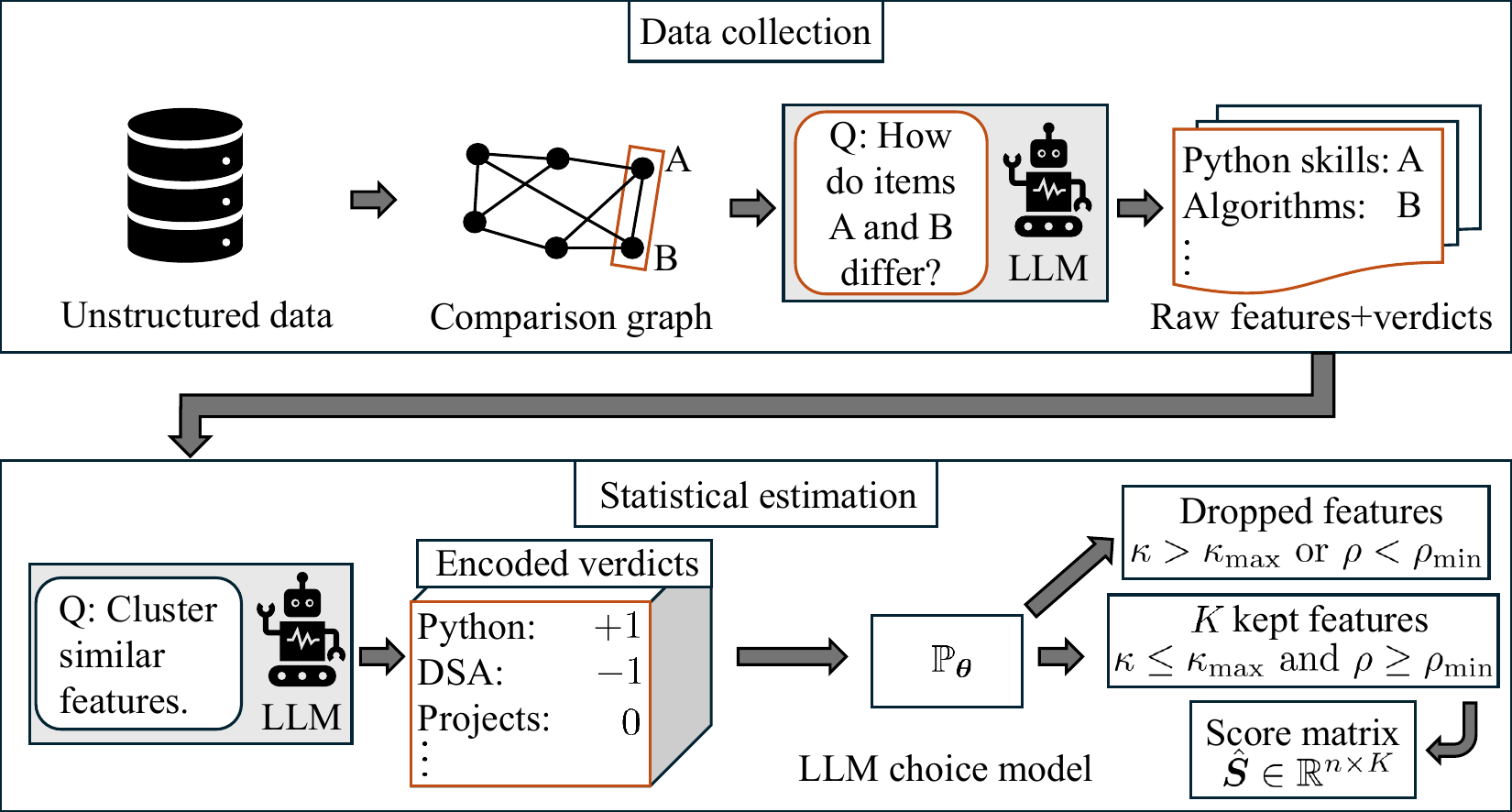}
    \caption{\textbf{Overview of the pipeline described in Section~\ref{sec:method}.} Items are first compared in pairs by the LLM, cited features are then clustered, and a statistical model learns the relevant features from the LLM's choices. Finally, statistical diagnostics identify the reliable scores. The orange running example follows one pair of cover letters (the setup of Section~\ref{sec:feature_extraction_empirical_results}) through the pipeline. }
    \label{fig:pipeline-feature-discovery}
\end{figure}

Our method proceeds in four steps, illustrated in Figure~\ref{fig:pipeline-feature-discovery}. First, LLMs compare pairs of items along a comparison graph, describing the features on which they differ and indicating which item shows more of each feature. These local comparisons require only a small context but are myopic, missing dataset-level information. Second, we group these descriptions into a shared set of features. Third, we fit a statistical choice model that aggregates these local observations into dataset-level, per-item feature scores, using both the cited differences and the features left unmentioned. Finally, we derive per-feature identifiability and reliability diagnostics that flag features whose scores are not pinned down by the observations.

\paragraph{Rationale collection.} Let $1,\dots,n$ denote the items. We compare them along a \emph{comparison graph}, a set of $N$ pairs $p=(i,j)$ where item $i$ is shown first. For each pair, an LLM judge is asked how the two items differ: it answers in free text, naming the features on which they differ and which item is stronger on each. These raw features, phrased differently across pairs, are then passed to a second LLM call that groups them into $K$ shared features. The outcome is a sparse \emph{verdict tensor} $\mT\in\{-1,0,+1\}^{N\times K}$: $T_{p,k}=0$ if feature $k$ was not invoked for pair $p$, and $T_{p,k}=+1$ (resp.\ $-1$) if it was invoked and the first (resp.\ second) item of the pair was cited as stronger.

\paragraph{Statistical choice model.} Given the verdicts $\mT$, our objective is to reverse-engineer the items' scores along each feature and the judge's biases. Our statistical model takes inspiration from Bradley--Terry models~\citep{bradleyterry1952} to estimate latent scores from pairwise comparisons (see Appendix~\ref{app:bt-ties} for related work).

We assume that each item has a \emph{latent score} along each feature, collected in a \emph{score matrix} $\mS \in \mathbb{R}^{n \times K}$. If pair $p$ compares items $i$ (shown first) and $j$, we write $\Delta_{p,k} = S_{i,k} - S_{j,k}$ for the true difference of their scores along feature $k$. Our model posits that these score differences drive the LLM's behavior in two ways: a) the more the items' scores differ on $k$, the more likely the LLM is to cite feature $k$; b) when feature $k$ is cited, the item having the larger score along it is more likely to be chosen. 

Importantly, we do not assume the LLM does any of this reliably: its citations and choices are noisy and may also carry the judge's biases. We hence model the citation of feature $k$ on pair $p$ as a coin flip of probability $\sigma(f(\Delta_{p,k}) + \gamma_k)$, where $\sigma$ is the sigmoid function and $f$ is an even function increasing in $\lvert\Delta\rvert$ (we use $f = \log(2\cosh)$, a smooth surrogate of the absolute value), so that the \emph{non-citation} probability is
\begin{equation}
    \label{eq:non-citation}
    \PP(T_{p,k} = 0) = 1 - \sigma(f(\Delta_{p,k}) + \gamma_k).
\end{equation}
Here $\gamma_k$ is the baseline citation propensity of feature $k$,
which accounts for a feature-specific eagerness or reluctance of the LLM to cite it. Under this model, silence is also informative: a non-citation suggests that the judge did not find the pair's difference on that feature salient enough to cite it. 

Our model also assumes that, conditioned on citing feature $k$, the LLM declares the first item the winner with probability $\sigma(\Delta_{p,k} + \beta)$, where $\beta$ captures the positional bias widely reported in the literature~\citep{zheng2023judgingllmasajudgemtbenchchatbot, ye2024justiceprejudicequantifyingbiases, gu2025surveyllmasajudge}, which we also observe in our experiments. Multiplying the two gives:
\begin{equation}
  \label{eq:citation}
  \begin{aligned}
  \PP(T_{p,k} = 1) &= \sigma(\Delta_{p,k} + \beta)\,
                       \sigma(f(\Delta_{p,k}) + \gamma_k) \\
  \PP(T_{p,k} = -1) &= \bigl(1 - \sigma(\Delta_{p,k} + \beta)\bigr)\,
                       \sigma(f(\Delta_{p,k}) + \gamma_k).
  \end{aligned}
\end{equation}
Hence, $\beta$ and $\gamma_k$ capture the LLM's biases (a preference for a position in the prompt, and feature-specific citation habits) so that they do not contaminate the estimated scores. The position coefficient $\beta$ is shared across features; as we show in Appendix~\ref{app:beta-orthogonality}, performing comparisons with presentation order randomized makes the scores essentially insensitive to this choice, and a per-feature refit confirms it empirically.

\paragraph{Parameter estimation.}
We estimate $\vtheta := (\operatorname{vec}(\mS), (\gamma_k)_{k=1}^{K},
\beta) \in \mathbb{R}^{nK + K + 1}$ from the verdict tensor $\mT$ by
penalized maximum likelihood,
\begin{equation}
  \label{eq:estimator}
  \widehat{\vtheta} \in \operatorname*{argmin}_{\vtheta}
  \Bigl\{\mathcal{L}(\vtheta) + \frac{\lambda}{2}\|\mS\|_F^2 \Bigr\},
\end{equation}
where $\mathcal{L}$ is the negative log-likelihood of $\mT$ and
$\lambda \ge 0$ is a ridge parameter. Under conditional independence of
the verdicts across pairs and features given $\vtheta$, $\mathcal{L}$ factorizes as
\begin{equation}
  \label{eq:nll}
  \begin{split}
  \mathcal{L}(\vtheta)
  = - \sum_{\text{pairs }p} \sum_{\text{features }k}
    \Bigl[
      & \1_{\{T_{p,k}\neq0\}}\log \sigma\bigl(f(\Delta_{p,k}) + \gamma_k\bigr)
       + \1_{\{T_{p,k}=0\}}\log\bigl(1 - \sigma(f(\Delta_{p,k}) + \gamma_k)\bigr) \\
      & + \1_{\{T_{p,k}=1\}}\log \sigma\bigl(\Delta_{p,k} + \beta\bigr)
      + \1_{\{T_{p,k}=-1\}}\log\bigl(1 - \sigma(\Delta_{p,k} + \beta)\bigr)\Bigr].
  \end{split}
\end{equation}
The likelihood is invariant to shifting all scores of a feature by a
common scalar; the ridge penalty resolves this by selecting, within each
invariance class, the centered representative $\vone\trans\vs_k = 0$, where $\vs_k$ is the $k$-th column of $\mS$.

\paragraph{Theoretical guarantees.} Our estimator raises a nonstandard
consistency question: the number of parameters grows with the number of
items $n$, and the negative log-likelihood~\eqref{eq:nll} is mildly
non-convex. We prove, under reasonable assumptions, that the likelihood
admits a local maximizer that is consistent for every item score at once
as $n$ grows with a bounded number of comparisons per pair, in the case
$K=1$ (single feature).

\begin{restatable}[Local MLE and uniform consistency, dense regime]{theorem}{mainconsistency}
\label{thm:consistency}
Assume A\ref{assump:dense}--A\ref{assump:gate} (every pair is compared at least once and at most a fixed number of times, independent of $n$; the true scores are bounded; $f$ is even and $C^3$) and A\ref{assump:info} (the population information for $(\gamma,\beta)$, after profiling out $\vs$, has smallest eigenvalue of order $n^2$). For every $c>0$ there exist constants $C, n_0$ depending only on $c$ and the constants in the assumptions such that for all $n \ge n_0$, with probability at least $1 - O(n^{-c})$, the negative log-likelihood \eqref{eq:nll} admits a stationary point $(\hat{\vs}, \hat\gamma, \hat\beta)$ on the hyperplane $\{\vone^\top \vs = 0\}$ satisfying
\begin{equation}
  \|\hat{\vs} - \vs^*\|_\infty
  \le C\sqrt{\frac{\log n}{n}},
  \qquad
  \max\left(|\hat\gamma - \gamma^*|, |\hat\beta - \beta^*|\right)
  \le C\sqrt{\frac{\log n}{n}}.
\end{equation}
Moreover, this point is the unique stationary point in a neighborhood
$\mathcal{B}_\infty(\vs^*, \varepsilon_n) \times
 \mathcal{B}(\gamma^*, r_n) \times \mathcal{B}(\beta^*, r_n)$
with $\varepsilon_n, r_n = \Theta\left(\sqrt{\frac{\log n}{n}}\right)$, and it maximizes the likelihood there: it is the local MLE.
\end{restatable}
In Appendix~\ref{app:model-theory}, we prove a more general version of
Theorem~\ref{thm:consistency} (Theorem~\ref{thm:joint}) in which both
channels carry arbitrary comparison-level covariates and $\gamma$ and $\beta$ are coefficient vectors.

In practice, we optimize the regularized likelihood~\eqref{eq:estimator}
numerically on sparse comparison graphs, and the fitted model is well
calibrated out of fold (Figures~\ref{fig:calibration_reliability_custom}
and~\ref{fig:calibration_skill_custom}). Theorem~\ref{thm:consistency}
assumes a dense design and does not cover this regime; what it
guarantees is that, near the truth, the non-convex likelihood has a
unique stationary point, which is a consistent local maximizer, so a
local optimizer targets a well-defined estimator rather than an
arbitrary critical point.

\paragraph{Extracting score reliability.}
Because $\hat{\vtheta}$ is an M-estimator, it satisfies a central
limit theorem as the number of comparisons grows, which yields
confidence intervals on every parameter (see Appendix~\ref{app:uq} for more details). The same machinery gives two per-feature diagnostics: a condition number
$\kappa_k$ of the score block of the Fisher information, which flags
features whose scores are poorly identified by the
data~\citep{rothenberg1971}, and a reliability coefficient $\rho_k$,
the fraction of the observed score spread attributable to true
between-item differences rather than estimation noise. Features with
$\kappa_k > \kappa_{\max}$ or $\rho_k < \rho_{\min}$, for thresholds
fixed a priori, are filtered out before a final refit.

\paragraph{Scoring new items.}
Running the full pipeline on the entire dataset allows the statistical
model to aggregate comparisons globally, correct the judge's biases, and
provide reliability diagnostics. With $d$ comparisons per item, this
requires $dn/2$ pairwise LLM calls. A cheaper alternative is to discover
features on a representative subset and then ask an LLM to score
additional items independently along those features, requiring one call
per item. This pointwise scoring requires specifying precisely how each
feature should be scored, which may require human contextual knowledge of
the dataset. It is also subject to well-documented biases of pointwise LLM
scoring~\citep{liusie2024comparative}. Alternatively, one can compare each
new item pairwise against a fixed number of already-scored items along the
fixed features and fit a Bradley--Terry model for each feature. This
retains dataset-level aggregation and position-bias correction without
repeating feature discovery. The choice trades cost against robustness to pointwise biases; both are evaluated in Section~\ref{sec:cmv}.

\section{Empirical Results on Interpretable Feature Extraction}
\label{sec:feature_extraction_empirical_results}

We first evaluate our approach in a synthetic hiring application where we control the ground-truth features, allowing us to test which features each method recovers and how accurately it recovers them. We construct a dataset of $n = 200$ cover letters responding to a real \textsc{Google} security software engineering job advertisement, with known features (which we call \emph{dials}) embedded in each letter. See Appendix~\ref{app:experiment-details} for details about the experiment.

\paragraph{Dataset.} Each letter realizes $18$ engineered ground-truth \emph{dials}: $8$ primary ones (\emph{years of experience}, \emph{education level}, \emph{school prestige}, \emph{number of languages}, \emph{production launches}, \emph{skill coverage}, \emph{employer prestige} and \emph{prose fluency}), $6$ binary role-skill presences (e.g.\ \emph{\textsc{Python}}, \emph{security assessments}), and $4$ three-level niche dials (e.g.\ \emph{open-source involvement}, \emph{security certifications}) exactly two of which are present in each letter.

Dials are drawn near-orthogonally and with capped correlation to letter length, so that each dial is a separately recoverable signal and length cannot proxy for quality. Of the 18 generated features, $D = 16$ are relevant recovery targets, while two serve as controls: \emph{skill coverage} is a redundant composite of the individual skill features, and \emph{prose fluency} is irrelevant to the job.

\paragraph{Metrics.} A discovered feature is said to \emph{track} a dial if the cover letters' scores along the feature correlate sufficiently with the dial. A discovered feature is said to be \emph{about} the dial if its name and the dial's name are sufficiently aligned in \textsc{text-embedding-3-large} embedding space. We will compare methods along the following metrics (defined mathematically in Appendix~\ref{app:metrics}).
\begin{itemize}[leftmargin=*]
    \item \textbf{Cost.} Total method cost in USD.
    \item \textbf{Statistical recovery (Tracked \& Stat.R.).} The fraction of tracked dials (Tracked), and how strongly on average each dial is tracked (Stat.R.).
    \item \textbf{Semantic recovery (Sem.R.).} The fraction of dials that have a discovered feature that is both about it and tracks it. By definition, it is at most Tracked: the gap between the two counts dials tracked under a misaligned name.
    \item \textbf{Attributable features (Attrib.).} The fraction of the discovered features that track a single dial.
    \item \textbf{\metricLeakage{} (LK).} The average correlation between a dial and the attributable features other than the one that best tracks it. Since dials are near-orthogonal by construction, a feature that measures one dial should be uncorrelated with the others; leakage quantifies how much of the other dials nonetheless bleeds into its scores (lower is better).
\end{itemize}

\paragraph{Benchmarks.} We compare our approach to direct LLM feature extraction and its variants. Specifically, we provide a powerful LLM (\textsc{GPT-5.4 (high reasoning)}) with the whole dataset or a subset and ask it to identify features and formulate dataset-specific guidelines for scoring them. We then use a cheaper model (\textsc{GPT-5.4 mini}) to score each cover letter on each feature. More precisely, we consider the following variants:

\begin{itemize}[leftmargin=*]
    \item \textbf{Direct LLM:} An LLM is queried on the full dataset to discover features, after which a cheaper LLM scores each item along each feature. We also consider \textbf{Direct LLM--Multi-score}, where multiple agents score each item with slightly different prompts, and their scores are averaged. This variant approximately matches our approach's cost, allowing us to evaluate whether the benefits arise from using multiple agents or from our specific way of orchestrating them.

    \item \textbf{Local LLMs:} Multiple agents each examine a randomly selected subset of the dataset and discover features. We pool their features and use an LLM to merge equivalent descriptions into a shared set (see Remark~\ref{rem:why-taxo} in Appendix~\ref{app:experiment-details} for the choice of merging procedure). A single agent then scores each item along these features. Analogously, we also consider \textbf{Local LLMs--Multi-score}. These variants help assess whether the performance benefits arise from distributing feature discovery across multiple agents or from our pairwise comparisons and statistical aggregation.

\end{itemize}

\paragraph{Ablations.} To examine the contributions of statistical scoring and reliability filtering, we retain the features discovered through our sub-agents' pairwise comparisons and consider two ablations, each with single-agent and multi-agent scoring.

\begin{itemize}[leftmargin=*]
    \item \textbf{Pairwise comparisons + LLM scoring:} LLMs directly score all our discovered features, replacing statistical scoring and its reliability filtering.

    \item \textbf{Reliable features + LLM scoring:} An LLM scores only the features retained by our statistical model's reliability filter. Comparing these two ablations isolates the contribution of filtering under LLM scoring.
\end{itemize}

\paragraph{Results.} Table~\ref{tab:main} reports the metrics of each
benchmark and ablation. Our approach outperforms direct LLM feature
extraction, even though that benchmark uses one of \textsc{OpenAI}'s most
capable models available at the time of the experiments. It tracks all
16 ground-truth features, compared with 13 for Direct LLM, and achieves
complete semantic recovery: the discovered features both track the
underlying signals and describe them meaningfully. The three missing
dials are a discovery failure rather than a scoring failure: at matched
cost, Direct LLM--Multi-score still tracks the same 13. More of our
features are attributable to a single ground-truth feature (16 of 21
versus 13 of 21), and leakage is lower ($0.090$ versus $0.109$). Direct
LLM is however cheaper (\$0.26 versus \$3.45).

Our approach also outperforms the multi-agent alternatives, whether
multiple agents are used for feature discovery (Local LLMs), scoring
(Multi-score), or both, with statistically significant gains in
tracking, attribution, and leakage over Local LLMs
(Figure~\ref{fig:paired}). Multi-agent scoring improves statistical
recovery but increases leakage, while multi-agent feature discovery
reduces recovery compared with a single agent with the full dataset.
The benefits therefore do not follow simply from using multiple
agents: how we divide their work into pairwise comparisons and
aggregate their outputs through a statistical model matters.

The ablations also achieve high tracking and semantic recovery,
consistent with their use of the same features as the full pipeline, but without statistical scoring and its reliability filter, attribution is
substantially lower (13--14 attributable features versus 16) and
leakage higher ($0.103$--$0.104$ for Multi-score methods versus $0.090$), despite a slightly higher
\metricStatR{} ($0.860$--$0.868$ versus $0.850$) and twice the cost. The
resulting scores therefore separate the ground-truth signals less
clearly. Filtering alone raises the attributable fraction under
multi-agent scoring from 14 of 32 to 14 of 21 while preserving
complete tracking and semantic recovery, showing that it removes
features that track no dial. Statistical scoring on the same 21
features then raises it to 16.

\begin{table}[t]
\centering
\small
\setlength{\tabcolsep}{4pt}
\resizebox{\linewidth}{!}{%
\begin{tabular}{@{}c l r r ccccc@{}}
\toprule
 & Method & Cost & $K$ & Tracked & Stat.R. & Sem.R. & Attrib. & LK$\downarrow$ \\
\midrule
 & \textbf{Ours} & 3.45 & 21 & \textbf{100.0\%} & 0.850 & \textbf{100.0\%} & \textbf{76.2\%} & \textbf{0.090} \\
\midrule
\multirow{4}{*}{\rotatebox{90}{\textcolor{black}{\scriptsize Benchmarks}}} & Direct LLM & \textbf{0.26} & 21 & 81.2\% & 0.732 & 81.2\% & 61.9\% & 0.109 \\
 & Direct LLM--Multi-score & 3.42 & 21 & 81.2\% & 0.787 & 81.2\% & 66.7\% & 0.118 \\
 & Local LLMs & 2.35 & 23 & 75.0\% & 0.702 & 68.8\% & 47.8\% & 0.110 \\
 & Local LLMs--Multi-score & 3.27 & 23 & 75.0\% & 0.737 & 68.8\% & 47.8\% & 0.113 \\
\midrule
\multirow{4}{*}{\rotatebox{90}{\textcolor{black}{\scriptsize Ablations}}} & Pairwise comparisons + LLM scoring & 3.67 & 32 & 93.8\% & 0.763 & 87.5\% & 40.6\% & 0.093 \\
 & Pairwise comparisons + LLM scoring--Multi-score & 6.55 & 32 & \textbf{100.0\%} & 0.860 & \textbf{100.0\%} & 43.8\% & 0.104 \\
 & Reliable features + LLM scoring & 3.60 & 21 & 93.8\% & 0.792 & 93.8\% & 66.7\% & 0.098 \\
 & Reliable features + LLM scoring--Multi-score & 6.54 & 21 & \textbf{100.0\%} & \textbf{0.868} & \textbf{100.0\%} & 66.7\% & 0.103 \\
\bottomrule
\end{tabular}%
}
\caption{Per-method estimates on the single full-data run. Costs are
Batch-API prices in USD; $K$ is the number of features the method reports. Tracked, Sem.R., Attrib. in \%; Stat.R. and LK are correlations.}
\label{tab:main}
\end{table}

\section{Application to a Real Dataset}
\label{sec:cmv}

We next evaluate our approach on a real-life dataset whose underlying features are unknown: online debates from Reddit's ChangeMyView forum. The central question is whether automatically discovered features provide interpretable insights into persuasion while retaining sufficient signal to predict which arguments change a participant's view.

\paragraph{The ChangeMyView dataset.}
In the /r/ChangeMyView (CMV) subreddit, a user (the original poster, OP) posts an opinion with their reasoning and invites counterarguments. The OP then awards an explicit delta ($\Delta$) to any reply that changed their view, providing labels of persuasion success. \citet{tan2016winning} turned this signal into a controlled paired prediction task: predicting which of two responses to a post actually convinced the OP. Hence, the items being assessed by our feature discovery pipeline will be \emph{exchanges}, namely pairs of the form (OP post, reply). See Appendix~\ref{app:experiment-details-cmv} for details about the experiment. Prior work on this task either hand-engineers features (word overlap, style and bag-of-words statistics in \citet{tan2016winning}; argumentation-theoretic features in \citet{wachsmuth2017computational}) or learns non-interpretable embeddings~\citep{durmus2018exploring}.

\begin{figure}[t]
  \begin{minipage}[c]{0.60\linewidth}\centering
    \includegraphics[width=\linewidth]{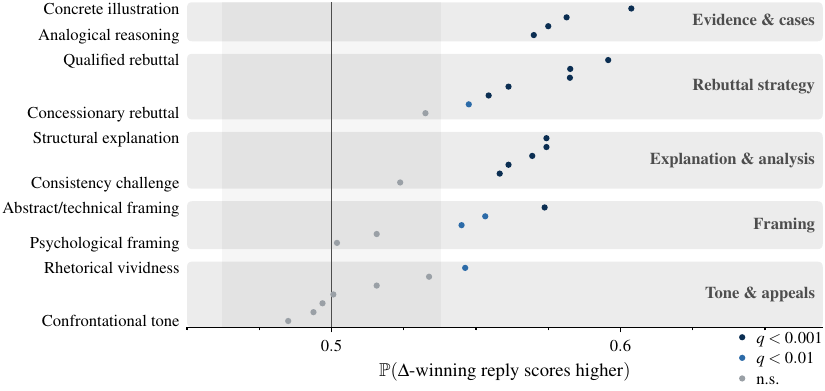}
    \caption{\textbf{Effectiveness of the discovered argumentation techniques.} Discovered features are gathered in five categories. For each discovered feature (D + PW), we report the \emph{effectiveness}, i.e. the fraction of the test pairs in which the reply that earned the $\Delta$ scores higher on that feature; $0.5$ is chance. BH $q$ is the Benjamini--Hochberg false-discovery rate across the $29$ tests. See Figure~\ref{fig:cmv-win-rates-full} for the figure showcasing every feature's effectiveness with confidence intervals.}
    \label{fig:cmv-win-rates}
  \end{minipage}\hfill
  \begin{minipage}[c]{0.38\linewidth}\centering\footnotesize
    \setlength{\tabcolsep}{4pt}
    \begin{tabular}{@{}l r l@{}}
    \toprule
    Arm & Dims & Accuracy \\
    \midrule
    Tan & 3{,}205 & 63.75\% \scriptsize{$[60.38, 67.00]$} \\
    D + PW & 29 & 62.63\% \scriptsize{$[59.38, 66.00]$} \\
    D$_{\mathrm{r}}$ + PW & 16 & 62.13\% \scriptsize{$[58.88, 65.50]$} \\
    Emb. & 3{,}072 & 61.63\% \scriptsize{$[58.25, 65.00]$} \\
    D$_{\mathrm{r}}$ + BT & 16 & 61.50\% \scriptsize{$[58.13, 64.88]$} \\
    D + BT & 29 & 61.38\% \scriptsize{$[58.00, 64.75]$} \\
    P + PW & 16 & 61.25\% \scriptsize{$[58.00, 64.63]$} \\
    P + BT & 16 & 61.13\% \scriptsize{$[57.75, 64.38]$} \\
    S + PW & 42 & 59.88\% \scriptsize{$[56.50, 63.25]$} \\
    \#words & 1 & 59.25\% \scriptsize{$[55.88, 62.75]$} \\
    Zero-shot & -- & 56.94\% \scriptsize{$[54.19, 59.88]$} \\
    \bottomrule
    \end{tabular}
    \captionof{table}{\textbf{Held-out accuracy of each method}, with $95\%$ bootstrap percentile intervals over the $800$ held-out pairs; chance $=50\%$. Emb.: \textsc{text-embedding-3-large} embeddings. Tan: hand-engineered features of \citet{tan2016winning}. Zero-shot: the judge predicts the $\Delta$-winning reply. \#words: reply length.}
    \label{tab:cmv-predictive}
  \end{minipage}
\end{figure}

\paragraph{Extracting interpretable features.} We apply our pairwise feature-discovery procedure to a subset of 300 training exchanges, grouping the resulting descriptions into 29 reply-side features, of which our reliability diagnostics retain $16$. These features capture human-recognizable aspects of argumentation, including \emph{concrete illustrations}, \emph{structured rebuttals}, \emph{conceptual clarification}, and \emph{confrontational tone}, without requiring hand-engineering. Table~\ref{tab:cmv-win-rates} in Appendix~\ref{app:discovered-features-cmv} lists all 29 features and their descriptions. Figure~\ref{fig:cmv-win-rates} illustrates their associations with persuasion: it reports how often a successful reply scores higher on each feature than its matched unsuccessful reply. For instance, features related to \emph{evidence and cases} show stronger positive associations with persuasion than those related to \emph{tone and appeals}. \emph{Confrontational tone} shows a negative association with persuasion, although this association is not statistically significant.

\begin{figure}[t]
  \centering
  \includegraphics[width=\linewidth]{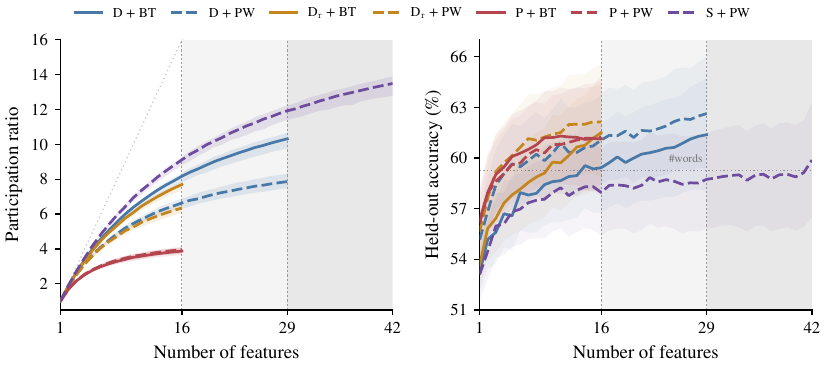}
  \caption{\textbf{Discovered features span more effective dimensions than prior ones while retaining predictive accuracy.} In both panels, each point is an average over random $k$-subsets of the feature set, so that sets of different sizes are compared at matched size. \emph{Left:} noise-corrected participation ratio of the feature-score correlation matrix (Spearman) over the $1{,}600$ held-out exchanges; dotted: $y=k$. \emph{Right:} held-out accuracy of the $\ell_1$ logistic regression on the $800$ held-out pairs; chance $=50\%$. D: discovered ($29$ features); D$_{\mathrm{r}}$: reliable subset of D ($16$); P: prior ($16$); S: sample-based ($42$); BT/PW: Bradley--Terry/pointwise scoring; \#words: reply length. Bands: $95\%$ bootstrap percentile intervals over held-out exchanges (left) and exchanges + pairs (right). Shaded: $k>16$, where P and D$_{\mathrm{r}}$ are not defined, and $k>29$, where only S is defined.
  Details in Appendix~\ref{app:additional-measures}.}
  \label{fig:cmv-dimension-curve}
\end{figure}

\paragraph{Benchmarks and ablations.}
We write D for the $29$ discovered features and D$_{\mathrm{r}}$ for the
$16$ that pass the reliability filter, and compare both to isolate the
contribution of the statistical model. In addition, we compare our
approach to features obtained from two types of direct LLM call. In the
first, similarly to the benchmarks of
Section~\ref{sec:feature_extraction_empirical_results},
\textsc{GPT-5.4 (high reasoning)} is given a sample of exchanges and asked
to propose distinct features, each with a name and a short definition (S,
for samples; $42$ features). In the second, the model is given no data,
only a brief description of the persuasion task (P, for prior; $16$
features).

For each feature set, the features are held fixed and the training and held-out exchanges are scored along them with \textsc{GPT-5.4 mini}, following the two scoring protocols of Section~\ref{sec:method} (\emph{Scoring new items}). In pointwise scoring (\textbf{PW}), one call per exchange assigns it a $0$--$10$ score on every
feature (Appendix~\ref{app:methods-cmv}). In pairwise scoring (\textbf{BT}), the LLM compares two exchanges on every feature at once and a Bradley--Terry model with a positional-bias term is fitted per feature: training exchanges are compared among themselves, and each held-out exchange is then compared against $16$ training exchanges only, whose scores are kept frozen, so that all scores
lie on a common scale and new items are scored at fixed cost. Since S has $42$ features, its pairwise calls would be prohibitively costly, so we report S under PW only.

\paragraph{Expressiveness of the discovered features.}
The left panel of Figure~\ref{fig:cmv-dimension-curve} reports the
noise-corrected participation ratio~\citep{Gao2017pr} of random subsets of each feature set, which summarizes the effective number of uncorrelated dimensions in the feature scores. Features derived from the data, whether from a sample of exchanges (S) or from our pairwise comparisons (D, D$_{\mathrm{r}}$), are far less redundant than features elicited from the model's prior (P): at $16$ features, D and D$_{\mathrm{r}}$ span about eight effective dimensions under BT scoring and six to seven under pointwise scoring, against about four for P, whose curve has flattened before $16$. The higher dimensionality under BT scoring is consistent with our intuition that aggregating comparisons at the dataset level preserves distinctions that myopic pointwise LLM scoring blurs.

\paragraph{Predictive signal.}
We fit an $\ell_1$-regularized logistic regression on the scores of each feature set to predict which of two replies to a post persuaded the OP. Table~\ref{tab:cmv-predictive} reports the held-out accuracy of each approach, alongside hand-engineered features, embeddings, and simpler baselines. Our approach (D and D$_{\mathrm{r}}$) achieves accuracy comparable to the best methods (see Table~\ref{tab:cmv-predictive-2} for full data estimates) while providing interpretable features.

Beyond raw accuracy, how much predictive signal a feature set carries relative to its size matters for interpretability. The right panel of Figure~\ref{fig:cmv-dimension-curve} therefore reports
held-out accuracy against the number $k$ of features, averaged over random
$k$-subsets of each set. Among the expressive, non-redundant feature sets identified in Figure~\ref{fig:cmv-dimension-curve} (S, D, D$_{\mathrm{r}}$), ours are the most accurate: the direct-LLM sample-elicited set S has more features and a high effective dimensionality, yet it is the least accurate at almost every $k$, $4.2$ points below D$_{\mathrm{r}}$\,+\,PW at $k = 16$ and $3.6$ points below D$_{\mathrm{r}}$\,+\,PW on average over $k$ (paired $95\%$ CI $[1.8, 5.4]$; see Figure~\ref{fig:cmv-diff-dr} in Appendix~\ref{app:additional-measures}). This indicates that part of the variation S captures is irrelevant to persuasion. Conversely, the features elicited from the LLM without data access (P) are among the most predictive at low $k$ but plateau quickly, illustrating the limits of data-blind feature elicitation.

The results also show that the statistical model plays an important role. The filtered set D$_{\mathrm{r}}$ has about half as many features as D, yet loses at most half a point of accuracy. Hence, the statistical model retains the informative features, yielding a compact, highly interpretable set with a strong predictive signal. The right panel of Figure~\ref{fig:cmv-dimension-curve} confirms this: at every $k$, random subsets of D$_{\mathrm{r}}$ are more accurate than equally sized subsets of D, under both scorings, by one point on average over $k \le 16$ (paired $95\%$ CIs $[-0.1, 2.2]$ for PW and $[0.0, 2.1]$ for BT), showing that features retained by the filter carry more signal than the unfiltered ones.

\section{Conclusion}
\label{sec:conclusion}

We introduced an approach that discovers human-interpretable features from unstructured data using LLMs: judges compare items pairwise, and a statistical choice model aggregates their rationales. On both a controlled experiment and real-life data, it recovers features significantly better than a direct LLM call, even with \textsc{GPT-5.4 (high reasoning)}. The recovered features capture most of the predictive signal with fewer, less redundant features, while matching the accuracy of established baselines. Appendix~\ref{app:verdict-extension} further extends the framework to auditing an LLM's own decisions.

\subsection*{Reproducibility statement}
Code to reproduce every experiment in the paper is available at the repository \url{https://github.com/ArnaudoB/Interpretable-Feature-Discovery-LLM}. The mathematical guarantees of the introduced statistical choice model are proved in Appendix~\ref{app:model-theory}, and full details of every experiment are given in Appendices~\ref{app:experiment-details} and~\ref{app:experiment-details-cmv}.

\section*{Statement on LLM Usage}

Because LLMs are the object of study, the judge outputs, pairwise comparisons and pointwise scores analyzed throughout, as well as the synthetic cover-letter corpus of Section~\ref{sec:feature_extraction_empirical_results}, were produced by LLMs through API calls.

Beyond this experimental role, our use of generative AI was as follows.
The text of the paper was written by the authors, with LLMs used for occasional rewording. Portions of the experimental code were written with
LLM assistance; all AI-assisted code was reviewed and tested by the authors, who take full responsibility for its correctness. The mathematical claims and proofs were developed by the authors without AI assistance; as an additional verification step, we used Claude Fable~5 (Anthropic) to audit the proofs, which surfaced an error; the authors diagnosed and corrected it by hand.

\newpage

\bibliographystyle{unsrtnat}
\bibliography{references}

\newpage
\appendix
\numberwithin{equation}{section}

\addtocontents{toc}{\protect\setcounter{tocdepth}{2}}   %
\renewcommand{\contentsname}{Contents of the Appendix}
\tableofcontents

\clearpage

\section{Extension: Interpretable LLM-based scoring}
\label{app:verdict-extension}

The goal of this small extension is to provide a method inspired by the feature discovery pipeline but well-suited to the goal of interpretable scoring. LLMs are increasingly used to score items (essays, résumés, or other LLMs) through pairwise comparisons, and the resulting verdicts are typically aggregated by fitting a Bradley--Terry model. This yields a single scalar score per item, whereas a decision-maker would want an overall score that decomposes into per-feature scores along interpretable dimensions. For example, for an LLM grading scientific research papers, have an overall score per-paper that aggregates scores on atomic features \emph{discovered from the data}: scientific rigor, novelty of the results, etc.

\paragraph{Method.} Established methods (e.g. \citet{li2024crowdsourceddatahighqualitybenchmarks}) have the LLM make pairwise comparisons and fit a Bradley--Terry model on their outputs to yield scores. Performing this on many features gives several scores that can be aggregated to one overall score that decomposes cleanly on atomic features, but it requires knowing the assessable features. 

Our method works by asking the LLM judge for a holistic verdict, and asking it to justify it by specifying \emph{along which feature} the winning item is better than the losing one. The canonicalization of cited qualities into $K$ shared features is identical to Section~\ref{sec:method}. The data for a compared pair $p$ are therefore the overall verdict $w_p=\1_{\text{first-presented item wins}}$ and the citation indicators $\vc_{p}\in\{0,1\}^K$, where $c_{p,k}=1$ if feature $k$ was cited as a quality the winner exhibits more; the direction of each citation is thus implied by $w_p$.

\paragraph{Statistical model.}
To infer per-item per-feature scores from the judge's justifications, our model links the holistic verdict to the per-feature scores by
setting:
\begin{equation}\label{eq:holistic}
\PP(w_p=1)=\sigma\left(\sum_{k=1}^{K}\Delta_{p,k}+\beta\right),
\end{equation}
where $\Delta_{p,k}$ is the score gap on feature $k$ (first minus
second) and $\beta$ is the position coefficient. Hence, the model simply posits that the larger the holistic score, the likelier an item is to win the comparisons it is involved in.

To model citations, the model posits that (conditionally on the overall verdict) each feature is cited independently, with a probability that
increases in the margin by which that feature favors the \emph{winner} (because we ask the LLM to justify its winning-item choice):
\begin{equation}\label{eq:justif}
\PP(c_{p,k}=1\mid w_p)=\sigma\big((2w_p-1)\Delta_{p,k}+\gamma_k\big).
\end{equation}
Here $2w_p - 1$ is $1$ when the first item won and $-1$ otherwise.

The likelihood of an observed pair is therefore:
\begin{equation}\label{eq:ext-lik}
\PP(w_p,\vc_p)=
\sigma\Big(\sum_{k = 1}^{K}\Delta_{p,k}+\beta\Big)^{w_p}
\Big(1-\sigma\Big(\sum_{k = 1}^{K}\Delta_{p,k}+\beta\Big)\Big)^{1-w_p}
\prod_{k=1}^{K}
\sigma(\eta_{p,k})^{c_{p,k}}\big(1-\sigma(\eta_{p,k})\big)^{1-c_{p,k}},
\end{equation}
with $\eta_{p,k}=(2w_p-1)\Delta_{p,k}+\gamma_k$, and the NLL sums
$-\log$ of \eqref{eq:ext-lik} over pairs, ridge-penalized as in
Eq.~\eqref{eq:estimator}.

The method therefore not only discovers the features and estimates each item's score on them, but also links these per-feature scores to a holistic verdict, so that the overall judgment decomposes additively into named, interpretable components.

\paragraph{Interpretability.} To ensure the model's identifiability, the fitted scores are mean-zero. If one sets $\sigma_k := \mathrm{std}(\mS_{\cdot,k})$, $\tilde{S}_{i, k} :=  \frac{S_{i, k}}{\sigma_k}$ and $\tilde{\Delta}_{p, k} = \frac{\Delta_{p, k}}{\sigma_k}$, Equation~\eqref{eq:holistic} becomes:
\begin{equation}\label{eq:holistic_z_score}
\PP(w_p=1)=\sigma\left(\sum_{k=1}^{K}\sigma_k\tilde{\Delta}_{p,k}+\beta\right),
\end{equation}
with each $\tilde{S}_{i,k}$ having mean zero and variance 1. One can hence interpret $\sigma_k$ as the judge's perceived importance of feature $k$ in the overall verdict, yielding interesting interpretability insights about what each judge rewards.

\begin{figure}[t]
    \centering
    \includegraphics[width=\linewidth]{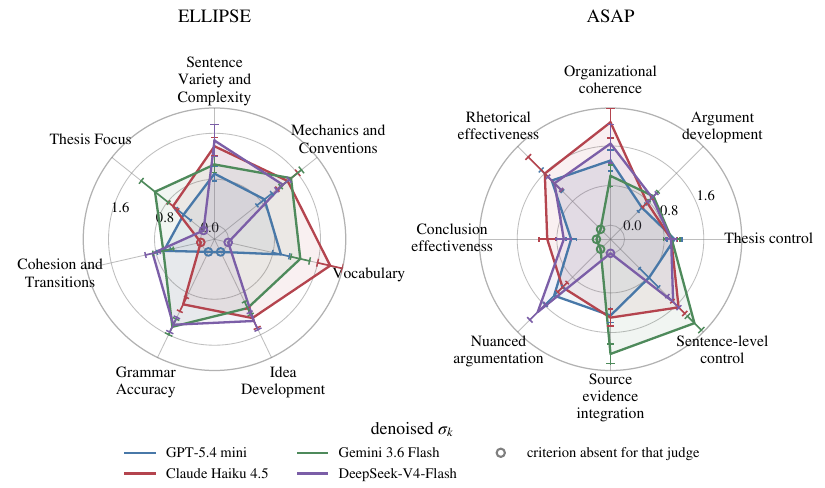}
    \caption{\textbf{Which features matter for each judge on each dataset.} Radial axis is $\sigma_k$ (with estimation noise removed) on each judge's reliable features ($\rho>0.5$, $\kappa<100$). Bars are 95\% delta-method intervals from the sandwich covariance. Axes are cross-judge clusters formed by meaning (\textsc{GPT-5.4 (high reasoning)}), restricted to constructs at least three of the four judges discovered (see Figures~\ref{fig:radar_ellipse} and~\ref{fig:radar_asap} for the full plots). An open marker denotes a construct absent from that judge's reliable basis (never surfaced, or removed by the identifiability screen), not a zero estimate.}
    \label{fig:cross_judge_radar}
\end{figure}

\paragraph{Application.}
We demonstrate the method on two essay corpora with human holistic scores: (i) $n=100$ essays from ELLIPSE~\citep{crossley2023ellipse}, a corpus of English-language-learner writing from grades 8--12 scored for overall language proficiency (as well as six analytic dimensions), and (ii) $n=100$ essays from ASAP~2.0~\citep{crossley2025asap}, a corpus of persuasive essays from grades 6--10 scored for overall writing quality. Each corpus is compared along a $d=40$-regular graph. To assess whether the discovered constructs and their importance depend on the judge, we run the pipeline with four judges from four model families: \textsc{GPT-5.4 mini}, \textsc{Claude Haiku 4.5}, \textsc{Gemini 3.6 Flash}, and \textsc{DeepSeek-V4-Flash}. Canonicalization is performed by \textsc{GPT-5.4 (high reasoning)} in every case, and every feature passes through reliability gates analogous to those of Section~\ref{sec:method}. To compare the judges' feature sets, we match features across judges semantically with \textsc{GPT-5.4 (high reasoning)} using Prompt~\ref{prm:cross-judge}; Figure~\ref{fig:cross_judge_radar} shows the resulting shared constructs.

\begin{figure}[ht]
    \centering
    \includegraphics[width=\linewidth]{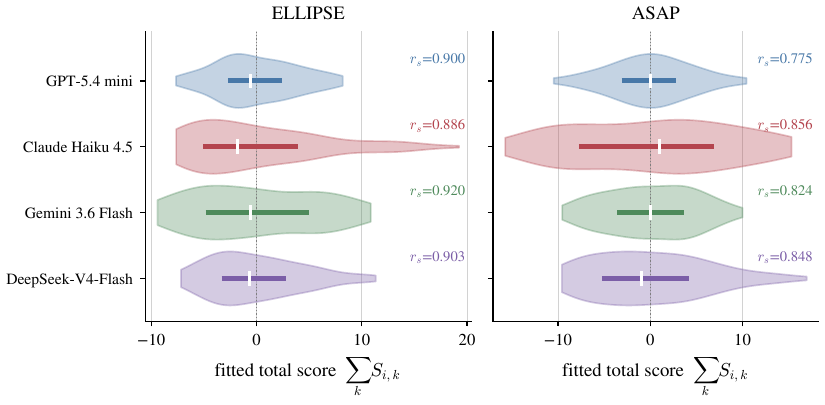}
    \caption{\textbf{Per-judge overall score distribution and its agreement (Spearman rank correlation) with human scores.}}
    \label{fig:total_score_distribution}
\end{figure}

\paragraph{Results.}
Figure~\ref{fig:cross_judge_radar} shows that the judges agree on
\emph{which} constructs structure their verdicts far more than on
\emph{how much} each one separates essays. On ELLIPSE, every
construct shared by three judges or more is lexical, mechanical, or
organizational, consistent with what varies in L2 learner writing; on
ASAP 2.0, the four-judge constructs are discourse-level (thesis control,
organizational coherence, argument development, sentence-level control). Since the same pipeline, pairs, and prompts (up to corpus
framing) produce both bases, this reorganization is attributable to the
corpora rather than to any judge's habits: the method recovers what the
task actually varies on. Within the shared constructs, magnitudes
disagree: \textsc{GPT-5.4 mini} is the most compressed judge throughout
(smallest $\sigma_k$ on $5$ of $7$ ELLIPSE and $4$ of $8$
ASAP 2.0 consensus axes), thesis control on ASAP 2.0 is the one construct on which all four judges essentially coincide ($\sigma_k\in[0.94,1.01]$), and sentence-level control is the most contested ($0.82$ to $2.11$). Interestingly, the largest spreads on both corpora sit on features surfaced by a \emph{single judge}: the sign of an idiosyncratic emphasis. Figure~\ref{fig:total_score_distribution} reports each judge's fitted total $\sum_k S_{i,k}$ against the human holistic score: rank agreement is high on both ELLIPSE (Spearman rank correlation $r_s = 0.89$--$0.92$) and ASAP 2.0 ($r_s = 0.77$--$0.86$). Each total decomposes exactly into named per-feature scores, so unlike a scalar Bradley--Terry fit, every verdict can be audited feature by feature: the property that motivated the extension.

\subsection{Details of the extension}
\label{sec:extension-appendix}

\paragraph{Data and comparison design.}
For each corpus we sample $n=100$ essays, band-balanced on the human
holistic score, and build $2{,}000$ pairs as the union of $40$ random
perfect matchings, so every essay appears in exactly $40$ comparisons.
The essay sample, the pair list, and the presentation order are generated
once and are byte-identical across the four judges of a corpus: within a corpus, only the judge varies. Between corpora, only the corpus framing of the elicitation prompt changes (ELLIPSE frames English-language proficiency of grade 8--12 learners; ASAP~2.0 frames source-based argumentative quality and inlines the source article and
assignment verbatim); schema, field order, parser, and all downstream
prompts are unchanged, so the ELLIPSE--ASAP 2.0 contrast is a corpus
contrast, not a prompt contrast.

\paragraph{Elicitation.}
One call per pair elicits a forced binary verdict (Prompts~\ref{prm:elicit-ellipse} and~\ref{prm:elicit-asap}) followed by the qualities the winner exhibits more than the loser, each with evidence, a
short label, and a one-sentence description. The prompt names no
features and contains no worked examples (to avoid priming the
constructs under study), forbids bundled qualities (``and'', ``/'',
``\&''), and rules out of scope the topic, the stance, length in itself,
and the corpora's anonymization placeholders. The output schema emits
the verdict \emph{before} the qualities, matching the factorization of
the likelihood \eqref{eq:ext-lik}. Reasoning modes are disabled on all four judges.

\paragraph{Canonicalization and reliability gates.}
Per judge, cited phrases with at least $3$ citations are grouped by
meaning into features by \textsc{GPT-5.4 (high reasoning)}, with a
catch-all ``Other'' bucket dropped downstream; group membership is
validated programmatically. The model of Equation~\eqref{eq:ext-lik} is fitted by joint MLE (L-BFGS, analytic gradients, ridge $10^{-3}$), and each feature passes a reliability screen built on the same principle as
Section~\ref{sec:method}: a condition number $\kappa_k$ of the score block of
the Fisher information of model~\eqref{eq:ext-lik}, and a signal-to-noise
ratio $\rho_k$ obtained by subtracting the sandwich estimation variance from
the observed score spread. The statistics differ from those of the main
pipeline because the likelihood does; we retain features with $\rho>0.50$ and
$\kappa<100$, then refit the model on the surviving features. Across the eight judge--corpus cells, the screen retains between $6$ and $16$ features.

\paragraph{Cross-judge consolidation.}
The four judges' reliable features are grouped into shared constructs by
a single \textsc{GPT-5.4 (high reasoning)} call per corpus, with at most
one feature per judge and per construct. The prompt requires
singletons to remain visible (a construct found by one judge is a
finding, not noise) and receives no statistics so semantic grouping cannot be contaminated by how essays scored. Figure~\ref{fig:cross_judge_radar} displays constructs recovered by at least three judges.

\paragraph{Reported quantity and denoising.}
The radial axis reports a noise-corrected spread: the observed
$\sigma_k$ mixes essay-to-essay variation with
estimation error, so we subtract the mean sandwich variance (Appendix~\ref{app:uq}) of the fitted scores, with
delta-method standard errors from the sandwich covariance. Denoising
shrinks ELLIPSE spreads by $10.5\%$ on average and reorders $10$ of $34$
features within their own judge.

\paragraph{Fit quality and position bias.}
Spearman correlation between the fitted total $\sum_k S_{i,k}$ and the human holistic score is $0.89$--$0.92$ on ELLIPSE and $0.77$--$0.86$ on ASAP 2.0. The fitted position coefficient $\beta$ varies in sign and magnitude across judges ($|\hat\beta|$ up to $2.6$) and tracks each judge's raw first-position win rate on both corpora: position bias is a genuine property that depends on the judge and on the dataset. 

\paragraph{Sensitivity and design caveats.}
We verified the sensitivity of the reliability gates: at $\rho>0.70,\kappa<50$ the ELLIPSE reliable feature sets shrink from $9/9/9/7$ to $6/6/6/7$ features (\textsc{GPT-5.4 mini}, \textsc{Claude Haiku 4.5}, \textsc{Gemini 3.6 Flash}, \textsc{DeepSeek-V4-Flash}) and the constructs shared by three judges or more fall from $7$ to $5$, so consensus counts depend on the gates and we report the setting used. Because different judges lose different features to $\kappa$, the consensus counts \emph{understate} agreement (e.g., Vocabulary is a $3/4$ construct on ELLIPSE only because the screen removed DeepSeek's). The citation floor interacts similarly: in a preliminary three-judge run, a floor of $5$ let thinly spread phrasing make constructs appear less shared than they were, which motivated the minimum-citation threshold of 3.

\begin{figure}[t]
    \centering
    \includegraphics[width=0.90\linewidth]{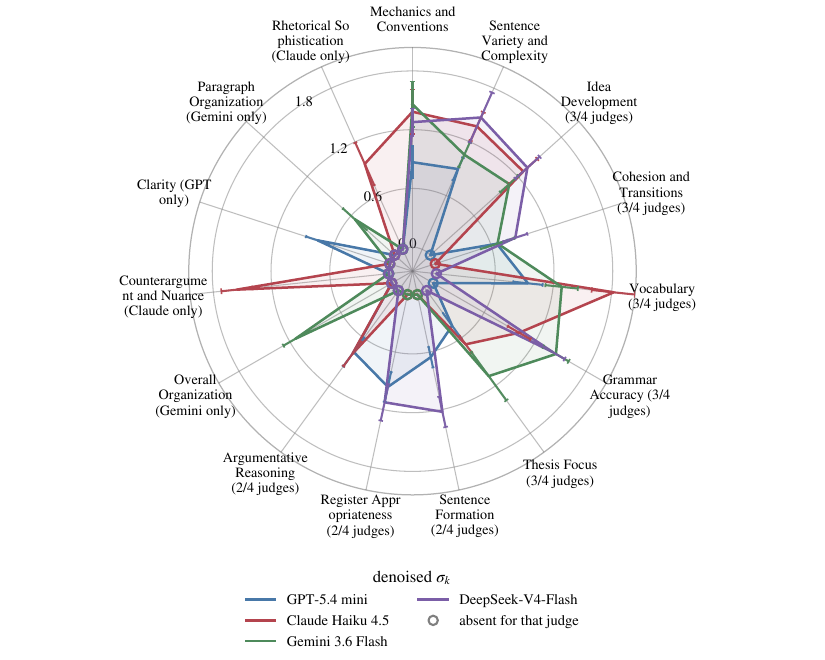}
    \caption{\textbf{Denoised per-feature spread by judge on ELLIPSE.}
Axes are cross-judge constructs formed by meaning (\textsc{GPT-5.4 (high reasoning)}); annotations give how many of the four judges discovered each construct. The radial coordinate is the denoised spread $\sigma_k$ of each judge's feature mapped to that construct, on per-judge standardized reliable feature sets ($\rho>0.5$, $\kappa<100$); bars are 95\% delta-method intervals from the sandwich covariance. An open marker denotes a construct absent from that judge's reliable basis (never surfaced, or removed by the identifiability screen), not a zero estimate.}
    \label{fig:radar_ellipse}
\end{figure}

\begin{figure}[t]
    \centering
    \includegraphics[width=0.90\linewidth]{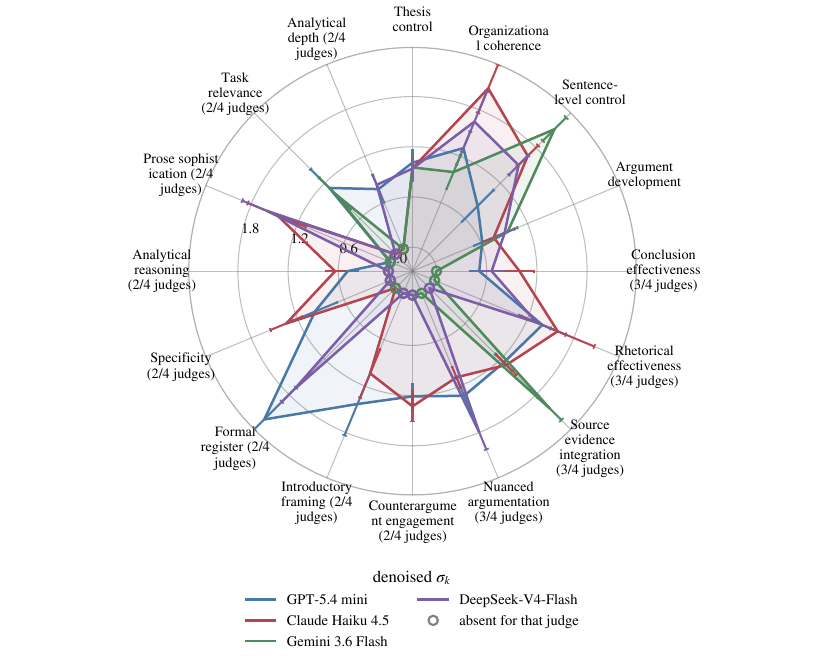}
    \caption{\textbf{Denoised per-feature spread by judge on ASAP~2.0.}
    Same construction, quantities, and conventions as
    Figure~\ref{fig:radar_ellipse}, on the ASAP 2.0 corpus.}
    \label{fig:radar_asap}
\end{figure}

\paragraph{Model calibration.}

The extension model assigns two probability forecasts of observable binary
events, one per channel. The winner gate, $\PP(w_p = 1) = \sigma\bigl(\sum_k \hat\Delta_{p,k} + \hat\beta\bigr)$, is the probability that the first-shown essay wins the overall verdict, scored once per comparison; the citation gate, $\PP(c_{p,k} = 1 \mid w_p) = \sigma\bigl((2w_p - 1)\hat\Delta_{p,k} + \hat\gamma_k\bigr)$, is the probability that feature $k$ is cited given the verdict, increasing in the margin by which $k$ favors the declared winner, scored on every (comparison, feature) cell of the reliable feature set. We score both as forecasts, out of fold ($5$-fold cross-validation over the $2{,}000$ comparisons, so that every prediction comes from parameters estimated without it, with $95\%$ bootstrap intervals over comparisons) independently for each of the four judges and on both corpora, so calibration is also a robustness check across judges.

\begin{figure}
    \centering
    \includegraphics[width=\linewidth]{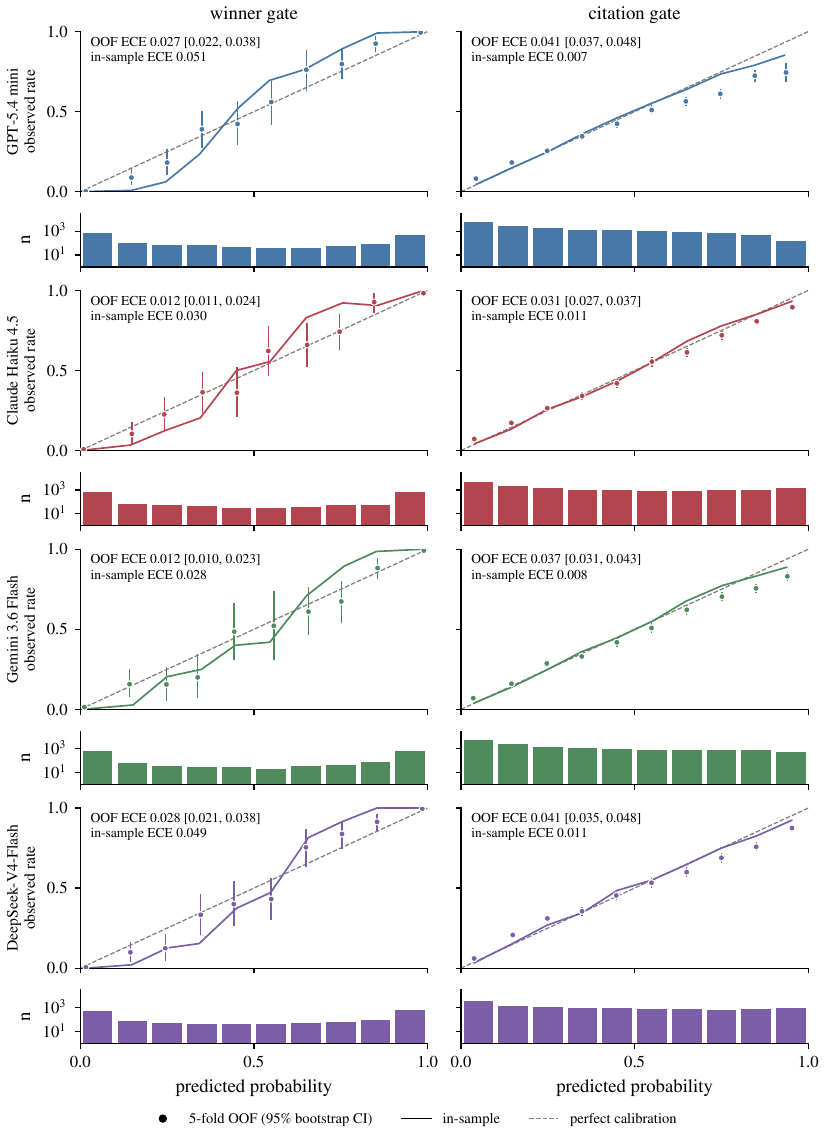}
    \caption{Out-of-fold reliability diagrams for the winner gate (left;
    unit: comparison) and the citation gate (right; unit: cell) of the
    model on ELLIPSE, one row per judge. Units are grouped into ten equal-width bins $B_1,\dots,B_{10}$ of predicted probability $\hat p_i$; points show the observed event rate $\bar y_b = \frac{1}{|B_b|}\sum_{i\in B_b} y_i$ against the mean prediction $\bar p_b = \frac{1}{|B_b|}\sum_{i\in B_b}\hat p_i$, with $95\%$ bootstrap intervals (resampling comparisons, resp.\ cells).
    Faint line: in-sample fit; dashed: perfect calibration
    ($\bar y_b=\bar p_b$). The expected calibration error is
    $\mathrm{ECE}=\sum_{b}\frac{|B_b|}{N}\lvert \bar y_b-\bar p_b\rvert$,
    with $N=\sum_b |B_b|$. Bottom strips: $|B_b|$ per bin, log scale.}
    \label{fig:calibration_reliability_ellipse}
\end{figure}

\begin{figure}
    \centering
    \includegraphics[width=\linewidth]{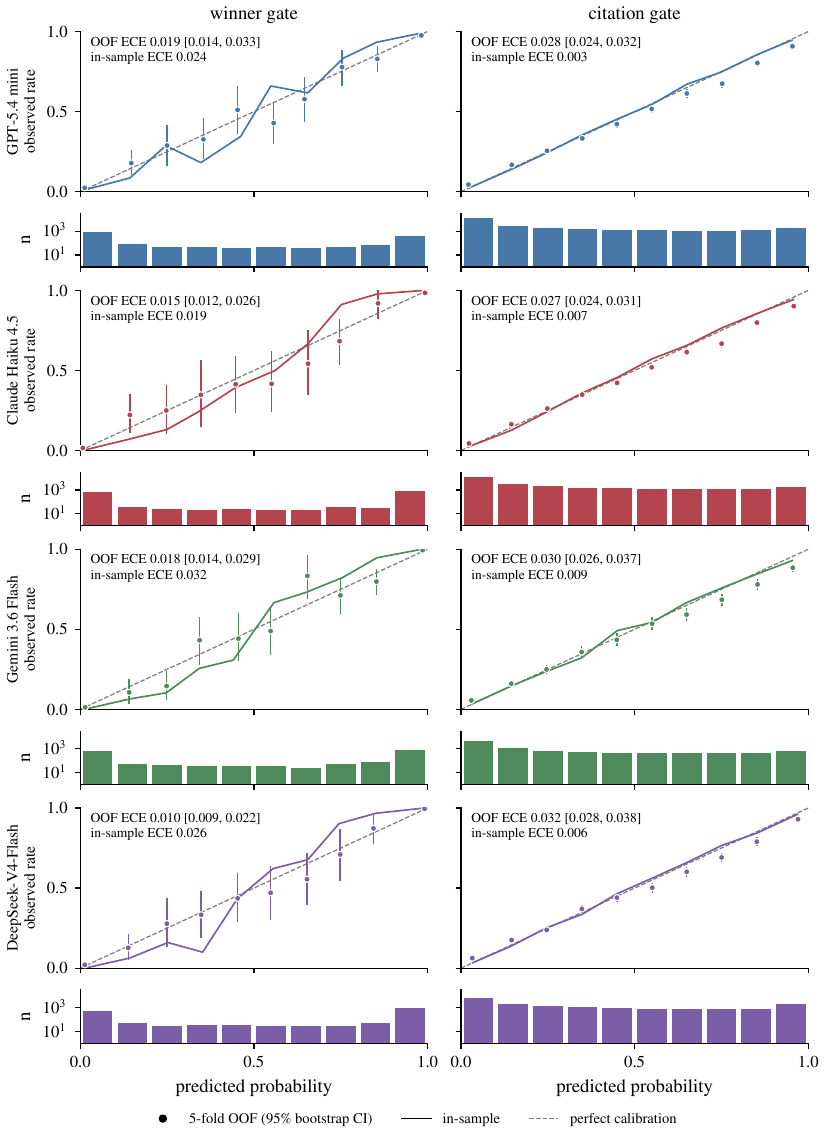}
    \caption{As Figure~\ref{fig:calibration_reliability_ellipse}, on
    ASAP 2.0.}
    \label{fig:calibration_reliability_asap}
\end{figure}

Figures~\ref{fig:calibration_reliability_ellipse}
and~\ref{fig:calibration_reliability_asap} show reliability diagrams for
ELLIPSE and ASAP 2.0: within ten equal-width bins of predicted probability,
the observed event rate is plotted against the predicted one, so a
perfectly calibrated gate lies on the diagonal. The expected calibration
error (ECE) summarizes each panel as the mean absolute gap between the two,
weighted by bin occupancy. The winner gate is calibrated for
every judge: out-of-fold ECE ranges over $0.012$--$0.028$ on ELLIPSE and
$0.010$--$0.019$ on ASAP 2.0, in no case worse than in sample. Its
predictions concentrate near the endpoints of the unit interval, so the
diagnostic is exercised where it is most demanding: the model is confident
on most comparisons and right about how confident it is. The citation gate
is accurate in sample (ECE $0.003$--$0.011$) but slightly degrades out of fold (ECE $0.027$--$0.041$), with the gap concentrated in the upper bins, where observed citation rates fall short of the predictions for all four
judges and on both corpora.

\begin{figure}
    \centering
    \includegraphics[width=\linewidth]{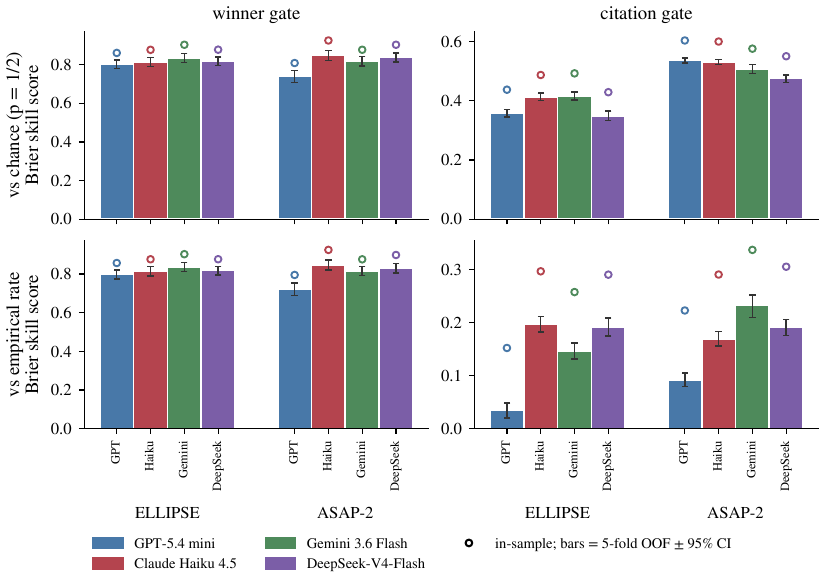}
    \caption{Out-of-fold Brier skill of the two gates, per judge and corpus, with in-sample values overlaid. With $\hat p_i$ the predicted
    probability and $y_i\in\{0,1\}$ the gated event, the Brier score is
    $\mathrm{BS}=N^{-1}\sum_i(\hat p_i-y_i)^2$ and the skill score against
    a reference forecast $q_i$ is $\mathrm{BSS}=1-\mathrm{BS}/\mathrm{BS}_{\mathrm{ref}}$, where $\mathrm{BS}_{\mathrm{ref}}=N^{-1}\sum_i(q_i-y_i)^2$; $\mathrm{BSS}=1$
    is perfect, $0$ matches the reference, and negative values are worse
    than it. Two references: chance ($q_i=\tfrac12$) and the empirical
    event rate of the training fold ($q_i=\bar y_{\mathrm{train}}$).
    Intervals as in Figure~\ref{fig:calibration_reliability_ellipse}.}
    \label{fig:calibration_skill}
\end{figure}

Figure~\ref{fig:calibration_skill} reports discrimination through the Brier
skill score against chance ($p = 1/2$, Brier $0.25$) and against the empirical rate estimated on the training folds (the rate at which the first-shown essay wins for the winner gate, the per-feature
citation rate for the citation gate). The second reference is the demanding
one for the citation channel: it is the maximum-likelihood fit of the
gap-independent submodel in which $\PP(c_{p,k} = 1 \mid w_p) = \sigma(\gamma_k)$, citation as pure per-feature habit, so skill over it is a direct out-of-fold test of whether citations track the winner-oriented gaps. 

The winner gate's skill scores against the two references coincide (slot balance keeps the first-shown rate near 1/2) and are large for every judge (0.80--0.84 on ELLIPSE, 0.72--0.85 on ASAP 2.0). The citation gate scores 0.03--0.23 against the habit-only submodel across judges and corpora, rejecting it throughout: citations do track the gaps, though the contribution is second-order next to the verdict channel and shrinks from its in-sample values in line with the reliability curves.

\clearpage

\paragraph{Prompts.}
The prompts used are reported below.

\begin{promptbox}{Comparative elicitation -- ELLIPSE.}{prm:elicit-ellipse}
=== SYSTEM ===
You are helping build an interpretable scoring framework by examining how student essays differ in English language proficiency. The framework works by collecting many pairwise comparisons: for each pair, a judgment of which essay is stronger overall is recorded together with the qualities that essay has more of. Aggregated across many pairs, these reveal the distinct dimensions along which student writing varies. There is no fixed rubric --- the qualities are recovered from comparisons like this one, not specified in advance.
 
These are essays written by English language learners in grades 8 to 12, during standardized writing assessments. They span a wide range of proficiency, from beginning to quite advanced English writers. Each essay responds to an independent writing prompt that requires no specialized background knowledge; different essays may respond to different prompts. Read and weigh the essays as you would when assessing the English writing proficiency of language learners at this level.
 
Your task. You will be shown two student essays, Essay A and Essay B. They may respond to different writing prompts. This pair is one comparison in a larger collection.
 
First, decide which essay demonstrates stronger English language proficiency overall. This is a forced choice: pick A or B even when the two are close.
 
Then identify the qualities that the essay you chose has more of than the other essay. Report the ones that genuinely stand out to you on reading this pair --- bring up whatever you find salient, and do not work from a predetermined list. Favor real, noticeable differences over minor ones, but do not force the list to be short: report as many as actually distinguish these two essays.
 
For each quality, first name the concrete, specific feature you actually saw in the essays --- a short phrase pointing to the specific evidence. Then, grounded in that evidence, name the quality and describe in one sentence what it captures.
 
Name only qualities on which the essay you chose is genuinely ahead. If the two essays are comparable on some aspect of proficiency, leave it out rather than listing it. Do not pad the list to justify your choice, and do not list a quality on which the other essay is in fact stronger.
 
Out of scope --- do not treat these as differences in proficiency:
- the writing prompt an essay was responding to, or the topic or difficulty of that prompt --- judge proficiency, not the task an essay happened to be assigned;
- the position or opinion an essay takes --- any stance can be expressed proficiently;
- length in itself;
- anonymization placeholders inserted by the corpus builders to remove identifying information, such as Generic_Name, Generic_City, Generic_School, PROPER_NAME, LOCATION_NAME, STUDENT_NAME, TEACHER_NAME or SCHOOL_NAME. Ignore them entirely: they are not the student's words. Never count them as spelling, capitalization, spacing or word-choice errors --- including when a placeholder is misspelled or run together with an adjacent word (for example "Generic_Namehad" or "wasGeneric_Name").
 
How to report. List each quality as exactly one item --- do not combine qualities with "and", "/", "&", or a comma; split related qualities into separate items. For the evidence field, give a short phrase naming the concrete feature you saw. For the quality field, give a concise general noun phrase naming the quality (the kind of short label you might use as a rubric column). For the description field, give one sentence saying what that quality captures.
 
Output only JSON of the form:
{"overall_winner": "A" | "B", "qualities": [{"evidence": "<short phrase>", "quality": "<concise noun phrase>", "description": "<one sentence>"}, ...]}
=== USER ===
Essay A:
(*@\ph{first essay}@*)
 
Essay B:
(*@\ph{second essay}@*)
 
Following the standard above, first give the essay that demonstrates stronger English language proficiency overall, then list the qualities that essay has more of than the other --- for each, first the concrete feature you saw, then the quality and a one-sentence description. Respond with the JSON object only.
\end{promptbox}

\begin{promptbox}{Comparative elicitation -- ASAP 2.0.}{prm:elicit-asap}
=== SYSTEM ===
You are helping build an interpretable scoring framework by examining how student argumentative essays differ in quality. The framework works by collecting many pairwise comparisons: for each pair, a judgment of which essay is stronger overall is recorded together with the qualities that essay has more of. Aggregated across many pairs, these reveal the distinct dimensions along which student writing varies. There is no fixed rubric --- the qualities are recovered from comparisons like this one, not specified in advance.
 
These are argumentative essays written by students in United States schools, primarily in grades 6, 8, 9 and 10. They span a wide range of writing ability and maturity, from beginning writers to quite proficient ones. Every essay responds to the same source-based argumentation task: the students read the article below and were asked to take and defend a position on it, using details from that article as evidence. Read and weigh the essays as you would when assessing argumentative writing at this level.
 
--- SOURCE ARTICLE ---
(*@\ph{source article, inlined verbatim (1{,}120 words, ``Driverless Cars Are Coming'')}@*)
 
--- ASSIGNMENT GIVEN TO STUDENTS ---
In the article "Driverless Cars are Coming," the author presents both positive and negative aspects of driverless cars. Using details from the article, create an argument for or against the development of these cars. Be sure to include: your position on driverless cars; appropriate details from the article that support your position; an introduction, a body, and a conclusion to your argumentative essay.
 
Your task. You will be shown two student essays, Essay A and Essay B, both responding to this assignment. This pair is one comparison in a larger collection.
 
First, decide which essay is the stronger argumentative essay overall. This is a forced choice: pick A or B even when the two are close.
 
Then identify the qualities that the essay you chose has more of than the other essay. Report the ones that genuinely stand out to you on reading this pair --- bring up whatever you find salient, and do not work from a predetermined list. Favor real, noticeable differences over minor ones, but do not force the list to be short: report as many as actually distinguish these two essays.
 
For each quality, first name the concrete, specific feature you actually saw in the essays --- a short phrase pointing to the specific evidence. Then, grounded in that evidence, name the quality and describe in one sentence what it captures.
 
Name only qualities on which the essay you chose is genuinely ahead. If the two essays are comparable on some aspect of quality, leave it out rather than listing it. Do not pad the list to justify your choice, and do not list a quality on which the other essay is in fact stronger.
 
Out of scope --- do not treat these as differences in quality:
- the side an essay takes, for or against the development of driverless cars --- either position can be argued well;
- the topic itself, or which particular details from the article an essay chose to use, except insofar as those choices affect the quality of the argument;
- length in itself;
- anonymization placeholders inserted by the corpus builders to remove identifying information, such as PROPER_NAME, OTHER_PII, STUDENT_NAME, LOCATION_NAME or SCHOOL_NAME. Ignore them entirely: they are not the student's words. Never count them as spelling, capitalization, spacing or word-choice errors --- including when a placeholder is misspelled or run together with an adjacent word.
 
How to report. List each quality as exactly one item --- do not combine qualities with "and", "/", "&", or a comma; split related qualities into separate items. For the evidence field, give a short phrase naming the concrete feature you saw. For the quality field, give a concise general noun phrase naming the quality (the kind of short label you might use as a rubric column). For the description field, give one sentence saying what that quality captures.
 
Output only JSON of the form:
{"overall_winner": "A" | "B", "qualities": [{"evidence": "<short phrase>", "quality": "<concise noun phrase>", "description": "<one sentence>"}, ...]}
=== USER ===
Essay A:
(*@\ph{first essay}@*)
 
Essay B:
(*@\ph{second essay}@*)
 
Following the standard above, first give the essay that is the stronger argumentative essay overall, then list the qualities that essay has more of than the other --- for each, first the concrete feature you saw, then the quality and a one-sentence description. Respond with the JSON object only.
\end{promptbox}

\begin{promptbox}{Cross-judge consolidation prompt.}{prm:cross-judge}
=== SYSTEM ===
You are a careful taxonomist of writing-assessment constructs. Output only valid JSON matching the requested schema.
=== USER ===
You are consolidating evaluation features discovered independently by several LLM judges into a set of shared constructs.
 
Each judge compared (*@\ph{corpus framing}@*) and, with no fixed rubric, named the qualities that distinguished the stronger essay. Each judge's raw phrasings have already been canonicalized separately, so within a single judge the features below are already distinct from one another. Your job is to identify, ACROSS judges, which features denote the same underlying construct.
 
Below is the full list. Each entry has an INDEX in square brackets, the JUDGE that produced it, its NAME, and a one-sentence DEFINITION.
 
(*@\ph{indexed feature list --- one entry per line: [i] (judge) name: definition}@*)
 
Instructions:
- Group entries that denote the SAME underlying evaluative construct, judging by MEANING, not by surface wording. Different words for the same idea belong in one group; the same word used for different ideas belongs in different groups.
- Keep genuinely DISTINCT constructs separate, even when related. Two constructs a writing expert would score on separate rubric lines should be separate groups.
- A group may contain AT MOST ONE feature per judge. The features within a judge were already canonicalized, so two of them never denote the same construct; if you find yourself wanting to place two features from one judge together, they belong in different groups.
- A feature with no counterpart in any other judge forms a group of ONE. Do not absorb it into a loosely related group to tidy up, and do not create a catch-all: a construct only one judge discovered is a real finding and must stand alone.
- Do NOT try to balance the groups. There is no target number of groups and no expectation that every group contains every judge. Groups of one, two, three, and four are all equally valid outcomes; forcing coverage would fabricate agreement that is not there.
- Do NOT merge two constructs just because they often co-occur or because strong essays tend to score well on both. Group only by shared MEANING.
- The judge labels are present only so you can respect the one-per-judge rule. They carry no information about meaning; never group or split on the basis of which judge produced an entry.
 
For each group, give:
- a canonical NAME: a concise noun phrase naming the single construct the group captures. Name the construct itself, not the union of its members' wordings; avoid conjunctions unless the construct genuinely has no single name.
- a one-sentence DEFINITION of that construct.
- the list of members, each as its INDEX, its JUDGE, and its NAME copied verbatim.
 
Rules on membership (these will be checked programmatically):
- Every input index must appear in exactly one group, exactly once. Place all of them; drop none; duplicate none.
- Do not invent indices that are not in the input.
- Each member's "judge" and "name" must be the exact input values at that index --- this is a cross-check, so copy them from the bracketed entry, do not paraphrase.
- No group may contain two members with the same judge.
 
Output only JSON:
{"clusters": [{"name": "<canonical noun phrase>", "definition": "<one sentence>", "members": [{"index": <int>, "judge": "<verbatim>", "name": "<verbatim>"}, ...]}, ...]}
\end{promptbox}

\clearpage

\section{Statistical Model Theory}
\label{app:model-theory}

\subsection{The Bradley--Terry with ties literature}
\label{app:bt-ties}

\paragraph{Classical tie models.}
The Bradley--Terry model \citep{bradleyterry1952} assigns each item a
latent merit and models a binary win/loss outcome through the logistic
link. Classical extensions accommodate a third outcome per comparison:
\citet{glenndavid1960} add a discrimination threshold to the
Thurstone--Mosteller model, \citet{raokupper1967} do the same on the
logistic scale, and \citet{davidson1970} models the tie probability as
proportional to the geometric mean of the two merits, governed by a tie
intensity $\nu$. \citet{davidsonbeaver1977} further incorporate
multiplicative within-pair order effects, the classical ancestor of our
position coefficient $\beta$; see \citet{cattelan2012} for a survey.
Since each of our per-feature observations is trinomial
(``$i$ wins'', ``$j$ wins'', ``not cited''), our model is naturally read
against this literature, and this subsection records the relationship.

\paragraph{Davidson's model is a gated model.}
Write $\pi_i = e^{s_i}$. Davidson's model puts
$\PP(i \text{ wins}) \propto e^{s_i}$,
$\PP(j \text{ wins}) \propto e^{s_j}$, and
$\PP(\text{tie}) \propto \nu\, e^{(s_i+s_j)/2}$. Dividing the three
weights by $e^{(s_i+s_j)/2}$ and writing $\Delta = s_i - s_j$,
\[
\PP(\text{decisive})
= \frac{2\cosh(\Delta/2)}{2\cosh(\Delta/2) + \nu}
= \sigma\Bigl(\log\bigl(2\cosh(\Delta/2)\bigr) - \log\nu\Bigr),
\qquad
\PP(i \mid \text{decisive}) = \sigma(\Delta)
\]
Davidson's model is exactly the single-feature instance of \eqref{eq:model} with gate link $f(u) = \log(2\cosh(u/2))$, gate intercept $\gamma = -\log\nu$, and no position effect. Our link $f = \log(2\cosh)$ is a variant of the same choice, and the general model of
Appendix~\ref{app:model-def} is a strict generalization of Davidson's.
The Rao--Kupper model admits no such factorization: it is the ordered
logit with symmetric cutpoints, $\PP(i \succ j) = \sigma(\Delta - \log\theta)$ and $\PP(i \succ j \text{ or tie}) = \sigma(\Delta + \log\theta)$, so the threshold couples the two channels and the direction given a decisive outcome is not of logistic form in $\Delta$.

\paragraph{Why tie models do not transfer off the shelf.}
Since Davidson's model is a gated model, a battery of $K$ independent
Davidson fits, one per feature, lies inside our model class
($\gamma_k = -\log\nu_k$) and is the natural classical baseline. Three
things separate our model from it. 

First, the regime and the semantics of the null outcome: the tie literature was developed where ties are a minority outcome among near-equal players and assert near-equality, whereas non-citation is our dominant outcome and signals lack of salience as much as similarity. The fitted intercepts of Table~\ref{tab:feats-ours} correspond to tie intensities $\nu_k = e^{-\gamma_k}$ ranging over two orders of magnitude, which is to our knowledge far outside anything that literature contemplates. 

Second, the position effect. The classical route
\citep{davidsonbeaver1977} multiplies the first-presented merit,
$\pi_i \mapsto e^{\beta}\pi_i$, leaving the tie weight
$\nu\sqrt{\pi_i\pi_j}$ unchanged; the model stays log-linear, with
\[
\PP(\text{decisive})
= \sigma\Bigl(\log 2\cosh\bigl(\tfrac{\Delta+\beta}{2}\bigr)
  + \tfrac{\beta}{2} - \log\nu\Bigr),
\qquad
\PP(i \mid \text{decisive}) = \sigma(\Delta+\beta).
\]
The conditional direction is exactly our direction channel, but the
gate argument is no longer even in $\Delta$: presentation order alters
the citation probability, and the null outcome is most likely at
$\Delta = -\beta$ rather than at a genuine tie. Attaching the order
effect symmetrically to both merits changes only the constant
$\beta/2$, absorbed into $\nu$; the non-evenness remains. Our parity structure (even $f$ and $\mY_{ijk}$, odd $\mZ_{ijk}$ defined in Appendix~\ref{app:model}) instead keeps the direction channel $\sigma(\Delta+\beta)$ while confining the position effect to it by construction. 

The trade-off is exact: an order effect on the merits is what keeps the trinomial model log-linear, so an order-invariant gate combined with a position-biased direction channel forces the likelihood \emph{outside} the multinomial-logit family -- which is precisely where concavity is lost (the signed curvature term of \eqref{eq:W}) and why the estimation theory of Appendices~\ref{app:conc}--\ref{app:stage2} cannot be imported from that literature.

Third, the $K$ features are estimated jointly with a shared position coefficient, so the joint likelihood is not a product of per-feature tie models, both channels carry general covariates, and the inferential targets of Section~\ref{sec:method} -- per-feature Laplacian condition numbers, reliability coefficients, and misspecification-robust intervals -- have no counterpart in the paired-comparison ties literature.

\paragraph{A discrete-choice reading, and scope beyond LLM judges.}
The gate/direction factorization can also be read as a two-level
discrete choice with an outside option: the judge first ``chooses''
whether the feature is worth mentioning, then chooses a winner,
analogous to a nested logit in which non-citation is the outside nest
\citep{mcfadden1978, borschsupan1990}. It is by this analogy that we elaborated the model in the first place. We do not claim a formal random-utility foundation -- our gate replaces the inclusive value by the even link $f$ -- but the analogy delimits the scope of the framework: it applies to any rater who, per comparison, reports verdicts only on the attributes salient enough to mention.

\paragraph{Estimation theory.}
Modern non-asymptotic theory for the binary Bradley--Terry model
studies the growing-dimension regime: \citet{simonsyao1999} establish
consistency and asymptotic normality as $n \to \infty$ in the dense
case, \citet{negahban2017rank} and \citet{shah2016estimation} analyze
spectral and minimax aspects, \citet{chen2019spectral} obtain
$\ell_\infty$ guarantees through the leave-one-out technique we adapt,
and \citet{yan2025inferencegeneralizedbradleyterrymodel} provides the uniform-consistency template our Stage-1 analysis builds on. For tie models, we are not aware of comparable growing-dimension $\ell_\infty$ guarantees. The technical landscape also differs: Davidson's model is a log-linear multinomial logit and Rao--Kupper an ordered logit, so both have concave log-likelihoods in the natural parametrization, whereas the free even link and the gate covariates make our likelihood non-concave -- the signed curvature term of \eqref{eq:W} -- which is the difficulty the
machinery of Appendices~\ref{app:conc}--\ref{app:stage2} is built to
handle.

\subsection{Why a shared position bias coefficient?}\label{app:beta-orthogonality}

The direction channel of the model includes a position coefficient
$\beta$: for a pair $p$ on feature $k$, given citation, the probability that the first-presented item wins is $d_{p,k}=\sigma(\Delta_{p,k}+\beta)$, where pairs are written in presentation order, so that $\Delta_{p,k}=s_{i,k}-s_{j,k}$ with $i$ the item shown first. The coefficient $\beta$ is shared across features. A natural alternative is a per-feature coefficient $\beta_k$; we first explain why this choice has almost no effect on the estimated scores, then verify it empirically.

\paragraph{Setup: how $\beta$ enters the score equations.}
Fix a feature $k$ (we drop it from the notation). The direction
log-likelihood is
\[
\ell(\vs,\beta)\;=\;\sum_{p\,:\,c_p=1}
\Big[\,w_p\log\sigma(\Delta_p+\beta)
+(1-w_p)\log\big(1-\sigma(\Delta_p+\beta)\big)\Big],
\]
where $w_p=1$ if the first-presented item of pair $p$ won. Consider an
item $i$, and let $P_i$ be the pairs containing $i$. In each
$p\in P_i$, the gap $\Delta_p$ contains $s_i$ with sign
\[
\varepsilon_p(i)\;=\;
\begin{cases}
+1 & \text{if $i$ was presented first in $p$},\\
-1 & \text{if $i$ was presented second},
\end{cases}
\]
The score equation for $s_i$ is therefore
\begin{equation}\label{eq:score-eq}
\frac{\partial\ell}{\partial s_i}
=\sum_{p\in P_i}\varepsilon_p(i)\big(w_p-d_p\big)\;=\;0 ,
\qquad d_p=\sigma(\Delta_p+\beta).
\end{equation}
This is the standard Bradley--Terry balance condition (observed wins
minus expected wins, counted from $i$'s side of each pair) except
that every expected win $d_p$ is tilted by the common offset $\beta$.

\paragraph{First-order effect of changing $\beta$.}
Suppose $\beta$ moves by an amount $\delta$ (for instance, from the
shared estimate $\hat\beta$ to a per-feature estimate $\hat\beta_k$,
so $\delta=\hat\beta_k-\hat\beta$). Each expected win moves by
$\mathrm{d}d_p = w_{\theta,p}\delta$ with
$w_{\theta,p}=d_p(1-d_p)$, so the score equation
\eqref{eq:score-eq} is thrown off balance by
\[
-\,\delta\sum_{p\in P_i}\varepsilon_p(i)\,w_{\theta,p}.
\]
To restore balance, $s_i$ must adjust. Differentiating
\eqref{eq:score-eq} in $s_i$ (and neglecting the adjustment of the
other items' scores, which enters only through the off-diagonal
Laplacian terms), the required adjustment is, to first order,
\begin{equation}\label{eq:shift}
\Delta\hat s_i\approx
-\delta
\frac{\displaystyle\sum_{p\in P_i}\varepsilon_p(i)\,w_{\theta,p}}
     {\displaystyle\sum_{p\in P_i} w_{\theta,p}} .
\end{equation}
We write $\approx$ because modifying $\beta$ also modifies the other $s_j$, which in turn modify $s_i$: we neglect these second-order perturbations.
The denominator is a sum of positive weights; the numerator is the
\emph{same} weights, but signed by presentation order. The score shift
is thus $\delta$ times a weighted average of $\pm1$'s: $+1$ for each
pair where $i$ was shown first, $-1$ where it was shown second.

\paragraph{Cancellation under the balanced design.}
Equation~\eqref{eq:shift} exposes exactly what the design controls. If
the weights were equal, the numerator would be
$n_{\mathrm{first}}(i)-n_{\mathrm{second}}(i)$: the imbalance in how
often item $i$ appeared in each presentation slot. Our comparison graph
assigns presentation order uniformly at random, independently of the
items; with $m$ comparisons per item, each item is first
$\mathrm{Binomial}(m,\tfrac12)$ times, so the slot fraction deviates
from $\tfrac12$ by $O_p(m^{-1/2})$ (for $m=40$: standard deviation
$0.079$, matching the realized imbalance). Hence
$\Delta\hat s_i = O_p(\delta/\sqrt m)$: the effect of any change in
$\beta$ on any score is reduced by the square root of the number of
comparisons, \emph{because presentation order is randomized}. 

Two remarks. First, the cancellation is not exact: under randomization the two presentation orders of a pair carry weights $\sigma'(\Delta+\beta)$ and $\sigma'(\Delta-\beta)$, whose signed average is $\tfrac12(\sigma'(\Delta+\beta)-\sigma'(\Delta-\beta))
=\beta\,\sigma''(\Delta)+O(\beta^3)$; the
numerator of \eqref{eq:shift} thus retains a deterministic residual of
order $O(\beta)$ per pair, on top of the $O_p(m^{-1/2})$ stochastic
imbalance. Both are negligible here ($\hat\beta$ and $\lvert\sigma''\rvert$ are small), but the orthogonality is first-order, not exact. Second, the citation channel contains no position term at all (the gate is $\sigma(f(\Delta)+\gamma_k)$), so the information that citations carry about score magnitudes is untouched by $\beta$ under either specification.

\paragraph{Empirical check.}
When we refit the data of Section~\ref{sec:feature_extraction_empirical_results} with per-feature $\beta_k$, the positional bias estimates differ significantly across features, yet the scores are \emph{essentially unchanged}: the median absolute shift is $1\%$ of the per-feature score spread (90th percentile: $5\%$). The residual movement is attributable to the imperfect slot balance: across features, the score shift correlates with $\lvert\hat\beta_k\rvert$ at $r=0.87$, as the cancellation argument predicts. 

We therefore retain the shared $\beta$ in the main model; the per-feature $\hat\beta_k$ remain interpretable as an audit of the judge (features differ in their susceptibility to position bias) but the design prevents this heterogeneity from reaching the scores.

\subsection{Uncertainty quantification}
\label{app:uq}

\subsubsection{Confidence intervals}

The estimator $\hat{\vtheta}$ is an M-estimator, which lets us adopt the weaker, standard view from robust statistics \citep{huber1967behavior,white1982maximum} to account for potential model misspecifications in our estimates. $\hat{\vtheta}$ estimates the \emph{pseudo-true} parameter $\vtheta^{\star}$ (which is the Kullback--Leibler projection of the judge's actual comparison distribution onto our model class). In the regime where the item set is held fixed and the number $N$ of edges of the comparison graph grows, the usual M-estimation argument yields
\begin{equation}
  \label{eq:sandwich-clt}
  \sqrt{N}\bigl(\hat{\vtheta} - \vtheta^{\star}\bigr)
  \xrightarrow[N\to\infty]{d}
  \mathcal{N}\bigl(0, \mH^{-1} \mG \mH^{-1}\bigr),
  \qquad
  \begin{aligned}
    \mH &= \mathbb{E}\bigl[\nabla^2_{\vtheta}\mathcal{L}(\vtheta^{\star})\bigr],\\
    \mG &= \mathbb{E}\bigl[\nabla_{\vtheta}\mathcal{L}(\vtheta^{\star})\,
           \nabla_{\vtheta}\mathcal{L}(\vtheta^{\star})^{\top}\bigr],
  \end{aligned}
\end{equation}
estimated by their empirical counterparts $\widehat{\mH}$ (average Hessian)
and $\widehat{\mG}$ (average outer product of per-comparison gradients) at
$\hat{\vtheta}$. The resulting \emph{sandwich} covariance $\widehat{\mSigma}^{\text{sand}} = \tfrac{1}{N}\widehat{\mH}^{-1}\widehat{\mG}\widehat{\mH}^{-1}$ coincides with the inverse Fisher information when the model is correct (by the information-matrix equality $\mG = \mH$), and remains an asymptotically valid covariance for the projection $\vtheta^{\star}$ when it is not. Hence, the confidence intervals on scores are statements about this projection, namely the score configuration that best summarizes the judge's comparisons within our model class.

\subsection{Reliability diagnostics} 

Since citation is sporadic, there may be features that are not cited enough or whose winners are inconsistent so that the scores are not precisely pinned down by the data. One advantage of our method is that it rests on a statistical model: we can hence inspect the \emph{condition numbers} of blocks of the Hessian to diagnose poorly identified features \citep{rothenberg1971}. Indeed, for each feature $k$, the unpenalized score block of the Fisher information is a
weighted Laplacian $\mL_k$. By computing:
\begin{equation}
    \label{eq:condition_number}
    \kappa_k = \frac{\lammax(\mL_k)}{\lambda_{2}(\mL_k)}
\end{equation}
and keeping features that have $\kappa \leq \kappa_{\text{max}}$ (where $\kappa_{\text{max}}$ is specified a priori) we can filter out poorly identified features. Features that have $\kappa > \kappa_{\text{max}}$ are called \emph{unidentified}.

Moreover, an interesting quantity to compute is the spread of the scores, as it tells us how markedly items differ along each feature in the dataset. However, a high spread could also be due to noisy estimates. We can hence denoise the estimated spreads of the scores by subtracting the analytic estimation variance:
\begin{equation}
    \label{eq:noise}
    \nu_k:=\frac{1}{n}\operatorname{Tr}\left(\widehat{\mSigma}_k^{\text{sand}}\right)
\end{equation}
 where $\widehat{\mSigma}^{\text{sand}}_k$ is the $k$-th score block of the sandwich covariance. Thus $\rho_k := \frac{\Var(\hat{\vs}_k) - \nu_k}{\Var(\hat{\vs}_k)}$ is the fraction of observed score spread attributable to genuine between-item signal. Keeping features with $\rho_k > \rho_{\text{min}}$ (where $\rho_{\text{min}}$ is specified a priori) filters out \emph{unreliable} features.
 
Having fitted the model a first time, spotted unidentified and unreliable features, and possibly chosen features that we want to drop or fuse, we can then refit the model and obtain scores on reliable (and chosen) features.

\subsection{Model, likelihood, score, and Hessian}\label{app:model}

\subsubsection{Model}\label{app:model-def}

We study the model where there is only one feature (case $K = 1$). There are $n+1$ items indexed by $\{0,1,\dots,n\}$. For each unordered pair
$\{i,j\}$ with $0\le i<j\le n$ we observe $m_{ij}$ comparisons, indexed
$k=1,\dots,m_{ij}$. Each comparison yields a trinomial outcome
\[
r_{ijk}\in\{0,+1,-1\}:\qquad
0=\text{``pair not cited''},\quad +1=\text{``}i\text{ wins''},\quad
-1=\text{``}j\text{ wins''.}
\]
We encode it by a \emph{citation} indicator and a \emph{direction} indicator,
\[
c_{ijk}:=\1_{\{r_{ijk}\neq0\}}\in\{0,1\},\qquad
a^{1}_{ijk}:=\1_{\{r_{ijk}=+1\}},\quad a^{-1}_{ijk}:=\1_{\{r_{ijk}=-1\}},
\]
so that $a^{1}_{ijk}+a^{-1}_{ijk}=c_{ijk}$.

Each item $i$ carries a latent merit $s_i\in\R$; we write
$\vs=(s_0,\dots,s_n)\trans\in\R^{n+1}$ for the merit vector (suppressing the
dependence on $n$ when clear) and $\Delta_{ij}:=s_i-s_j$ for the gap. There are
two fixed-dimension structural parameters: the gate coefficient
$\vgamma\in\R^{p_{\gamma}}$ and the direction (position-bias) coefficient
$\vbeta\in\R^{p_{\beta}}$. Each comparison carries covariates
$\mY_{ijk}\in\R^{p_{\gamma}}$ (gate) and $\mZ_{ijk}\in\R^{p_{\beta}}$
(direction), with parities
\begin{equation}\label{eq:parity}
\mY_{jik}=\mY_{ijk}\ \text{(even)},\qquad \mZ_{jik}=-\mZ_{ijk}\ \text{(odd)},
\qquad \Delta_{ji}=-\Delta_{ij}.
\end{equation}
Let $\sigma(x)=e^{x}/(1+e^{x})$ and let $f:\R\to\R$ be a fixed \emph{even}
link. Define the two linear predictors and their sigmoids,
\begin{equation}\label{eq:predictors}
\eta_{ijk}:=f(\Delta_{ij})+\mY_{ijk}\trans\vgamma,\quad
g_{ijk}:=\sigma(\eta_{ijk});
\qquad
\theta_{ijk}:=\Delta_{ij}+\mZ_{ijk}\trans\vbeta,\quad
d_{ijk}:=\sigma(\theta_{ijk}).
\end{equation}
The model is, conditionally independently across all $(i,j,k)$,
\begin{equation}\label{eq:model}
\PP(r_{ijk}=0)=1-g_{ijk},\quad
\PP(r_{ijk}=+1)=g_{ijk}\,d_{ijk},\quad
\PP(r_{ijk}=-1)=g_{ijk}\,(1-d_{ijk}).
\end{equation}
Equivalently $c_{ijk}\sim\mathrm{Bernoulli}(g_{ijk})$, and conditionally on
$c_{ijk}=1$, $a^{1}_{ijk}\sim\mathrm{Bernoulli}(d_{ijk})$. The parities
\eqref{eq:parity} make \eqref{eq:model} invariant under $i\leftrightarrow j$
(the gate is invariant; the direction flips), since $f$ is even while $\Delta$
and $\mZ$ are odd.

\paragraph{Identification (the $\vone^\perp$ gauge).} Adding a constant to
every $s_i$ leaves \eqref{eq:model} unchanged (only gaps enter), so we impose
the gauge $\vs\in\vone^\perp:=\{\vx\in\R^{n+1}:\vone\trans\vx=0\}$ throughout.
This choice is natural here: every merit-direction score and Hessian block
below is a linear combination of the centered vectors
$\vomega_{ij}:=\ve_i-\ve_j\in\vone^\perp$, so the gauge is preserved
automatically by the estimating equations and by the gradient iterations of
Appendix~\ref{app:consistency}, with no explicit projection.

\subsubsection{Log-likelihood}
From \eqref{eq:model}, the per-comparison log-likelihood is
\[
\ell_{ijk}=c_{ijk}\log g_{ijk}+(1-c_{ijk})\log(1-g_{ijk})
+a^{1}_{ijk}\log d_{ijk}+a^{-1}_{ijk}\log(1-d_{ijk}),
\]
and the total is
\begin{equation}\label{eq:log-likelihood}
\ell(\vs,\vgamma,\vbeta)=\sum_{0\le i<j\le n}\sum_{k=1}^{m_{ij}}\ell_{ijk}.
\end{equation}

\subsubsection{Scores}
Using $\sigma'=\sigma(1-\sigma)$ and the chain rule through
\eqref{eq:predictors}
($\frac{\partial\eta_{ijk}}{\partial\Delta_{ij}}=f'(\Delta_{ij})$,
$\frac{\partial\theta_{ijk}}{\partial\Delta_{ij}}=1$), the per-comparison
scores are
\[
\frac{\partial\ell_{ijk}}{\partial\eta_{ijk}}=c_{ijk}-g_{ijk},\qquad
\frac{\partial\ell_{ijk}}{\partial\theta_{ijk}}=a^{1}_{ijk}-c_{ijk}d_{ijk}.
\]
The three score blocks are
\begin{align}
\nabla_{\vs}\ell
&=\sum_{i<j}\sum_{k}\Big[(c_{ijk}-g_{ijk})\,f'(\Delta_{ij})
   +\big(a^{1}_{ijk}-c_{ijk}d_{ijk}\big)\Big]\,\vomega_{ij}
   \ \in\R^{n+1},\label{eq:score-s}\\[2pt]
\nabla_{\vgamma}\ell&=\sum_{i<j}\sum_{k}(c_{ijk}-g_{ijk})\,\mY_{ijk}
   \ \in\R^{p_{\gamma}},\label{eq:score-g}\\[2pt]
\nabla_{\vbeta}\ell&=\sum_{i<j}\sum_{k}\big(a^{1}_{ijk}-c_{ijk}d_{ijk}\big)\,
   \mZ_{ijk}\ \in\R^{p_{\beta}}.\label{eq:score-b}
\end{align}
Define the per-coordinate merit score
\begin{equation}\label{eq:score}
G_a(\vs,\vgamma,\vbeta):=-\frac{\partial\ell}{\partial s_a}
=-\sum_{j\neq a}\sum_{k}\psi_{ajk},\qquad
\psi_{ajk}:=(c_{ajk}-g_{ajk})f'(\Delta_{aj})+\big(a^{1}_{ajk}-c_{ajk}d_{ajk}\big).
\end{equation}
Collecting the coordinates, the \emph{merit score map} is
\begin{equation}\label{eq:score-map}
\mG(\vs,\vgamma,\vbeta):=\nabla_{\vs}(-\ell)=(G_0,\dots,G_n)\trans
=-\sum_{i<j}\sum_{k}\Big[(c_{ijk}-g_{ijk})f'(\Delta_{ij})
+\big(a^1_{ijk}-c_{ijk}d_{ijk}\big)\Big]\vomega_{ij},
\end{equation}
which takes values in $\vone^\perp$ because every
$\vomega_{ij}=\ve_i-\ve_j$ is centered. When $(\vgamma,\vbeta)$ are held fixed
we write $\mG_{\vgamma,\vbeta}(\vs):=\mG(\vs,\vgamma,\vbeta)$, viewed as a map
$\R^{n+1}\to\vone^\perp\subset\R^{n+1}$. The MLE solves
$\mG(\vs,\vgamma,\vbeta)=\vzero$ on $\vone^\perp$ together with
\eqref{eq:score-g} and \eqref{eq:score-b}.

\subsubsection{Hessian}
Set $w_{\eta,ijk}:=g_{ijk}(1-g_{ijk})$ and $w_{\theta,ijk}:=d_{ijk}(1-d_{ijk})$.
Differentiating once more ($\theta$ is linear so $\partial^2\theta=0$, while
$\frac{\partial^2\eta}{\partial\Delta^2}=f''$) gives the per-comparison merit
curvature
\begin{equation}\label{eq:W}
W_{ijk}:=-\frac{\partial^2\ell_{ijk}}{\partial\Delta^2}
=\underbrace{w_{\eta,ijk}\,f'(\Delta_{ij})^2+c_{ijk}\,w_{\theta,ijk}}_{\text{PSD part}}
\;-\;\underbrace{(c_{ijk}-g_{ijk})\,f''(\Delta_{ij})}_{\text{``saboteur''}},
\end{equation}
and the full Hessian of $-\ell$ has the block structure (all sums
$\sum_{i<j}\sum_k$)
\begin{align}
\mJ(\vs,\vgamma,\vbeta):=\nabla^2_{\vs}(-\ell)
   &=\textstyle\sum W_{ijk}\,\vomega_{ij}\vomega_{ij}\trans
   &&(\text{full merit block, }(n{+}1)\times(n{+}1)),\label{eq:Jss}\\
\mA_{\vgamma\vgamma}:=\nabla^2_{\vgamma\vgamma}(-\ell)
   &=\textstyle\sum w_{\eta,ijk}\,\mY_{ijk}\mY_{ijk}\trans,\label{eq:Hgg}\\
\mA_{\vbeta\vbeta}:=\nabla^2_{\vbeta\vbeta}(-\ell)
   &=\textstyle\sum c_{ijk}\,w_{\theta,ijk}\,\mZ_{ijk}\mZ_{ijk}\trans,\label{eq:Hbb}\\
\mB_{\vgamma}:=\nabla^2_{\vs\vgamma}(-\ell)
   &=\textstyle\sum w_{\eta,ijk}\,f'(\Delta_{ij})\,\vomega_{ij}\mY_{ijk}\trans,\label{eq:Hsg}\\
\mB_{\vbeta}:=\nabla^2_{\vs\vbeta}(-\ell)
   &=\textstyle\sum c_{ijk}\,w_{\theta,ijk}\,\vomega_{ij}\mZ_{ijk}\trans,\label{eq:Hsb}\\
\nabla^2_{\vgamma\vbeta}(-\ell)&=\vzero.\label{eq:Hgb}
\end{align}
Blocks \eqref{eq:Hgg}--\eqref{eq:Hbb} are positive semidefinite; the cross
block \eqref{eq:Hgb} vanishes because, given the predictors, the two channels
share no randomness. The merit block \eqref{eq:Jss} is \emph{not} PSD
pathwise: the saboteur in \eqref{eq:W} is signed, so a single $W_{ijk}$ can be
negative, e.g.\ on a cited near-tie, where $f'\approx0$ and $c=1$,
$W\approx w_\theta-(1-g)f''$. This non-convexity is the central difficulty.
The two off-diagonal blocks $\mB_{\vgamma},\mB_{\vbeta}$ are named here for
later use (Lemma~\ref{lem:cross-jac} and Corollary~\ref{cor:mle}).

\subsection{Notation and external tools}\label{app:tools}

\paragraph{Notation.} Following the template conventions, bold lowercase
letters denote vectors ($\vs,\vu,\vomega_{ij}$) and bold uppercase letters
denote matrices ($\mJ,\mA,\mB,\mV,\mSigma$); in keeping with the
Bradley--Terry literature we also write in bold uppercase the per-comparison
covariate vectors $\mY_{ijk}\in\R^{p_\gamma}$, $\mZ_{ijk}\in\R^{p_\beta}$ and
the vector-valued score maps $\mG,\mD,\mT$. Calligraphic letters denote balls
($\gB$), events ($\gE$), $\sigma$-algebras ($\gF$) and nets ($\gN$). Starred
quantities ($\vs^*,\vgamma^*,\vbeta^*,g^*_{ijk},\dots$) are evaluated at the
truth. Universal constants may change from line to line.

\paragraph{Norms.} For $\vx\in\R^d$, $\norm{\vx}_\infty=\max_a\abs{x_a}$ and
$\norm{\vx}_2=(\sum_a x_a^2)^{1/2}$. For a matrix $\mA=(A_{ab})$ we write
$\norm{\mA}_{\max}=\max_{a,b}\abs{A_{ab}}$ (entrywise) and use triple bars for
the induced operator norms: $\opnorm{\mA}_2$ is the spectral norm (induced by
$\norm{\cdot}_2$) and $\opnorm{\mA}_\infty=\max_a\sum_b\abs{A_{ab}}$ is the
maximum absolute row sum (induced by $\norm{\cdot}_\infty$). For symmetric
$\mA$, $\opnorm{\mA}_2\le\opnorm{\mA}_\infty$. For symmetric $\mA$ acting on
the centered subspace, $\lamperp(\mA)$ is the smallest eigenvalue restricted to
$\vone^\perp=\{\vx:\vone\trans\vx=0\}$.

\paragraph{Restricted spectral norm.} For a symmetric matrix $\mM$ that maps
$\vone^\perp$ into itself, write
\[
\opnorm{\mM}_{2,\perp}:=\sup_{\vx\in\vone^\perp,\ \norm{\vx}_2=1}\norm{\mM\vx}_2
\]
for its spectral norm restricted to the centered subspace. In particular, if
$\mM=\mI-\eta\mN$ with $\mN$ symmetric and PSD on $\vone^\perp$ and
$0\le\eta\lambda\le1$ for every eigenvalue $\lambda$ of $\mN|_{\vone^\perp}$,
then
\begin{equation}\label{eq:contraction-norm}
\opnorm{\mI-\eta\mN}_{2,\perp}=1-\eta\,\lamperp(\mN).
\end{equation}

\paragraph{Balls.} For $\vx\in\R^{d}$ and $r>0$, $\gB_2(\vx,r)$ and
$\gB_\infty(\vx,r)$ denote the \emph{open} $\ell_2$ and $\ell_\infty$ balls of
radius $r$ centered at $\vx$, and $\overline{\gB_2},\overline{\gB_\infty}$
their closures. For the fixed-dimensional structural parameters we write
$\gB(\vx,r)$ for the open $\ell_2$ ball.

\paragraph{Sigmoid bounds.} For all $x$:
$0<\sigma'(x)=\sigma(1-\sigma)\le\frac14$, $\abs{\sigma''}\le\frac14$, and
$\frac{d}{dx}[\sigma(1-\sigma)]=\sigma(1-\sigma)(1-2\sigma)$ has absolute
value $\le\frac14$.

\medskip
We import the following standard tools.

\begin{fact}[Vector Bernstein]\label{fact:vector-bernstein}
For independent mean-zero $\vx_{\ell}\in\R^d$ with $\norm{\vx_\ell}_2\le R$
and $\sum_\ell\E\norm{\vx_\ell}_2^2\le\nu$,
\[
\PP\left(\norm{\sum_\ell\vx_\ell}_2\ge t\right)
\le 2(d+1)\exp\!\Big(-\frac{t^2}{2\nu+2Rt/3}\Big).
\]
\end{fact}

\begin{fact}[Matrix Bernstein]\label{fact:matrix-bernstein}
For independent symmetric mean-zero $\mX_\ell\in\R^{d\times d}$ with
$\opnorm{\mX_\ell}_2\le R$ and $\opnorm{\sum_\ell\E\mX_\ell^2}_2\le\nu$,
\[
\PP\!\left(\opnorm{\sum_\ell\mX_\ell}_2\ge t\right)
\le 2d\,\exp\!\left(-\frac{t^2}{2\nu+2Rt/3}\right).
\]
\end{fact}

\begin{fact}[Hoeffding]\label{fact:hoeffding}
For independent $x_\ell\in[\alpha_\ell,\beta_\ell]$,
\[
\PP\!\left(\Big\lvert\sum_\ell(x_\ell-\E x_\ell)\Big\rvert\ge t\right)
\le 2\exp\!\left(-\frac{2t^2}{\sum_\ell(\beta_\ell-\alpha_\ell)^2}\right).
\]
\end{fact}

\begin{fact}[Degree-theoretic existence; Ortega--Rheinboldt 1970]
\label{fact:ortega-rheinboldt}
If $\Omega\subset\R^p$ is open and bounded, $\mF:\overline\Omega\to\R^p$ is
continuous, and there is $\vx_0\in\Omega$ with
$(\vx-\vx_0)\trans\mF(\vx)\ge0$ for all $\vx\in\partial\Omega$, then $\mF$ has
a zero in $\overline\Omega$.
\end{fact}

\begin{fact}[Gershgorin disc theorem]\label{fact:gershgorin}
If $\mA=(A_{ab})_{a,b=1}^{p}$ is a $p\times p$ matrix, then
\[
\operatorname{spec}(\mA)\ \subseteq\
\bigcup_{a=1}^{p}\Big\{z\in\mathbb{C}:\
\abs{z-A_{aa}}\le\textstyle\sum_{b\ne a}\abs{A_{ab}}\Big\}.
\]
In particular,
$\lammax(\mA)\le\max_{a}\big(A_{aa}+\sum_{b\ne a}\abs{A_{ab}}\big)$ when
$\mA$ is symmetric.
\end{fact}

\begin{fact}[Schur complement criterion for positive definiteness;
Haynsworth 1968]\label{fact:schur}
Let $\mM$ be a symmetric matrix in block form
\[
\mM=\begin{pmatrix}\mP&\mQ\\[2pt] \mQ\trans&\mR\end{pmatrix},
\]
with $\mP$ symmetric and invertible, and let
$\mM/\mP:=\mR-\mQ\trans\mP^{-1}\mQ$ denote the Schur complement of $\mP$ in
$\mM$. Then the inertias add,
\[
\operatorname{In}(\mM)=\operatorname{In}(\mP)+\operatorname{In}(\mM/\mP),
\]
where $\operatorname{In}(\cdot)=(n_+,n_-,n_0)$ counts the positive, negative
and zero eigenvalues. In particular, if $\mP\succ0$ then
\[
\mM\succ0\quad\Longleftrightarrow\quad \mM/\mP\succ0.
\]
\end{fact}

\subsection{Assumptions and the population curvature floor}\label{app:assump}

\begin{assumption}[Dense bounded design]\label{assump:dense}
$1\le m_{ij}\le m^\ast$ for all $0\le i<j\le n$, with $m^\ast$ a fixed
constant.
\end{assumption}

\begin{assumption}[Compactness]\label{assump:compact}
There exist absolute constants $C_s,\kappa>0$ such that, for all $n$,
\[
\norm{\vs^{*}}_\infty \le C_s,\qquad
\sup_{0\le i,j\le n,\ k}\norm{\mY_{ijk}}_2 \le \kappa,\qquad
\sup_{0\le i,j\le n,\ k}\norm{\mZ_{ijk}}_2 \le \kappa.
\]
\end{assumption}

\begin{assumption}[Smooth even gate]\label{assump:gate}
$f$ is even and $C^3$, so that
\[
F_1:=\sup_{\abs{u}\le 2C_s+1}\abs{f'(u)},\qquad
F_2:=\sup_{\abs{u}\le 2C_s+1}\abs{f''(u)},\qquad
F_3:=\sup_{\abs{u}\le 2C_s+1}\abs{f'''(u)}
\]
are finite.
\end{assumption}

\begin{remark}[Working region]\label{rem:nbhd}
The supremum range $\abs{u}\le 2C_s+1$ in A\ref{assump:gate} is a fixed
interval strictly containing the range $[-2C_s,2C_s]$ of the population gaps.
Throughout, the \emph{working region} is the set
$\{\vs:\norm{\vs}_\infty\le C_s+\frac12\}$: every merit vector $\vs$ that
arises below lies within $o(1)$ of $\vs^*$, hence lies in the working region
for $n$ large; its gaps then obey $\abs{\Delta_{ij}}\le 2C_s+1$, so the bounds
$F_1,F_2,F_3$ apply to every gap encountered. The Hessian-geometry lemmas of
Appendix~\ref{app:lip} are stated for $\norm{\vs}_\infty\le C_s+\frac12$
accordingly.
\end{remark}

Under A\ref{assump:dense}--A\ref{assump:gate} the predictors
$\eta^*_{ijk},\theta^*_{ijk}$ lie in a fixed compact interval, so there are
absolute constants $0<\underline g\le\overline g<1$ with
$\underline g\le g^*_{ijk}\le\overline g$, and likewise for $d^*_{ijk}$; hence
$w^*_{\eta,ijk},w^*_{\theta,ijk}\in[\underline w,\frac14]$ for an absolute
$\underline w>0$. Define the \emph{population merit weight}
\begin{equation}\label{eq:wbar}
\bar w_{ijk}:=w^*_{\eta,ijk}\,f'(\Delta^*_{ij})^2
+g^*_{ijk}\,w^*_{\theta,ijk}\ \ (\ge0).
\end{equation}

\begin{lemma}[Curvature floor]\label{lem:floor}
Under A\ref{assump:compact}--A\ref{assump:gate} there are absolute constants
$0<b_0\le b_1$ with $b_0\le\bar w_{ijk}\le b_1$ for all $i,j,k$. One may take
$b_0=\underline g\,\underline w$ and $b_1=\frac14(F_1^2+1)$.
\end{lemma}

\begin{proof}
The upper bound is immediate: $w^*_\eta\le\frac14$,
$f'(\Delta^*)^2\le F_1^2$, $g^*\le1$, $w^*_\theta\le\frac14$, so
$\bar w_{ijk}\le\frac14 F_1^2+\frac14$. For the lower bound, both terms in
\eqref{eq:wbar} are nonnegative, so it suffices to bound the second:
\[
\bar w_{ijk}\ \ge\ g^*_{ijk}\,w^*_{\theta,ijk}\ \ge\
\underline g\,\underline w\ =:\ b_0\ >\ 0.
\]
The point is that at a tie ($\Delta^*_{ij}=0$) the gate channel vanishes,
since $f$ even implies $f'(0)=0$; the floor is then carried entirely by the
direction channel $g^*w^*_\theta$, which is positive precisely because tied
pairs are still cited with probability $\ge\underline g$ under compactness.
\end{proof}

\begin{remark}
If $\norm{\vs^*}_\infty$ is allowed to grow, $\underline g\to0$ and $b_0\to0$,
which is the regime where the rates stated below degrade.
\end{remark}

\begin{assumption}[Gate design]\label{assump:gatedesign}
There is a constant $c_0>0$ such that
\[
\lammin\!\Big(\sum_{i<j}\sum_k
w^*_{\eta,ijk}\,\mY_{ijk}\mY_{ijk}\trans\Big)\ge c_0 n^2.
\]
\end{assumption}

\begin{assumption}[Direction design]\label{assump:dirdesign}
There is a constant $c_0>0$ such that
\[
\lammin\!\Big(\sum_{i<j}\sum_k
g^*_{ijk}w^*_{\theta,ijk}\,\mZ_{ijk}\mZ_{ijk}\trans\Big)\ge c_0 n^2.
\]
\end{assumption}

\begin{remark}
A\ref{assump:gatedesign} and A\ref{assump:dirdesign} posit constants that may
differ; without loss of generality we use a single $c_0$ for both, namely the
smaller of the two. A further assumption, on the efficient information for the
structural block, is stated as A\ref{assump:info} in Appendix~\ref{app:stage2},
where the relevant Schur complement is introduced; it implies
A\ref{assump:gatedesign}--A\ref{assump:dirdesign}
(Remark~\ref{rem:A6-implies}), which we nevertheless retain for their
interpretability.
\end{remark}

\subsubsection{The population Hessian floor}\label{app:Vfloor}

At the truth, decompose the realized merit Hessian
$\mJ^*:=\mJ(\vs^*,\vgamma^*,\vbeta^*)$ as
\begin{equation}\label{eq:VE}
\mJ^*=\mV+\mE,\qquad
\mV:=\E[\mJ^*]=\sum_{i<j}\sum_k\bar w_{ijk}\,\vomega_{ij}\vomega_{ij}\trans,
\qquad
\mE:=\sum_{i<j}\sum_k\xi_{ijk}\,\vomega_{ij}\vomega_{ij}\trans,
\end{equation}
where, using $\E[c_{ijk}]=g^*_{ijk}$ and $w^*_\theta$ deterministic given
$\vs^*,\vbeta^*$,
\begin{equation}\label{eq:xi}
\xi_{ijk}:=W^*_{ijk}-\bar w_{ijk}
=(c_{ijk}-g^*_{ijk})\big(w^*_{\theta,ijk}-f''(\Delta^*_{ij})\big),
\qquad \E[\xi_{ijk}]=0,\quad \abs{\xi_{ijk}}\le B:=\frac14+F_2.
\end{equation}
$\mV$ is the expected (population) Hessian; $\mE$ is the mean-zero
``saboteur''. The symbol $\xi_{ijk}$ denotes this \emph{curvature} fluctuation
throughout; the unrelated leave-one-out \emph{score} fluctuation introduced in
\eqref{eq:loo-identity} is written $\chi_{imk}$.

\begin{lemma}[Population floor]\label{lem:Vfloor}
Under A\ref{assump:dense}--A\ref{assump:gate}, $\lamperp(\mV)\ge nb_0$, with
$b_0$ as in Lemma~\ref{lem:floor}.
\end{lemma}

\begin{proof}
$\mV$ is the Laplacian of the comparison graph with edge weights
$\sum_k\bar w_{ijk}$. For centered $\vx$ ($\vone\trans\vx=0$),
\[
\vx\trans\mV\vx=\sum_{i<j}\sum_k\bar w_{ijk}(x_i-x_j)^2
\ \ge\ b_0\sum_{i<j}(x_i-x_j)^2=b_0\,\vx\trans\mL\,\vx,
\]
using $\bar w_{ijk}\ge b_0$ (Lemma~\ref{lem:floor}) and $m_{ij}\ge1$
(A\ref{assump:dense}), where $\mL$ is the unweighted Laplacian of $K_{n+1}$
(the complete graph on $\{0,\dots,n\}$). Since
$\lamperp(\mL)=\lambda_2(\mL)=n+1$ for $K_{n+1}$,
$\vx\trans\mV\vx\ge nb_0\norm{\vx}_2^2$. Minimizing over centered unit $\vx$
gives the claim.
\end{proof}

\subsection{The leave-one-out construction}\label{app:loo}

The $\ell_\infty$ analysis of Appendix~\ref{app:consistency} compares each
gradient iterate with a \emph{leave-one-out (LOO)} proxy, defined here once
and for all. Fix an item $m\in\{0,1,\dots,n\}$, and let $\Fm$ be the
$\sigma$-algebra generated by the outcomes of all comparisons \emph{not}
involving item $m$,
\[
\Fm:=\sigma\big(\{c_{ijk},a^1_{ijk}\ :\ 0\le i<j\le n,\
m\notin\{i,j\},\ k\}\big).
\]
The \emph{LOO objective} replaces every comparison involving $m$ by its
expectation under the truth, leaving the remaining comparisons untouched:
\begin{equation}\label{eq:ll_loo}
\ell^{(m)}(\vs,\vgamma,\vbeta)
=\sum_{\substack{i<j\\ i\neq m,\ j\neq m}}\sum_k
\ell_{ijk}(\vs,\vgamma,\vbeta)
\;+\;\sum_{i\neq m}\sum_k \bar\ell_{imk}(\vs,\vgamma,\vbeta),
\end{equation}
where
\[
\begin{aligned}
\bar\ell_{imk}(\vs,\vgamma,\vbeta)
:={}& g^*_{imk}\log g_{imk}+(1-g^*_{imk})\log(1-g_{imk})\\
&+ g^*_{imk}\big[d^*_{imk}\log d_{imk}
+(1-d^*_{imk})\log(1-d_{imk})\big].
\end{aligned}
\]
Its merit gradient $\mG^{(m)}_{\vgamma,\vbeta}:=\nabla_{\vs}(-\ell^{(m)})$ is
\begin{equation}\label{eq:score_loo}
\begin{aligned}
\mG^{(m)}_{\vgamma,\vbeta}(\vs)
&= -\sum_{\substack{i<j \\ i\neq m,\ j\neq m}} \sum_k
   \big[(c_{ijk}-g_{ijk})f'(\Delta_{ij})
   + (a^1_{ijk}-c_{ijk}d_{ijk})\big]\vomega_{ij} \\
&\quad - \sum_{i\neq m}\sum_k
   \big[(g^*_{imk}-g_{imk})f'(\Delta_{im})
   + g^*_{imk}(d^*_{imk}-d_{imk})\big]\vomega_{im},
\end{aligned}
\end{equation}
and its merit Hessian $\mJ^{(m)}:=\nabla^2_{\vs}(-\ell^{(m)})$ is the graph
Laplacian
\begin{equation}\label{eq:hessian_loo}
\mJ^{(m)}(\vs,\vgamma,\vbeta)
=\sum_{i<j}W^{(m)}_{ij}\,\vomega_{ij}\vomega_{ij}\trans,
\qquad W^{(m)}_{ij}=\sum_k W^{(m)}_{ijk},
\end{equation}
with edge weights
\begin{equation}\label{eq:hessian_loo_weights}
W^{(m)}_{ijk}=
\begin{cases}
W_{ijk}, & m \notin \{i, j\},\\[4pt]
w_{\eta,imk}\,f'(\Delta_{im})^2+g^*_{imk}\,w_{\theta,imk}
-(g^*_{imk}-g_{imk})\,f''(\Delta_{im}), & m \in \{i, j\}.
\end{cases}
\end{equation}

\begin{remark}[Symmetry of the weights]\label{rem:parity}
Every per-comparison curvature weight is symmetric under
$i\leftrightarrow j$: indeed $g_{jik}=g_{ijk}$ and $w_{\eta,jik}=w_{\eta,ijk}$
(the gate predictor is even), $w_{\theta,jik}=w_{\theta,ijk}$ (since
$d_{jik}=1-d_{ijk}$), $f'(\Delta_{ji})^2=f'(\Delta_{ij})^2$ and
$f''(\Delta_{ji})=f''(\Delta_{ij})$ ($f'$ odd, $f''$ even), and
$c_{jik}=c_{ijk}$; hence $W_{jik}=W_{ijk}$ and likewise
$W^{(m)}_{mik}=W^{(m)}_{imk}$ (using $g^*_{mik}=g^*_{imk}$). The score
summands are instead antisymmetric, $\psi_{jik}=-\psi_{ijk}$, matching
$\vomega_{ji}=-\vomega_{ij}$. We use the two orderings of a pair
interchangeably below.
\end{remark}

Like $\mG$ and $\mJ$, the objects $\mG^{(m)}_{\vgamma,\vbeta}$ and $\mJ^{(m)}$
take values in / act on $\vone^\perp$ (every $\vomega_{im}$ is centered). Two
structural facts are used repeatedly.

\medskip
\noindent\emph{Measurability.} The construction \eqref{eq:ll_loo} strips out
all of item $m$'s randomness: in the $m$-edge terms $\bar\ell_{imk}$ the
outcomes $c_{imk},a^1_{imk}$ have been replaced by their population
expectations $g^*_{imk}$ and $g^*_{imk}d^*_{imk}$, so $\bar\ell_{imk}$ is a
\emph{non-random} function of $(\vs,\vgamma,\vbeta)$ (the covariates and the
population quantities are fixed under
A\ref{assump:dense}--A\ref{assump:gate}). The remaining terms depend on the
data only through the outcomes of comparisons disjoint from $m$, i.e.\ the
generators of $\Fm$. Hence $\ell^{(m)}(\cdot,\vgamma,\vbeta)$,
$\mG^{(m)}_{\vgamma,\vbeta}$, and $\mJ^{(m)}$ are $\Fm$-measurable as random
functions of $\vs$ (a non-random function being measurable with respect to
any $\sigma$-algebra). Consequently any iterative scheme driven by
$\mG^{(m)}_{\vgamma,\vbeta}$ and started at the deterministic point $\vs^*$
produces $\Fm$-measurable iterates; in particular the LOO trajectory
$\{\vs^{t,(m)}\}_t$ of Algorithm~\ref{alg:gd_loo} below is $\Fm$-measurable at
every step.

\medskip
\noindent\emph{The score discrepancy.} Subtracting \eqref{eq:score-map} from
\eqref{eq:score_loo}, all pairs disjoint from $m$ cancel and the $m$-edges
combine into
\begin{equation}\label{eq:loo-identity}
\begin{gathered}
\mG^{(m)}_{\vgamma,\vbeta}(\vs)-\mG_{\vgamma,\vbeta}(\vs)
=\mD^{(m)}_{\vgamma,\vbeta}(\vs)
:=\sum_{i\ne m}\sum_k \chi_{imk}\,\vomega_{im},\\
\chi_{imk}:=(c_{imk}-g^*_{imk})\big(f'(\Delta_{im})-d_{imk}\big)
+\big(a^1_{imk}-g^*_{imk}d^*_{imk}\big).
\end{gathered}
\end{equation}
Note that $\mD^{(m)}_{\vgamma,\vbeta}$ depends on the structural parameters
only through $d_{imk}=\sigma(\Delta_{im}+\mZ_{imk}\trans\vbeta)$, i.e.\ on
$\vbeta$ and not on $\vgamma$.

\subsection{Concentration}\label{app:conc}

\begin{lemma}[Score bounds]\label{lem:score}
Under A\ref{assump:dense}--A\ref{assump:gate}, for every $c>0$, with
probability at least $1-O(n^{-c})$,
\begin{align}
\text{(merit)}\quad &\max_{0\le a\le n}\abs{G_a(\vs^*,\vgamma^*,\vbeta^*)}
\le C_\psi\sqrt{n\log n},\qquad
C_\psi:=2(F_1+1)\sqrt{(c+1)\,m^\ast},\label{eq:Escore}\\
\text{(structural)}\quad
&\norm{\nabla_{\vgamma}\ell(\vs^*,\vgamma^*,\vbeta^*)}_2
\le C_\kappa\,n\sqrt{\log n},\qquad
\norm{\nabla_{\vbeta}\ell(\vs^*,\vgamma^*,\vbeta^*)}_2
\le C_\kappa\,n\sqrt{\log n},\label{eq:Estruct}
\end{align}
where $C_\kappa:=\kappa\sqrt{2\,m^\ast(c+1)\max\{p_{\gamma},p_{\beta}\}}$.
\end{lemma}

\begin{proof}
\emph{Merit.} Fix $a$. By \eqref{eq:score}, $G_a=-\sum_{j\ne a}\sum_k\psi_{ajk}$
at the truth, a sum of $m_a:=\sum_{j\ne a}m_{aj}\in[n,nm^\ast]$ independent
terms. Each $\psi_{ajk}$ is a bounded function of $(c_{ajk},a^1_{ajk})$ with
$\abs{\psi_{ajk}}\le F_1+1$ and is mean-zero: $\E[(c-g^*)f']=0$ since
$\E c=g^*$, and
$\E[a^1-cd^*]=\E[\E[a^1\mid c]-cd^*]=\E[cd^*-cd^*]=0$. Hoeffding
(Fact~\ref{fact:hoeffding}) with per-term range $2(F_1+1)$ gives
\[
\PP\big(\abs{G_a}\ge t\big)
\le2\exp\!\left(-\frac{2t^2}{m_a(2(F_1+1))^2}\right)
=2\exp\!\left(-\frac{t^2}{2m_a(F_1+1)^2}\right).
\]
Take $t=C_\psi\sqrt{n\log n}$ with $C_\psi=2(F_1+1)\sqrt{(c+1)m^\ast}$. Since
$m_a\le nm^\ast$, the exponent is
\[
-\frac{C_\psi^2\,n\log n}{2m_a(F_1+1)^2}
\le-\frac{4(F_1+1)^2(c+1)m^\ast\,n\log n}{2\,nm^\ast(F_1+1)^2}
=-2(c+1)\log n,
\]
so $\PP(\abs{G_a}\ge t)\le2n^{-2(c+1)}$. A union bound over $a=0,\dots,n$
gives the merit claim with probability
$\ge1-2(n+1)n^{-2(c+1)}=1-O(n^{-(2c+1)})=1-O(n^{-c})$.

\medskip
\emph{Structural.} Each coordinate of $\nabla_{\vgamma}\ell$ (and of
$\nabla_{\vbeta}\ell$) is a sum of
$M:=\sum_{i<j}m_{ij}\le\binom{n+1}{2}m^\ast$ independent mean-zero terms; the
$l$-th coordinate of $\nabla_{\vgamma}\ell$ is
$\sum_{i<j,k}(c_{ijk}-g^*_{ijk})Y_{ijk,l}$, with each term bounded in absolute
value by $\kappa$ since $\abs{c-g^*}\le1$ and
$\abs{Y_{ijk,l}}\le\norm{\mY_{ijk}}_2\le\kappa$ (and likewise for $\vbeta$).
Hoeffding with per-term range $2\kappa$ and $t=C_\kappa'\sqrt{M\log n}$,
$C_\kappa':=\kappa\sqrt{2(c+1)}$, gives, for each fixed coordinate,
\[
\PP\big(\abs{\text{coord}}\ge t\big)
\le 2\exp\!\Big(-\frac{2t^2}{M(2\kappa)^2}\Big)
=2\exp\!\big(-(c+1)\log n\big)=2n^{-(c+1)}.
\]
A union bound over the fixed $p_{\gamma}+p_{\beta}$ coordinates yields
$\max\{\norm{\nabla_{\vgamma}\ell}_\infty,\norm{\nabla_{\vbeta}\ell}_\infty\}
\le C_\kappa'\sqrt{M\log n}$ with probability $1-O(n^{-(c+1)})=1-O(n^{-c})$.
Since $p_{\gamma},p_{\beta}$ are fixed and
$M\le\binom{n+1}{2}m^\ast\le n^2 m^\ast$ (using $\binom{n+1}{2}\le n^2$ for
$n\ge1$), so $\sqrt M\le n\sqrt{m^\ast}$,
\[
\norm{\nabla_{\vgamma}\ell}_2\le\sqrt{p_{\gamma}}\,C_\kappa'\sqrt{M\log n}
\le \kappa\sqrt{2 p_{\gamma} m^\ast(c+1)}\;n\sqrt{\log n}
\le C_\kappa\,n\sqrt{\log n},
\]
with $C_\kappa:=\kappa\sqrt{2 m^\ast(c+1)\max\{p_{\gamma},p_{\beta}\}}$, and
identically for $\nabla_{\vbeta}\ell$.
\end{proof}

\begin{lemma}[Spectral control of the saboteur]\label{lem:saboteur}
Under A\ref{assump:dense}--A\ref{assump:gate}, for every $c>0$ there is an
absolute constant $C_E$ such that the event
\[
\gE_{\mathrm{sab}}:=\big\{\opnorm{\mE}_2\le C_E\sqrt{n\log n}\big\}
\]
has probability at least $1-O(n^{-c})$.
\end{lemma}

\begin{proof}
Write $\mE=\sum_{i<j,k}\mX_{ijk}$ with
$\mX_{ijk}=\xi_{ijk}\,\vomega_{ij}\vomega_{ij}\trans$, independent, symmetric,
mean-zero ($\E\xi_{ijk}=0$ by \eqref{eq:xi}). Since
$\norm{\vomega_{ij}}_2^2=2$ and $\abs{\xi_{ijk}}\le B$,
$\opnorm{\mX_{ijk}}_2\le 2B=:R$. Using
$(\vomega_{ij}\vomega_{ij}\trans)^2=2\,\vomega_{ij}\vomega_{ij}\trans$ and
$\E[\xi_{ijk}^2]\le B^2$,
\[
\sum_{i<j,k}\E\mX_{ijk}^2
=\sum_{i<j,k}2\,\E[\xi_{ijk}^2]\,\vomega_{ij}\vomega_{ij}\trans
\ \preceq\ 2B^2m^\ast\sum_{i<j}\vomega_{ij}\vomega_{ij}\trans,
\]
and the right-hand side of the last display is $2B^2m^\ast$ times the
Laplacian of $K_{n+1}$, of spectral norm $n+1$, so
$\opnorm{\sum_{i<j,k}\E\mX_{ijk}^2}_2\le 2B^2m^\ast(n+1)=:\nu$. With
$t=C_E\sqrt{n\log n}$, the matrix-Bernstein
(Fact~\ref{fact:matrix-bernstein}) exponent satisfies
\[
-\frac{t^2}{2\nu+2Rt/3}\ \underset{n\to\infty}{\sim}\
-\frac{C_E^2}{4B^2m^\ast}\log n,
\]
so $\PP(\opnorm{\mE}_2\ge t)\le
2(n+1)\,n^{-C_E^2/(4B^2m^\ast)}=O(n^{-c})$ for $C_E$ large enough.
\end{proof}

\begin{lemma}[Leave-one-out perturbation bound]\label{lem:loo-fluct}
Assume A\ref{assump:dense}--A\ref{assump:gate}, and fix
$m\in\{0,1,\dots,n\}$, a pair $(\vgamma,\vbeta)$, and an $\Fm$-measurable
random point $\vs\in\R^{n+1}$. With $\mD^{(m)}_{\vgamma,\vbeta}$ and
$\chi_{imk}$ as in \eqref{eq:loo-identity}, for every $c>0$, with $K:=2(F_1+2)\sqrt{2(c+1)\,m^\ast}$, there is $n_0=n_0(c)$ such that for all $n\ge n_0$,
\[
\PP\Big(\Big\{\norm{\mD^{(m)}_{\vgamma,\vbeta}(\vs)}_2> K\sqrt{n\log n}\Big\}
\cap\Big\{\norm{\vs}_\infty\le C_s+\frac12\Big\}\Big)\ =\ O(n^{-c}).
\]
In particular, if $\norm{\vs}_\infty\le C_s+\frac12$ almost surely, then
$\norm{\mD^{(m)}_{\vgamma,\vbeta}(\vs)}_2\le K\sqrt{n\log n}$ with probability
at least $1-O(n^{-c})$.
\end{lemma}

\begin{proof}
Condition on $\Fm$. By hypothesis $\vs$ is then deterministic, hence so are
$f'(\Delta_{im})$ and $d_{imk}$, and the confinement event
$\{\norm{\vs}_\infty\le C_s+\frac12\}$ is $\Fm$-measurable. On the
$\Fm$-atoms where confinement fails, the conditional probability of the
displayed intersection is zero, so it suffices to prove the conditional bound
on the atoms where $\norm{\vs}_\infty\le C_s+\frac12$; there, every gap
satisfies $\abs{\Delta_{im}}\le 2C_s+1$, so the bounds of A\ref{assump:gate}
apply. The pairs $\{(c_{imk},a^1_{imk})\}_{i\ne m,k}$ are independent across
$(i,k)$ with the law \eqref{eq:model} and independent of $\Fm$. Set
$\mX_{imk}:=\chi_{imk}\,\vomega_{im}\in\R^{n+1}$ and verify vector Bernstein
(Fact~\ref{fact:vector-bernstein}) for $\{\mX_{imk}\}_{i\ne m,k}$ under the
conditional law.

\emph{(a) Mean-zero.} $\E[c_{imk}-g^*_{imk}]=0$ and
$\E[a^1_{imk}-g^*_{imk}d^*_{imk}]=0$. Since $f'(\Delta_{im})-d_{imk}$ is
deterministic under the conditioning, $\E[\chi_{imk}\mid\Fm]=0$, hence
$\E[\mX_{imk}\mid\Fm]=\vzero$.

\emph{(b) Uniform bound.} $\abs{c-g^*}\le1$, $\abs{a^1-g^*d^*}\le1$,
$\abs{f'}\le F_1$ (A\ref{assump:gate}, on the confinement atoms),
$d\in[0,1]$, so $\abs{\chi_{imk}}\le(F_1+1)+1=F_1+2$. As
$\norm{\vomega_{im}}_2=\sqrt2$,
$\norm{\mX_{imk}}_2=\sqrt2\,\abs{\chi_{imk}}\le\sqrt2\,(F_1+2)=:R$.

\emph{(c) Total second moment.} There are $n$ indices $i\ne m$ in
$\{0,\dots,n\}$, each with $\le m^\ast$ comparisons, so
\[
\sum_{i\ne m,k}\E\big[\norm{\mX_{imk}}_2^2\mid\Fm\big]
=\sum_{i\ne m,k}2\,\E[\chi_{imk}^2\mid\Fm]
\le 2(F_1+2)^2 m^\ast\,n=:\nu.
\]
By vector Bernstein in dimension $d=n+1$, for any $t\ge0$,
\[
\PP\Big(\norm{\mD^{(m)}_{\vgamma,\vbeta}(\vs)}_2\ge t\ \Big|\ \Fm\Big)
\le 2(n+2)\exp\!\left(-\frac{t^2}{2\nu+2Rt/3}\right).
\]
Take $t=K\sqrt{n\log n}$ with $K=2(F_1+2)\sqrt{2(c+1)m^\ast}$. Then
\[
\frac{Rt}{3\nu}
=\frac{\sqrt2(F_1+2)\,K\sqrt{n\log n}}{3\cdot 2(F_1+2)^2 m^\ast n}
=\frac{K}{3\sqrt2\,(F_1+2)m^\ast}\sqrt{\frac{\log n}{n}}\ \le\ 1
\]
for $n\ge n_0$, so $2\nu+2Rt/3\le 4\nu$ and
\[
\PP\Big(\norm{\mD^{(m)}_{\vgamma,\vbeta}(\vs)}_2\ge t\ \Big|\ \Fm\Big)
\le 2(n+2)\exp\!\Big(-\frac{t^2}{4\nu}\Big)
=2(n+2)\exp\!\Big(-\frac{K^2\,n\log n}{8(F_1+2)^2 m^\ast n}\Big).
\]
With $K^2=8(c+1)(F_1+2)^2 m^\ast$ the exponent is $-(c+1)\log n$, so the
conditional probability is at most $2(n+2)\,n^{-(c+1)}=O(n^{-c})$ on the
confinement atoms. Taking expectations over $\Fm$ completes the proof.
\end{proof}

\begin{remark}[Why the confinement dichotomy]\label{rem:dichotomy}
The uniform bound in step (b) is the only place where the position of $\vs$
enters, through $\abs{f'(\Delta_{im})}\le F_1$; outside the working region
$f'$ need not be bounded. Lemma~\ref{lem:loo-fluct} is therefore stated in
the ``dichotomy'' form: the exceptional event is small \emph{on the part of
the space where $\vs$ is confined}. In the applications below, $\vs$ is an
LOO iterate whose confinement is exactly what the coupled induction of
Lemma~\ref{lem:coupled-induction} establishes on the good event, so the two
statements mesh with no circularity: the good events are defined in dichotomy
form, and the induction supplies the confinement that activates the score
bound at each step.
\end{remark}

\subsection{Hessian estimates}\label{app:lip}

\begin{lemma}[Hessian Lipschitz bound in $\vs$]\label{lem:lip}
Under A\ref{assump:dense}--A\ref{assump:gate}, for all $\vx,\vy$ with
$\norm{\vx}_\infty,\norm{\vy}_\infty\le C_s+\frac12$ and any fixed
$\vgamma,\vbeta$,
\[
\opnorm{\mJ(\vx,\vgamma,\vbeta)-\mJ(\vy,\vgamma,\vbeta)}_\infty
\le 4\,L_f\,(\max_a m_a)\,\norm{\vx-\vy}_\infty
\le 4\,L_f\,m^\ast n\,\norm{\vx-\vy}_\infty,
\]
where $L_f=\frac14 F_1^3+\frac34 F_1F_2+F_3+\frac14$ and
$m_a:=\sum_{b\ne a}m_{ab}$. The same bound holds verbatim for the
leave-one-out Hessian $\mJ^{(m)}$ of \eqref{eq:hessian_loo}.
\end{lemma}

\begin{proof}
Since $\mJ$ depends on $\vs$ only through the gaps, it suffices to bound the
scalar derivative $W'_{ijk}(\Delta):=\frac{dW_{ijk}}{d\Delta}$ from
\eqref{eq:W}. Using
\[
\frac{dg}{d\Delta}=w_\eta f',\quad
\frac{dw_\eta}{d\Delta}=w_\eta(1-2g)f',\quad
\frac{dd}{d\Delta}=w_\theta,\quad
\frac{dw_\theta}{d\Delta}=w_\theta(1-2d),\quad
\frac{dc}{d\Delta}=0,
\]
the three terms of $W$ differentiate to
\[
\begin{aligned}
\frac{d}{d\Delta}\big[w_\eta f'^2\big]&=w_\eta(1-2g)f'^3+2w_\eta f'f'',\\
\frac{d}{d\Delta}\big[c\,w_\theta\big]&=c\,w_\theta(1-2d),\\
\frac{d}{d\Delta}\big[-(c-g)f''\big]&=w_\eta f'f''-(c-g)f'''.
\end{aligned}
\]
Adding (the two $w_\eta f'f''$ terms combine to $3w_\eta f'f''$),
\[
W'(\Delta)=w_\eta(1-2g)f'^3+3w_\eta f'f''+c\,w_\theta(1-2d)-(c-g)f'''.
\]
Bounding with $w_\eta,w_\theta\le\frac14$, $\abs{1-2g},\abs{1-2d}\le1$,
$c\in[0,1]$, $\abs{c-g}\le1$, and
$\abs{f'}\le F_1,\abs{f''}\le F_2,\abs{f'''}\le F_3$ (valid since
$\norm{\vx}_\infty,\norm{\vy}_\infty\le C_s+\frac12$ gives gaps of magnitude
at most $2C_s+1$, the range of A\ref{assump:gate}) gives
$\abs{W'(\Delta)}\le\frac14 F_1^3+\frac34 F_1F_2+\frac14+F_3=L_f$.
To assemble: $\mJ$ is a Laplacian with edge weights $W_{ij}=\sum_k W_{ijk}$;
set $\delta W_{ij}=W_{ij}(\vx)-W_{ij}(\vy)$. Row $a$ of $\mJ(\vx)-\mJ(\vy)$
has diagonal $\sum_{b\ne a}\delta W_{ab}$ and off-diagonals
$-\delta W_{ab}$, so
\[
\norm{\big(\mJ(\vx)-\mJ(\vy)\big)_{a\cdot}}_1
=\Big\lvert\textstyle\sum_{b\ne a}\delta W_{ab}\Big\rvert
+\sum_{b\ne a}\abs{\delta W_{ab}}
\le 2\sum_{b\ne a}\abs{\delta W_{ab}}.
\]
By the mean value theorem and
$\abs{\Delta^{\vx}_{ab}-\Delta^{\vy}_{ab}}
=\abs{(x_a-y_a)-(x_b-y_b)}\le2\norm{\vx-\vy}_\infty$,
each comparison contributes
$\abs{W'_{abk}}\,\abs{\Delta^{\vx}_{ab}-\Delta^{\vy}_{ab}}
\le 2L_f\norm{\vx-\vy}_\infty$, so
$\abs{\delta W_{ab}}\le m_{ab}\,L_f\cdot2\norm{\vx-\vy}_\infty$. Summing over
$b$,
\[
\norm{(\mJ(\vx)-\mJ(\vy))_{a\cdot}}_1
\le 4L_f\Big(\sum_{b\ne a}m_{ab}\Big)\norm{\vx-\vy}_\infty
=4L_f m_a\norm{\vx-\vy}_\infty,
\]
and maximizing over $a$ and using $m_a\le m^\ast n$ (A\ref{assump:dense})
gives the claim. For $\mJ^{(m)}$ the $m$-edge weight $W^{(m)}_{imk}$ has, by
the same calculation,
\[
\Big|\frac{dW^{(m)}_{imk}}{d\Delta}\Big|
=\abs{w_\eta(1-2g)f'^3+3w_\eta f'f''+g^*w_\theta(1-2d)-(g^*-g)f'''}\le L_f
\]
(now $g^*\le1$ replaces $c\in[0,1]$ and $\abs{g^*-g}\le1$ replaces
$\abs{c-g}\le1$), so the identical bound holds.
\end{proof}

\begin{lemma}[Hessian Lipschitz bound in $(\vgamma,\vbeta)$]\label{lem:lip-zeta}
Under A\ref{assump:dense}--A\ref{assump:gate}, there is an absolute constant
$L_\zeta$ such that for all $\vs$ with $\norm{\vs}_\infty\le C_s+\frac12$ and
all $\vgamma,\vgamma'\in\R^{p_{\gamma}}$, $\vbeta,\vbeta'\in\R^{p_{\beta}}$,
\[
\opnorm{\mJ(\vs,\vgamma,\vbeta)-\mJ(\vs,\vgamma',\vbeta')}_\infty
\le L_\zeta\, n\,\big(\norm{\vgamma-\vgamma'}_2+\norm{\vbeta-\vbeta'}_2\big).
\]
The same bound holds verbatim for the leave-one-out Hessian $\mJ^{(m)}$ of
\eqref{eq:hessian_loo}.
\end{lemma}

\begin{proof}
As in Lemma~\ref{lem:lip}, $\mJ$ is a Laplacian with edge weights
$W_{ij}=\sum_k W_{ijk}$, and the row-sum norm is controlled by
$\sum_{b\ne a,k}\abs{W_{abk}(\vgamma,\vbeta)-W_{abk}(\vgamma',\vbeta')}$ up
to the factor $2$ from the diagonal-plus-off-diagonal structure. It therefore
suffices to bound the per-comparison increment of $W_{ijk}$ in
$(\vgamma,\vbeta)$ at fixed $\vs$.

Recall $W_{ijk}=w_\eta f'(\Delta_{ij})^2+c\,w_\theta-(c-g)f''(\Delta_{ij})$
with $g=\sigma(f(\Delta_{ij})+\mY_{ijk}\trans\vgamma)$, $w_\eta=g(1-g)$,
$d=\sigma(\Delta_{ij}+\mZ_{ijk}\trans\vbeta)$, $w_\theta=d(1-d)$.
Differentiating through the predictors $\eta=f+\mY\trans\vgamma$ and
$\theta=\Delta+\mZ\trans\vbeta$:
\[
\nabla_{\vgamma} g=w_\eta\,\mY,\quad
\nabla_{\vgamma} w_\eta=w_\eta(1-2g)\,\mY,\quad
\nabla_{\vbeta} d=w_\theta\,\mZ,\quad
\nabla_{\vbeta} w_\theta=w_\theta(1-2d)\,\mZ,
\]
while $\nabla_{\vgamma}\Delta=\nabla_{\vbeta}\Delta=\vzero$ at fixed $\vs$
and $\nabla_{\vbeta}g=\nabla_{\vgamma}d=\vzero$. Hence
\[
\begin{aligned}
\nabla_{\vgamma} W_{ijk}
&=\big[w_\eta(1-2g)f'(\Delta_{ij})^2+f''(\Delta_{ij})\,w_\eta\big]\,\mY_{ijk},\\
\nabla_{\vbeta} W_{ijk}
&=c\,w_\theta(1-2d)\,\mZ_{ijk}.
\end{aligned}
\]
Using $w_\eta,w_\theta\le\frac14$, $\abs{1-2g},\abs{1-2d}\le1$, $c\in[0,1]$,
$\abs{f'}\le F_1$, $\abs{f''}\le F_2$ on the working region
(A\ref{assump:gate}), and $\norm{\mY_{ijk}}_2,\norm{\mZ_{ijk}}_2\le\kappa$
(A\ref{assump:compact}),
\[
\norm{\nabla_{\vgamma} W_{ijk}}_2\le \frac14(F_1^2+F_2)\,\kappa,\qquad
\norm{\nabla_{\vbeta} W_{ijk}}_2\le \frac14\,\kappa.
\]
By the mean value theorem along the segment from $(\vgamma',\vbeta')$ to
$(\vgamma,\vbeta)$,
\[
\abs{W_{abk}(\vgamma,\vbeta)-W_{abk}(\vgamma',\vbeta')}
\le \frac14(F_1^2+F_2)\kappa\,\norm{\vgamma-\vgamma'}_2
+\frac14\kappa\,\norm{\vbeta-\vbeta'}_2.
\]
Summing over the $\le m^\ast$ comparisons per pair and over $b\ne a$,
applying the factor $2$ from the Laplacian row structure, and maximizing over
$a$,
\begin{align*}
\opnorm{\mJ(\vs,\vgamma,\vbeta)-\mJ(\vs,\vgamma',\vbeta')}_\infty
&\le \frac12 m^\ast\kappa\,n
\Big((F_1^2+F_2)\norm{\vgamma-\vgamma'}_2+\norm{\vbeta-\vbeta'}_2\Big)\\
&\le L_\zeta\,n\big(\norm{\vgamma-\vgamma'}_2+\norm{\vbeta-\vbeta'}_2\big),
\end{align*}
with $L_\zeta:=\frac12 m^\ast\kappa\,\max\{F_1^2+F_2,\,1\}$, an absolute
constant under A\ref{assump:dense}--A\ref{assump:gate}.

For $\mJ^{(m)}$, the $m$-edge weight $W^{(m)}_{imk}$ of
\eqref{eq:hessian_loo_weights} has, by the same differentiation (the saboteur
$-(g^*_{imk}-g_{imk})f''(\Delta_{im})$ contributing
$+\nabla_{\vgamma} g_{imk}\, f''(\Delta_{im})
= w_{\eta,imk} f''(\Delta_{im})\,\mY_{imk}$, exactly as $-(c_{ijk}-g_{ijk})f''$
does above),
\begin{align*}
\nabla_{\vgamma} W^{(m)}_{imk}
  &= \bigl[\, w_{\eta,imk}(1-2g_{imk})\, f'(\Delta_{im})^2
    + w_{\eta,imk}\, f''(\Delta_{im}) \,\bigr] \mY_{imk},\\
\nabla_{\vbeta} W^{(m)}_{imk}
  &= g^*_{imk}\, w_{\theta,imk}(1-2d_{imk})\, \mZ_{imk},
\end{align*}
so with $g^*_{imk}\le 1$ replacing $c_{ijk}\in[0,1]$ the bounds
$\norm{\nabla_{\vgamma} W^{(m)}_{imk}}_2 \le \frac14 (F_1^2+F_2)\kappa$ and
$\norm{\nabla_{\vbeta} W^{(m)}_{imk}}_2 \le \frac14 \kappa$ are unchanged,
and the row-sum assembly gives the identical bound.
\end{proof}

\begin{lemma}[Cross-Jacobian bound]\label{lem:cross-jac}
Assume A\ref{assump:dense}--A\ref{assump:gate} and $n\ge2$. There is an
absolute constant $C_M$ such that, for all
$(\vgamma,\vbeta)\in\R^{p_{\gamma}}\times\R^{p_{\beta}}$ and all $\vs$ with
$\norm{\vs}_\infty\le C_s+\frac12$,
\[
\opnorm{\frac{\partial \mG_{\vgamma,\vbeta}(\vs)}
{\partial(\vgamma,\vbeta)}}_2\ \le\ C_M\, n^{3/2}.
\]
The same bound holds verbatim for the leave-one-out cross-Jacobian
$\partial\mG^{(m)}_{\vgamma,\vbeta}(\vs)/\partial(\vgamma,\vbeta)$, with
$\mG^{(m)}_{\vgamma,\vbeta}$ as in \eqref{eq:score_loo}.
\end{lemma}

\begin{proof}
Since $\mG_{\vgamma,\vbeta}=\nabla_{\vs}(-\ell)$ (the merit score
\eqref{eq:score-map}),
\[
\frac{\partial \mG_{\vgamma,\vbeta}}{\partial(\vgamma,\vbeta)}
=\nabla^2_{\vs,(\vgamma,\vbeta)}(-\ell)
=\big[\,\mB_{\vgamma}\ \big|\ \mB_{\vbeta}\,\big],
\]
with $\mB_{\vgamma},\mB_{\vbeta}$ as in \eqref{eq:Hsg}--\eqref{eq:Hsb}. We
bound $\opnorm{\mB_{\vgamma}}_2$ and $\opnorm{\mB_{\vbeta}}_2$ directly.

For $\mB_{\vgamma}$, take unit $\vu\in\vone^\perp\subset\R^{n+1}$ and unit
$\vv\in\R^{p_{\gamma}}$ (the columns of $\mB_{\vgamma}$ lie in
$\vone^\perp$, so this restriction loses nothing):
\[
\vu\trans \mB_{\vgamma} \vv
=\sum_{i<j,k} w_{\eta,ijk}f'(\Delta_{ij})\,(u_i-u_j)\,(\mY_{ijk}\trans\vv).
\]
By Cauchy--Schwarz,
\[
\big|\vu\trans \mB_{\vgamma} \vv\big|
\le\underbrace{\Big(\sum_{i<j,k}\big(w_{\eta,ijk}f'(\Delta_{ij})\big)^2
(u_i-u_j)^2\Big)^{1/2}}_{=:T_1}
\underbrace{\Big(\sum_{i<j,k}(\mY_{ijk}\trans\vv)^2\Big)^{1/2}}_{=:T_2}.
\]
For $T_1$: $\abs{w_{\eta,ijk}f'(\Delta_{ij})}\le\frac14 F_1$, and summing the
$\le m^\ast$ comparisons per pair,
$\sum_{i<j,k}(u_i-u_j)^2\le m^\ast\sum_{i<j}(u_i-u_j)^2\le m^\ast(n+1)$, so
$T_1\le\frac14 F_1\sqrt{m^\ast(n+1)}$. For $T_2$:
$(\mY_{ijk}\trans\vv)^2\le\kappa^2$ and there are
$\le\binom{n+1}{2}m^\ast$ terms, so
$T_2\le\kappa\sqrt{\binom{n+1}{2}m^\ast}
\le\frac{1}{\sqrt2}\kappa\sqrt{m^\ast}(n+1)$. Taking the supremum over
$\vu,\vv$,
\[
\opnorm{\mB_{\vgamma}}_2\le T_1T_2
\le \frac{1}{4\sqrt2}\,F_1\kappa\, m^\ast\,(n+1)^{3/2}.
\]
We bound $\mB_{\vbeta}$ identically, with $\abs{c_{ijk}w_{\theta,ijk}}\le\frac14$
(so $F_1$ is replaced by $1$) and $\mZ_{ijk}$ in place of $\mY_{ijk}$:
\[
\opnorm{\mB_{\vbeta}}_2\le \frac{1}{4\sqrt2}\,\kappa\, m^\ast\,(n+1)^{3/2}.
\]
Using
$\opnorm{[\mB_{\vgamma}\mid\mB_{\vbeta}]}_2
\le\opnorm{\mB_{\vgamma}}_2+\opnorm{\mB_{\vbeta}}_2$ and
$(n+1)^{3/2}\le 2n^{3/2}$ for $n\ge2$,
\[
\opnorm{\frac{\partial \mG_{\vgamma,\vbeta}(\vs)}
{\partial(\vgamma,\vbeta)}}_2
\le \frac{1}{2\sqrt2}\,(F_1+1)\,\kappa\, m^\ast\,n^{3/2}=C_M\,n^{3/2},
\qquad C_M:=\frac{1}{2\sqrt2}(F_1+1)\kappa m^\ast.
\]
For the leave-one-out variant, differentiating the $m$-edge terms of
\eqref{eq:score_loo} at fixed $\vs$ (using
$\nabla_{\vgamma}(g^*_{imk}-g_{imk})=-w_{\eta,imk}\mY_{imk}$ and
$\nabla_{\vbeta}(d^*_{imk}-d_{imk})=-w_{\theta,imk}\mZ_{imk}$, the starred
quantities being constants) gives
\[
\mB^{(m)}_{\vgamma}
:=\partial_{\vgamma}\mG^{(m)}_{\vgamma,\vbeta}
=\sum_{\substack{i<j\\ m\notin\{i,j\}}}\sum_k
   w_{\eta,ijk}f'(\Delta_{ij})\,\vomega_{ij}\mY_{ijk}\trans
 +\sum_{i\ne m}\sum_k
   w_{\eta,imk}f'(\Delta_{im})\,\vomega_{im}\mY_{imk}\trans,
\]
and $\mB^{(m)}_{\vbeta}$ analogously with $g^*_{imk}w_{\theta,imk}$ in place
of $c_{ijk}w_{\theta,ijk}$. Every column again lies in $\vone^\perp$, each
coefficient obeys the same bounds
($\abs{w_{\eta,imk}f'(\Delta_{im})}\le\frac14 F_1$,
$\abs{g^*_{imk}w_{\theta,imk}}\le\frac14$, with $g^*_{imk}\le1$ replacing
$c_{ijk}\in[0,1]$), and the sums range over each unordered pair exactly once
with $\le m^\ast$ comparisons, so the Cauchy--Schwarz argument applies
verbatim and yields the identical bound $C_M n^{3/2}$.
\end{proof}

\begin{lemma}[Cross-block Lipschitz bound]\label{lem:cross-lip}
Assume A\ref{assump:dense}--A\ref{assump:gate} and $n\ge2$. Write
$\mB(\vs,\vgamma,\vbeta):=[\,\mB_{\vgamma}\mid\mB_{\vbeta}\,]$ for the
merit--structural cross block of \eqref{eq:Hsg}--\eqref{eq:Hsb}. There is an
absolute constant $L_{B}$ such that, for all $\vs,\vs'$ with
$\norm{\vs}_\infty,\norm{\vs'}_\infty\le C_s+\frac12$ and all
$\vzeta=(\vgamma,\vbeta)$, $\vzeta'=(\vgamma',\vbeta')$,
\[
\opnorm{\mB(\vs,\vzeta)-\mB(\vs',\vzeta')}_2
\ \le\ L_{B}\,n^{3/2}\Big(\norm{\vs-\vs'}_\infty+\norm{\vzeta-\vzeta'}_2\Big).
\]
\end{lemma}

\begin{proof}
Both blocks share the structure
$\sum_{i<j}\sum_k\phi_{ijk}\,\vomega_{ij}\vv_{ijk}\trans$ with a scalar
per-comparison coefficient $\phi_{ijk}$ and a fixed covariate $\vv_{ijk}$:
for $\mB_{\vgamma}$, $\phi_{ijk}=w_{\eta,ijk}f'(\Delta_{ij})$ and
$\vv_{ijk}=\mY_{ijk}$; for $\mB_{\vbeta}$, $\phi_{ijk}=c_{ijk}w_{\theta,ijk}$
and $\vv_{ijk}=\mZ_{ijk}$. The difference of two such blocks has the same
structure with coefficient
$\delta_{ijk}:=\phi_{ijk}(\vs,\vzeta)-\phi_{ijk}(\vs',\vzeta')$, so the
Cauchy--Schwarz factorization of Lemma~\ref{lem:cross-jac} applies verbatim:
for unit $\vu\in\vone^\perp$ and unit $\vv$,
\begin{multline*}
\abs{\vu\trans\big(\mB_{\vgamma}(\vs,\vzeta)
-\mB_{\vgamma}(\vs',\vzeta')\big)\vv}
\le\Big(\sum_{i<j,k}\delta_{ijk}^2(u_i-u_j)^2\Big)^{1/2}
   \Big(\sum_{i<j,k}(\mY_{ijk}\trans\vv)^2\Big)^{1/2}\\
\le\ \max_{i<j,k}\abs{\delta_{ijk}}\cdot
\frac{1}{\sqrt2}\,\kappa\,m^\ast\,(n+1)^{3/2}.
\end{multline*}
It remains to bound the coefficient increments. For $\mB_{\vgamma}$: on the
working region,
\begin{align*}
\Big|\frac{\partial(w_\eta f')}{\partial\Delta}\Big|
&=\abs{w_\eta(1-2g)f'^2+w_\eta f''}\le\frac14(F_1^2+F_2),\\
\norm{\nabla_{\vgamma}(w_\eta f')}_2
&=\abs{w_\eta(1-2g)f'}\,\norm{\mY_{ijk}}_2\le\frac14 F_1\kappa,
\end{align*}
and $\nabla_{\vbeta}(w_\eta f')=\vzero$; with
$\abs{\Delta_{ij}(\vs)-\Delta_{ij}(\vs')}\le2\norm{\vs-\vs'}_\infty$ and the
mean value theorem along the segment (which stays in the working region, the
$\ell_\infty$ ball being convex),
\[
\abs{\delta_{ijk}}
\ \le\ \frac12(F_1^2+F_2)\norm{\vs-\vs'}_\infty
+\frac14 F_1\kappa\,\norm{\vgamma-\vgamma'}_2.
\]
For $\mB_{\vbeta}$: at fixed data $c_{ijk}\in\{0,1\}$,
$\abs{\partial(c\,w_\theta)/\partial\Delta}=\abs{c\,w_\theta(1-2d)}\le\frac14$
and $\norm{\nabla_{\vbeta}(c\,w_\theta)}_2\le\frac14\kappa$, with
$\nabla_{\vgamma}(c\,w_\theta)=\vzero$, giving
$\abs{\delta_{ijk}}\le\frac12\norm{\vs-\vs'}_\infty
+\frac14\kappa\norm{\vbeta-\vbeta'}_2$. Combining the two blocks by
$\opnorm{[\,\mB_{\vgamma}\mid\mB_{\vbeta}\,]}_2
\le\opnorm{\mB_{\vgamma}}_2+\opnorm{\mB_{\vbeta}}_2$ and using
$(n+1)^{3/2}\le2n^{3/2}$ for $n\ge2$,
\[
\opnorm{\mB(\vs,\vzeta)-\mB(\vs',\vzeta')}_2
\ \le\ L_{B}\,n^{3/2}\Big(\norm{\vs-\vs'}_\infty
+\norm{\vzeta-\vzeta'}_2\Big),
\]
with $L_{B}:=\sqrt2\,\kappa\,m^\ast\big(F_1^2+F_2+F_1\kappa+1+\kappa\big)$,
an absolute constant under A\ref{assump:dense}--A\ref{assump:gate}.
\end{proof}

\begin{lemma}[Structural-block Lipschitz bound]\label{lem:struct-curv}
Assume A\ref{assump:dense}--A\ref{assump:gate}. Write
$\mA:=\operatorname{diag}(\mA_{\vgamma\vgamma},\mA_{\vbeta\vbeta})$ for the
block-diagonal structural Hessian of \eqref{eq:Hgg} and \eqref{eq:Hbb}. There
is an absolute constant $L_{A}$ such that, for all $\vs,\vs'$ with
$\norm{\vs}_\infty,\norm{\vs'}_\infty\le C_s+\frac12$ and all
$\vzeta=(\vgamma,\vbeta)$, $\vzeta'=(\vgamma',\vbeta')$,
\[
\opnorm{\mA(\vs,\vzeta)-\mA(\vs',\vzeta')}_2
\ \le\ L_{A}\,n^{2}\Big(\norm{\vs-\vs'}_\infty+\norm{\vzeta-\vzeta'}_2\Big).
\]
\end{lemma}

\begin{proof}
$\mA$ is block-diagonal, so it suffices to bound each block. For the gate
block, each weight $w_{\eta,ijk}=g(1-g)$ obeys, on the working region,
\[
\abs{\partial_\Delta w_{\eta,ijk}}
=\abs{w_\eta(1-2g)f'(\Delta_{ij})}\le\frac14 F_1,
\qquad
\norm{\nabla_{\vgamma}w_{\eta,ijk}}_2
=\abs{w_\eta(1-2g)}\,\norm{\mY_{ijk}}_2\le\frac14\kappa,
\]
and $\nabla_{\vbeta}w_{\eta,ijk}=\vzero$; with
$\abs{\Delta_{ij}(\vs)-\Delta_{ij}(\vs')}\le2\norm{\vs-\vs'}_\infty$ and the
mean value theorem along the segment (contained in the working region by
convexity),
\[
\abs{w_{\eta,ijk}(\vs,\vgamma)-w_{\eta,ijk}(\vs',\vgamma')}
\le\frac12 F_1\norm{\vs-\vs'}_\infty
+\frac14\kappa\norm{\vgamma-\vgamma'}_2.
\]
Since $\opnorm{\mY_{ijk}\mY_{ijk}\trans}_2\le\kappa^2$ and there are
$\le\binom{n+1}{2}m^\ast\le n^2m^\ast$ summands, the triangle inequality
gives
\[
\opnorm{\mA_{\vgamma\vgamma}(\vs,\vgamma)
-\mA_{\vgamma\vgamma}(\vs',\vgamma')}_2
\le\kappa^2m^\ast n^2\big(\frac12 F_1\norm{\vs-\vs'}_\infty
+\frac14\kappa\norm{\vgamma-\vgamma'}_2\big).
\]
For the direction block, at fixed data $c_{ijk}\in\{0,1\}$,
$\abs{\partial_\Delta(c\,w_{\theta})}\le\frac14$,
$\norm{\nabla_{\vbeta}(c\,w_{\theta})}_2\le\frac14\kappa$,
$\nabla_{\vgamma}(c\,w_\theta)=\vzero$, and the same assembly applies with
$\mZ_{ijk}$ in place of $\mY_{ijk}$. Summing the two blocks yields the claim
with $L_{A}:=\kappa^2m^\ast\max\{F_1+1,\ \kappa\}$.
\end{proof}

\begin{lemma}[Upper curvature bound]\label{lem:curv-ceiling}
Assume A\ref{assump:dense}--A\ref{assump:gate}. With
$W_{\max}:=\frac14 F_1^2+\frac14+F_2$ and $C_{\max}:=2m^\ast W_{\max}$, the
merit Hessian satisfies, for \emph{every} $\vs$ with
$\norm{\vs}_\infty\le C_s+\frac12$ and every fixed $(\vgamma,\vbeta)$,
\[
\opnorm{\mJ(\vs,\vgamma,\vbeta)}_2\ \le\ C_{\max}\,n.
\]
The same bound holds for the leave-one-out Hessian $\mJ^{(m)}$ of
\eqref{eq:hessian_loo}.
\end{lemma}

\begin{proof}
$\mJ$ is the graph Laplacian with edge weights $W_{ij}=\sum_k W_{ijk}$,
where, from \eqref{eq:W}, the first two terms of $W_{ijk}$ are nonnegative
and the saboteur is bounded by $F_2$, so
\[
-W_{\max} \le\ -\abs{c_{ijk}-g_{ijk}}\abs{f''(\Delta_{ij})}
 \le\ W_{ijk}
 \le\ \frac14 F_1^2+\frac14+F_2 = W_{\max},
\qquad\text{i.e.}\quad \abs{W_{ijk}}\le W_{\max}.
\]
For a graph Laplacian $\mJ=\sum_{i<j}W_{ij}\,\vomega_{ij}\vomega_{ij}\trans$
with weights $W_{ij}$, row $a$ has diagonal $\sum_{b\ne a}W_{ab}$ and
off-diagonals $-W_{ab}$, so by Gershgorin (Fact~\ref{fact:gershgorin}) every
eigenvalue $z$ lies in a disc of centre $\sum_{b\ne a}W_{ab}$ and radius
$\sum_{b\ne a}\abs{W_{ab}}$ for some $a$; since $\mJ$ is symmetric, its
eigenvalues are real and
\[
\opnorm{\mJ}_2\ =\ \max\abs{z}
\ \le\ \max_{a}\Big(\Big|\sum_{b\ne a}W_{ab}\Big|
+\sum_{b\ne a}\abs{W_{ab}}\Big)
\ \le\ 2\max_a\sum_{b\ne a}\abs{W_{ab}}.
\]
Each $W_{ab}=\sum_k W_{abk}$ is a sum of $m_{ab}\le m^\ast$ comparisons, each
$\abs{W_{abk}}\le W_{\max}$, with at most $n$ neighbours $b$, so
$\sum_{b\ne a}\abs{W_{ab}}\le m^\ast n\,W_{\max}$. Hence
$\opnorm{\mJ}_2\le 2m^\ast W_{\max}\,n=C_{\max}\,n$. For $\mJ^{(m)}$ the
$m$-edge weight also satisfies
$\abs{W^{(m)}_{imk}}=\abs{w_\eta f'^2+g^*w_\theta-(g^*-g)f''}
\le\frac14 F_1^2+\frac14+F_2=W_{\max}$, so the identical bound holds.
\end{proof}

\begin{lemma}[Local positive-definiteness of the Hessian]\label{lem:localpd}
Assume A\ref{assump:dense}--A\ref{assump:gate}, with $b_0$ as in
Lemma~\ref{lem:floor}. For any sequences $\varepsilon_n,\rho_n=o(1)$ the following
holds: for every $c>0$ there exists $n_0$ such that for all $n\ge n_0$, on 
$\gE_{\mathrm{sab}}$ as in Lemma~\ref{lem:saboteur}, for every $(\vgamma,\vbeta)$ with
$\norm{\vgamma-\vgamma^*}_2,\norm{\vbeta-\vbeta^*}_2\le\rho_n$ and every
$\vs\in\overline{\gB_\infty(\vs^*,\varepsilon_n)}$,
\[
\lamperp\big(\mJ(\vs,\vgamma,\vbeta)\big)\ \ge\ \frac12\, n b_0.
\]
\end{lemma}

\begin{proof}
We establish the bound at $\vs^*$, where the only randomness is used, then
extend to the ball deterministically.

\medskip
\noindent\emph{Step 1: positive definiteness at $\vs^*$.}
With $\mJ^*=\mV+\mE$ as in \eqref{eq:VE}, Lemma~\ref{lem:Vfloor} gives
$\lamperp(\mV)\ge nb_0$ and Lemma~\ref{lem:saboteur} gives
$\opnorm{\mE}_2\le C_E\sqrt{n\log n}$ on an event of probability at least
$1-O(n^{-c})$. Weyl's inequality on the centered subspace yields
\[
\lamperp(\mJ^*)=\lamperp(\mV+\mE)\ \ge\ \lamperp(\mV)-\opnorm{\mE}_2
\ \ge\ nb_0-C_E\sqrt{n\log n}.
\]

\medskip
\noindent\emph{Step 2: extension to the ball.}
Fix $(\vgamma,\vbeta)$ in the stated balls and
$\vs\in\overline{\gB_\infty(\vs^*,\varepsilon_n)}$. For any centered unit vector
$\vz$ ($\vone\trans\vz=0$, $\norm{\vz}_2=1$),
\[
\vz\trans \mJ(\vs,\vgamma,\vbeta)\,\vz
\ \ge\ \lamperp(\mJ^*)-\opnorm{\mJ(\vs,\vgamma,\vbeta)-\mJ^*}_2.
\]
Using $\opnorm{\cdot}_2\le\opnorm{\cdot}_\infty$ for symmetric matrices, the
triangle inequality, Lemma~\ref{lem:lip} (Lipschitz in $\vs$) and
Lemma~\ref{lem:lip-zeta} (Lipschitz in $(\vgamma,\vbeta)$),
\begin{align*}
\opnorm{\mJ(\vs,\vgamma,\vbeta)-\mJ^*}_2
&\le \opnorm{\mJ(\vs,\vgamma,\vbeta)-\mJ(\vs^*,\vgamma,\vbeta)}_\infty
   + \opnorm{\mJ(\vs^*,\vgamma,\vbeta)-\mJ(\vs^*,\vgamma^*,\vbeta^*)}_\infty\\
&\le 4L_f m^\ast n\,\norm{\vs-\vs^*}_\infty
   + L_\zeta n\big(\norm{\vgamma-\vgamma^*}_2+\norm{\vbeta-\vbeta^*}_2\big)\\
&\le 4L_f m^\ast n\,\varepsilon_n + 2L_\zeta n\,\rho_n\ =\ o(n),
\end{align*}
since $\varepsilon_n,\rho_n=o(1)$ (the segment $[\vs^*,\vs]$ lies in the working
region for $n$ large, so Lemma~\ref{lem:lip} applies).

\medskip
\noindent\emph{Conclusion.} On the event above,
\[
\lamperp\big(\mJ(\vs,\vgamma,\vbeta)\big)
=\inf_{\vz:\ \vone\trans\vz=0,\ \norm{\vz}_2=1}\vz\trans\mJ(\vs,\vgamma,\vbeta)\vz
\ \ge\ nb_0-C_E\sqrt{n\log n}-o(n)\ \ge\ \frac12\, nb_0
\]
for $n\ge n_0$, since $\sqrt{n\log n}=o(n)$.
\end{proof}

\subsection{Stage 1: existence of a local consistent estimator and
\texorpdfstring{$\ell_\infty$}{l-infinity} rates}\label{app:consistency}
\label{app:stage1}

Recall the merit score map $\mG_{\vgamma,\vbeta}=\nabla_{\vs}(-\ell)$ from
\eqref{eq:score-map}, valued in $\vone^\perp$. For fixed $(\vgamma,\vbeta)$
we write $\widehat\vs_{\vgamma,\vbeta}$ for a zero of
$\mG_{\vgamma,\vbeta}$ in $\vone^\perp$; its existence and uniqueness near
$\vs^*$ are the content of Lemma~\ref{lem:stage1}.

\begin{lemma}[Stage-1 $\ell_\infty$ rate]\label{lem:stage1}
Assume A\ref{assump:dense}--A\ref{assump:gate}. Fix any constant
$C_{\gamma\beta}\ge0$ and put
\[
\rho_n:=C_{\gamma\beta}\sqrt{\frac{\log n}{n}},
\qquad
\varepsilon_n:=C_5\sqrt{\frac{\log n}{n}},
\qquad C_5:=C_4+1,
\]
with $C_4=C_4(C_{\gamma\beta})$ the constant of
Lemma~\ref{lem:coupled-induction} below. Then for every $c>0$ there is an
integer $n_0=n_0(c)$ and an event of probability at least $1-O(n^{-c})$ on
which, for all $n\ge n_0$ and every $(\vgamma,\vbeta)$ with
$\max\{\norm{\vgamma-\vgamma^*}_2,\norm{\vbeta-\vbeta^*}_2\}\le\rho_n$, the
merit equation $\mG_{\vgamma,\vbeta}(\vs)=\vzero$ has a solution
$\widehat\vs_{\vgamma,\vbeta}$, unique in
$\overline{\gB_\infty(\vs^*,\varepsilon_n)}\cap(\vs^*+\vone^\perp)$, and
\[
\big\|\widehat\vs_{\vgamma,\vbeta}-\vs^*\big\|_\infty
\ \le\ C_5\sqrt{\frac{\log n}{n}}.
\]
\end{lemma}

The proof rests on the leave-one-out technique. Fix $(\vgamma,\vbeta)$ near
the truth and run gradient descent on $-\ell(\cdot,\vgamma,\vbeta)$,
initialized at $\vs^*$ (Algorithm~\ref{alg:gd}); for each item $m$, run the
same scheme on the LOO objective $\ell^{(m)}$ of \eqref{eq:ll_loo}
(Algorithm~\ref{alg:gd_loo}).

\begin{algorithm}[ht]
\caption{Gradient descent on the merit block (fixed $\vgamma,\vbeta$)}
\label{alg:gd}
\begin{algorithmic}[1]
\Require step size $\eta=(C_{\max}n)^{-1}$, fixed $(\vgamma,\vbeta)$
\State $\vs^{0}\gets \vs^*$
\For{$t=0,1,\dots,T-1$}
  \State $\vs^{t+1}\gets \vs^{t}-\eta\,\mG_{\vgamma,\vbeta}(\vs^{t})$
    \Comment{$\mG_{\vgamma,\vbeta}=\nabla_{\vs}(-\ell)$ is the merit score
    \eqref{eq:score-map}}
\EndFor
\State \Return $\vs^{T}$
\end{algorithmic}
\end{algorithm}

\begin{algorithm}[ht]
\caption{LOO gradient descent on the merit block (fixed $\vgamma,\vbeta$,
item $m$)}
\label{alg:gd_loo}
\begin{algorithmic}[1]
\Require step size $\eta=(C_{\max}n)^{-1}$, fixed $(\vgamma,\vbeta)$
\State $\vs^{0,(m)}\gets \vs^*$
\For{$t=0,1,\dots,T-1$}
  \State $\vs^{t+1,(m)}\gets \vs^{t,(m)}-\eta\,
  \mG^{(m)}_{\vgamma,\vbeta}(\vs^{t,(m)})$
    \Comment{$\mG^{(m)}_{\vgamma,\vbeta}$ as in \eqref{eq:score_loo}}
\EndFor
\State \Return $\vs^{T,(m)}$
\end{algorithmic}
\end{algorithm}

\paragraph{The gauge is preserved automatically.}
Because $\mG_{\vgamma,\vbeta}$ and $\mG^{(m)}_{\vgamma,\vbeta}$ are valued in
$\vone^\perp$ (every $\vomega_{ij}$ is centered) and both trajectories start
at $\vs^0=\vs^{0,(m)}=\vs^*\in\vone^\perp$, every iterate
$\vs^t,\vs^{t,(m)}$ remains in $\vone^\perp$; no explicit projection is
required. In particular $\vs^t-\vs^*\in\vone^\perp$ and
$\vs^t-\vs^{t,(m)}\in\vone^\perp$ for all $t$.

\medskip
Two facts drive the argument. First, because the realized merit Hessian is
positive definite on a neighborhood of $\vs^*$ (Lemma~\ref{lem:localpd}),
the iterates contract toward the local maximizer
$\widehat\vs_{\vgamma,\vbeta}$ and, after $\Theta(\log n)$ steps, lie within
negligible distance of it; it therefore suffices to control the iterates
themselves. Second, the iterates remain close to their starting point
$\vs^*$, which yields consistency.

The difficulty is that this confinement is naturally controlled in $\ell_2$,
whereas the target is the sharper, per-coordinate $\ell_\infty$ rate: an
$\ell_2$ bound permits the entire error to concentrate on a single
coordinate. To rule this out we use, for each coordinate $m$, the
leave-one-out proxy of Appendix~\ref{app:loo}: by \eqref{eq:loo-identity} the
proxy is $\Fm$-measurable, which restores the concentration needed to bound
the $m$-th coordinate individually. The genuine iterate and its proxy are
then shown to remain close, and combining the two gives the uniform
$\ell_\infty$ bound.

Concretely, the argument is a single induction carrying four coupled
estimates simultaneously along the trajectory: the $\ell_2$ confinement, the
per-coordinate accuracy of the proxy, the closeness of iterate and proxy, and
the $\ell_\infty$ bound, which is the invariant that keeps the trajectory
inside the region where positive definiteness holds (Lemma~\ref{lem:localpd})
and where the LOO score bound of Lemma~\ref{lem:loo-fluct} is active
(Remark~\ref{rem:dichotomy}).

\begin{lemma}[Coupled trajectory induction]\label{lem:coupled-induction}
Assume A\ref{assump:dense}--A\ref{assump:gate}. Fix a horizon $T=T(n)\ge1$, a
constant $C_{\gamma\beta}\ge0$, and a pair $(\vgamma,\vbeta)$ with
\[
\norm{\vgamma-\vgamma^*}_2\le C_{\gamma\beta}\sqrt{\frac{\log n}{n}},
\qquad
\norm{\vbeta-\vbeta^*}_2\le C_{\gamma\beta}\sqrt{\frac{\log n}{n}}.
\]
Run Algorithm~\ref{alg:gd} from $\vs^0=\vs^*$, and for each
$m\in\{0,1,\dots,n\}$ run Algorithm~\ref{alg:gd_loo} from
$\vs^{0,(m)}=\vs^*$, both with step size $\eta=(C_{\max}n)^{-1}$. Define the
events
\begin{align*}
\gE_{\mathrm{score}}
  &:= \Big\{ \max_{0\le a\le n}\big|G_a(\vs^*,\vgamma^*,\vbeta^*)\big|
  \le C_\psi\sqrt{n\log n} \Big\}, \\
\gE_{\mathrm{sab}}
  &:= \big\{\, \opnorm{\mE}_2 \le C_E\sqrt{n\log n} \,\big\}, \\
\gE_{\mathrm{loo}}(\vgamma,\vbeta)
  &:= \bigcap_{m=0}^{n}\ \bigcap_{t=0}^{T}
  \left(\Big\{ \norm{\mD^{(m)}_{\vgamma,\vbeta}\!\big(\vs^{t,(m)}\big)}_2
  \le 2K\sqrt{n\log n} \Big\}
  \cup\Big\{\norm{\vs^{t,(m)}-\vs^*}_\infty>\frac12\Big\}\right),
\end{align*}
with $\mD^{(m)}_{\vgamma,\vbeta}$ as in \eqref{eq:loo-identity} and $K$ as in
Lemma~\ref{lem:loo-fluct}, and set
$\gE:=\gE_{\mathrm{score}}\cap\gE_{\mathrm{sab}}
\cap\gE_{\mathrm{loo}}(\vgamma,\vbeta)$. Then for every $c>0$ there exist
constants $C_1,C_2,C_3,C_4>0$, depending only on $c$, on the constants of
A\ref{assump:dense}--A\ref{assump:gate}, and on $C_{\gamma\beta}$ (not on $n$
or $t$), and an integer $n_0=n_0(c)$, such that for all $n\ge n_0$,
\[
\PP(\gE)\ \ge\ 1-O\big((T+1)\,n^{1-c}\big),
\]
and on $\gE$ the following hold simultaneously for all $0\le t\le T$:
\begin{subequations}\label{eq:induction}
\begin{align}
\norm{\vs^{t}-\vs^*}_2 &\le C_1\sqrt{\log n}, \label{eq:ind-l2}\\[2pt]
\max_{0\le m\le n}\big|\,s^{t,(m)}_m-s^*_m\,\big|
&\le C_2\sqrt{\frac{\log n}{n}}, \label{eq:ind-proxy}\\[2pt]
\max_{0\le m\le n}\norm{\vs^{t}-\vs^{t,(m)}}_2
&\le C_3\sqrt{\frac{\log n}{n}}, \label{eq:ind-close}\\[2pt]
\norm{\vs^{t}-\vs^*}_\infty &\le C_4\sqrt{\frac{\log n}{n}}.
\label{eq:ind-linf}
\end{align}
\end{subequations}
\end{lemma}

\begin{proof}
We proceed by induction on $t$. The constants are fixed at the outset, in the
acyclic order $C_1\to C_2\to C_3\to C_4$ recorded in the closing paragraph;
they depend only on the stated quantities and not on $n_0$, so $n_0$ may then
be chosen large enough for the finitely many $o(1)$ absorptions below (in
particular, for every invocation of Lemma~\ref{lem:localpd} at the radii
$C_4\sqrt{\log n/n}$, $(C_3+C_4)\sqrt{\log n/n}$ and $\rho_n$, all $o(1)$,
whose defining event is $\gE_{\mathrm{sab}}\supseteq\gE$). Let the induction
hypothesis $\gH(t)$ assert that
\eqref{eq:ind-l2}--\eqref{eq:ind-linf} hold for all $0\le\tau\le t$. Since
both trajectories start at $\vs^*$, $\gH(0)$ holds trivially. Assume
$\gH(t)$ for some $0\le t<T$. Throughout,
$\vs^t,\vs^{t,(m)}\in\vone^\perp$ and the differences
$\vs^t-\vs^*,\ \vs^t-\vs^{t,(m)}\in\vone^\perp$ (gauge preservation). Note
also that $\gH(t)$ activates the LOO score bound: by \eqref{eq:ind-close} and
\eqref{eq:ind-linf},
$\norm{\vs^{\tau,(m)}-\vs^*}_\infty\le(C_3+C_4)\sqrt{\log n/n}\le\frac12$
for all $\tau\le t$ and $n\ge n_0$, so on $\gE_{\mathrm{loo}}$ the first
alternative of the dichotomy is in force and
\begin{equation}\label{eq:active-loo}
\norm{\mD^{(m)}_{\vgamma,\vbeta}\big(\vs^{\tau,(m)}\big)}_2
\le 2K\sqrt{n\log n},
\qquad 0\le m\le n,\ 0\le\tau\le t.
\end{equation}

\medskip
\noindent\emph{Step 1: $\norm{\vs^{t+1}-\vs^*}_2\le C_1\sqrt{\log n}$.}
By Algorithm~\ref{alg:gd},
\[
\vs^{t+1}-\vs^*
=\vs^{t}-\vs^*-\eta\mG_{\vgamma,\vbeta}(\vs^*)
-\eta\big(\mG_{\vgamma,\vbeta}(\vs^{t})-\mG_{\vgamma,\vbeta}(\vs^*)\big).
\]
Writing $\bar\mJ:=\int_0^1\mJ(\vs^*+\tau(\vs^{t}-\vs^*))\,d\tau$, the
fundamental theorem of calculus gives
$\mG_{\vgamma,\vbeta}(\vs^{t})-\mG_{\vgamma,\vbeta}(\vs^*)
=\bar\mJ(\vs^{t}-\vs^*)$, so
\[
\vs^{t+1}-\vs^*=(\mI-\eta\bar\mJ)(\vs^{t}-\vs^*)
-\eta\mG_{\vgamma,\vbeta}(\vs^*).
\]
By \eqref{eq:ind-linf} the segment $[\vs^*,\vs^t]$ lies in
$\overline{\gB_\infty(\vs^*,C_4\sqrt{\log n/n})}$, hence in the working
region for $n\ge n_0$; Lemma~\ref{lem:curv-ceiling} gives
$\eta\le1/\lammax(\bar\mJ)$ and Lemma~\ref{lem:localpd} gives
$\lamperp(\bar\mJ)\ge\frac12 nb_0$. Since $\vs^t-\vs^*\in\vone^\perp$,
\eqref{eq:contraction-norm} gives
\[
\opnorm{\mI-\eta\bar\mJ}_{2,\perp}
=1-\eta\,\lamperp(\bar\mJ)\le 1-\frac12\eta nb_0.
\]
Hence, with \eqref{eq:ind-l2},
\[
\norm{\vs^{t+1}-\vs^*}_2
\le\big(1-\frac12\eta nb_0\big)C_1\sqrt{\log n}
+\eta\,\norm{\mG_{\vgamma,\vbeta}(\vs^*)}_2.
\]
For the residual, by the fundamental theorem of calculus in
$(\vgamma,\vbeta)$,
\[
\mG_{\vgamma,\vbeta}(\vs^*)=\mG_{\vgamma^*,\vbeta^*}(\vs^*)
+\Big(\int_0^1\frac{\partial\mG(\vs^*,\cdot)}{\partial(\vgamma,\vbeta)}
\big((\vgamma^*,\vbeta^*)+\tau((\vgamma,\vbeta)-(\vgamma^*,\vbeta^*))\big)
d\tau\Big)\big((\vgamma,\vbeta)-(\vgamma^*,\vbeta^*)\big),
\]
so
\[
\norm{\mG_{\vgamma,\vbeta}(\vs^*)}_2
\le\norm{\mG_{\vgamma^*,\vbeta^*}(\vs^*)}_2
+\sup_{(\vgamma',\vbeta')}
\opnorm{\frac{\partial\mG(\vs^*,\cdot)}{\partial(\vgamma,\vbeta)}
(\vgamma',\vbeta')}_2\,
\norm{(\vgamma,\vbeta)-(\vgamma^*,\vbeta^*)}_2.
\]
On $\gE_{\mathrm{score}}$,
$\norm{\mG_{\vgamma^*,\vbeta^*}(\vs^*)}_\infty\le C_\psi\sqrt{n\log n}$,
hence
$\norm{\mG_{\vgamma^*,\vbeta^*}(\vs^*)}_2
\le\sqrt{n+1}\,C_\psi\sqrt{n\log n}\le\sqrt2\,C_\psi\,n\sqrt{\log n}$. By
Lemma~\ref{lem:cross-jac} (applied at $\vs^*$, which lies in the working
region, uniformly over all $(\vgamma',\vbeta')$) the supremum is
$\le C_M n^{3/2}$, and
$\norm{(\vgamma,\vbeta)-(\vgamma^*,\vbeta^*)}_2
\le\norm{\vgamma-\vgamma^*}_2+\norm{\vbeta-\vbeta^*}_2
\le 2C_{\gamma\beta}\sqrt{\log n/n}$. Therefore
\begin{equation}\label{eq:residual-score}
\norm{\mG_{\vgamma,\vbeta}(\vs^*)}_2\le C_6\,n\sqrt{\log n},
\qquad C_6:=\sqrt2\,C_\psi+2C_M C_{\gamma\beta},
\end{equation}
and, with $\eta=(C_{\max}n)^{-1}$,
\[
\norm{\vs^{t+1}-\vs^*}_2
\le\big(1-\frac12\eta nb_0\big)C_1\sqrt{\log n}+\eta\,C_6\,n\sqrt{\log n}.
\]
This is $\le C_1\sqrt{\log n}$ provided
\[
C_1\ \ge\ \frac{2C_6}{b_0}.
\]

\medskip
\noindent\emph{Step 2:
$\max_{0\le m\le n}\big|\,s^{t+1,(m)}_m-s^*_m\,\big|
\le C_2\sqrt{\log n/n}$.}
Fix $m\in\{0,\dots,n\}$. By Algorithm~\ref{alg:gd_loo}, the $m$-th coordinate
updates as
$s^{t+1,(m)}_m=s^{t,(m)}_m-\eta\,
[\mG^{(m)}_{\vgamma,\vbeta}(\vs^{t,(m)})]_m$, where the only terms of the LOO
score involving coordinate $m$ are
\[
[\mG^{(m)}_{\vgamma,\vbeta}(\vs)]_m
=-\sum_{i\ne m}\sum_k\big[(g^*_{mik}-g_{mik})f'(\Delta_{mi})
+g^*_{mik}(d^*_{mik}-d_{mik})\big].
\]
Since $[\mG^{(m)}_{\vgamma^*,\vbeta^*}(\vs^*)]_m=0$,
\[
s^{t+1,(m)}_m-s^*_m
=\big(s^{t,(m)}_m-s^*_m\big)
-\eta\underbrace{\big[\mG^{(m)}_{\vgamma,\vbeta}(\vs^{t,(m)})
-\mG^{(m)}_{\vgamma,\vbeta}(\vs^*)\big]_m}_{=:S}
-\eta\underbrace{\big[\mG^{(m)}_{\vgamma,\vbeta}(\vs^*)
-\mG^{(m)}_{\vgamma^*,\vbeta^*}(\vs^*)\big]_m}_{=:Z}.
\]
\emph{Control of $S$.} By the mean value theorem applied to the scalar map
$[\mG^{(m)}_{\vgamma,\vbeta}]_m$, there is $\bar\vs$ on
$[\vs^*,\vs^{t,(m)}]$ with
$S=\big(\nabla_{\vs}[\mG^{(m)}_{\vgamma,\vbeta}]_m(\bar\vs)\big)\trans
(\vs^{t,(m)}-\vs^*)$. Since
$\mG^{(m)}_{\vgamma,\vbeta}=\nabla_{\vs}(-\ell^{(m)})$, one has
$\nabla_{\vs}[\mG^{(m)}_{\vgamma,\vbeta}]_m=(\mJ^{(m)})_{m\cdot}$, the $m$-th
row of \eqref{eq:hessian_loo}. Hence
\[
S=(\mJ^{(m)})_{mm}(\bar\vs)\,\big(s^{t,(m)}_m-s^*_m\big)
   +\sum_{i\ne m}(\mJ^{(m)})_{mi}(\bar\vs)\,\big(s^{t,(m)}_i-s^*_i\big).
\]
At the \emph{true} parameters the saboteur in
\eqref{eq:hessian_loo_weights} vanishes, hence
$W^{(m)}_{imk}(\vs^*,\vgamma^*,\vbeta^*)=\bar w_{imk}\ge b_0$
(Lemma~\ref{lem:floor} and Remark~\ref{rem:parity}) and, summing over the
$\ge n$ comparisons meeting $m$,
\[
\bigl(\mJ^{(m)}\bigr)_{mm}(\vs^*,\vgamma^*,\vbeta^*) \ge nb_0.
\]
At the working pair $(\vgamma,\vbeta)\neq(\vgamma^*,\vbeta^*)$ the saboteur
does \emph{not} vanish at $\vs^*$, so we instead transfer the floor by
perturbation. By \eqref{eq:ind-close} and \eqref{eq:ind-linf} the segment
point $\bar\vs$ lies in
$\overline{\gB_\infty\bigl(\vs^*,(C_3+C_4)\sqrt{\log n/n}\bigr)}$ (the ball
being convex), hence in the working region for $n\ge n_0$. Since any diagonal
entry of a symmetric matrix $\mA$ satisfies
$\abs{A_{mm}}\le\opnorm{\mA}_\infty$, the leave-one-out variants of
Lemma~\ref{lem:lip} and Lemma~\ref{lem:lip-zeta}, applied at $\bar\vs$ with
$\norm{\vgamma-\vgamma^*}_2,\norm{\vbeta-\vbeta^*}_2
\le C_{\gamma\beta}\sqrt{\log n/n}$, give
\begin{align*}
\Bigl| \bigl(\mJ^{(m)}\bigr)_{mm}(\bar\vs,\vgamma,\vbeta)
      &- \bigl(\mJ^{(m)}\bigr)_{mm}(\vs^*,\vgamma^*,\vbeta^*) \Bigr|\\
&\le 4L_f m^\ast n\, \norm{\bar\vs-\vs^*}_\infty
      + L_\zeta n \bigl( \norm{\vgamma-\vgamma^*}_2
      + \norm{\vbeta-\vbeta^*}_2 \bigr)\\
&\le \bigl[\, 4L_f m^\ast(C_3+C_4) + 2L_\zeta C_{\gamma\beta} \,\bigr]
        \sqrt{n\log n} = o(n),
\end{align*}
so that
\[
\bigl(\mJ^{(m)}\bigr)_{mm}(\bar\vs,\vgamma,\vbeta) \ge n b_0 - o(n)
\;\ge\; \frac12\, n b_0
\qquad \text{for } n \ge n_0.
\]
For the off-diagonal part,
$\abs{(\mJ^{(m)})_{mi}(\bar\vs)}=\abs{W^{(m)}_{mi}(\bar\vs)}
\le m^\ast W_{\max}$ for each $i\ne m$, so
\[
\norm{(\mJ^{(m)})_{m,-m}(\bar\vs)}_2
:=\Big(\sum_{i\ne m}\abs{(\mJ^{(m)})_{mi}(\bar\vs)}^2\Big)^{1/2}
\le \sqrt{n}\,m^\ast W_{\max}=\frac{C_{\max}}{2}\sqrt{n}.
\]
By the triangle inequality with \eqref{eq:ind-l2} and \eqref{eq:ind-close},
\[
\big\|\vs^{t,(m)}-\vs^*\big\|_2
\le\norm{\vs^{t,(m)}-\vs^t}_2+\norm{\vs^t-\vs^*}_2
\le C_3\sqrt{\frac{\log n}{n}}+C_1\sqrt{\log n}
=(C_1+o(1))\sqrt{\log n}.
\]
Hence, by Cauchy--Schwarz,
\[
\Big|\sum_{i\ne m}(\mJ^{(m)})_{mi}(\bar\vs)\,(s^{t,(m)}_i-s^*_i)\Big|
\le\norm{(\mJ^{(m)})_{m,-m}(\bar\vs)}_2\,\big\|\vs^{t,(m)}-\vs^*\big\|_2
\le\frac{C_{\max}}{2}(C_1+o(1))\sqrt{n\log n}.
\]
\emph{Control of $Z$.} By the mean value theorem in $(\vgamma,\vbeta)$ there
is $(\vgamma',\vbeta')$ on $[(\vgamma^*,\vbeta^*),(\vgamma,\vbeta)]$ with
$\abs{Z}\le\norm{\partial_{(\vgamma,\vbeta)}
[\mG^{(m)}]_m(\vgamma',\vbeta')}_2\,
\norm{(\vgamma,\vbeta)-(\vgamma^*,\vbeta^*)}_2$. The per-coordinate Jacobian
rows (evaluated at $\vs^*$, in the working region) are
\[
\partial_{\vgamma}[\mG^{(m)}]_m
=-\sum_{i\ne m,k}w_{\eta,mik}f'(\Delta^*_{mi})\mY_{mik}\trans,
\qquad
\partial_{\vbeta}[\mG^{(m)}]_m
=-\sum_{i\ne m,k}g^*_{mik}w_{\theta,mik}\mZ_{mik}\trans,
\]
each of the $\le m^\ast n$ summands bounded in $\ell_2$ norm by
$\frac14 F_1\kappa$ (resp.\ $\frac14\kappa$), so
$\norm{\partial_{(\vgamma,\vbeta)}[\mG^{(m)}]_m}_2
\le\frac14(F_1+1)\kappa m^\ast n=:C'n$. With
$\norm{(\vgamma,\vbeta)-(\vgamma^*,\vbeta^*)}_2
\le 2C_{\gamma\beta}\sqrt{\log n/n}$,
\[
\abs{Z}\ \le\ 2C' C_{\gamma\beta}\,n\sqrt{\frac{\log n}{n}}
=2C' C_{\gamma\beta}\sqrt{n\log n}.
\]
Note also the ceiling:
$(\mJ^{(m)})_{mm}(\bar\vs)=\sum_{i\ne m,k}W^{(m)}_{imk}(\bar\vs)
\le m^\ast n\,W_{\max}=\frac12 C_{\max}n$, so
$\eta\,(\mJ^{(m)})_{mm}(\bar\vs)\le\frac12<1$; combined with the floor
$(\mJ^{(m)})_{mm}(\bar\vs)\ge\frac12 nb_0$, the coefficient of the leading
term below satisfies
$0<1-\eta(\mJ^{(m)})_{mm}(\bar\vs)\le 1-\frac12\eta nb_0$. Collecting the
contributions,
\[
\big|s^{t+1,(m)}_m-s^*_m\big|
\le\big(1-\frac12\eta nb_0\big)\big|s^{t,(m)}_m-s^*_m\big|
   +\frac{\eta C_{\max}}{2}(C_1+o(1))\sqrt{n\log n}
   +\eta\,2C' C_{\gamma\beta}\sqrt{n\log n}.
\]
By \eqref{eq:ind-proxy} and $\eta=(C_{\max}n)^{-1}$, and using
$C_1+o(1)\le C_1+1$ for $n\ge n_0$, the right-hand side is
$\le C_2\sqrt{\log n/n}$ provided
\[
C_2 \ge \frac{C_{\max}(C_1+1)+4C' C_{\gamma\beta}}{b_0}.
\]

\medskip
\noindent\emph{Step 3:
$\max_{0\le m\le n}\norm{\vs^{t+1}-\vs^{t+1,(m)}}_2\le C_3\sqrt{\log n/n}$.}
Writing the two updates and inserting
$\pm\mG_{\vgamma,\vbeta}(\vs^{t,(m)})$,
\[
\vs^{t+1}-\vs^{t+1,(m)}
=\big(\mI-\eta\bar\mJ\big)(\vs^t-\vs^{t,(m)})
-\eta\big(\mG_{\vgamma,\vbeta}(\vs^{t,(m)})
-\mG^{(m)}_{\vgamma,\vbeta}(\vs^{t,(m)})\big),
\]
where now
$\bar\mJ:=\int_0^1\mJ(\vs^{t,(m)}+\tau(\vs^t-\vs^{t,(m)}))\,d\tau$. By
\eqref{eq:ind-close} and \eqref{eq:ind-linf} the whole segment
$[\vs^{t,(m)},\vs^t]$ lies in
$\overline{\gB_\infty(\vs^*,(C_3+C_4)\sqrt{\log n/n})}$, so
Lemmas~\ref{lem:curv-ceiling} and~\ref{lem:localpd} apply to $\bar\mJ$ and,
since $\vs^t-\vs^{t,(m)}\in\vone^\perp$, \eqref{eq:contraction-norm} gives
$\opnorm{\mI-\eta\bar\mJ}_{2,\perp}=1-\eta\lamperp(\bar\mJ)
\le 1-\frac12\eta nb_0$. For the second term, the identity
\eqref{eq:loo-identity} and the activated LOO bound \eqref{eq:active-loo}
give
$\norm{\mG_{\vgamma,\vbeta}(\vs^{t,(m)})
-\mG^{(m)}_{\vgamma,\vbeta}(\vs^{t,(m)})}_2
=\norm{\mD^{(m)}_{\vgamma,\vbeta}(\vs^{t,(m)})}_2\le 2K\sqrt{n\log n}$ on
$\gE_{\mathrm{loo}}$. Hence
\[
\norm{\vs^{t+1}-\vs^{t+1,(m)}}_2
\le\big(1-\frac12\eta nb_0\big)\norm{\vs^t-\vs^{t,(m)}}_2
+2\eta K\sqrt{n\log n}.
\]
With $\eta=(C_{\max}n)^{-1}$, $\frac12\eta nb_0=\frac{b_0}{2C_{\max}}$ and
$2\eta K\sqrt{n\log n}=\frac{2K}{C_{\max}}\sqrt{\log n/n}$; using
\eqref{eq:ind-close},
\[
\norm{\vs^{t+1}-\vs^{t+1,(m)}}_2
\le\Big(\big(1-\frac{b_0}{2C_{\max}}\big)C_3
+\frac{2K}{C_{\max}}\Big)\sqrt{\frac{\log n}{n}}.
\]
The bracket is $\le C_3$ iff
$\frac{2K}{C_{\max}}\le\frac{b_0}{2C_{\max}}C_3$, i.e.\
$C_3\ge\frac{4K}{b_0}$.

\medskip
\noindent\emph{Step 4: $\norm{\vs^{t+1}-\vs^*}_\infty\le C_4\sqrt{\log n/n}$.}
For each $m$,
\begin{align*}
\big|s^{t+1}_m-s^*_m\big|
&\le\big|s^{t+1}_m-s^{t+1,(m)}_m\big|+\big|s^{t+1,(m)}_m-s^*_m\big|\\
&\le\max_{0\le m\le n}\norm{\vs^{t+1}-\vs^{t+1,(m)}}_2
+\max_{0\le m\le n}\big|s^{t+1,(m)}_m-s^*_m\big|,
\end{align*}
so by Steps 2--3,
$\norm{\vs^{t+1}-\vs^*}_\infty\le(C_2+C_3)\sqrt{\log n/n}
\le C_4\sqrt{\log n/n}$ provided $C_4\ge C_2+C_3$.

\medskip
\noindent\emph{Closing the constants and the probability.}
Choose, in order, $C_1\ge2C_6/b_0$ with
$C_6=\sqrt2\,C_\psi+2C_MC_{\gamma\beta}$, then
$C_2\ge(C_{\max}(C_1+1)+4C'C_{\gamma\beta})/b_0$, then $C_3\ge4K/b_0$, then
$C_4\ge C_2+C_3$; this is the acyclic order $C_1\to C_2\to C_3\to C_4$, so
the choices are consistent and depend only on the stated quantities. Each
step preserved \eqref{eq:ind-l2}--\eqref{eq:ind-linf}, completing the
induction. Finally, by Lemma~\ref{lem:score},
$\PP(\gE_{\mathrm{score}})\ge1-O(n^{-c})$; by Lemma~\ref{lem:saboteur},
$\PP(\gE_{\mathrm{sab}})\ge1-O(n^{-c})$; and for
$\gE_{\mathrm{loo}}(\vgamma,\vbeta)$, each iterate $\vs^{t,(m)}$ is
$\Fm$-measurable (gauge plus Appendix~\ref{app:loo}), and
$\norm{\vs^{t,(m)}-\vs^*}_\infty\le\frac12$ implies
$\norm{\vs^{t,(m)}}_\infty\le C_s+\frac12$ (A\ref{assump:compact}), so
Lemma~\ref{lem:loo-fluct} bounds each term of the complement,
\[
\PP\Big(\Big\{\norm{\mD^{(m)}_{\vgamma,\vbeta}(\vs^{t,(m)})}_2
>2K\sqrt{n\log n}\Big\}\cap
\Big\{\norm{\vs^{t,(m)}-\vs^*}_\infty\le\frac12\Big\}\Big)=O(n^{-c}),
\]
a fortiori at threshold $2K\ge K$; a union bound over the $(n+1)(T+1)$ pairs
$(m,t)$ gives
$\PP(\gE_{\mathrm{loo}}(\vgamma,\vbeta))\ge1-O((T+1)n^{1-c})$.
\end{proof}

Lemma~\ref{lem:coupled-induction} holds at a fixed $(\vgamma,\vbeta)$. To
obtain Lemma~\ref{lem:stage1} we run Algorithm~\ref{alg:gd} for
$T=\Theta(\log n)$ steps and feed the coupled bounds into a degree-theoretic
existence argument. This requires the LOO event to hold uniformly over all
$(\vgamma,\vbeta)$ in the structural ball, which is the next lemma.

\begin{lemma}[Uniformity of $\gE_{\mathrm{loo}}$ over the structural ball]
\label{lem:loo-uniform}
Assume A\ref{assump:dense}--A\ref{assump:gate}, and let $T=T(n)$ be any
horizon with $T\le C_T\log n$ for a constant $C_T$. For
$\vzeta=(\vgamma,\vbeta)$ in the balls of
Lemma~\ref{lem:coupled-induction}, let $\{\vs^{t,(m)}(\vzeta)\}_{t\le T}$ be
the LOO trajectories of Algorithm~\ref{alg:gd_loo}, and write
$C_4':=C_3+C_4$ with $C_3,C_4$ the constants of
Lemma~\ref{lem:coupled-induction}. Then for every $c>0$ there is
$n_0=n_0(c)$ such that for all $n\ge n_0$, on a single event
$\widetilde{\gE}_{\mathrm{loo}}$ with
$\PP(\widetilde{\gE}_{\mathrm{loo}})\ge1-O(n^{-c})$, the following hold
simultaneously, with
$\Theta:=\gB(\vgamma^*,\rho_n)\times\gB(\vbeta^*,\rho_n)$:
\begin{enumerate}
\item[(a)] (uniform LOO bound)
$\displaystyle\sup_{\vzeta\in\Theta}\ \max_{0\le m\le n}\ \max_{0\le t\le T}
\norm{\mD^{(m)}_{\vzeta}\big(\vs^{t,(m)}(\vzeta)\big)}_2
\ \le\ 2K\sqrt{n\log n}$;
\item[(b)] (uniform confinement)
$\displaystyle\sup_{\vzeta\in\Theta}\ \max_{0\le m\le n}\ \max_{0\le t\le T}
\norm{\vs^{t,(m)}(\vzeta)-\vs^*}_\infty\ \le\ (C_4'+1)\sqrt{\log n/n}$.
\end{enumerate}
Moreover
$\widetilde{\gE}_{\mathrm{loo}}\subseteq
\gE_{\mathrm{score}}\cap\gE_{\mathrm{sab}}$, and on
$\widetilde{\gE}_{\mathrm{loo}}$ the event $\gE$ of
Lemma~\ref{lem:coupled-induction} holds at every $\vzeta$ in the stated
balls.
\end{lemma}

\begin{proof}
Write $p:=p_{\gamma}+p_{\beta}$ (fixed). Throughout, ``the working region''
means $\{\vs:\norm{\vs}_\infty\le C_s+\frac12\}$, on which the bounds
$F_1,F_2,F_3$ of A\ref{assump:gate} apply to every gap
(Remark~\ref{rem:nbhd}).

\medskip\noindent
\emph{Step A: sensitivity bounds on the working region.}
Fix $m$ and $\vzeta,\vzeta'\in\Theta$.

(i) \emph{Point sensitivity of $\mD^{(m)}$.} For $\vs,\vs'$ in the working
region, each summand $\chi_{imk}$ of \eqref{eq:loo-identity} is, at fixed
$\vzeta$, a function of the gap $\Delta_{im}$ alone, with
\[
\Big|\frac{\partial\chi_{imk}}{\partial\Delta_{im}}\Big|
=\abs{(c_{imk}-g^*_{imk})\big(f''(\Delta_{im})-w_{\theta,imk}\big)}
\le F_2+\frac14=:L_\chi
\]
on the region; hence
$\abs{\chi_{imk}(\vs)-\chi_{imk}(\vs')}
\le L_\chi\abs{\Delta_{im}(\vs)-\Delta_{im}(\vs')}
\le2L_\chi\norm{\vs-\vs'}_\infty$. The difference
$\mD^{(m)}_{\vzeta}(\vs)-\mD^{(m)}_{\vzeta}(\vs')$ has $m$-th coordinate a
sum of $\le m^\ast n$ increments and, for each $i\ne m$, an $i$-th
coordinate consisting of one pair's $\le m^\ast$ increments; summing in
$\ell_2$ and using $\norm{\cdot}_\infty\le\norm{\cdot}_2$,
\[
\norm{\mD^{(m)}_{\vzeta}(\vs)-\mD^{(m)}_{\vzeta}(\vs')}_2
\ \le\ 2L_\chi m^\ast\sqrt{n^2+n}\,\norm{\vs-\vs'}_\infty
\ \le\ C_{D}\,n\,\norm{\vs-\vs'}_2
\]
with $C_{D}:=4L_\chi m^\ast$.

(ii) \emph{Coefficient sensitivity of $\mD^{(m)}$.} At fixed $\vs$ in the
working region, $\mD^{(m)}_{\vzeta}$ depends on $\vzeta$ only through
$d_{imk}=\sigma(\Delta_{im}+\mZ_{imk}\trans\vbeta)$ (no $\vgamma$-dependence,
by \eqref{eq:loo-identity}); as $\sigma$ is $\frac14$-Lipschitz and
$\norm{\mZ_{imk}}_2\le\kappa$, each summand moves by at most
$\abs{c_{imk}-g^*_{imk}}\cdot\frac14\kappa\norm{\vbeta-\vbeta'}_2
\le\frac14\kappa\norm{\vzeta-\vzeta'}_2$, and the same $\ell_2$ assembly as
in (i) gives
\[
\norm{\mD^{(m)}_{\vzeta}(\vs)-\mD^{(m)}_{\vzeta'}(\vs)}_2
\ \le\ C_{\mathrm{coef}}\,n\,\norm{\vzeta-\vzeta'}_2,
\qquad C_{\mathrm{coef}}:=\kappa m^\ast.
\]

(iii) \emph{One-step map.} Writing one LOO step as
$\vs^{t+1,(m)}=\Phi^{(m)}_{\vzeta}(\vs^{t,(m)})$ with
$\Phi^{(m)}_{\vzeta}(\vs)=\vs-\eta\,\mG^{(m)}_{\vzeta}(\vs)$: on the working
region, $\partial_{\vs}\Phi^{(m)}_{\vzeta}=\mI-\eta\mJ^{(m)}$ has
$\opnorm{\cdot}_2\le2$ (by Lemma~\ref{lem:curv-ceiling} and
$\eta=(C_{\max}n)^{-1}$, the spectrum of the symmetric matrix
$\eta\mJ^{(m)}$ lies in $[-1,1]$, hence that of $\mI-\eta\mJ^{(m)}$ in
$[0,2]$), and
$\opnorm{\partial_{\vzeta}\Phi^{(m)}_{\vzeta}}_2
=\eta\,\opnorm{\partial_{\vzeta}\mG^{(m)}_{\vzeta}}_2\le\eta\,C_M n^{3/2}
=C_M C_{\max}^{-1}\sqrt n$ (Lemma~\ref{lem:cross-jac}, leave-one-out
variant). All three bounds require the argument(s) to lie in the working
region; establishing that they do, along the trajectories at
\emph{arbitrary} $\vzeta\in\Theta$, is the purpose of Step~C.

\medskip\noindent
\emph{Step B: net, good event, and behavior at net points.}
Since $T\le C_T\log n$, we have $2^{\,T+1}\le n^{\kappa_T}$ with
$\kappa_T:=2C_T$ for $n\ge n_0$. Let $\gN\subset\Theta$ be a $\delta$-net in
$\ell_2$ with
\[
\delta:= n^{-(\kappa_T+2)},
\qquad\text{so}\qquad
\abs{\gN}\le\Big(\frac{3\rho_n}{\delta}\Big)^{p}=n^{O(1)},
\quad \log\abs{\gN}=O(\log n).
\]
Define
\begin{multline*}
\widetilde{\gE}_{\mathrm{loo}} :=
\gE_{\mathrm{score}} \cap \gE_{\mathrm{sab}}\ \cap \\
\bigcap_{\vzeta_j\in\gN} \bigcap_{m=0}^{n} \bigcap_{t=0}^{T}
\Big(\Big\{\norm{\mD^{(m)}_{\vzeta_j}\big(\vs^{t,(m)}(\vzeta_j)\big)}_2
\le K\sqrt{n\log n}\Big\}
\cup\Big\{\norm{\vs^{t,(m)}(\vzeta_j)-\vs^*}_\infty>\frac12\Big\}
\Big).
\end{multline*}
Each $\vzeta_j\in\gN$ is deterministic, so $\vs^{t,(m)}(\vzeta_j)$ is
$\Fm$-measurable (Appendix~\ref{app:loo}), and Lemma~\ref{lem:loo-fluct}
bounds the probability of each complementary event
$\{\norm{\mD^{(m)}_{\vzeta_j}}_2>K\sqrt{n\log n}\}\cap
\{\norm{\vs^{t,(m)}(\vzeta_j)-\vs^*}_\infty\le\frac12\}$ by $O(n^{-c'})$:
running the lemma at grade $c'$ and union-bounding over
$\gN\times\{0,\dots,n\}\times\{0,\dots,T\}$, a set of cardinality
$n^{O(1)}$ (with exponent depending only on $p$, $\kappa_T$), gives
$\PP(\widetilde{\gE}_{\mathrm{loo}})\ge1-O(n^{-c})$ for $c'$ large enough
(depending on $c$, $p$, $C_T$), after intersecting with the events of
Lemmas~\ref{lem:score} and~\ref{lem:saboteur}.

On $\widetilde{\gE}_{\mathrm{loo}}$, for every net point $\vzeta_j$ the
event
$\gE=\gE_{\mathrm{score}}\cap\gE_{\mathrm{sab}}\cap\gE_{\mathrm{loo}}(\vzeta_j)$
of Lemma~\ref{lem:coupled-induction} holds: the net dichotomy at threshold
$K$ implies the defining dichotomy at threshold $2K$ a fortiori (same escape
set, weaker bound). Hence \eqref{eq:ind-l2}--\eqref{eq:ind-linf} are in
force at every $\vzeta_j$ for all $t\le T$; combining \eqref{eq:ind-close}
(with $\norm{\cdot}_\infty\le\norm{\cdot}_2$) and \eqref{eq:ind-linf} yields
the \emph{net-point confinement of the LOO trajectories},
\begin{equation}\label{eq:net-confinement}
\norm{\vs^{t,(m)}(\vzeta_j)-\vs^*}_\infty \le C_4'\sqrt{\log n/n},
\qquad \vzeta_j\in\gN,\ 0\le m\le n,\ 0\le t\le T,
\end{equation}
with $C_4'=C_3+C_4$ as in the statement.

\medskip\noindent
\emph{Step C: confinement induction off the net.}
For $\vzeta\in\Theta$ fix a nearest net point
$\vzeta_j=\vzeta_j(\vzeta)$, $\norm{\vzeta-\vzeta_j}_2\le\delta$. Set
$\Lambda_t:=(2^{\,t}-1)\,C_M C_{\max}^{-1}\sqrt n$, and note
\begin{equation}\label{eq:slack-choice}
\Lambda_T\delta \le 2^{T}C_MC_{\max}^{-1}\sqrt n\,n^{-(\kappa_T+2)}
 \le C_MC_{\max}^{-1}n^{-3/2} \le \sqrt{\log n/n}
\qquad(n\ge n_0).
\end{equation}
We prove by induction on $t$ the hypothesis
\[
H(t):\qquad
\norm{\vs^{\tau,(m)}(\vzeta)-\vs^{\tau,(m)}(\vzeta_j)}_2 \le \Lambda_\tau\delta
\quad\text{for all }\vzeta\in\Theta,\ 0\le m\le n,\ 0\le\tau\le t.
\]
$H(t)$ has the following consequence, by \eqref{eq:net-confinement},
\eqref{eq:slack-choice} and $\norm{\cdot}_\infty\le\norm{\cdot}_2$: for all
$\tau\le t$,
\begin{equation}\label{eq:uniform-confinement}
\norm{\vs^{\tau,(m)}(\vzeta)-\vs^*}_\infty
\le C_4'\sqrt{\log n/n}+\Lambda_\tau\delta
 \le (C_4'+1)\sqrt{\log n/n},
\end{equation}
so that every point $\vs^{\tau,(m)}(\vzeta)$, every net-point iterate, and
(by convexity of the $\ell_\infty$ ball) every segment between them lies in
the working region for $n\ge n_0$.

$H(0)$ holds trivially, both trajectories starting at $\vs^*$. Assume
$H(t)$, $t<T$. Decompose one step as
\begin{multline*}
\vs^{t+1,(m)}(\vzeta)-\vs^{t+1,(m)}(\vzeta_j)
=\underbrace{\Phi^{(m)}_{\vzeta}\big(\vs^{t,(m)}(\vzeta)\big)
             -\Phi^{(m)}_{\vzeta}\big(\vs^{t,(m)}(\vzeta_j)\big)}_{=:P}\\
+\underbrace{\Phi^{(m)}_{\vzeta}\big(\vs^{t,(m)}(\vzeta_j)\big)
             -\Phi^{(m)}_{\vzeta_j}\big(\vs^{t,(m)}(\vzeta_j)\big)}_{=:Q}.
\end{multline*}
For $P$: by \eqref{eq:uniform-confinement} the segment between the two
arguments lies in the working region, so Step~A(iii) and the mean value
theorem give
$\norm{P}_2\le2\,\norm{\vs^{t,(m)}(\vzeta)-\vs^{t,(m)}(\vzeta_j)}_2
\le2\Lambda_t\delta$.
For $Q$: the argument $\vs^{t,(m)}(\vzeta_j)$ lies in the working region by
\eqref{eq:net-confinement}, and the segment $[\vzeta_j,\vzeta]$ lies in the
convex set $\Theta$, so Step~A(iii) gives
$\norm{Q}_2\le C_MC_{\max}^{-1}\sqrt n\,\delta$. Hence
\[
\norm{\vs^{t+1,(m)}(\vzeta)-\vs^{t+1,(m)}(\vzeta_j)}_2
\ \le\ \big(2\Lambda_t+C_MC_{\max}^{-1}\sqrt n\big)\delta
\ =\ \Lambda_{t+1}\,\delta,
\]
by the recursion $\Lambda_{t+1}=2\Lambda_t+C_MC_{\max}^{-1}\sqrt n$. This
proves $H(t+1)$, and the induction delivers $H(T)$; consequence
\eqref{eq:uniform-confinement} at $\tau\le T$ is claim~(b).

\medskip\noindent
\emph{Step D: off-net extension of the LOO bound.}
Fix $\vzeta\in\Theta$, $m$, $t$, and $\vzeta_j=\vzeta_j(\vzeta)$. By
\eqref{eq:net-confinement} the net-point iterate is confined,
$\norm{\vs^{t,(m)}(\vzeta_j)-\vs^*}_\infty\le C_4'\sqrt{\log n/n}
\le\frac12$ for $n\ge n_0$, so on $\widetilde{\gE}_{\mathrm{loo}}$ the
first alternative of the net dichotomy is active:
$\norm{\mD^{(m)}_{\vzeta_j}(\vs^{t,(m)}(\vzeta_j))}_2\le K\sqrt{n\log n}$.
All remaining points lie in the working region by
\eqref{eq:net-confinement} and \eqref{eq:uniform-confinement}, so
Steps~A(i)--(ii) apply:
\begin{multline*}
\norm{\mD^{(m)}_{\vzeta}\big(\vs^{t,(m)}(\vzeta)\big)}_2
\ \le\ \norm{\mD^{(m)}_{\vzeta_j}\big(\vs^{t,(m)}(\vzeta_j)\big)}_2\\
+C_{D}\,n\,\norm{\vs^{t,(m)}(\vzeta)-\vs^{t,(m)}(\vzeta_j)}_2
+C_{\mathrm{coef}}\,n\,\norm{\vzeta-\vzeta_j}_2.
\end{multline*}
The first term is $\le K\sqrt{n\log n}$; by $H(T)$ and
\eqref{eq:slack-choice} the second is
$\le C_{D}\,n\cdot C_MC_{\max}^{-1}n^{-3/2}=O(n^{-1/2})$; the third is
$\le C_{\mathrm{coef}}\,n^{1-(\kappa_T+2)}=O(n^{-1})$. For $n\ge n_0$ the
last two are together $\le K\sqrt{n\log n}$, whence the total is
$\le2K\sqrt{n\log n}$. Taking suprema over $\vzeta\in\Theta$ and maxima over
$m,t$ gives claim~(a). The final sentence of the statement restates that
$\gE$ of Lemma~\ref{lem:coupled-induction} holds at every $\vzeta$: its
first two constituents hold by construction, and
$\gE_{\mathrm{loo}}(\vzeta)$ holds because the actual bound of claim~(a)
implies the dichotomy at threshold $2K$ a fortiori.
\end{proof}

\begin{proof}[Proof of Lemma~\ref{lem:stage1}]
Let $D\ge1$ be a constant and set
\[
\varrho:=1-\frac{b_0}{2C_{\max}}\in(0,1),
\qquad
T:=\Big\lceil\frac{(D+2)\log n}{\log(1/\varrho)}\Big\rceil=\Theta(\log n),
\]
so that $T+1\le C_T\log n$ with $C_T:=(D+3)/\log(1/\varrho)$ for $n\ge n_0$.
Set
$\widetilde{\gE}:=\gE_{\mathrm{score}}\cap\gE_{\mathrm{sab}}\cap
\widetilde{\gE}_{\mathrm{loo}}$, with $\widetilde{\gE}_{\mathrm{loo}}$ from
Lemma~\ref{lem:loo-uniform} at horizon $T+1$. As
$\gE_{\mathrm{score}},\gE_{\mathrm{sab}}$ are evaluated at
$(\vgamma^*,\vbeta^*)$, the union bound gives
$\PP(\widetilde{\gE})\ge1-O(n^{-c})$.

Fix any $(\vgamma,\vbeta)$ with
$\max\{\norm{\vgamma-\vgamma^*}_2,\norm{\vbeta-\vbeta^*}_2\}\le\rho_n$. By
the final assertion of Lemma~\ref{lem:loo-uniform},
$\widetilde{\gE}$ implies the event $\gE$ of
Lemma~\ref{lem:coupled-induction} \emph{for this} $(\vgamma,\vbeta)$; run
Algorithm~\ref{alg:gd} for $T+1$ steps, applying
Lemma~\ref{lem:coupled-induction} with horizon $T+1$. All statements below
hold on $\gE$; the four bounds \eqref{eq:ind-l2}--\eqref{eq:ind-linf} are in
force for $0\le t\le T+1$.

\medskip
\noindent\emph{Step 1 (near-stationarity).}
Since $\vs^{t+1}-\vs^t=-\eta\mG_{\vgamma,\vbeta}(\vs^t)$, consecutive
differences satisfy
$\vs^{t+1}-\vs^t=(\mI-\eta\bar\mJ_t)(\vs^t-\vs^{t-1})$ with
$\bar\mJ_t:=\int_0^1\mJ(\vs^{t-1}+\tau(\vs^t-\vs^{t-1}))\,d\tau$. By
\eqref{eq:ind-linf} the segment
$[\vs^{t-1},\vs^t]\subset\overline{\gB_\infty(\vs^*,C_4\sqrt{\log n/n})}$, so
Lemma~\ref{lem:curv-ceiling} gives $\eta\le1/\lammax(\bar\mJ_t)$ and
Lemma~\ref{lem:localpd} gives $\lamperp(\bar\mJ_t)\ge\frac12 nb_0$; with
$\vs^t-\vs^{t-1}\in\vone^\perp$, \eqref{eq:contraction-norm} gives
$\opnorm{\mI-\eta\bar\mJ_t}_{2,\perp}\le\varrho$ for $1\le t\le T+1$.
Iterating from $\vs^0=\vs^*$ and using
$\norm{\mG_{\vgamma,\vbeta}(\vs^*)}_2\le C_6\,n\sqrt{\log n}$ on $\gE$
(display \eqref{eq:residual-score} in the proof of
Lemma~\ref{lem:coupled-induction}, with
$C_6=\sqrt2\,C_\psi+2C_M C_{\gamma\beta}$),
\[
\norm{\vs^1-\vs^0}_2=\eta\norm{\mG_{\vgamma,\vbeta}(\vs^*)}_2
\le\frac{C_6}{C_{\max}}\sqrt{\log n},
\qquad
\norm{\vs^{t+1}-\vs^t}_2\le\varrho^{\,t}\frac{C_6}{C_{\max}}\sqrt{\log n}.
\]
Therefore, using $\vs^{T+1}-\vs^T=-\eta\mG_{\vgamma,\vbeta}(\vs^T)$ and
$\eta=(C_{\max}n)^{-1}$,
\[
\norm{\mG_{\vgamma,\vbeta}(\vs^T)}_2
=\frac1\eta\norm{\vs^{T+1}-\vs^T}_2
\le C_6\,n\sqrt{\log n}\,\varrho^{\,T}\le n^{-D},
\]
the last step by the choice of $T$ (for $n$ large).

\medskip
\noindent\emph{Step 2 (existence, via degree theory on a small ball).}
If $\mG_{\vgamma,\vbeta}(\vs^T)=\vzero$, set
$\widehat\vs_{\vgamma,\vbeta}:=\vs^T$ and pass directly to Step~3.
Otherwise, put
\[
r:=\frac{4\norm{\mG_{\vgamma,\vbeta}(\vs^T)}_2}{nb_0}
\in\Big(0,\ \frac{4n^{-D}}{b_0}\Big],
\qquad
\Omega:=\gB_2(\vs^T,r)\cap(\vs^*+\vone^\perp).
\]
Because $\norm{\vs^T-\vs^*}_\infty\le C_4\sqrt{\log n/n}$ (by
\eqref{eq:ind-linf}) and $r=o(\sqrt{\log n/n})$, the whole ball satisfies
$\overline{\Omega}\subset\gB_\infty(\vs^*,C_5\sqrt{\log n/n})$ for
$n\ge n_0$, since $C_5=C_4+1$. For $\vx\in\partial\Omega$ write
$\vv:=\vx-\vs^T\in\vone^\perp$ (as $\vs^T\in\vone^\perp$), $\norm{\vv}_2=r$,
and $\bar\mJ:=\int_0^1\mJ(\vs^T+\tau\vv)\,d\tau$. Lemma~\ref{lem:localpd}
applies to $\bar\mJ$ (segment inside
$\overline{\gB_\infty(\vs^*,\varepsilon_n)}$), so
\begin{multline*}
\vv\trans\mG_{\vgamma,\vbeta}(\vx)
=\vv\trans\mG_{\vgamma,\vbeta}(\vs^T)+\vv\trans\bar\mJ\,\vv
\ge-r\norm{\mG_{\vgamma,\vbeta}(\vs^T)}_2+\frac12 nb_0 r^2\\
=\frac12 nb_0\,
r\Big(r-\frac{2\norm{\mG_{\vgamma,\vbeta}(\vs^T)}_2}{nb_0}\Big)>0.
\end{multline*}
By the degree-theoretic existence criterion
(Fact~\ref{fact:ortega-rheinboldt}, with anchor $\vs^T$ and
$\mF=\mG_{\vgamma,\vbeta}$ restricted to the affine slice
$\vs^*+\vone^\perp$, identified with $\R^{n}$ and on which
$\mG_{\vgamma,\vbeta}$ takes values in $\vone^\perp$),
$\mG_{\vgamma,\vbeta}$ has a zero
$\widehat\vs_{\vgamma,\vbeta}\in\overline{\Omega}$.

\medskip
\noindent\emph{Step 3 (rate).}
Since $\widehat\vs_{\vgamma,\vbeta}\in\overline{\gB_2(\vs^T,r)}$,
\[
\norm{\widehat\vs_{\vgamma,\vbeta}-\vs^*}_\infty
\le\norm{\vs^T-\vs^*}_\infty+\norm{\widehat\vs_{\vgamma,\vbeta}-\vs^T}_\infty
\le C_4\sqrt{\frac{\log n}{n}}+r
\ \le\ C_5\sqrt{\frac{\log n}{n}}
\]
for $n\ge n_0$, as $r=O(n^{-D})=o(\sqrt{\log n/n})$ for $D\ge1$ and
$C_5=C_4+1$; in particular
$\widehat\vs_{\vgamma,\vbeta}\in\gB_\infty(\vs^*,\varepsilon_n)$.

\medskip
\noindent\emph{Step 4 (uniqueness).}
The set $\overline{\gB_\infty(\vs^*,\varepsilon_n)}\cap(\vs^*+\vone^\perp)$ is
convex, and for distinct $\vx,\vy$ in it, with
$\bar\mJ:=\int_0^1\mJ(\vy+\tau(\vx-\vy))\,d\tau$ and
$\vx-\vy\in\vone^\perp$, Lemma~\ref{lem:localpd} gives
\[
\big(\mG_{\vgamma,\vbeta}(\vx)-\mG_{\vgamma,\vbeta}(\vy)\big)\trans(\vx-\vy)
=(\vx-\vy)\trans\bar\mJ\,(\vx-\vy)
\ge\frac12 nb_0\norm{\vx-\vy}_2^2>0.
\]
Hence $\mG_{\vgamma,\vbeta}$ is strictly monotone there and has at most one
zero; with Step 2, $\widehat\vs_{\vgamma,\vbeta}$ is the unique zero in
$\overline{\gB_\infty(\vs^*,\varepsilon_n)}\cap(\vs^*+\vone^\perp)$.
\end{proof}

\subsection{Stage 2: the profiled likelihood and joint consistency}
\label{app:stage2}

\paragraph{The population information.}
At the truth, the Fisher information of the model is the expected Hessian
$\gI_n:=\E\big[\nabla^2(-\ell)(\vs^*,\vgamma^*,\vbeta^*)\big]$, whose blocks
are obtained from \eqref{eq:Hgg}--\eqref{eq:Hsb} by replacing $c_{ijk}$ with
$g^*_{ijk}$ (the remaining quantities being deterministic at the truth) and
whose merit block is $\mV$ of \eqref{eq:VE}. Explicitly, with
\begin{align*}
\mA^{\popsup}&:=\begin{pmatrix}\mA_{\vgamma\vgamma}(\vs^*,\vgamma^*)&0\\[2pt]
0&\sum_{i<j}\sum_k
g^*_{ijk}w^*_{\theta,ijk}\,\mZ_{ijk}\mZ_{ijk}\trans\end{pmatrix},\\
\mB^{\popsup}&:=\Big[\ \mB_{\vgamma}(\vs^*,\vgamma^*)\ \Big|\
\sum_{i<j}\sum_k
g^*_{ijk}w^*_{\theta,ijk}\,\vomega_{ij}\mZ_{ijk}\trans\ \Big]
\end{align*}
(note $\mA_{\vgamma\vgamma}$ and $\mB_{\vgamma}$ carry no data, hence equal
their population versions at the truth), define the \emph{efficient
information for the structural block} as the Schur complement of the merit
block in $\gI_n$,
\begin{equation}\label{eq:sigma-pop}
\mSigma^{\popsup}\ :=\
\mA^{\popsup}-(\mB^{\popsup})\trans\,\mV^{-1}\,\mB^{\popsup},
\end{equation}
with $\mV^{-1}$ taken on $\vone^\perp$ (legitimate by
Lemma~\ref{lem:Vfloor}).

\begin{assumption}[Efficient information floor]\label{assump:info}
There is a constant $c_\Sigma>0$ such that
$\lammin\big(\mSigma^{\popsup}\big)\ge c_\Sigma\,n^2$ for all $n$ large
enough.
\end{assumption}

\begin{remark}\label{rem:A6-implies}
Since $(\mB^{\popsup})\trans\mV^{-1}\mB^{\popsup}\succeq0$ (the columns of
$\mB^{\popsup}$ lie in $\vone^\perp$, where $\mV\succ0$),
A\ref{assump:info} forces $\lammin(\mA^{\popsup})\ge c_\Sigma n^2$; as
$\mA^{\popsup}$ is block-diagonal with blocks exactly the matrices of
A\ref{assump:gatedesign} and A\ref{assump:dirdesign}, Assumption
A\ref{assump:info} implies both with $c_0=c_\Sigma$. We nevertheless retain
A\ref{assump:gatedesign}--A\ref{assump:dirdesign} as standing conditions for
their interpretability. For asymptotic normality, which is beyond the scope
of this appendix, A\ref{assump:info} would be strengthened to the existence
of the limit $\bar\mSigma:=\lim_n N^{-1}\mSigma^{\popsup}\succ0$,
$N=\binom{n+1}{2}$; only the floor is needed for consistency.
\end{remark}

\begin{lemma}[Information concentration at the truth]\label{lem:info-conc}
Assume A\ref{assump:dense}--A\ref{assump:gate}. For every $c>0$ there is an
absolute constant $C_\Sigma$ and an event $\gE_\Sigma$ with
$\PP(\gE_\Sigma)\ge1-O(n^{-c})$ on which
\[
\opnorm{\mA_{\vbeta\vbeta}(\vs^*,\vbeta^*)-\mA^{\popsup}_{\vbeta\vbeta}}_2
\ \le\ C_\Sigma\,n\sqrt{\log n},
\qquad
\opnorm{\mB_{\vbeta}(\vs^*,\vgamma^*,\vbeta^*)-\mB^{\popsup}_{\vbeta}}_2
\ \le\ C_\Sigma\,n\sqrt{\log n},
\]
where $\mA^{\popsup}_{\vbeta\vbeta}$ and $\mB^{\popsup}_{\vbeta}$ denote the
direction blocks of $\mA^{\popsup}$ and $\mB^{\popsup}$.
\end{lemma}

\begin{proof}
\emph{Direction block of $\mA$.} Write
$\mA_{\vbeta\vbeta}(\vs^*,\vbeta^*)-\mA^{\popsup}_{\vbeta\vbeta}
=\sum_{i<j,k}\mX_{ijk}$ with
$\mX_{ijk}:=(c_{ijk}-g^*_{ijk})\,w^*_{\theta,ijk}\,\mZ_{ijk}\mZ_{ijk}\trans$,
independent, symmetric, mean-zero, of dimension $p_{\beta}$ (fixed). Then
$\opnorm{\mX_{ijk}}_2\le\frac14\kappa^2=:R$ and
$\opnorm{\sum\E\mX_{ijk}^2}_2\le\sum\opnorm{\E\mX_{ijk}^2}_2
\le\frac{1}{16}\kappa^4\,m^\ast n^2=:\nu$, so matrix Bernstein
(Fact~\ref{fact:matrix-bernstein}) with $t=C\,n\sqrt{\log n}$ gives exponent
$\lesssim-\,C^2\log n/(2\cdot\frac{1}{16}\kappa^4m^\ast)$ for $n$ large,
hence probability $O(n^{-(c+1)})$ for $C$ large enough.

\emph{Direction block of $\mB$.} Fix a column index $l\le p_{\beta}$. The
$l$-th column of the difference is
$\sum_{i<j,k}\varsigma_{ijk}\,\vomega_{ij}\in\vone^\perp$ with
$\varsigma_{ijk}:=(c_{ijk}-g^*_{ijk})\,w^*_{\theta,ijk}\,Z_{ijk,l}$,
independent and mean-zero across $(i,j,k)$;
$\norm{\varsigma_{ijk}\vomega_{ij}}_2\le\frac{\sqrt2}{4}\kappa=:R$ and
$\sum\E\norm{\varsigma_{ijk}\vomega_{ij}}_2^2
\le\frac18\kappa^2 m^\ast n^2=:\nu$. Vector Bernstein
(Fact~\ref{fact:vector-bernstein}) in dimension $n+1$ with
$t=C'n\sqrt{\log n}$ gives, as in Lemma~\ref{lem:loo-fluct}, probability
$O(n^{-(c+1)})$ for $C'$ large. A union bound over the $p_{\beta}$ columns
and $\opnorm{\mM}_2\le\sqrt{p_{\beta}}\max_l\norm{\mM_{\cdot l}}_2$ complete
the proof, with $C_\Sigma:=\max\{C,\sqrt{p_{\beta}}\,C'\}$.
\end{proof}

\begin{lemma}[Uniform information floor on the joint neighborhood]
\label{lem:sigma-floor}
Assume A\ref{assump:dense}--A\ref{assump:gate} and A\ref{assump:info}, and
let
$\gK:=\overline{\gB_\infty(\vs^*,\varepsilon_n)}\times\overline{\gB(\vgamma^*,\rho_n)}
\times\overline{\gB(\vbeta^*,\rho_n)}$ with
$\varepsilon_n,\rho_n=O(\sqrt{\log n/n})$. Write, for $(\vs,\vzeta)\in\gK$,
\begin{gather*}
\mSigma(\vs,\vzeta)\ :=\
\mA(\vs,\vzeta)-\mB(\vs,\vzeta)\trans\,\mJ(\vs,\vzeta)^{-1}\,\mB(\vs,\vzeta),\\
\mA:=\begin{pmatrix}\mA_{\vgamma\vgamma}&0\\0&\mA_{\vbeta\vbeta}\end{pmatrix},
\qquad
\mB:=[\,\mB_{\vgamma}\mid\mB_{\vbeta}\,],
\end{gather*}
with $\mJ^{-1}$ taken on $\vone^\perp$. For every $c>0$ there is $n_0$ such
that for all $n\ge n_0$, on $\gE_\Sigma\cap\gE_{\mathrm{sab}}$,
\[
\mSigma(\vs,\vzeta)\ \succeq\ \frac12\,c_\Sigma\,n^2\,\mI_p
\qquad\text{for every }(\vs,\vzeta)\in\gK,
\]
where $p=p_\gamma+p_\beta$.
\end{lemma}

\begin{proof}
Throughout, $\mJ^{-1}$ is well defined on $\vone^\perp$ with
$\opnorm{\mJ(\vs,\vzeta)^{-1}}_{2,\perp}\le2/(nb_0)$ for all
$(\vs,\vzeta)\in\gK$, by Lemma~\ref{lem:localpd} (whose event is
$\gE_{\mathrm{sab}}$); likewise
$\opnorm{\mV^{-1}}_{2,\perp}\le1/(nb_0)$ by Lemma~\ref{lem:Vfloor}. Also
$\opnorm{\mB(\vs,\vzeta)}_2\le C_M n^{3/2}$ on $\gK$ by
Lemma~\ref{lem:cross-jac}, and $\opnorm{\mB^{\popsup}}_2\le C_M n^{3/2}$ by
the identical Cauchy--Schwarz argument (with
$g^*_{ijk}w^*_{\theta,ijk}\le\frac14$).

\emph{Step 1: from $\mSigma^{\popsup}$ to $\mSigma(\vs^*,\vzeta^*)$.} Set
$\delta\mA:=\mA(\vs^*,\vzeta^*)-\mA^{\popsup}$ and
$\delta\mB:=\mB(\vs^*,\vzeta^*)-\mB^{\popsup}$; only their direction blocks
are nonzero, and on $\gE_\Sigma$ both have spectral norm
$\le C_\Sigma n\sqrt{\log n}$ (Lemma~\ref{lem:info-conc}). Telescoping,
\[
\mB\trans\mJ^{-1}\mB-(\mB^{\popsup})\trans\mV^{-1}\mB^{\popsup}
=\delta\mB\trans\mJ^{-1}\mB
+(\mB^{\popsup})\trans\mJ^{-1}\delta\mB
+(\mB^{\popsup})\trans\big(\mJ^{-1}-\mV^{-1}\big)\mB^{\popsup},
\]
all matrices evaluated at $(\vs^*,\vzeta^*)$. On $\gE_{\mathrm{sab}}$,
$\mJ^{-1}-\mV^{-1}=-\mJ^{-1}\mE\,\mV^{-1}$ with
$\opnorm{\mE}_2\le C_E\sqrt{n\log n}$ (Lemma~\ref{lem:saboteur}), so the
three terms are bounded in spectral norm by
\[
2\cdot C_M n^{3/2}\cdot\frac{2}{nb_0}\cdot C_\Sigma n\sqrt{\log n}
\ +\ C_M^2n^3\cdot\frac{2C_E\sqrt{n\log n}}{n^2b_0^2}
\ =\ O\big(n^{3/2}\sqrt{\log n}\big)\ =\ o(n^2),
\]
whence, with $\opnorm{\delta\mA}_2=O(n\sqrt{\log n})$ and Weyl,
$\mSigma(\vs^*,\vzeta^*)\succeq\mSigma^{\popsup}-o(n^2)\,\mI_p
\succeq\frac34 c_\Sigma n^2\,\mI_p$ for $n\ge n_0$, by
A\ref{assump:info}.

\emph{Step 2: across $\gK$.} Fix $(\vs,\vzeta)\in\gK$. For the $\mA$-block,
Lemma~\ref{lem:struct-curv} gives
$\opnorm{\mA(\vs,\vzeta)-\mA(\vs^*,\vzeta^*)}_2
\le L_{A}n^2(\varepsilon_n+2\rho_n)=o(n^2)$. For the cross block,
Lemma~\ref{lem:cross-lip} gives
$\opnorm{\mB(\vs,\vzeta)-\mB(\vs^*,\vzeta^*)}_2
\le L_{B}n^{3/2}(\varepsilon_n+2\rho_n)=o(n^{3/2})$. For the inverse,
\begin{align*}
\opnorm{\mJ(\vs,\vzeta)^{-1}-\mJ(\vs^*,\vzeta^*)^{-1}}_{2,\perp}
&\le\frac{4}{n^2b_0^2}\,\opnorm{\mJ(\vs,\vzeta)-\mJ(\vs^*,\vzeta^*)}_2\\
&\le\frac{4}{n^2b_0^2}\big(4L_fm^\ast n\,\varepsilon_n+2L_\zeta n\,\rho_n\big)
=o(n^{-1})
\end{align*}
by Lemmas~\ref{lem:lip} and~\ref{lem:lip-zeta}. Telescoping
$\mB\trans\mJ^{-1}\mB$ between the two points as in Step~1, each of the
three terms is $O(n^{3/2})\cdot O(n^{-1})\cdot o(n^{3/2})$ or
$O(n^3)\cdot o(n^{-1})$, hence $o(n^2)$; with the $\mA$-increment and Weyl,
$\mSigma(\vs,\vzeta)\succeq\mSigma(\vs^*,\vzeta^*)-o(n^2)\,\mI_p
\succeq\frac12 c_\Sigma n^2\,\mI_p$ for $n\ge n_0$, uniformly over $\gK$
since every $o(n^2)$ bound above depends on $(\vs,\vzeta)$ only through
$\varepsilon_n,\rho_n$.
\end{proof}

\paragraph{Structural scores and Hessian blocks.}
Write the stacked structural score of $-\ell$ as
\[
\mT(\vs,\vzeta):=\nabla_{\vzeta}(-\ell)(\vs,\vzeta)
=\big(\mG_{\vgamma}(\vs,\vgamma)\trans,\
\mG_{\vbeta}(\vs,\vbeta)\trans\big)\trans\in\R^{p},
\qquad p=p_{\gamma}+p_{\beta},
\]
where, from \eqref{eq:score-g}--\eqref{eq:score-b},
\begin{align*}
\mG_{\vgamma}(\vs,\vgamma)&:=\nabla_{\vgamma}(-\ell)
=\sum_{i<j}\sum_k (g_{ijk}-c_{ijk})\,\mY_{ijk},\\
\mG_{\vbeta}(\vs,\vbeta)&:=\nabla_{\vbeta}(-\ell)
=\sum_{i<j}\sum_k (c_{ijk}d_{ijk}-a^1_{ijk})\,\mZ_{ijk},
\end{align*}
with the corresponding curvature blocks $\mA_{\vgamma\vgamma}$ and
$\mA_{\vbeta\vbeta}$ as in \eqref{eq:Hgg} and \eqref{eq:Hbb}.
Since $g_{ijk}=\sigma(f(\Delta_{ij})+\mY_{ijk}\trans\vgamma)$ carries no
$\vbeta$ and $d_{ijk}=\sigma(\Delta_{ij}+\mZ_{ijk}\trans\vbeta)$ carries no
$\vgamma$ (the data $c,a^1$ being fixed), at a fixed merit $\vs$ the map
$\mG_{\vgamma}$ depends on $\vgamma$ alone and $\mG_{\vbeta}$ on $\vbeta$
alone---equivalently $\nabla^2_{\vgamma\vbeta}(-\ell)=0$ by
\eqref{eq:Hgb}---so the structural block $\mA$ of the Hessian is
block-diagonal. The two channels are nonetheless \emph{not} independent
estimation problems: both share the merit vector, and profiling out $\vs$
re-couples them through the cross blocks, the profile Hessian being the
Schur complement $\mSigma=\mA-\mB\trans\mJ^{-1}\mB$ of
Lemma~\ref{lem:profile}, whose off-diagonal block
$\mB_{\vgamma}\trans\mJ^{-1}\mB_{\vbeta}$ is nonzero in general. The joint
argument below therefore treats $\vzeta=(\vgamma,\vbeta)$ as a single
$p$-dimensional parameter.

\paragraph{The profiled likelihood.}
Let $\Theta_2:=\gB(\vgamma^*,2\rho_n)\times\gB(\vbeta^*,2\rho_n)$ denote the
doubled structural ball. Lemmas~\ref{lem:stage1}
and~\ref{lem:loo-uniform} hold verbatim with $\rho_n$ replaced by $2\rho_n$
(their hypotheses require only a radius of the form
$\mathrm{const}\cdot\sqrt{\log n/n}$, now with constant
$2C_{\gamma\beta}$); we henceforth work on the corresponding event
$\widetilde{\gE}_{\mathrm{loo}}$ of Lemma~\ref{lem:loo-uniform} run at
radius $2\rho_n$ (and horizon $T+1$ as in the proof of
Lemma~\ref{lem:stage1}), on which, by the final assertion of
Lemma~\ref{lem:loo-uniform}, the conclusions of Lemma~\ref{lem:stage1} hold
simultaneously for every $\vzeta\in\Theta_2$---including the degenerate
radius $C_{\gamma\beta}=0$ at $\vzeta=\vzeta^*:=(\vgamma^*,\vbeta^*)$. For
$\vzeta\in\Theta_2$ write $\widehat\vs_{\vzeta}$ for the unique zero of
$\mG_{\vzeta}$ in
$\overline{\gB_\infty(\vs^*,\varepsilon_n)}\cap(\vs^*+\vone^\perp)$ furnished by
Lemma~\ref{lem:stage1}, with $\varepsilon_n=C_5\sqrt{\log n/n}$ and $C_5$ the
Stage-1 constant at the doubled radius. The \emph{profiled negative
log-likelihood} is
\begin{equation}\label{eq:profile-def}
\varphi(\vzeta)\ :=\ -\ell\big(\widehat\vs_{\vzeta},\vzeta\big),
\qquad \vzeta\in\Theta_2.
\end{equation}
Since $-\ell(\cdot,\vzeta)$ is strictly convex on
$\overline{\gB_\infty(\vs^*,\varepsilon_n)}\cap(\vs^*+\vone^\perp)$
(Lemma~\ref{lem:localpd}) with interior critical point
$\widehat\vs_{\vzeta}$, one has equivalently
$\varphi(\vzeta)=\min\{-\ell(\vs,\vzeta):
\vs\in\overline{\gB_\infty(\vs^*,\varepsilon_n)}\cap(\vs^*+\vone^\perp)\}$, so
$\varphi$ is the merit-profiled objective in the usual sense.

\begin{lemma}[Profile regularity and envelope identities]\label{lem:profile}
Assume A\ref{assump:dense}--A\ref{assump:gate}. For every $c>0$ there is
$n_0$ such that for all $n\ge n_0$, on
$\widetilde{\gE}_{\mathrm{loo}}$, the following hold.
\begin{enumerate}
\item[(a)] (\emph{Regularity of the profile root.}) The map
$\vzeta\mapsto\widehat\vs_{\vzeta}$ is $C^1$ on the interior of $\Theta_2$,
with values in $\gB_\infty(\vs^*,\varepsilon_n)\cap(\vs^*+\vone^\perp)$ and
Jacobian
\[
\frac{d\widehat\vs_{\vzeta}}{d\vzeta}\ =\
-\,\mJ\big(\widehat\vs_{\vzeta},\vzeta\big)^{-1}\,
\mB\big(\widehat\vs_{\vzeta},\vzeta\big)\ \in\ \R^{(n+1)\times p},
\]
where $\mJ^{-1}$ is taken on $\vone^\perp$ and
$\mB=[\,\mB_{\vgamma}\mid\mB_{\vbeta}\,]$ as in Lemma~\ref{lem:cross-lip}.
\item[(b)] (\emph{Envelope identities.}) $\varphi$ is $C^2$ on the interior
of $\Theta_2$, with
\begin{align*}
\nabla\varphi(\vzeta)\ &=\ \mT\big(\widehat\vs_{\vzeta},\vzeta\big),\\
\nabla^2\varphi(\vzeta)\ &=\ \mSigma\big(\widehat\vs_{\vzeta},\vzeta\big)
\ =\ \mA\big(\widehat\vs_{\vzeta},\vzeta\big)
-\mB\big(\widehat\vs_{\vzeta},\vzeta\big)\trans\,
\mJ\big(\widehat\vs_{\vzeta},\vzeta\big)^{-1}\,
\mB\big(\widehat\vs_{\vzeta},\vzeta\big).
\end{align*}
\item[(c)] (\emph{Profile curvature floor.}) If in addition
A\ref{assump:info} holds then, on
$\widetilde{\gE}_{\mathrm{loo}}\cap\gE_\Sigma$,
\[
\nabla^2\varphi(\vzeta)\ \succeq\ \frac12\,c_\Sigma\,n^2\,\mI_p
\qquad\text{for every $\vzeta$ in the interior of }\Theta_2.
\]
\end{enumerate}
\end{lemma}

\begin{proof}
\emph{(a).} Parametrize the gauge slice by $\vone^\perp$: writing
$\vs=\vs^*+\vu$ with $\vu\in\vone^\perp$, the map
$\mF(\vu,\vzeta):=\mG_{\vzeta}(\vs^*+\vu)\in\vone^\perp$ is $C^2$ on an open
neighborhood of its domain (under A\ref{assump:gate}, $f\in C^3$ makes
$\ell$ three times continuously differentiable in all arguments, and
$\mG_{\vzeta}=\nabla_{\vs}(-\ell)$). Fix $\vzeta_0$ in the interior of
$\Theta_2$ and set $\vu_0:=\widehat\vs_{\vzeta_0}-\vs^*$, so
$\mF(\vu_0,\vzeta_0)=\vzero$ with $\norm{\vu_0}_\infty<\varepsilon_n$ (the root is
interior to the $\ell_\infty$ ball since, by the proof of
Lemma~\ref{lem:stage1},
$\norm{\widehat\vs_{\vzeta}-\vs^*}_\infty
\le C_4\sqrt{\log n/n}+o(\sqrt{\log n/n})$ and $C_5=C_4+1$). The partial
derivative $\partial_{\vu}\mF=\mJ(\vs^*+\vu,\vzeta)\big|_{\vone^\perp}$ is
invertible at $(\vu_0,\vzeta_0)$ with $\lamperp\ge\frac12 nb_0$
(Lemma~\ref{lem:localpd}, whose event
$\gE_{\mathrm{sab}}\supseteq\widetilde{\gE}_{\mathrm{loo}}$ is in force,
its radii $\varepsilon_n$, $2\rho_n$ being $o(1)$). By the implicit function
theorem on $\vone^\perp\times\R^{p}$, there is a neighborhood of
$\vzeta_0$ and a $C^1$ map $\vzeta\mapsto\vu(\vzeta)$ with
$\mF(\vu(\vzeta),\vzeta)=\vzero$ and $\vu(\vzeta_0)=\vu_0$; shrinking the
neighborhood keeps $\norm{\vu(\vzeta)}_\infty<\varepsilon_n$, so by the
uniqueness in Lemma~\ref{lem:stage1} the branch coincides with
$\widehat\vs_{\vzeta}-\vs^*$ there. Hence
$\vzeta\mapsto\widehat\vs_{\vzeta}$ is $C^1$ at every interior $\vzeta_0$.
Differentiating $\mG_{\vzeta}(\widehat\vs_{\vzeta})=\vzero$ in $\vzeta$,
\[
\mJ\big(\widehat\vs_{\vzeta},\vzeta\big)\,
\frac{d\widehat\vs_{\vzeta}}{d\vzeta}
+\frac{\partial\mG_{\vzeta}}{\partial\vzeta}\big(\widehat\vs_{\vzeta}\big)
\ =\ \vzero,
\]
and $\partial\mG_{\vzeta}/\partial\vzeta=\mB$ (proof of
Lemma~\ref{lem:cross-jac}); solving on $\vone^\perp$ (both the columns of
$\mB$ and the range of $d\widehat\vs_{\vzeta}/d\vzeta$ lie in
$\vone^\perp$, the root staying in the gauge slice) gives the stated
Jacobian.

\emph{(b).} By the chain rule and part (a),
\[
\nabla\varphi(\vzeta)
=\Big(\frac{d\widehat\vs_{\vzeta}}{d\vzeta}\Big)\trans\,
\underbrace{\nabla_{\vs}(-\ell)\big(\widehat\vs_{\vzeta},\vzeta\big)}_{=\,
\mG_{\vzeta}(\widehat\vs_{\vzeta})\,=\,\vzero}
+\ \nabla_{\vzeta}(-\ell)\big(\widehat\vs_{\vzeta},\vzeta\big)
\ =\ \mT\big(\widehat\vs_{\vzeta},\vzeta\big):
\]
the merit direction contributes exactly zero, by the definition of the
profile root. For the Hessian, differentiate once more:
\[
\nabla^2\varphi(\vzeta)
=\frac{\partial\mT}{\partial\vs}\big(\widehat\vs_{\vzeta},\vzeta\big)\,
\frac{d\widehat\vs_{\vzeta}}{d\vzeta}
+\frac{\partial\mT}{\partial\vzeta}\big(\widehat\vs_{\vzeta},\vzeta\big)
=\mB\trans\big(-\mJ^{-1}\mB\big)+\mA
=\mSigma\big(\widehat\vs_{\vzeta},\vzeta\big),
\]
using $\partial\mT/\partial\vs=\nabla^2_{\vzeta\vs}(-\ell)=\mB\trans$ and
$\partial\mT/\partial\vzeta=\nabla^2_{\vzeta\vzeta}(-\ell)=\mA$, the latter
block-diagonal by \eqref{eq:Hgb}. In particular $\nabla^2\varphi$ is
symmetric, as it must be, and $\varphi$ is $C^2$ since all blocks are $C^1$
compositions of $C^1$ maps.

\emph{(c).} By part (a),
$\big(\widehat\vs_{\vzeta},\vzeta\big)\in
\gB_\infty(\vs^*,\varepsilon_n)\times\Theta_2\subseteq\gK$ with the radii of
Lemma~\ref{lem:sigma-floor} (there $\rho_n$ is read as
$2\rho_n=O(\sqrt{\log n/n})$), so that lemma applies at
$\big(\widehat\vs_{\vzeta},\vzeta\big)$ and gives
$\nabla^2\varphi(\vzeta)=\mSigma(\widehat\vs_{\vzeta},\vzeta)
\succeq\frac12 c_\Sigma n^2\,\mI_p$.
\end{proof}

\begin{lemma}[Structural score at the profile anchor]\label{lem:struct-score}
Assume A\ref{assump:dense}--A\ref{assump:gate} and $n\ge2$. Let
$C_5'$ be the Stage-1 constant of Lemma~\ref{lem:stage1} run with
$C_{\gamma\beta}=0$, i.e.\ $C_5'=C_4(0)+1$ with $C_4(0)$ from
Lemma~\ref{lem:coupled-induction} at $C_{\gamma\beta}=0$; in particular
$C_5'$ does not depend on $C_{\gamma\beta}$, and on
$\widetilde{\gE}_{\mathrm{loo}}$ the merit root
$\widehat\vs_{\vzeta^*}$ satisfies
$\norm{\widehat\vs_{\vzeta^*}-\vs^*}_\infty\le C_5'\sqrt{\log n/n}$. Then,
on $\widetilde{\gE}_{\mathrm{loo}}$,
\[
\norm{\mT\big(\widehat\vs_{\vzeta^*},\vzeta^*\big)}_2
\ \le\ C_\star\,n^{3/2}\sqrt{\log n},
\qquad
C_\star\ :=\ \sqrt2\,\big(C_\kappa+C_M\,C_5'\big).
\]
\end{lemma}

\begin{proof}
At the truth, the two blocks of $\mT(\vs^*,\vzeta^*)$ are
$\mG_{\vgamma}(\vs^*,\vgamma^*)$ and $\mG_{\vbeta}(\vs^*,\vbeta^*)$, each of
$\ell_2$ norm $\le C_\kappa n\sqrt{\log n}$ on
$\gE_{\mathrm{score}}\supseteq\widetilde{\gE}_{\mathrm{loo}}$
(Lemma~\ref{lem:score}, structural part), so the stacked vector obeys
$\norm{\mT(\vs^*,\vzeta^*)}_2\le\sqrt2\,C_\kappa\,n\sqrt{\log n}$. For the
increment, $\partial\mT/\partial\vs=\mB\trans$ (proof of
Lemma~\ref{lem:profile}), so the fundamental theorem of calculus along
$\vs^*+\tau(\widehat\vs_{\vzeta^*}-\vs^*)$---a segment contained in
$\overline{\gB_\infty(\vs^*,C_5'\sqrt{\log n/n})}$, hence in the working
region---and Lemma~\ref{lem:cross-jac} give
\begin{align*}
\norm{\mT(\widehat\vs_{\vzeta^*},\vzeta^*)-\mT(\vs^*,\vzeta^*)}_2
&\le\ C_M n^{3/2}\,\norm{\widehat\vs_{\vzeta^*}-\vs^*}_2\\
&\le\ C_M n^{3/2}\sqrt{n+1}\;C_5'\sqrt{\frac{\log n}{n}}
\ \le\ \sqrt2\,C_M C_5'\,n^{3/2}\sqrt{\log n}.
\end{align*}
Adding the two bounds and using
$n\sqrt{\log n}\le n^{3/2}\sqrt{\log n}$ gives the claim.
\end{proof}

\begin{theorem}[Joint $\ell_\infty/\ell_2$ consistency]\label{thm:joint}
Assume A\ref{assump:dense}--A\ref{assump:gate} and A\ref{assump:info}. For
every $c>0$ there is $n_0=n_0(c)$ and an event of probability at least
$1-O(n^{-c})$ on which, for all $n\ge n_0$, there is a point
$(\widehat\vs,\widehat\vgamma,\widehat\vbeta)$ with
$\nabla_{\vs}\ell=\nabla_{\vgamma}\ell=\nabla_{\vbeta}\ell=\vzero$
satisfying
\[
\norm{\widehat\vs-\vs^*}_\infty=O\Big(\sqrt{\frac{\log n}{n}}\Big),\qquad
\norm{\widehat\vgamma-\vgamma^*}_2=O\Big(\sqrt{\frac{\log n}{n}}\Big),\qquad
\norm{\widehat\vbeta-\vbeta^*}_2=O\Big(\sqrt{\frac{\log n}{n}}\Big).
\]
Moreover $\widehat\vzeta=(\widehat\vgamma,\widehat\vbeta)$ is the unique
zero of $\nabla\varphi$ in $\overline{\gB(\vzeta^*,\rho_n)}$.
\end{theorem}

\begin{proof}
\emph{Constants, in acyclic order.} First run
Lemma~\ref{lem:coupled-induction} at the fixed pair
$(\vgamma^*,\vbeta^*)$, i.e.\ with $C_{\gamma\beta}=0$, and let
$C_5'$ be the resulting Stage-1 constant of
Lemma~\ref{lem:struct-score}; $C_5'$ depends only on the constants of
A\ref{assump:dense}--A\ref{assump:gate} and on $c$. Next set
\[
C_\star:=\sqrt2\,(C_\kappa+C_M C_5'),\qquad
C_{\gamma\beta}:=\frac{3C_\star}{c_\Sigma},\qquad
\rho_n:=C_{\gamma\beta}\sqrt{\frac{\log n}{n}}.
\]
Finally run Lemmas~\ref{lem:coupled-induction} and~\ref{lem:loo-uniform} at
the doubled radius $2\rho_n$ (constant $2C_{\gamma\beta}$), obtain the
Stage-1 constant $C_5$ $(\ge C_5'$ without loss of generality$)$, and set
$\varepsilon_n:=C_5\sqrt{\log n/n}$. The chain is
$C_5'\to C_\star\to C_{\gamma\beta}\to C_5$, and $C_5$ is consumed below
only through the $o(1)$ radius $\varepsilon_n$ (via Lemmas~\ref{lem:sigma-floor}
and~\ref{lem:profile}), never through a score bound; there is no cycle.

\emph{Event.} Work on
$\gE:=\widetilde{\gE}_{\mathrm{loo}}\cap\gE_\Sigma$, with
$\widetilde{\gE}_{\mathrm{loo}}$ the event of Lemma~\ref{lem:loo-uniform}
at radius $2\rho_n$ (it contains $\gE_{\mathrm{score}}$ and
$\gE_{\mathrm{sab}}$) and $\gE_\Sigma$ that of Lemma~\ref{lem:info-conc};
the union bound gives $\PP(\gE)\ge1-O(n^{-c})$. On $\gE$, the profile root
$\widehat\vs_{\vzeta}$ and the profile $\varphi$ of \eqref{eq:profile-def}
are defined on $\Theta_2$, and Lemma~\ref{lem:profile} is in force. Note
that the root $\widehat\vs_{\vzeta^*}$ of Lemma~\ref{lem:struct-score}
coincides with the profile root at $\vzeta=\vzeta^*$: both are zeros of
$\mG_{\vgamma^*,\vbeta^*}$ in
$\overline{\gB_\infty(\vs^*,\varepsilon_n)}\cap(\vs^*+\vone^\perp)$ (using
$C_5'\le C_5$), where Lemma~\ref{lem:stage1} guarantees uniqueness.

\emph{Existence of a structural root.} Apply the degree-theoretic criterion
(Fact~\ref{fact:ortega-rheinboldt}) to $\mF=\nabla\varphi$ on
$\Omega=\gB(\vzeta^*,\rho_n)$ with anchor $\vzeta^*$; $\mF$ is continuous
on $\overline\Omega$ since $\overline\Omega$ lies in the interior of
$\Theta_2$, where $\varphi$ is $C^2$ (Lemma~\ref{lem:profile}(b)). For
$\vzeta\in\partial\Omega$ write $\vu:=\vzeta-\vzeta^*$,
$\norm{\vu}_2=\rho_n$. By the fundamental theorem of calculus along
$[\vzeta^*,\vzeta]\subset\overline\Omega$, the envelope identity
$\nabla\varphi(\vzeta^*)=\mT(\widehat\vs_{\vzeta^*},\vzeta^*)$, and the
curvature floor of Lemma~\ref{lem:profile}(c),
\[
\vu\trans\nabla\varphi(\vzeta)
=\vu\trans\nabla\varphi(\vzeta^*)
+\vu\trans\Big(\int_0^1\nabla^2\varphi(\vzeta^*+\tau\vu)\,d\tau\Big)\vu
\ \ge\ -\rho_n\,C_\star n^{3/2}\sqrt{\log n}
\ +\ \frac12\,c_\Sigma\,n^2\rho_n^2,
\]
using Lemma~\ref{lem:struct-score} for the first term. Hence
\[
\vu\trans\nabla\varphi(\vzeta)
\ \ge\ \frac12\,c_\Sigma\,n^2\rho_n
\Big(\rho_n-\frac{2C_\star}{c_\Sigma}\sqrt{\frac{\log n}{n}}\Big)\ >\ 0,
\]
the bracket being strictly positive since
$C_{\gamma\beta}=3C_\star/c_\Sigma>2C_\star/c_\Sigma$. By
Fact~\ref{fact:ortega-rheinboldt}, $\nabla\varphi$ has a zero
$\widehat\vzeta\in\overline\Omega$; since $\vu\trans\nabla\varphi>0$ on
$\partial\Omega$, the zero is interior,
$\widehat\vzeta\in\gB(\vzeta^*,\rho_n)$.

\emph{Uniqueness of the structural root.} For distinct
$\vzeta_1,\vzeta_2\in\overline\Omega$, with
$\bar\mSigma:=\int_0^1\nabla^2\varphi(\vzeta_2+\tau(\vzeta_1-\vzeta_2))\,d\tau
\succeq\frac12 c_\Sigma n^2\,\mI_p$ (Lemma~\ref{lem:profile}(c), the
segment lying in $\overline\Omega$),
\[
\big(\nabla\varphi(\vzeta_1)-\nabla\varphi(\vzeta_2)\big)\trans
(\vzeta_1-\vzeta_2)
=(\vzeta_1-\vzeta_2)\trans\,\bar\mSigma\,(\vzeta_1-\vzeta_2)
\ \ge\ \frac12\,c_\Sigma\,n^2\norm{\vzeta_1-\vzeta_2}_2^2\ >\ 0,
\]
so $\nabla\varphi$ is strictly monotone on $\overline\Omega$ and
$\widehat\vzeta$ is its unique zero there.

\emph{Conclusion.} Set $\widehat\vs:=\widehat\vs_{\widehat\vzeta}$ and
$(\widehat\vgamma,\widehat\vbeta):=\widehat\vzeta$. Then
$\nabla_{\vs}\ell(\widehat\vs,\widehat\vzeta)=\vzero$ by the definition of
the profile root, and
$\nabla_{\vzeta}\ell(\widehat\vs,\widehat\vzeta)
=-\mT(\widehat\vs_{\widehat\vzeta},\widehat\vzeta)
=-\nabla\varphi(\widehat\vzeta)=\vzero$ by the envelope identity
(Lemma~\ref{lem:profile}(b)). The rates hold with the constants built into
the radii:
$\norm{\widehat\vs-\vs^*}_\infty\le\varepsilon_n=C_5\sqrt{\log n/n}$ and
$\max\{\norm{\widehat\vgamma-\vgamma^*}_2,
\norm{\widehat\vbeta-\vbeta^*}_2\}
\le\norm{\widehat\vzeta-\vzeta^*}_2\le\rho_n
=C_{\gamma\beta}\sqrt{\log n/n}$.
\end{proof}

\begin{corollary}[Uniqueness and the MLE]\label{cor:mle}
On the event and under the assumptions of Theorem~\ref{thm:joint}, the
point $(\widehat\vs,\widehat\vgamma,\widehat\vbeta)$ is the unique
stationary point of $\ell$ in
$\gK=\overline{\gB_\infty(\vs^*,\varepsilon_n)}\times
\overline{\gB(\vgamma^*,\rho_n)}\times\overline{\gB(\vbeta^*,\rho_n)}$
intersected with the gauge slice $\vs^*+\vone^\perp$, and it maximizes
$\ell$ over that set; it is in this sense the MLE.
\end{corollary}

\begin{proof}
On $\vone^\perp\times\R^{p}$ the full Hessian of $-\ell$ is the bordered
matrix $\mM=\begin{pmatrix}\mJ&\mB\\ \mB\trans&\mA\end{pmatrix}$.
Throughout $\gK$, $\mJ\succeq\frac12 nb_0$ on $\vone^\perp$
(Lemma~\ref{lem:localpd}) and
$\mM/\mJ=\mSigma\succeq\frac12 c_\Sigma n^2\,\mI_p$
(Lemma~\ref{lem:sigma-floor}), so by the Schur criterion
(Fact~\ref{fact:schur}) $\mM\succ0$ throughout $\gK$. Hence $-\ell$ is
strictly convex on the convex set
$\gK\cap[(\vs^*+\vone^\perp)\times\R^{p_{\gamma}}\times\R^{p_{\beta}}]$ and
admits at most one stationary point there, which is its minimizer if
interior. Theorem~\ref{thm:joint} exhibits an interior stationary point; it
is therefore unique and maximizes $\ell$ over
$\gK\cap[(\vs^*+\vone^\perp)\times\R^{p_{\gamma}}\times\R^{p_{\beta}}]$.
\end{proof}

\clearpage

\section{Experiment Details on Interpretable Feature Recovery}
\label{app:experiment-details}

This appendix collects the full experimental details for the synthetic
cover-letter study of Section~\ref{sec:feature_extraction_empirical_results}. Appendix~\ref{app:dataset} describes the seeded corpus generation: the
planted dials, their levels and sampling distributions, and the generation
prompts. Appendix~\ref{app:methods-details} gives the formal definitions of the recovery metrics (Tracked, Stat.R., Sem.R., Attrib., LK), including the matching procedure between discovered features and planted dials. It also specifies the four baseline configurations (Direct LLM, Direct LLM--Multi-score, Local LLMs, Local LLMs--Multi-score), their prompts, and the API budget of each method. Appendix~\ref{app:results} provides additional results such as the full list of surfaced features with their fitted parameters and diagnostics, and the judge-written rubric. Finally, Appendix~\ref{app:variance} details the paired-bootstrap protocol
behind Figure~\ref{fig:paired} and Table~\ref{tab:paired}, and gathers
additional recovery figures (dial-recovery heatmaps and the
feature-merging threshold analysis).

\subsection{Dataset}
\label{app:dataset}

\begin{figure}[ht]
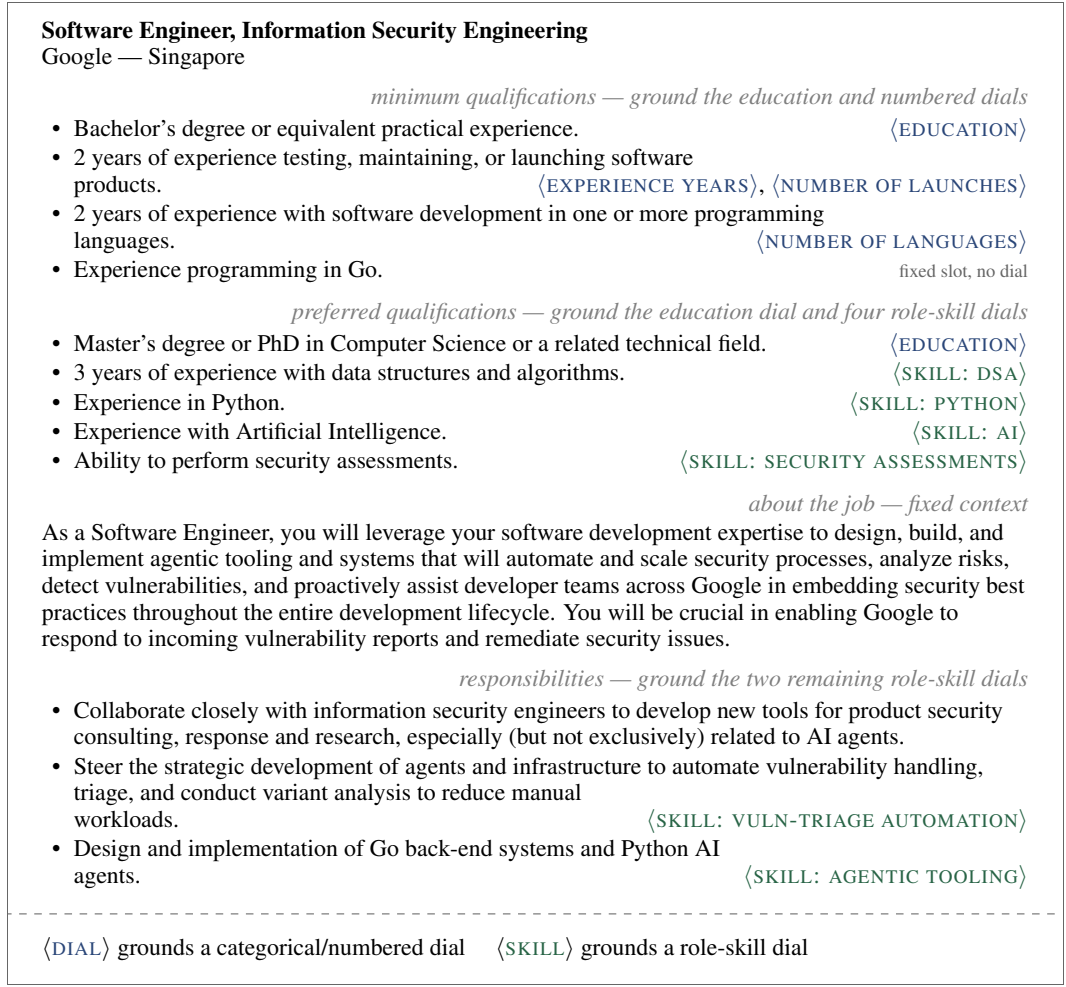

\centering
\begin{tcolorbox}[enhanced, width=\linewidth, colback=white,
  colframe=black!60, boxrule=0.3pt, sharp corners,
  left=10pt, right=10pt, top=4pt, bottom=6pt, fontupper=\small]
\raggedright\setlength{\parindent}{0pt}
\textbf{Software Engineer, Information Security Engineering}\\
Google --- Singapore
\blocktag{minimum qualifications --- ground the education and numbered dials}
\begin{adlist}
\item Bachelor's degree or equivalent practical experience.
      \tagright{\dialtag{education}}
\item 2 years of experience testing, maintaining, or launching software
      products. \tagright{\dialtag{experience years}, \dialtag{number of launches}}
\item 2 years of experience with software development in one or more
      programming languages. \tagright{\dialtag{number of languages}}
\item Experience programming in Go. \tagright{\band{fixed slot, no dial}}
\end{adlist}
\blocktag{preferred qualifications --- ground the education dial and four role-skill dials}
\begin{adlist}
\item Master's degree or PhD in Computer Science or a related technical
      field. \tagright{\dialtag{education}}
\item 3 years of experience with data structures and algorithms.
      \tagright{\coretag{skill: dsa}}
\item Experience in Python. \tagright{\coretag{skill: python}}
\item Experience with Artificial Intelligence.
      \tagright{\coretag{skill: ai}}
\item Ability to perform security assessments.
      \tagright{\coretag{skill: security assessments}}
\end{adlist}
\blocktag{about the job --- fixed context}
As a Software Engineer, you will leverage your software development expertise to design, build, and implement agentic tooling and systems that will automate and scale security processes, analyze risks, detect vulnerabilities, and proactively assist developer teams across Google in embedding security best practices throughout the entire development lifecycle. You will be crucial in enabling Google to respond to incoming vulnerability reports and remediate security issues.
\blocktag{responsibilities --- ground the two remaining role-skill dials}
\begin{adlist}
\item Collaborate closely with information security engineers to develop new tools for product security consulting, response and research, especially (but not exclusively) related to AI agents.
\item Steer the strategic development of agents and infrastructure to automate vulnerability handling, triage, and conduct variant analysis to reduce manual workloads.
      \tagright{\coretag{skill: vuln-triage automation}}
\item Design and implementation of Go back-end systems and Python AI agents.
      \tagright{\coretag{skill: agentic tooling}}
\end{adlist}
\tcbline
{\footnotesize
$\langle$\textcolor{dialC}{\textsc{dial}}$\rangle$ grounds a categorical/numbered dial
\quad
$\langle$\textcolor{coreC}{\textsc{skill}}$\rangle$ grounds a role-skill dial}
\end{tcolorbox}
\caption{\textbf{The job advertisement used as the fixed evaluation context.} Colored tags mark which requirement grounds which dial, using the same visual vocabulary as the letter template (Figure~\ref{fig:template-skeleton}): the qualifications ground the education and numbered dials, and the preferred qualifications and responsibilities ground the six role-skill dials. Go is a mandatory requirement realized identically in every letter, not a dial. The remaining dials of Table~\ref{tab:dials} (school and employer prestige, prose fluency, and the four niche dials) have no anchor in the ad and are described in Appendix~\ref{app:dataset}.}
\label{fig:job-ad}
\end{figure}

\paragraph{Job ad.} The real job ad was archived \href{https://web.archive.org/web/20260706204238/https://www.google.com/about/careers/applications/jobs/results/111928378944037574-software-engineer-information-security-engineering}{here}. The job ad obtained after trimming boilerplate content is shown in Figure~\ref{fig:job-ad}.

\paragraph{Example of template.}
\label{app:template}

Figure~\ref{fig:template-skeleton} shows one of the ten templates used for the cover letter generation.

\begin{figure}[ht]
\centering
\begin{tcolorbox}[enhanced, width=\linewidth, colback=white,
  colframe=black!60, boxrule=0.3pt, sharp corners,
  left=10pt, right=10pt, top=4pt, bottom=6pt, fontupper=\small]
\raggedright\setlength{\parindent}{0pt}

Dear Google Hiring Team,
{\footnotesize\itshape\color{black!50}\hfill fixed greeting}

\blocktag{fixed opening prose}
Security work is the rare kind of engineering where the best outcome is that nothing happens. That inversion is what pulls me toward the Software Engineer role on the Information Security Engineering team: the chance to build systems whose success is measured in incidents that never occur.

\blocktag{background clause --- categorical and numbered dials}
My foundation is a \dialslot{education} in computer science from
\dialslot{school} and \dialslot{experience years} years of professional
software development at \dialslot{employer}, where I have tested,
maintained, and launched \dialslot{number of launches} production services, writing Go alongside \dialslot{number of languages} other languages.

\blocktag{body of six core units: $k$ skill sentences $+ (6-k)$ fillers, paragraphed $[2,2,2]$}
\coreslot{core unit 1} \coreslot{core unit 2}\par\smallskip
\coreslot{core unit 3} \coreslot{core unit 4}\par\smallskip
\coreslot{core unit 5} \coreslot{core unit 6}

\blocktag{niche paragraph: exactly two active niche dials}
\nicheslot{niche unit 1} \nicheslot{niche unit 2}

\blocktag{fixed closing, ending with verbatim authorization sentence}
I would be glad to talk in more depth about the team's challenges and where I could contribute from day one. I am fully authorized to work in Singapore and do not require visa sponsorship.\par\smallskip
Sincerely,\par
\nuisslot{name}

\tcbline
{\footnotesize
$\langle$\textcolor{dialC}{\textsc{dial}}$\rangle$ ground-truth dial value
\quad
$\langle$\textcolor{coreC}{\textsc{core}}$\rangle$ skill/filler unit
\quad
$\langle$\textcolor{nicheC}{\textsc{niche}}$\rangle$ niche unit
}
\end{tcolorbox}
\caption{\textbf{Example cover-letter template.} Roman text is fixed template prose; angle-bracketed slots are filled from the ground-truth dial vector. Skill coverage sets how many core units are capability sentences; paragraph count is constant. Fluency is imposed afterward by injecting $0/5/12$ surface errors.}
\label{fig:template-skeleton}
\end{figure}

\begin{table}[t]
\centering
\footnotesize
\begin{threeparttable}
\begin{tabular}{@{}llll@{}}
\toprule
Dial & Type & Levels & Group \\
\midrule
Experience (years) & ordered & 3 / 5 / 7 & primary \\
Education level & ordered & BS / MS / PhD & primary \\
School selectivity & ordered & low / mid / high & primary \\
Employer prestige & ordered & startup / mid / big tech & primary \\
Languages known & ordered & 2 / 5 / 9 & primary \\
Production launches & ordered & 2 / 4 / 6 & primary \\
Skill coverage\tnote{$\dagger$} & ordered & 2 / 4 / 6 & primary \\
Prose fluency\tnote{$\ddagger$} & ordered & 0 / 5 / 12 mistakes & primary \\
\midrule
DSA depth & binary & absent / present & per-skill \\
Python & binary & absent / present & per-skill \\
AI/agents experience & binary & absent / present & per-skill \\
Security assessments & binary & absent / present & per-skill \\
Vuln-triage automation & binary & absent / present & per-skill \\
Agentic tooling design & binary & absent / present & per-skill \\
\midrule
Side projects & niche\tnote{$\ast$} & absent / toy / hard & niche \\
Open-source & niche\tnote{$\ast$} & absent / patches / maintainer & niche \\
Security certification & niche\tnote{$\ast$} & absent / entry / advanced & niche \\
Capture-the-flag (CTF) & niche\tnote{$\ast$} & absent / participated / strong & niche \\
\bottomrule
\end{tabular}
\begin{tablenotes}
\item[$\dagger$] Counts the number of skills mentioned; a composite of the
per-skill dials, hence not counted as a separate recovery target.
\item[$\ddagger$] Deliberately irrelevant to the role; serves as a leakage
check. Hence $D=16$ relevant dials.
\item[$\ast$] Exactly 2 of the 4 niche dials are non-absent in any given
letter.
\end{tablenotes}
\end{threeparttable}
\caption{The 18 engineered ground-truth dials of the cover-letter corpus.}
\label{tab:dials}
\end{table}

In what follows, we describe each dial in more detail: the levels it takes and how those levels are realized in the assembled letter. Everything is also gathered in Table~\ref{tab:dials}.

\paragraph{Experience.}
The experience dial takes one of three values (3, 5 or 7 years) and is stated in the background clause as a tenure at the current employer. It is realized as a single digit and is one of the three numbers on the letter's digit whitelist.

\paragraph{Education.}
The education dial takes three levels, BS $<$ MS $<$ PhD, rendered in the opening as a ``\{EDUCATION\} in computer science from \{SCHOOL\}''. The three degrees are written as \emph{two-word} phrases: ``Bachelor's degree'', ``Master's degree'', ``Doctoral degree'', so that the dial adds no length signal (the bare token ``PhD'' would make higher education correlate with shorter letters).

\paragraph{School prestige.}
The school dial fills the \{SCHOOL\} slot and takes three ascending
selectivity bands: \emph{less selective} (Cal State East Bay, University of Toledo, Kennesaw State University, Central Michigan University), \emph{moderately selective} (strong general publics: Penn State, Ohio State, Arizona State, Virginia Tech), and \emph{highly selective} (MIT, Stanford, UC Berkeley, Princeton). All names are \emph{real} institutions, including the bottom band, whose intended reading is ``an ordinary accredited school''. Band membership is anchored to a published selectivity classification.

\paragraph{Employer prestige.}
The employer dial fills the \{EMPLOYER\} slot and takes three ascending
recognizability bands: an \emph{unknown-startup} band of fictional,
unrecognizable firms (Northlake, Bluegrove, Quarrypoint, Milldam), a \emph{medium} band of real mid-recognizability companies (Zillow, Etsy, Expedia, Wayfair), and a \emph{big-tech} band (Meta, Amazon, Microsoft, Apple). Unlike the school dial, the bottom band is fictional on purpose: invented names give a clean ``no prestige prior'' floor, whereas a real but obscure firm would carry its own idiosyncratic associations.

\paragraph{Programming languages.}
This numbered dial takes values $\{2,5,9\}$ and is stated in the background clause as ``writing Go alongside \{NUMBER OF LANGUAGES\} other languages.'' Go is a fixed, mandatory slot (a minimum qualification); only the count varies. The count is realized as a single whitelisted digit in a clause present in every letter, so it grades a numbered attribute without introducing any length variation across its levels.

\paragraph{Production launches.}
This numbered dial takes values $\{2,4,6\}$ and appears in the background clause as ``tested, maintained, and launched \{NUMBER OF LAUNCHES\} production services.'' Like the language count it is a self-contained count (never a duration), realized as a whitelisted digit in a fixed clause. It is assigned orthogonally to the experience dial so that it cannot re-enter the model as an experience proxy.

\paragraph{Skill coverage.}
The skill-coverage dial sets \emph{how many} of the six role-skills the letter demonstrates, taking values $k\in\{2,4,6\}$. Its effect is to determine how many of the six core body units are capability sentences: exactly $k$ core units render a present skill, and the remaining $6-k$ are \emph{fillers}: role/domain-motivation sentences that never claim a skill, a niche, or communication ability. Because the body always contains six core units laid out in a fixed paragraph plan regardless of $k$, coverage changes the \emph{content} of the body but not its length or paragraph count, so it cannot ride the verbosity bias.

\paragraph{Skill identity.}
Which skills are present (as opposed to how many) is a constrained set rather than an ordered dial. The pool has six letter-addressable items derived from the ad's preferred qualifications and responsibilities: data-structures depth, Python, applied-AI experience, security assessments, vulnerability-triage automation, and agentic-tooling design. Each present skill is rendered as one concrete capability sentence drawn from the current template's phrasing, with no numbers and no durations.

\paragraph{Niche dials.}
There are four niche dials: side project, open-source involvement, security
certification, and capture-the-flag (CTF). Each is a three-level ordered factor: \emph{absent}, a middle level, and a strong level (side project: toy $<$ hard; open source: patches $<$ maintainer; certification: entry (Security+) $<$ advanced (OSCP); CTF: participated $<$ strong). They are subject to a presence constraint: exactly two of the four are non-absent in every letter, so ``number of extras mentioned'' is held constant and cannot become a richness or length axis. The two active dials are realized, at their assigned level, as evidence clauses in a single dedicated niche paragraph placed at a fixed position after the body.

\paragraph{Fluency.}
The fluency dial controls surface prose quality and takes three levels anchored to an exact injected-error count: clean ($0$), mid ($5$), and degraded ($12$). Fluency is not a slot but a deterministic post-assembly pass that injects the target number of surface spelling errors into the finished letter; the word count is re-validated in band after injection, so the only thing that varies across fluency levels is the density of errors.

\paragraph{Nuisance factors.}
Two factors are balanced across the corpus but are not part of the ground truth and are not scored. Each letter is assembled from one of ten templates, which vary the greeting, opening, closing, and the exact wording of every skill, niche, and filler unit while holding the underlying dial content fixed. The applicant's name, drawn from a fixed pool, appears only in the signature and carries no signal.

\paragraph{Dataset creation.} The dataset comprises $n=200$ cover letters. Of these, $184$ form a balanced pool: every level of every dial appears an equal number of times, and the joint assignment of levels to letters is chosen as the most orthogonal of $256$ independently drawn candidate designs (the candidate minimizing the largest non-structural pairwise $|r|$). The remaining $16$ are deliberate anchors: $8$ polished-but-thin and $8$ rough-gem profiles that anti-correlate the ``prestige'' dials with the substance dials, included so the judge sees the full range of the design, and excluded from the balance and orthogonality gates below. 

Each letter is then assembled deterministically from one of 10 templates keyed to its levels. For example, experience, education, and school/employer prestige are stated in the opening; the $k\in\{2,4,6\}$ present role-skills are rendered as capability sentences drawn from a pool (one pool per template), the identity of the $k$ present skills being chosen by a greedy usage-balancing rule so that skill identity is orthogonal to coverage $k$; production launches is stated in the background clause; the two active niche dials appear as evidence clauses; and prose fluency is imposed by injecting a target number of surface spelling errors ($0$/$5$/$12$). 

On the balanced pool the realized design satisfies $|r|\le0.15$ over all non-structural dial pairs (max $0.146$) and $|r|\le0.22$ between each dial and letter length (max $0.202$); over the full corpus, where the anchors enter, these are $0.195$ and $0.219$. Letter length is not part of the design objective but it is verified after assembly. Generation is seeded and fully reproducible.

\subsection{Methods Details}
\label{app:methods-details}

\subsubsection{Metrics}
\label{app:metrics}
Each method yields a set of discovered features that we will need to match to the dials. 

\paragraph{Matching the discovered features to the dials.} For each method, we perform a Hungarian matching assigning each dial to a distinct feature so as to maximize total cosine similarity in \textsc{text-embedding-3-large}'s embedding space. The pairs it rejects (features and dials sharing domain vocabulary but, in the assignment's judgment, not describing the same construct) form an empirical background of unrelated pairs, and we define a threshold $\eta$ as the 95th percentile of their alignment ($\eta = 0.406$): the level that unrelated pairs almost never attain. A matched pair clearing $\eta$ is therefore semantically exceptional relative to background similarity, not merely the least-bad assignment; cosine similarities carry no absolute scale, so the reference distribution, rather than a fixed constant, is what gives the threshold meaning. 

Having defined this threshold, we map a method's set of discovered features to the dial set by performing semantic Hungarian matching, but only keeping matches whose cosine similarity is greater than $\eta$. 

In what follows, we also say that a feature $k$ \emph{tracks} a dial $d$ if the absolute Pearson correlation $|r_{k,d}|$ between the two is greater than a threshold $\tau = 0.4$ fixed a priori and well above the permutation floor ($\approx 0.15$ under letter-shuffled nulls; results are stable for $\tau \in [0.3, 0.5]$). The metrics mentioned in Section~\ref{sec:feature_extraction_empirical_results} are mathematically defined as follows:

\paragraph{Statistical recovery.} For each dial, does \emph{some} feature track it? We report the fraction of dials with $\max_k \lvert r_{k,d}\rvert > \tau$ (Tracked), and the mean of
$\max_k \lvert r_{k,d}\rvert$ over all $D$ dials, misses included (Stat.R.).

\paragraph{Semantic recovery.} For each dial, is there a feature that is both \emph{about} it and tracks it? Let $M_m$ assign features to dials by Hungarian matching on semantic alignment, and $c_{m,d}$ the alignment of dial $d$ with its match $M_m(d)$:
\begin{equation}
  \label{eq:semrec}
  \operatorname{Sem.R.}(m)
  = \frac{1}{D}\sum_{d=1}^{D}
    \1_{\{c_{m,d} \ge \eta\}}
    \1_{\{\lvert r_{M_m(d),\,d}\rvert > \tau\}}.
\end{equation}

\paragraph{Attributable features.} Analogously, what fraction of a
method's $K_m$ discovered features measure a single dial? A feature is
\emph{attributable} if it tracks some dial and that dial dominates its
attribution, i.e.\ carries a majority of its squared correlation,
$\pi_k = \max_d r_{k,d}^2/ \sum_d r_{k,d}^2 > \tfrac12$:
\begin{equation}
  \label{eq:attrib}
  \operatorname{Attrib.}(m)
  = \frac{1}{K_m}\sum_{k=1}^{K_m}
    \1_{\{\max_d \lvert r_{k,d}\rvert > \tau\}}\1_{\left\{\pi_k > \tfrac12\right\}}
\end{equation}

\paragraph{Leakage.} Do scores leak across dials? Since dials are near-orthogonal by construction, off-target correlations of a valid feature set should be small. With $\mathcal{A}_m$ the attributable features and $k_m(d) = \argmax_k \lvert r_{k,d}\rvert$ the tracker of dial $d$,
\begin{equation}
  \label{eq:leak}
  \operatorname{LK}(m)
  = \frac{1}{D}\sum_{d=1}^{D}
    \operatorname*{mean}_{k \in \mathcal{A}_m \setminus \{k_m(d)\}} \lvert r_{k,d} \rvert
\end{equation}
(lower is better). Restricting to attributable features prevents noise features from diluting the average.

\subsubsection{Methods Description}

\paragraph{Our approach.}

For our method, the comparison graph on the $n = 200$ cover letters is a union of $d=40$ balanced subgraphs (called \emph{matchings}, each a perfect matching in which every item is compared exactly once). The number $d = 40$ was chosen to guarantee high precision.

Prompt~\ref{prm:comp-system} was used for the pairwise comparisons (\textsc{GPT-5.4 mini}), and Prompt~\ref{prm:taxo-main} was used for the feature taxonomy (\textsc{GPT-5.4 (high reasoning)}). The cost was estimated using the API logs.

\paragraph{Other methods.} Other methods feed the entirety of the dataset (Direct LLM)/a sample of it (Local LLMs) to the judge (\textsc{GPT-5.4 (high reasoning)}) and have it draw up a rubric tailored to the samples seen (Prompt~\ref{prm:pw-discover}). This rubric is then used to have \textsc{GPT-5.4 mini} score each cover letter pointwise on each feature (Prompt~\ref{prm:pw-score}).

For Local LLMs, only $40$ letters of the dataset are given to the judge during the feature elicitation step, enabling the LLM to focus its attention on a smaller sample of data. To ensure that each cover letter is seen at least once, we repeat the elicitation with $E = 27$ different samples, which is the average number of times one would have to repeat the step so that each cover letter is seen at least once. We then consider the union of the features, performed with \textsc{GPT-5.4 (high reasoning)} and Prompt~\ref{prm:taxo-pooled} (see Remark~\ref{rem:why-taxo} for why we used LLM-based taxonomy instead of semantic hierarchical clustering).

To match the cost of our approach, the scoring is repeated $20$ times for Direct LLM--Multi-score, and $6$ times for Local LLMs--Multi-score. Each scoring is performed by slightly modifying the prompt, changing the instruction variant (Prompt~\ref{prm:pw-variants}) and permuting the features inside the rubric.

\clearpage

\subsubsection{Prompts}

\begin{promptbox}{Comparative method (ours) -- pairwise comparison prompt.}{prm:comp-system}
You are helping build an interpretable scoring framework by examining how job applicants' cover letters differ. The framework works by collecting many pairwise comparisons: for each pair, the ways the two letters differ are articulated in plain terms, and aggregated across many pairs these reveal the distinct dimensions along which the applications vary. There is no fixed rubric — the dimensions are recovered from comparisons like this one, not specified in advance.
 
The letters below are applications for the same software-engineering role. The full job posting they respond to is:
=== JOB POSTING ===
(*@\ph{full job posting}@*)
=== END JOB POSTING ===
 
Your task. You will be shown two cover letters, Letter A and Letter B, both written for the role above. This pair is one comparison in a larger collection.
 
Identify the dimensions on which the two letters meaningfully differ as applications for this role. Report the differences that genuinely stand out to you on reading this pair — bring up whatever you find salient, and do not work from a predetermined list. Favor real, noticeable differences over minor ones, but do not force the list to be short: report as many as actually distinguish these two letters.
 
For each dimension, first name the concrete, specific feature you actually saw in the letters — a short phrase pointing to the specific evidence. Then, grounded in that evidence, name the dimension, describe in one sentence what it captures, and indicate which letter shows more of it.
 
Judge each dimension on its own terms, from its own evidence. Applications have uneven profiles: it is entirely normal for one letter to be stronger on some dimensions and the other letter stronger on others, so let the letter that shows more of each dimension fall wherever the evidence for that dimension points.
 
Out of scope — do not treat these as dimensions of difference:
- the applicant's name or any personal identifier;
- the closing sentence about work authorization / visa sponsorship, which is identical in every letter;
- length in itself.
 
How to report. List each dimension as exactly one item — do not combine dimensions with "and", "/", "&", or a comma; split related dimensions into separate items. For the evidence field, give a short phrase naming the concrete feature you saw. For the dimension field, give a concise general noun phrase naming the dimension (the kind of short label you might use as a rubric column). For the description field, give one sentence saying what that dimension captures. Indicate which letter shows more of it.
 
Output only JSON of the form:
{"differences": [{"evidence": "<short phrase>", "dimension": "<concise noun phrase>", "description": "<one sentence>", "winner": "A" | "B"}, ...]}
\end{promptbox}

\begin{promptbox}{Comparative method (ours) -- feature taxonomy.}{prm:taxo-main}
You are organizing a flat list of evaluation dimensions into a set of canonical features.
 
These dimensions were produced by judges comparing job applicants' cover letters for the same software-engineering role. Each judge, on each comparison, named the dimensions on which two letters differed. Aggregated across many comparisons, the same underlying construct was named in many slightly different ways. Your job is to group the different phrasings of the SAME underlying construct together.
 
Below is the full list. Each entry has an INDEX in square brackets, a short NAME, a one-sentence DESCRIPTION of what it captures, and CITATIONS (how many times it was named — for your awareness only).
 
(*@\ph{indexed dimension list}@*)
 
Instructions:
- Group entries that denote the SAME underlying evaluative construct, judging by MEANING, not by surface wording. Different words for the same idea belong in one group; the same word used for different ideas belongs in different groups.
- Do NOT force everything into a small number of groups, but do NOT split one construct into its facets either. A dimension that does not clearly share a construct with any other does NOT become its own feature: place all such genuinely-apart, idiosyncratic dimensions together in a single catch-all group whose canonical name is exactly "Other". Do not create many single-member groups, and do not absorb a genuinely-apart dimension into a loosely-related group to tidy up — send it to "Other".
- Do NOT merge two constructs just because they often co-occur or because strong applicants tend to score well on both. Group only by shared MEANING, never by how applicants might score.
- Citations are context only. Do not merge a frequently-named dimension into a rarely-named one or vice versa on the basis of counts; counts do not determine grouping.
 
For each group, give:
- a canonical NAME (a concise noun phrase suitable as a rubric column; use exactly "Other" for the catch-all group),
- a one-sentence DEFINITION of the construct the group captures,
- the list of members, each as its INDEX and its NAME copied verbatim from that index.
 
Rules on membership (these will be checked programmatically):
- Every input index must appear in exactly one group, exactly once. Place all of them; drop none; duplicate none.
- Do not invent indices that are not in the input.
- Each member's "name" must be the exact input NAME at that index — this is a cross-check, so copy it from the bracketed entry, do not paraphrase.
 
Output only JSON:
{"features": [{"name": "<canonical noun phrase>", "definition": "<one sentence>", "members": [{"index": <int>, "name": "<verbatim input name at that index>"}, ...]}, ...]}
\end{promptbox}

\begin{promptbox}{Pointwise baselines (Direct LLM/Local LLMs) -- rubric discovery prompt.}{prm:pw-discover}
You are helping build an interpretable scoring framework by examining how job applicants' cover letters differ. There is no fixed rubric — the dimensions are recovered from what actually distinguishes these letters, not specified in advance.
=== JOB POSTING ===
(*@\ph{full job posting}@*)
=== END JOB POSTING ===
Below is a sample of cover letters submitted for this role, selected to span the full range of applicant quality.
=== COVER LETTERS ===
(*@\ph{cover letters}@*)
Reading across them, attend to the ways the letters differ from one another — the axes along which you can see one letter showing more, less, or something different from another. Identify the dimensions on which these letters meaningfully differ as applications for this role. Report whatever genuinely stands out to you; do not work from a predetermined list, and do not force the list to be short — report as many dimensions as actually distinguish these letters.
A dimension may be something a letter shows more or less of along a range, or something a letter either shows or does not.
Keep genuinely distinct dimensions separate even when they are related or tend to co-occur — do not merge two dimensions because strong letters tend to show both. List each dimension as exactly one item — do not combine dimensions with "and", "/", "&", or a comma; split related dimensions into separate items.
For each dimension, first name the concrete, specific feature you actually saw in the letters — a short phrase pointing to the evidence. Then, grounded in that evidence, name the dimension and describe in one sentence what it captures. Then write a short assessment guideline a future reader will use to score a new letter on this dimension on a scale you need to specify: say what to look for, and describe what distinguishes each score of the scale you come up with.
Out of scope — do not treat these as dimensions of difference:
- the applicant's name or any personal identifier;
- the closing sentence about work authorization / visa sponsorship, which is identical in every letter;
- length in itself.
Output only JSON:
{"features": [{"evidence": "<short phrase>", "name": "<concise noun phrase>", "description": "<one sentence>", "levels": [{"score": 1, "descriptor": "<a letter at the weakest point of your scale>"}, {"score": 2, "descriptor": "<...>"}, ... up to the top score of the scale you chose]}, ...]}
— for each dimension choose your own scale (score 1 = weakest, up to a top point N you decide) and give one descriptor per score point, from 1 to N.
\end{promptbox}

\begin{promptbox}{Local LLMs -- feature fusion prompt.}{prm:taxo-pooled}
You are organizing a flat list of evaluation dimensions into a set of canonical features.
 
These dimensions were produced by independent rubric elicitations over job applicants' cover letters for the same software-engineering role. Each elicitation read its own random sample of letters and, working from scratch, named the dimensions on which those letters meaningfully differed. Aggregated across many independent elicitations, the same underlying construct was named in many slightly different ways. Your job is to group the different phrasings of the SAME underlying construct together.
 
Below is the full list. Each entry has an INDEX in square brackets, a short NAME, a one-sentence DESCRIPTION of what it captures, and RUN (which elicitation produced it — for your awareness only).
 
(*@\ph{indexed dimension list}@*)
 
Instructions:
- Group entries that denote the SAME underlying evaluative construct, judging by MEANING, not by surface wording. Different words for the same idea belong in one group; the same word used for different ideas belongs in different groups.
- Do NOT force everything into a small number of groups, but do NOT split one construct into its facets either. A dimension that does not clearly share a construct with any other does NOT become its own feature: place all such genuinely-apart, idiosyncratic dimensions together in a single catch-all group whose canonical name is exactly "Other". Do not create many single-member groups, and do not absorb a genuinely-apart dimension into a loosely-related group to tidy up — send it to "Other".
- Do NOT merge two constructs just because they often co-occur or because strong applicants tend to score well on both. Group only by shared MEANING, never by how applicants might score.
- Run labels are context only. Do not merge a dimension named by many runs into one named by few, or vice versa, on the basis of how often it recurs; recurrence does not determine grouping. Two entries from the SAME run were named as distinct dimensions by that run, so they are unlikely to be the same construct — but judge by MEANING, and merge them if they plainly are.
 
For each group, give:
- a canonical NAME (a concise noun phrase suitable as a rubric column; use exactly "Other" for the catch-all group),
- a one-sentence DEFINITION of the construct the group captures,
- the list of members, each as its INDEX and its NAME copied verbatim from that index.
 
Rules on membership (these will be checked programmatically):
- Every input index must appear in exactly one group, exactly once. Place all of them; drop none; duplicate none.
- Do not invent indices that are not in the input.
- Each member's "name" must be the exact input NAME at that index — this is a cross-check, so copy it from the bracketed entry, do not paraphrase.
 
Output only JSON:
{"features": [{"name": "<canonical noun phrase>", "definition": "<one sentence>", "members": [{"index": <int>, "name": "<verbatim input name at that index>"}, ...]}, ...]}
\end{promptbox}
 
\begin{promptbox}{Pointwise baselines (Direct LLM/Local LLMs) -- scoring prompt template.}{prm:pw-score}
You are scoring a cover letter against a fixed rubric, for the software-engineering role below.
=== JOB POSTING ===
(*@\ph{full job posting}@*)
=== END JOB POSTING ===
=== COVER LETTER ===
(*@\ph{letter}@*)
(*@\ph{instruction variant}@*)
A dimension you cannot fully assess from this letter should still receive your best estimate, not a default middle score. Score each dimension independently, and do not let the score on one dimension affect the scores on others. Applications have uneven profiles: it is possible for a letter to score low on some dimensions and high on others.
=== RUBRIC DIMENSIONS (score in the order listed; each lists the descriptors for its own score scale) ===
(*@\ph{rubric, in scoring order}@*)
Out of scope — do not let these affect any score:
- the applicant's name or any personal identifier;
- the closing sentence about work authorization / visa sponsorship, which is identical in every letter;
- length in itself.
(*@\ph{output specification}@*)
\end{promptbox}
 
\begin{promptbox}{Pointwise scoring -- the four reasoning-path instruction variants.}{prm:pw-variants}
(*@\textnormal{\bfseries [direct]}@*)
Score the letter on each rubric dimension below independently. Each dimension lists a descriptor for every point of its own score scale; assign the integer score whose descriptor best matches the letter on that dimension. Judge each dimension on its own.
 
(*@\textnormal{\bfseries [evidence\_first]}@*)
For each rubric dimension below, briefly note what in the letter bears on that dimension, then assign the integer score, from that dimension's own listed scale, whose descriptor best matches the letter on that dimension. Treat each dimension independently.
 
(*@\textnormal{\bfseries [standard\_referenced]}@*)
For each rubric dimension below, compare the letter against the score descriptors listed for that dimension, then assign the integer score whose descriptor it matches most closely. Judge each dimension on its own.
 
(*@\textnormal{\bfseries [independent]}@*)
Go through the rubric dimensions one at a time, in the order listed. For each dimension, consider only what the letter shows on that specific dimension and assign the integer score, from that dimension's own listed scale, whose descriptor best matches; do not form an overall impression of the letter and do not let the other dimensions influence the score.
\end{promptbox}

\subsection{Additional Results}
\label{app:results}

\subsubsection{Model calibration}
\label{sec:synthetic-calibration}

In this subsection, we study whether or not the statistical model is well-calibrated by computing its Expected Calibration Error (ECE) and its Brier skill score against various baselines.

The fitted model assigns to each (comparison, feature) cell two probability forecasts of observable binary events: the citation gate
$\hat g_{p,k} = \sigma\bigl(f(\hat\Delta_{p,k}) + \hat\gamma_k\bigr)$, the
probability that feature $k$ is cited on comparison $p$, defined on all
$84{,}000$ cells ($K = 21$ features), and the direction gate $\hat d_{p,k} = \sigma\bigl(\hat\Delta_{p,k} + \hat\beta\bigr)$, the probability that the first-shown letter is favored given citation, defined
on the $29{,}030$ cited cells. We score both as forecasts, out of fold:
$5$-fold cross-validation over comparisons, so that every cell is predicted
by parameters estimated without it, with $95\%$ bootstrap intervals over
comparisons throughout; in-sample counterparts quantify the optimism of the
fit.

\begin{figure}
    \centering
    \includegraphics[width=\linewidth]{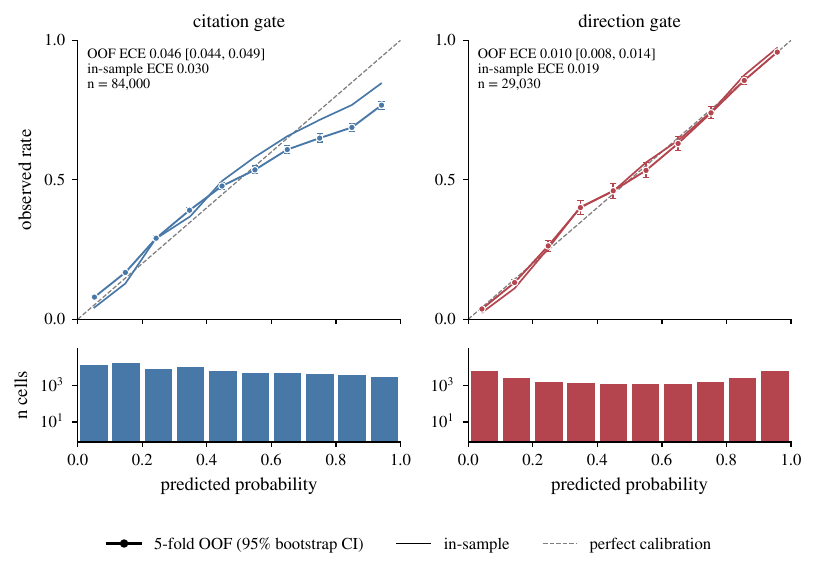}
    \caption{Out-of-fold reliability of the citation gate (left, all cells)
    and direction gate (right, cited cells) of the feature-discovery model.
    Points: observed rate per bin of predicted probability, with $95\%$
    bootstrap intervals over comparisons; faint line: in-sample; dashed:
    perfect calibration; bottom: cells per bin (log scale).}
    \label{fig:calibration_reliability_custom}
\end{figure}

Figure~\ref{fig:calibration_reliability_custom} reports reliability: cells are grouped into $10$ equal-width bins of predicted probability, the observed event rate in each bin is compared against the diagonal of perfect
calibration, and the expected calibration error (ECE) is the occupancy-weighted mean absolute gap. The direction gate is calibrated essentially everywhere: its out-of-fold ECE is $0.010$ $[0.008, 0.014]$ (in sample: $0.019$), and its predictions populate the full unit interval including the extreme bins, so the diagnostic is exercised where it is most demanding. The citation gate's out-of-fold ECE is $0.046$ $[0.044, 0.049]$
against $0.030$ in sample, and its reliability curve is flatter than the
diagonal (observed rates exceed predictions in the low bins and fall short in the high ones) so the fitted gate is too extreme at both ends,
consistent with the amplitude of $f$ slightly overshooting out of fold rather than the gap dependence being absent.

\begin{figure}
    \centering
    \includegraphics[width=\linewidth]{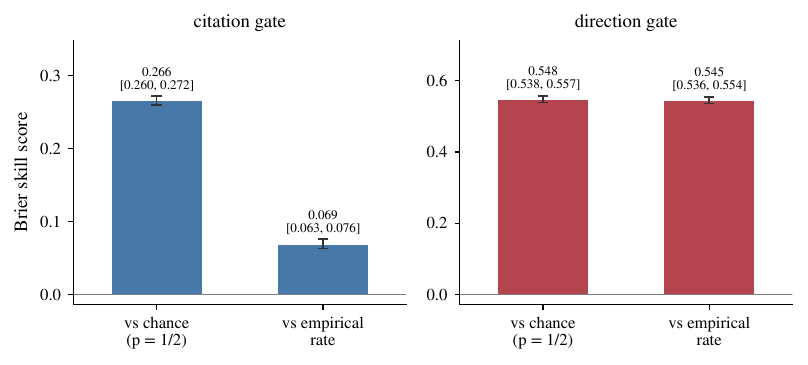}
    \caption{Out-of-fold Brier skill of the two gates against chance and
    against the training-fold empirical rates; intervals as in
    Figure~\ref{fig:calibration_reliability_custom}.}
    \label{fig:calibration_skill_custom}
\end{figure}

Figure~\ref{fig:calibration_skill_custom} reports discrimination through the Brier skill score, $1 - \mathrm{Brier}(\text{model})/\mathrm{Brier}(\text{reference})$, against two references estimated on the training folds: chance ($p = 1/2$, Brier $0.25$) and the empirical rate (the per-feature citation rate for the citation gate, the per-feature
rate of favoring the first-shown letter among cited cells for the direction
gate). The second reference is the demanding one. For the citation gate it is the maximum-likelihood fit of the gap-independent submodel $f = 0$, whose gate $\sigma(\gamma_k)$ is constant within each feature, so skill over it is a direct out-of-fold test of whether the merit gap carries information about citation. It does: the gate scores $0.069$ $[0.063, 0.076]$ against this submodel, rejecting $f = 0$, down from $0.168$ in sample; its $0.266$ $[0.260, 0.272]$ against chance partly rewards knowing the citation base rate rather than per-cell discrimination. The direction gate scores $0.548$ $[0.538, 0.557]$ against chance and $0.545$ $[0.536, 0.554]$ against the empirical rate (the two coincide because slot balance keeps the first-shown rates near $1/2$) and retains most of its in-sample skill ($0.623$). The channel that carries the scores is thus both calibrated and strongly predictive out of sample; the gap effect on citation is significant but second-order, estimated with an amplitude that overshoots out of fold.

\subsubsection{Discovered Features}
\label{app:discovered-features}

Tables~\ref{tab:feats-ours} and~\ref{tab:feats-m1} report the features discovered by each method described in Table~\ref{tab:main}. More precisely, Table~\ref{tab:feats-ours} presents each feature with its reliability coefficients and its fitted $\gamma$ coefficients, Table~\ref{tab:feats-m1} shows the detailed rubric created by \textsc{GPT-5.4 (high reasoning)}.

As we can see in Table~\ref{tab:feats-ours}, our approach yields fine-grained features, as the judge distinguishes the existence of side projects (``Side-project initiative'') from their magnitude (``Side-project ambition'', which almost passed the reliability thresholds). Moreover, the reliability diagnostics correctly filter features such as ``Narrative framing and tone'', which is a well-identified feature ($\kappa = 4.9$) but whose citations are inconsistent with the estimated scores ($\rho = 0.72 < \rho_{\min}$): a sign of a ``boilerplate'' feature that the judge cites independently from the items it is comparing. This is partly captured by the $\gamma$ coefficient. For example, ``Employer prestige'' has a smaller $\gamma$ than ``Narrative framing and tone'' but is reliable: a sign that the former is discriminative while the latter is a boilerplate feature.

We also notice in Table~\ref{tab:feats-m1} that the rubric correctly captures some of the dials' levels. For example, the judge correctly identifies that for the Education dial, BS $<$ MS $<$ PhD: the rubric is correctly tailored to the samples that the judge sees.

\begin{table}[!htbp]
\centering
\scriptsize
\setlength{\tabcolsep}{5pt}
\renewcommand{\arraystretch}{1.05}
\begin{tabular}{@{}l r r r r c@{}}
\toprule
feature & Cite rate & $\gamma_k$ & $\kappa$ & $\rho$ & Kept \\
\midrule
AI experience & 0.613 & $-1.36$ {\tiny$[-1.45, -1.27]$} & 3.6 & 0.937 & \cmark \\
Security assessment experience & 0.596 & $-1.37$ {\tiny$[-1.46, -1.28]$} & 2.9 & 0.935 & \cmark \\
Security automation experience & 0.562 & $-1.31$ {\tiny$[-1.40, -1.22]$} & 2.4 & 0.917 & \cmark \\
Agentic systems experience & 0.544 & $-1.52$ {\tiny$[-1.61, -1.43]$} & 2.7 & 0.924 & \cmark \\
Professional experience & 0.531 & $-2.43$ {\tiny$[-2.55, -2.31]$} & 3.8 & 0.960 & \cmark \\
Algorithmic strength & 0.524 & $-1.48$ {\tiny$[-1.57, -1.39]$} & 2.7 & 0.917 & \cmark \\
Security certification & 0.445 & $-2.54$ {\tiny$[-2.66, -2.43]$} & 4.0 & 0.942 & \cmark \\
Education level & 0.414 & $-2.75$ {\tiny$[-2.86, -2.64]$} & 3.3 & 0.943 & \cmark \\
Open-source contribution & 0.392 & $-2.45$ {\tiny$[-2.55, -2.34]$} & 3.9 & 0.924 & \cmark \\
Off-hours security practice & 0.382 & $-2.11$ {\tiny$[-2.21, -2.02]$} & 3.3 & 0.891 & \cmark \\
Python experience & 0.365 & $-2.05$ {\tiny$[-2.14, -1.95]$} & 2.7 & 0.873 & \cmark \\
Programming language breadth & 0.293 & $-3.13$ {\tiny$[-3.26, -3.01]$} & 2.6 & 0.925 & \cmark \\
Prestige signaling & 0.271 & $-2.57$ {\tiny$[-2.68, -2.46]$} & 3.8 & 0.855 & \cmark \\
Production delivery scope & 0.264 & $-2.84$ {\tiny$[-2.96, -2.72]$} & 2.8 & 0.884 & \cmark \\
Mission and company motivation & 0.258 & $-2.37$ {\tiny$[-2.48, -2.27]$} & 3.7 & 0.770 & \cmark \\
Platform and developer enablement orientation & 0.192 & $-2.96$ {\tiny$[-3.08, -2.83]$} & 5.8 & 0.771 & \cmark \\
Go experience & 0.191 & $-2.98$ {\tiny$[-3.11, -2.85]$} & 5.0 & 0.799 & \cmark \\
Security philosophy & 0.133 & $-3.79$ {\tiny$[-3.97, -3.61]$} & 9.4 & 0.801 & \cmark \\
Motivation narrative & 0.121 & $-4.16$ {\tiny$[-4.37, -3.95]$} & 8.8 & 0.806 & \cmark \\
Side-project initiative & 0.094 & $-4.74$ {\tiny$[-5.02, -4.47]$} & 8.1 & 0.850 & \cmark \\
Employer prestige & 0.074 & $-5.13$ {\tiny$[-5.45, -4.80]$} & 8.3 & 0.820 & \cmark \\
\midrule
Side-project ambition & 0.194 & $-3.40$ {\tiny$[-3.55, -3.24]$} & 10.5 & 0.866 & \xmark \\
Narrative framing and tone & 0.159 & $-3.10$ {\tiny$[-3.24, -2.97]$} & 4.9 & 0.721 & \xmark \\
Systems and technical depth & 0.112 & $-3.72$ {\tiny$[-3.89, -3.54]$} & 11.6 & 0.699 & \xmark \\
General security grounding & 0.093 & $-3.88$ {\tiny$[-4.07, -3.69]$} & 7.8 & 0.652 & \xmark \\
Role alignment & 0.091 & $-3.80$ {\tiny$[-3.98, -3.62]$} & 10.3 & 0.580 & \xmark \\
Impact quantification & 0.068 & $-4.74$ {\tiny$[-5.02, -4.46]$} & 12.8 & 0.731 & \xmark \\
Maintenance orientation & 0.053 & $-8.55$ {\tiny$[-9.87, -7.23]$} & 198.5 & 0.862 & \xmark \\
Closing posture & 0.050 & $-4.67$ {\tiny$[-4.95, -4.38]$} & 10.5 & 0.564 & \xmark \\
Learning posture & 0.047 & $-6.61$ {\tiny$[-7.22, -6.00]$} & 56.1 & 0.762 & \xmark \\
Engineering rigor & 0.028 & $-8.86$ {\tiny$[-10.20, -7.51]$} & 228.0 & 0.831 & \xmark \\
Other & 0.012 & $-8.54$ {\tiny$[-9.60, -7.49]$} & 92.0 & 0.036 & \xmark \\
Writing polish & 0.009 & $-9.51$ {\tiny$[-11.26, -7.76]$} & 156.6 & 0.440 & \xmark \\
\bottomrule
\end{tabular}
\caption{Features surfaced by our comparative method in the point estimate of Table~\ref{tab:main}. Of $32$ surfaced features, we keep the features that have $\rho \geq \rho_{\min} := 0.75$ and $\kappa \leq \kappa_{\max} := 10$, leaving $21$ features. Fitted $\hat\beta = +0.005$ (95\% CI $[-0.033, +0.043]$). CIs for $\hat\beta$ and $\gamma$ coefficients were computed using~\eqref{eq:sandwich-clt}.}
\label{tab:feats-ours}
\end{table}

\begingroup
\setlength{\tabcolsep}{4pt}
\renewcommand{\arraystretch}{0.75}
\setlength{\LTcapwidth}{\linewidth}
\begin{longtable}{@{}>{\scriptsize\raggedright\arraybackslash}p{0.13\linewidth} >{\scriptsize\raggedright\arraybackslash}p{0.27\linewidth} >{\scriptsize\raggedright\arraybackslash}p{0.555\linewidth}@{}}
\toprule
Feature & Description & Scoring guidelines \\
\midrule
\endfirsthead
\multicolumn{3}{@{}l}{\scriptsize\itshape Table~\ref{tab:feats-m1} continued}\\
\toprule
Feature & Description & Scoring guidelines \\
\midrule
\endhead
\midrule
\multicolumn{3}{r@{}}{\scriptsize\itshape continued on next page}\\
\endfoot
\bottomrule
\noalign{\vspace{6pt}}
\caption{The rubric elicited by Direct LLM. Descriptions and the per-score guidelines are reproduced verbatim as the judge produced them; the open elicitation lets the judge choose its own scale, and it selected a $1$--$4$ scale for every feature.}
\label{tab:feats-m1}\\
\endlastfoot
Academic degree & Captures the formal level of academic preparation or equivalent technical grounding the applicant presents. & \textbf{1}~No clear qualifying degree or equivalent preparation is shown. \newline \textbf{2}~Shows a bachelor's degree or equivalent practical preparation. \newline \textbf{3}~Shows a master's degree in a relevant technical field. \newline \textbf{4}~Shows a doctoral degree in a relevant technical field. \\
\addlinespace[2pt]
Adversarial practice & Captures ongoing offensive-security practice through CTF participation and competitive results. & \textbf{1}~No ongoing offensive-security practice is shown. \newline \textbf{2}~Mentions occasional CTF participation or light adversarial practice. \newline \textbf{3}~Shows regular or serious CTF participation as continuing skill development. \newline \textbf{4}~Shows strong competitive results, such as near-top or top placements in substantial CTF fields. \\
\addlinespace[2pt]
Agent boundary design & Captures experience designing agentic systems with explicit control boundaries between Python agents and Go services. & \textbf{1}~No agentic tooling or agent architecture is described. \newline \textbf{2}~Mentions agents or agentic systems only at a high level. \newline \textbf{3}~Describes building agentic systems with Python agents and Go back-end support. \newline \textbf{4}~Describes explicit guardrails such as bounded actions, permissions, auditing, or enforced boundaries between agents and services. \\
\addlinespace[2pt]
Algorithmic depth & Captures depth of data-structure and algorithm reasoning as evidenced by concrete redesign or optimization examples. & \textbf{1}~Does not provide evidence of algorithmic or data-structure skill. \newline \textbf{2}~Claims strong algorithms knowledge, but with little concrete illustration. \newline \textbf{3}~Provides a clear example of using better algorithms or data structures to improve a system. \newline \textbf{4}~Provides a strong example with technical specificity and visible impact, such as removing a bottleneck or making a previously hard problem tractable. \\
\addlinespace[2pt]
Applied AI deployment & Captures whether the applicant has shipped AI or LLM features in production and understands reliability or evaluation concerns. & \textbf{1}~No meaningful AI or LLM experience is shown. \newline \textbf{2}~Mentions AI interest or exposure without a clear shipped production example. \newline \textbf{3}~Shows direct experience shipping AI or LLM-backed functionality in production. \newline \textbf{4}~Shows shipped AI or LLM systems plus concrete attention to evaluation, reliability, or failure modes. \\
\addlinespace[2pt]
Developer enablement mindset & Captures whether the applicant frames security tooling as a way to help developers do the right thing with less friction. & \textbf{1}~Does not frame security work in terms of helping other engineers. \newline \textbf{2}~Mentions helping teams only in broad or generic language. \newline \textbf{3}~Clearly emphasizes making secure behavior easier or more natural for developers. \newline \textbf{4}~Explicitly focuses on reducing friction, building good defaults, or serving busy developer teams through usable tooling. \\
\addlinespace[2pt]
Evidence specificity & Captures how consistently the letter backs claims with specific examples, mechanisms, or project details. & \textbf{1}~Relies mostly on generic assertions and unsubstantiated claims. \newline \textbf{2}~Mixes some specifics with many broad claims. \newline \textbf{3}~Uses several concrete examples, technologies, or project details to support claims. \newline \textbf{4}~Consistently grounds claims in precise examples, mechanisms, or scenario-level detail throughout the letter. \\
\addlinespace[2pt]
Impact articulation & Captures whether achievements are described in terms of concrete effect on team capacity, risk reduction, or manual workload. & \textbf{1}~Mostly describes activities performed, with little indication of why they mattered. \newline \textbf{2}~Hints at benefits, but leaves the concrete effect largely implicit. \newline \textbf{3}~Explains clear outcomes such as reduced manual work, better reliability, or improved throughput. \newline \textbf{4}~Makes impact vivid and credible through concrete consequences, strong before-after framing, or quantified value. \\
\addlinespace[2pt]
Maintainer mindset & Captures whether the applicant brings a maintainer's perspective shaped by testing, inherited systems, failure modes, and long-term operability. & \textbf{1}~Does not show a maintenance or reliability perspective. \newline \textbf{2}~Mentions testing, maintenance, or inherited systems without drawing lessons from them. \newline \textbf{3}~Shows that maintenance or failure-mode experience informs how the applicant designs systems. \newline \textbf{4}~Strongly presents a maintainer identity centered on operability, inheritance, failure handling, and long-term resilience. \\
\addlinespace[2pt]
Mechanical polish & Captures proofreading quality and whether spelling or typing errors distract from credibility. & \textbf{1}~Contains frequent spelling, typing, or proofreading errors that materially distract. \newline \textbf{2}~Contains several noticeable errors that reduce polish. \newline \textbf{3}~Contains minor occasional errors, but remains generally professional. \newline \textbf{4}~Is mechanically clean, polished, and free of distracting errors. \\
\addlinespace[2pt]
Open-source stewardship & Captures contribution level to open-source software, from occasional patches to maintaining a community-used project. & \textbf{1}~No open-source contribution is mentioned. \newline \textbf{2}~Mentions occasional fixes or patches to open-source dependencies. \newline \textbf{3}~Shows regular stewardship, such as maintaining a small project or handling releases. \newline \textbf{4}~Shows active maintenance of a community-used project, library, or framework with reviews, releases, and external users. \\
\addlinespace[2pt]
Personal systems initiative & Captures self-directed engineering initiative shown through side projects, especially ambitious distributed-systems or reliability-heavy builds. & \textbf{1}~No meaningful side project or independent build is described. \newline \textbf{2}~Shows a modest personal project, utility, or app that demonstrates follow-through. \newline \textbf{3}~Shows a substantial self-directed engineering project with nontrivial design or reliability concerns. \newline \textbf{4}~Shows an ambitious systems project, especially distributed or failure-heavy work chosen to tackle hard engineering problems. \\
\addlinespace[2pt]
Platform leverage orientation & Captures whether the letter explicitly values foundational platform work with compounding downstream impact. & \textbf{1}~Does not articulate interest in platform or org-wide leverage. \newline \textbf{2}~Mentions scale or impact in generic terms. \newline \textbf{3}~Clearly values foundational work whose benefits reach many teams or systems. \newline \textbf{4}~Repeatedly frames the role in terms of compounding downstream impact across a large organization like Google. \\
\addlinespace[2pt]
Production ownership & Captures the breadth of concrete production service launch and maintenance ownership the letter presents. & \textbf{1}~Mentions production work vaguely or only limited ownership without clear launches. \newline \textbf{2}~Shows ownership of a small number of production services, with some testing, maintenance, or launch responsibility. \newline \textbf{3}~Shows clear responsibility across several production services, indicating repeated launch and maintenance experience. \newline \textbf{4}~Shows broad ownership across many production services, suggesting repeated end-to-end delivery in production. \\
\addlinespace[2pt]
Python fluency & Captures how strongly the letter shows Python as a real working language for services, automation, or tooling. & \textbf{1}~Does not show meaningful Python experience. \newline \textbf{2}~Mentions Python, but only briefly or without clear production use. \newline \textbf{3}~Shows regular Python use for services, automation, analysis, or internal tooling. \newline \textbf{4}~Shows strong production-grade Python ownership, with concrete services or tooling others rely on. \\
\addlinespace[2pt]
Security assessment practice & Captures hands-on experience performing security reviews, threat modeling, code review, or validating remediation. & \textbf{1}~No clear security assessment work is described. \newline \textbf{2}~Mentions security reviews generically, without concrete assessment methods. \newline \textbf{3}~Shows concrete assessment work such as threat modeling, code review, attack-surface mapping, or service reviews. \newline \textbf{4}~Shows repeated or formal assessment practice, including documented findings and validation of remediation or fixes. \\
\addlinespace[2pt]
Security credentialing & Captures external validation of security knowledge through named certifications and their rigor. & \textbf{1}~No security certification is mentioned. \newline \textbf{2}~Shows a baseline security certification such as Security+. \newline \textbf{3}~Shows a stronger hands-on certification such as OSCP. \newline \textbf{4}~Shows multiple or especially rigorous certifications that materially strengthen the security case. \\
\addlinespace[2pt]
Security operations systems-thinking & Captures whether the letter sees security response as a scalable systems-design problem rather than ad hoc manual toil. & \textbf{1}~Treats security interest generically, without a clear systems view of the problem. \newline \textbf{2}~Mentions automation or scale, but without a crisp model of the underlying bottleneck. \newline \textbf{3}~Frames vulnerability response or security toil as a systems-design or capacity problem. \newline \textbf{4}~Articulates a sharp leverage model, such as scaling scarce expert judgment, variant analysis, or turning heroic work into routine automation. \\
\addlinespace[2pt]
Security rigor ethos & Captures whether the letter articulates a clear high-standard philosophy about building security-critical software. & \textbf{1}~Shows generic enthusiasm for security, but no clear quality philosophy. \newline \textbf{2}~Signals seriousness about correctness or standards in broad terms. \newline \textbf{3}~Articulates a distinct philosophy about the bar security tooling should meet. \newline \textbf{4}~Presents a memorable, role-relevant engineering creed that clearly matches high-assurance security work. \\
\addlinespace[2pt]
Software tenure & Captures how much relevant professional software experience in years the letter claims. & \textbf{1}~Less than the role minimum or unclear how much professional software experience the applicant has. \newline \textbf{2}~Shows roughly entry-level relevant experience, such as about 2 to 3 years in professional software development. \newline \textbf{3}~Shows solid mid-level experience, such as about 4 to 5 years building and maintaining software. \newline \textbf{4}~Shows substantial experience, such as 6 or more years of directly relevant software development work. \\
\addlinespace[2pt]
Vulnerability workflow automation & Captures direct experience automating vulnerability intake, triage, deduplication, routing, or prioritization workflows. & \textbf{1}~Shows no direct vulnerability-handling automation experience. \newline \textbf{2}~Expresses interest in automating vulnerability work, but without direct evidence of having done it. \newline \textbf{3}~Shows hands-on automation of vulnerability intake or triage, such as routing or queue handling. \newline \textbf{4}~Shows deep workflow automation with concrete functions like classification, deduplication, prioritization, routing, or major manual-load reduction. \\
\end{longtable}
\endgroup

\subsection{Run-variance analysis}
\label{app:variance}

\subsubsection{Per-method run-variance bands}
\label{app:variance-bands}

\begin{figure}[t]
\centering
\includegraphics{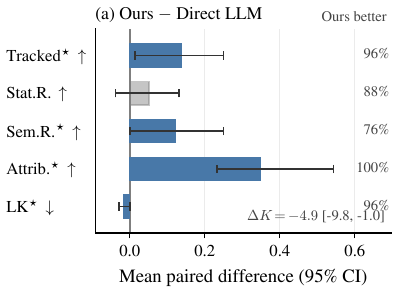}\hfill
\includegraphics{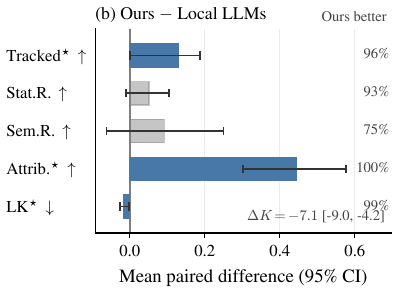}
\caption{Paired comparisons over $B=50$ bootstrap draws (with
replacement) of the $200$-letter corpus; every arm sees the same draws,
and differences are taken within replicate. Each row shows the mean
paired difference with its $95\%$ percentile CI ($\star$: excludes zero); the right column gives \emph{Ours better}, the share of replicates our method wins ($\Delta K$, the difference of discovered features, is annotated below each panel and carries no share). \textbf{The panels
bracket discovery variance from opposite sides and are not comparable
to one another}: (a) re-runs discovery per replicate on both sides,
(b) freezes discovery on both sides, pooled over fusion and corpus
draws. \textbf{Both panels score our method at a reduced budget}
(average degree $25.3$ against the designed $40$); it leads on several
metrics regardless. Exact values: Table~\ref{tab:paired}.}
\label{fig:paired}
\end{figure}

Table~\ref{tab:main-bands} repeats the point estimates of
Table~\ref{tab:main} together with each method's run-variance band. The
gray bands are per-method run-variance summaries; each arm resamples the
unit its own pipeline randomizes over, so band widths reflect the design
as much as the method and carry no cross-method meaning.

\begin{table}[t]
\centering
\small
\setlength{\tabcolsep}{3pt}
\begin{tabular}{l r r ccccc}
\toprule
Method & Cost & $K$ & Tracked & Stat.R. & Sem.R. & Attrib. & LK$\downarrow$ \\
\midrule
\textbf{Ours} & 3.45 & 21 & \textbf{100.0\%} & \textbf{0.850} & \textbf{100.0\%} & \textbf{76.2\%} & \textbf{0.090} \\
\band{\quad resampling matchings} & & \band{[16.2, 22.0]} & \band{[87.5, 93.8]} & \band{[0.737, 0.805]} & \band{[81.2, 93.8]} & \band{[71.4, 87.5]} & \band{[0.087, 0.095]} \\
\addlinespace
Direct LLM & \textbf{0.26} & 21 & 81.2\% & 0.732 & 81.2\% & 61.9\% & 0.109 \\
\band{\quad resampling letters} & & \band{[17.2, 25.0]} & \band{[68.8, 87.5]} & \band{[0.649, 0.825]} & \band{[62.5, 87.5]} & \band{[40.6, 69.6]} & \band{[0.109, 0.135]} \\
\addlinespace
Direct LLM--Multi-score & 3.42 & 21 & 81.2\% & 0.787 & 81.2\% & 66.7\% & 0.118 \\
\band{\quad resampling letters} & & \band{[21.0, 21.0]} & \band{[81.2, 81.2]} & \band{[0.788, 0.807]} & \band{[81.2, 81.2]} & \band{[57.1, 66.7]} & \band{[0.119, 0.145]} \\
\addlinespace
Local LLMs & 2.35 & 23 & 75.0\% & 0.702 & 68.8\% & 47.8\% & 0.110 \\
\band{\quad resampling elicitation subsets} & & \band{[22.2, 27.8]} & \band{[75.0, 87.5]} & \band{[0.696, 0.789]} & \band{[62.5, 87.5]} & \band{[44.0, 55.7]} & \band{[0.103, 0.120]} \\
\addlinespace
Local LLMs--Multi-score & 3.27 & 23 & 75.0\% & 0.737 & 68.8\% & 47.8\% & 0.113 \\
\band{\quad resampling letters} & & \band{[23.0, 23.0]} & \band{[75.0, 75.0]} & \band{[0.737, 0.761]} & \band{[68.8, 68.8]} & \band{[39.1, 52.2]} & \band{[0.115, 0.136]} \\
\bottomrule
\end{tabular}
\caption{Per-method point estimates with run-variance bands. Tracked,
Sem.R., Attrib. in \%; Stat.R. and LK are correlations; bands are in the
same units as the column. Below each method's point estimates, the gray
row gives that method's \textbf{own} run-variance band
($[2.5,\,97.5]$ percentiles across replicates), obtained by resampling the
unit named at the left of the row. These bands are \textbf{not} confidence
intervals for the point estimates above them -- the point estimates come
from the single full-data run, the bands from re-runs on resampled data,
so a point estimate may fall outside its band -- and they \textbf{must not
be compared across methods}, since the arms resample different units.
Paired, comparable contrasts: Table~\ref{tab:paired}.}
\label{tab:main-bands}
\end{table}

\begin{table}[t]
\centering
\small
\begin{tabular}{l r c r}
\toprule
& Difference & 95\% CI & Ours better \\
\midrule
\multicolumn{4}{l}{\textbf{(a) Ours $-$ Direct LLM} \; (discovery re-run per replicate on \emph{both} sides)} \\
\midrule
$\Delta K$ & $-4.90$ & $[-9.78,\ -1.00]$ & --- \\
$\Delta$ \textbf{Tracked}$^{*}$ & $+0.138$ & $[+0.014,\ +0.250]$ & $96\%$ \\
$\Delta$ Stat.R. & $+0.051$ & $[-0.038,\ +0.132]$ & $88\%$ \\
$\Delta$ \textbf{Sem.R.}$^{*}$ & $+0.122$ & $[+0.000,\ +0.250]$ & $76\%$ \\
$\Delta$ \textbf{Attrib.}$^{*}$ & $+0.349$ & $[+0.233,\ +0.543]$ & $100\%$ \\
$\Delta$ \textbf{LK}$^{*}$ & $-0.017$ & $[-0.028,\ -0.000]$ & $96\%$ \\
\midrule
\multicolumn{4}{l}{\textbf{(b) Ours $-$ Local LLMs} \; (discovery frozen on \emph{both} sides; pooled over fusion \emph{and} corpus draws)} \\
\midrule
$\Delta K$ & $-7.08$ & $[-9.00,\ -4.23]$ & --- \\
$\Delta$ \textbf{Tracked}$^{*}$ & $+0.129$ & $[+0.000,\ +0.188]$ & $96\%$ \\
$\Delta$ Stat.R. & $+0.051$ & $[-0.010,\ +0.104]$ & $93\%$ \\
$\Delta$ Sem.R. & $+0.092$ & $[-0.062,\ +0.250]$ & $75\%$ \\
$\Delta$ \textbf{Attrib.}$^{*}$ & $+0.444$ & $[+0.302,\ +0.577]$ & $100\%$ \\
$\Delta$ \textbf{LK}$^{*}$ & $-0.016$ & $[-0.027,\ -0.003]$ & $99\%$ \\
\bottomrule
\end{tabular}
\caption{Paired comparisons over $B=50$ bootstrap draws (with replacement)
of the $200$-letter corpus; every arm sees the same draws, and differences
are taken within replicate. Each row reports the mean difference, its
$95\%$ percentile CI ($^{*}$: excludes zero), and \emph{Ours better}: the
share of replicates in which our method wins in the metric's preferred
direction ($\Delta K$ is descriptive and carries no share). \textbf{The panels bracket discovery variance from opposite sides and are not comparable to one another}. \textbf{Both panels score our method at a reduced budget (average degree $25.3$ against the designed
$40$)}; it leads on several metrics regardless.
Design rationale, the conditional-fusion alternative, and budget details:
Appendix~\ref{app:paired-design}.}
\label{tab:paired}
\end{table}

Ours resamples the 40 matchings that make up the comparison graph, and then re-runs the taxonomy on every replicate (the judging of the pairs is not rerun as it is too costly); the $K$ band covers the reliable count after the $\kappa/\rho$ filter, from $28.9$ discovered features on average. Direct LLM resamples the letters and re-elicits its rubric per replicate.
Direct LLM--Multi-score shares Direct LLM's rubric but its discovery step is deterministic, so its band reflects letter resampling only; the degenerate entries ($K$, Tracked, Sem.R.) are exact. Local LLMs redraws which $E{=}27$ elicitations are pooled, so its band carries fusion variance but holds the corpus fixed -- unlike Direct LLM, whose band varies both. Local LLMs--Multi-score is instead the direct counterpart of Direct LLM--Multi-score: a single fusion with deterministic discovery, scored with $6$ repeated passes averaged before resampling, so its band too reflects letter resampling only, with $K$, Tracked and Sem.R. again exact.

\subsubsection{Design of the paired comparisons}
\label{app:paired-design}

Panel (b) of Figure~\ref{fig:paired} cannot mirror panel (a). Local LLMs
discovery replicates are indexed by the choice of $E{=}27$ pooled
elicitations rather than by letters, so they admit no pairing with corpus
draws, and re-eliciting on each resampled corpus would be prohibitively
costly. In (a) both arms re-run discovery on every replicate, so the
intervals include discovery variance on both sides; in (b) our taxonomy is held at the canonical fit while Local LLMs' fusion is varied over all $50$ draws, the $50\times50$ (fusion, corpus) differences being pooled; the intervals therefore carry corpus variance for both arms and fusion variance for Local LLMs, but no re-discovery for ours.

Conditioning instead on the canonical fusion (the first
$E{=}27$ elicitations) gives uniformly larger, all-significant gaps
(Tracked $+0.177$, Stat.R. $+0.080$, Sem.R. $+0.177$, Attrib. $+0.466$, LK $-0.016$), that fusion being among the weaker draws; the pooled figures in
Table~\ref{tab:paired} are the \textbf{conservative} choice. Across fusions, the mean differences span Tracked $[+0.052, +0.177]$, Stat.R. $[-0.003, +0.093]$, Sem.R. $[-0.010, +0.240]$, Attrib. $[+0.362, +0.505]$, LK $[-0.023, -0.007]$. Resampling the corpus also reduces our method's comparison budget: a replicate retains only judged pairs with both endpoints drawn ($1{,}603$ of $4{,}000$ on average, consistent with the ${\approx}0.632^{2}$ both-endpoints survival rate of an $n$-out-of-$n$ bootstrap), giving an average degree of $25.3$ against the designed $40$, on which our method reports $\approx 16$ features on average rather than $21$. \textbf{The advantage does not rest on the larger budget}: at degree $25$ the comparative arm costs ${\approx}\$2.24$, against $\$3.45$ at degree $40$ and $\$2.35$ for Local LLMs.

\subsubsection{Dials Recovered}
\label{app:dials-recovered}

\begin{figure}
    \centering
    \includegraphics[width=\linewidth]{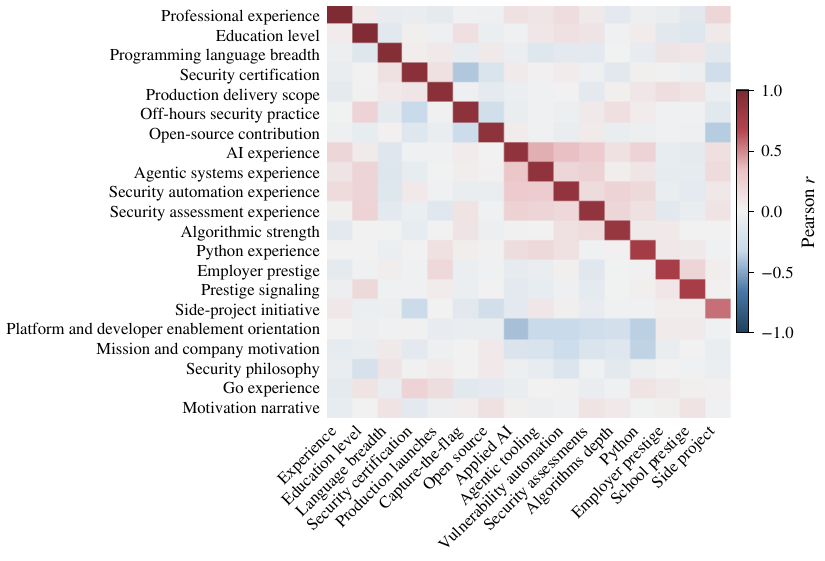}
    \caption{Dial-recovery heatmap yielded by the features discovered by our method. The rows are the discovered features and the columns are the dials.}
    \label{fig:recovery-heatmap-comparative}
\end{figure}

\begin{figure}
    \centering
    \includegraphics[width=\linewidth]{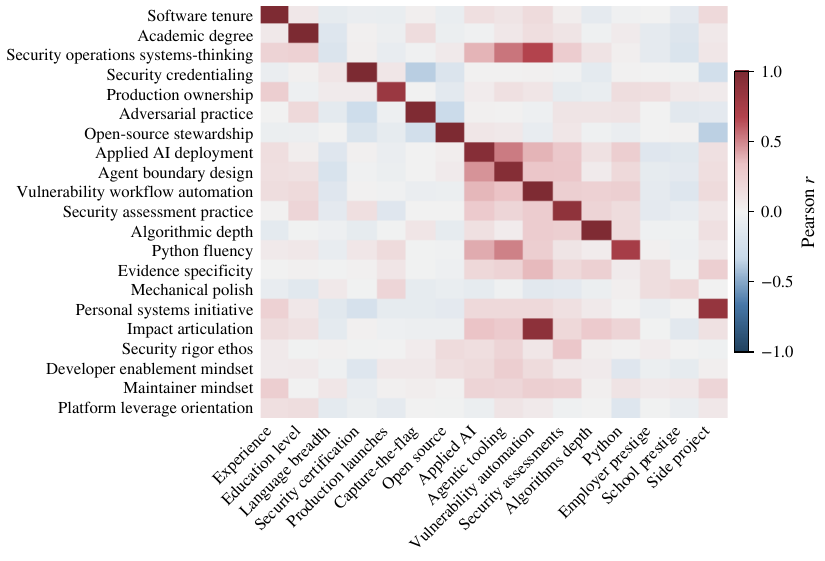}
    \caption{Dial-recovery heatmap yielded by the features discovered by Direct LLM. The rows are the discovered features and the columns the dials. The scores leak more between one another and the dial recovery is less clean than in Figure~\ref{fig:recovery-heatmap-comparative} as some dials are not recovered.}
    \label{fig:recovery-heatmap-m1}
\end{figure}

Figures~\ref{fig:recovery-heatmap-comparative} and~\ref{fig:recovery-heatmap-m1} compare the dial-recovery heatmap between our method and Direct LLM. As we can see, our method yields a strong diagonal recovery with only a few discovered features tracking no known dial, while Direct LLM clearly misses 3 dials and has dials tracked by several discovered features.

\begin{figure}
    \centering
    \includegraphics[width=\linewidth]{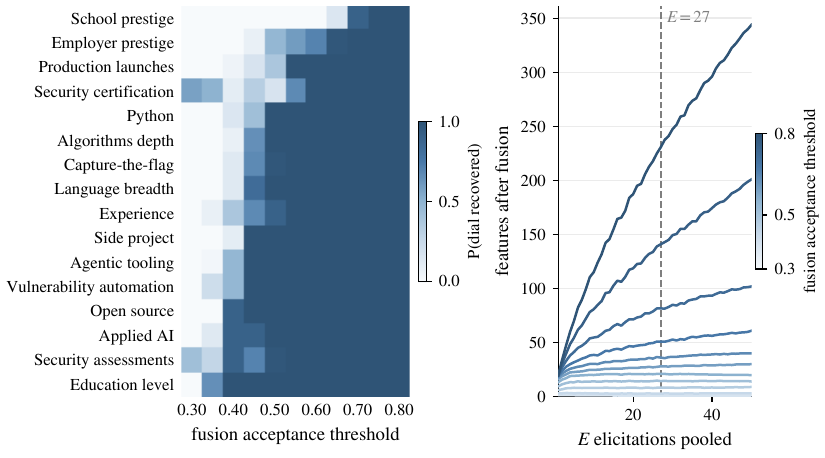}
    \caption{Left: probability that a dial is recovered as a function of the acceptance threshold used to merge similar features across the $E=27$ elicitations of Local LLMs (dials are matched to the merged features by thresholded Hungarian assignment at the calibrated $\eta$). Right: number of features surviving the merge, against the number of pooled elicitations, one curve per acceptance threshold. At $E=27$, recovering every dial demands a high threshold, which leaves an unusable number of near-duplicate features, hence the need for an LLM taxonomy.}
    \label{fig:threshold_pair}
\end{figure}

\begin{remark}
    \label{rem:why-taxo}
    As we can see in Figure~\ref{fig:threshold_pair}, after $E=27$ rubric elicitations, recovering every dial demands a high acceptance threshold, which leaves an unusable number of near-duplicate features. It is the same problem that led us to use LLM taxonomy instead of semantic clustering in our own standardization step: the judge naturally merges duplicate features and excludes idiosyncratic ones.
\end{remark}

\newpage

\section{Experiment Details on the ChangeMyView Dataset}
\label{app:experiment-details-cmv}

\subsection{Dataset}
\label{app:cmv-dataset}

\subsubsection{Samples used}

Within each discussion tree, \citet{tan2016winning} matched each $\Delta$-winning argument to the non-winning argument from the same tree whose opening reply is most topically similar, yielding a balanced task ($3{,}456$ training and $807$ held-out \emph{pairs} of $\Delta$-winning/non $\Delta$-winning answers) in which topic is held fixed and the goal is to identify the persuasive member of each pair. 

To mitigate length bias, we restricted to pairs whose two root replies both fall within $[20, 1200]$ words, leaving $3{,}411$ pairs for the training set and $800$ eligible held-out pairs. For our study, we used a sub-sample of $400$ pairs from the training set. 

\subsubsection{Memorization probe}
\label{app:cmv-contamination}

Because the CMV corpus predates the judge's training cutoff, a skeptical
reading of Table~\ref{tab:cmv-predictive} is that the judge recalls
which replies earned a $\Delta$ rather than inferring it. To test this, we use $1{,}600$ exchanges in four cells of $400$ ($\{\Delta, \text{no-}\Delta\} \times \{\text{train}, \text{held-out}\}$, base rate $50\%$ by construction). Each exchange is judged under two framings on identical de-identified text (Prompts~\ref{prm:cmv-contam-task} and~\ref{prm:cmv-contam-named}): an \emph{unnamed} framing (P1) asking
whether the reply changed the poster's view, with platform identifiers
scrubbed, and a \emph{named} framing (P2) that presents the same
scrubbed text but identifies the source as r/ChangeMyView and explains
the $\Delta$ mechanic. A memorized outcome is easier to retrieve when
the source is named, so the lift $\mathrm{P2} - \mathrm{P1}$ on
constant text is the recall signal.

\begin{table}[t]
\centering
\footnotesize
\setlength{\tabcolsep}{4pt}
\begin{tabular}{@{}l l@{}}
\toprule
Framing & Accuracy \\
\midrule
P1 (unnamed, scrubbed) & $53.6\%$ \scriptsize{$[51.2,\ 56.1]$} \\
P2 (named)             & $53.6\%$ \scriptsize{$[51.1,\ 56.1]$} \\
\midrule
Lift $\mathrm{P2}-\mathrm{P1}$ & $-0.1$\,pp \scriptsize{$[-2.4,\ +2.3]$} \\
\bottomrule
\end{tabular}
\caption{Memorization probe on $1{,}600$ CMV exchanges pinned to the
data of Table~\ref{tab:cmv-predictive} (judge: \textsc{GPT-5.4 mini};
chance $=50\%$). The lift is assessed by a paired McNemar test on the
same exchanges ($b/c = 185/184$, $p = 1.000$).}
\label{tab:cmv-contamination}
\end{table}

Naming the source buys the judge nothing: the lift is $-0.1$\,pp with a
$95\%$ confidence interval of $[-2.4, +2.3]$\,pp
(Table~\ref{tab:cmv-contamination}). Finally, the residual above-chance accuracy of
the unnamed framing ($53.6\%$, CI $[51.2, 56.1]$) is not
recall-shaped: it is driven by a strong asymmetry across the outcome
split ($69.4\%$ on no-$\Delta$ exchanges versus $37.9\%$ on $\Delta$
exchanges under P1), a systematic prior toward predicting ``no
change''. Consistently, naming the source (P2) does not sharpen either cell but shifts the threshold, raising $\Delta$-cell accuracy to $43$--$46\%$ while lowering no-$\Delta$ accuracy to $62$--$64\%$.

We conclude that the predictive result of Table~\ref{tab:cmv-predictive} is not explained by outcome recall: on the very exchanges it is measured on, the judge shows no recall signal under source naming.

\begin{promptbox}{Contamination probe -- shared task framing (P1; unnamed).}{prm:cmv-contam-task}
You are shown one online discussion. It begins with a person's stated
VIEW (a position they hold and are willing to reconsider), followed by
ONE reply from someone arguing against that view. Your job: decide
whether that reply actually changed the original person's view. Answer
strictly yes or no.
=== VIEW ===
(*@\ph{OP view, de-identified}@*)
=== REPLY ===
(*@\ph{root reply, de-identified}@*)
(*@\ph{output specification}@*)
\end{promptbox}

\begin{promptbox}{Contamination probe -- named-source preamble (P2).}{prm:cmv-contam-named}
This is a post from the r/ChangeMyView (CMV) forum on Reddit. There, an
original poster states a view and awards a "delta" ((*@$\Delta$@*)) to
any reply that succeeds in changing their view.
(*@\textnormal{\itshape [followed by the shared task framing of
Prompt~\ref{prm:cmv-contam-task}, on identical de-identified text]}@*)
\end{promptbox}

\subsection{Methods Details}
\label{app:methods-cmv}

This subsection gives all the details of each method.

\paragraph{Shared evaluation protocol.}
All arms score the same $2{,}400$ exchanges (each an (OP view, root reply) pair), which come from $1{,}200$ of Tan's matched pairs, each contributing the delta-winning reply and its matched loser: $400$ pairs ($800$ exchanges) drawn from the training split and all $800$ eligible held-out pairs ($1{,}600$ exchanges), with replies of $20$--$1{,}200$ words. Given a feature set's per-exchange scores, the paired task is an $\ell_1$ logistic regression on the within-pair score difference $\mathrm{sign}\cdot(s_{\text{winner}} - s_{\text{loser}})$ with no intercept, the sign dealt from a balanced random sign per pair so chance is exactly $50\%$. The penalty $\lambda$ in the $\ell_1$ model is selected by $5$-fold \textsc{GroupKFold} cross-validation grouped by original poster over $13$ log-spaced values $\lambda \in \{10^{-3}, \dots, 10^{3}\}$, fitted on the $400$ training pairs and evaluated once on the $800$ held-out pairs. Selected values are $\lambda = 10$ (D\,+\,BT), $10$ (D\,+\,PW), $31.6$ (D$_{\mathrm{r}}$\,+\,BT), $31.6$ (D$_{\mathrm{r}}$\,+\,PW), $31.6$ (P\,+\,BT), $3.2$ (P\,+\,PW), $3.2$ (S\,+\,PW), $31.6$ (Tan), $31.6$ (Emb.) and $31.6$ (\#words).

\paragraph{D: feature discovery (shared by D\,+\,BT, D\,+\,PW, D$_{\mathrm{r}}$\,+\,BT, D$_{\mathrm{r}}$\,+\,PW).} Features are mined from $150$ training pairs ($300$ exchanges) drawn from the training split only. The $300$ exchanges are arranged in a $40$-regular comparison graph built as a union of $40$ random perfect matchings, giving $6{,}000$ cross-topic comparisons performed by \textsc{GPT-5.4 mini} using Prompt~\ref{prm:cmv-comp}; a thread's own winner--loser edge is forbidden, so the judge never compares two replies to the same post. For each comparison the judge names, in free text, the dimensions on which the two exchanges differ in ways bearing on persuasion, with an evidence phrase and which side shows more of it ($7.65$ dimensions named per comparison; $45{,}894$ citations over $17{,}724$ distinct phrases). Phrases cited at least $10$ times are grouped into canonical features by one \textsc{GPT-5.4 (high reasoning)} call (Prompt~\ref{prm:cmv-taxo}), yielding $45$ features. Excluding OP-side features (they describe the shared post and so cancel within a matched pair), the catch-all ``Other'', and one singleton label leaves the \textbf{29 challenger-side features} used in the (D) feature set. Fitting the statistical model of Section~\ref{sec:method} and keeping features having $\rho \geq \rho_{\min} := 0.75$ and $\kappa \leq \kappa_{\max} := 50$ yields the feature set D$_{\mathrm{r}}$ of 16 features. Feature set: Table~\ref{tab:cmv-win-rates}.

\paragraph{P: prior-elicited rubric (shared by P\,+\,BT and P\,+\,PW).}
A 16-feature set is elicited from the model's prior, with no data. \textsc{GPT-5.4 (high reasoning)} is given a two-sentence setting (that people state views online and others reply to change them) and asked to name dimensions for assessing persuasiveness, each a matter of degree and distinct from the others, with a name and a one-sentence definition (Prompt~\ref{prm:cmv-prior-elicit}). Asked instead to choose the count itself, the model produces $15.65 \pm 1.42$ features over two independent runs of $10$ draws, so for a canonical run, we asked it to produce 16 dimensions, which is hence not a binding constraint and matches the size of the D$_{\mathrm{r}}$ set. Feature set: Table~\ref{tab:cmv-features-p}.

\paragraph{S: sample-elicited features (S\,+\,PW).}
A feature set is elicited from the same model after it has seen data. In a single call, \textsc{GPT-5.4 (high reasoning)} reads $752$ pairs ($1{,}504$ replies, about $860$k tokens of input), each presented as an OP view followed by two replies written independently to it, and is asked to name the dimensions along which the replies differ as attempts to change the poster's view, giving for each a short evidence phrase, a name and a one-sentence definition (Prompt~\ref{prm:cmv-sample-elicit}). Features must be properties of a reply rather than of the poster's view, so that they can vary within a matched pair. The number of features is left to the model: with the same open-count instruction but no data, it names about $16$ (see P); with data, it names $42$. The exchanges keep one pair per distinct original poster, cover the range of reply lengths, and are drawn from both the training and the held-out pairs, without their labels. Feature set: Table~\ref{tab:cmv-features-s}.

\paragraph{D\,+\,BT and P\,+\,BT: pairwise Bradley--Terry scoring.} Each feature is scored by its own Bradley--Terry model on a frozen comparison graph (Prompt~\ref{prm:cmv-score-bt}). The $800$ training exchanges form a $16$-regular anchor graph ($6{,}400$ comparisons); each of the $1{,}600$ test exchanges is then compared against $16$ anchors ($25{,}600$ comparisons), for $32{,}000$ judge calls. Every call presents two exchanges with the full feature list and forces a winner on \emph{each} feature, with a short evidence phrase. Anchor scores are fitted per feature by plain Bradley--Terry with a position-bias term $\beta$ (ridge $10^{-3}$); each test exchange is then scored by a one-dimensional convex solve against the \emph{frozen} anchor scores, so test items never influence each other. P\,+\,BT uses a graph verified identical to D\,+\,BT's, so the two arms differ only in the features named in the prompt.

\paragraph{D\,+\,PW, D$_{\mathrm{r}}$\,+\,PW and P\,+\,PW: pointwise scoring.}
The same features are instead rated absolutely (Prompt~\ref{prm:cmv-score-pw}): one call per exchange rates it $0$--$10$ on every feature so the feature set costs $2{,}400$ calls rather than $32{,}000$. The prompt is a deliberate near-copy of the pairwise one; two differences are unavoidable and consequential. A scale must be defined, anchored on the population of replies, because ratings must be comparable across exchanges rated in separate calls --- pairwise never needs this, which is precisely the property this arm gives up; and the prompt must push against central tendency, since without it ratings pile up at $5$--$7$ and the arm loses to compression rather than to anything substantive. Because a pointwise score is a single draw, each feature set was additionally rated a \emph{second} time with the features reshuffled into different order, which supplies the independent replicate the noise correction of Figure~\ref{fig:cmv-dimension-curve} requires; the same replicate exists for the BT arms for free, by refitting on disjoint halves of the $32{,}000$ comparisons. The accuracy reported in Table~\ref{tab:cmv-predictive} is the one attained by the average of these two scores.

\subsection{Additional results}
\label{app:additional-results}

\subsubsection{Additional measures}
\label{app:additional-measures}

\begin{figure}
    \centering
    \includegraphics[width=\linewidth]{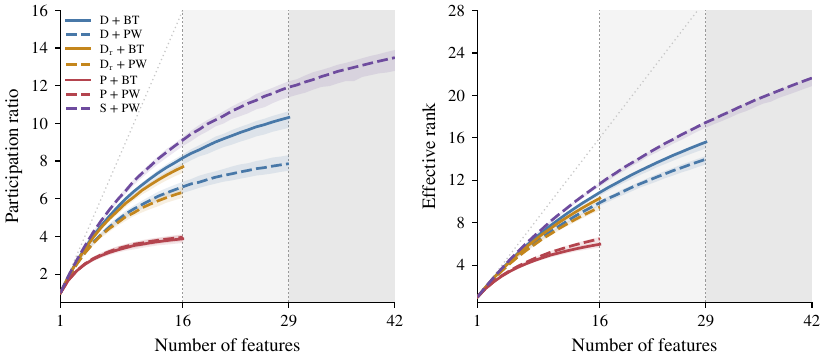}
   \caption{\textbf{Both dimensionality measures agree on the ordering of the feature sets.} Noise-corrected effective dimensionality against feature set size $k$, averaged over $k$-subsets of each feature set. \emph{Left:} participation ratio, as in Figure~\ref{fig:cmv-dimension-curve}. \emph{Right:} effective rank~\citep{roy2007erank}, $\exp H(p)$ with $p_i = \sigma_i / \sum_j \sigma_j$ the normalized singular values. Both are computed on the disattenuated correlation matrix. The order holds under both measures. Bands, shading and dotted line as in Figure~\ref{fig:cmv-dimension-curve}.}
    \label{fig:cmv-dimension-curve-full}
\end{figure}

This subsection gathers additional results, notably accuracy measures (Table~\ref{tab:cmv-predictive-2}) and dimensionality measures (Figure~\ref{fig:cmv-dimension-curve-full}). 

\paragraph{Other runs.} Table~\ref{tab:cmv-predictive-2} shows results about other runs that were made. Methods that specify ``full'' were fitted on the whole training set. \textbf{BOW} and \textbf{POS} are two surface baselines of \citet{tan2016winning}, built identically except for what a reply is reduced to: BOW represents each reply by its bag of lowercased word tokens (vocabulary fitted on the training replies, tokens occurring more than five times kept), while POS discards the words entirely and represents the reply by the sequence of its fine-grained Penn part-of-speech tags as assigned by spaCy, so that only syntactic style remains ($42$ tags).

\begin{figure}[t]
  \centering
  \includegraphics[width=\linewidth]{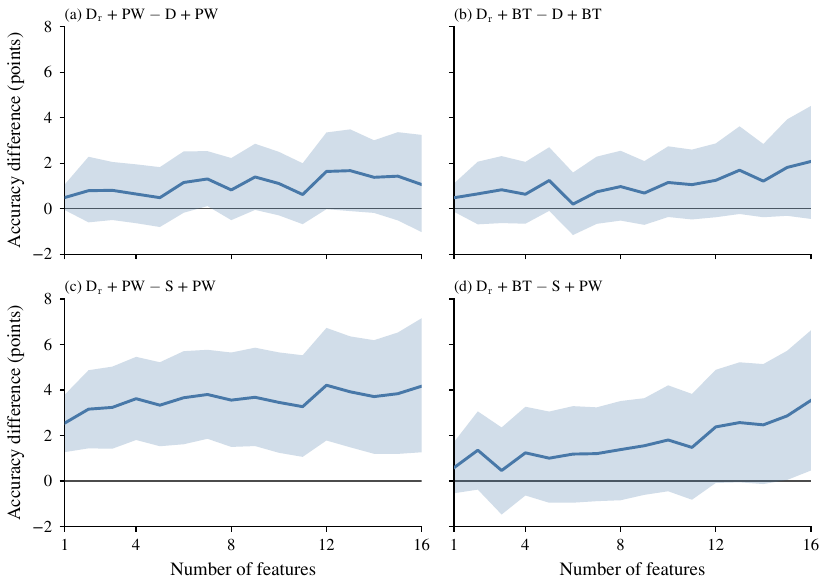}
  \caption{\textbf{The reliable discovered features are more predictive, feature for feature, than the full discovered set and than the sample-elicited set.} Each panel shows, as a function of the number of features $k$, the difference in held-out accuracy between D$_{\mathrm{r}}$ (the $16$ reliable discovered features) and another feature set. At each $k$, an $\ell_1$-regularized logistic regression is fitted on up to $50$ random $k$-subsets of each set (all subsets when there are fewer), and accuracy is averaged over subsets; PW scores are averaged over two scoring runs. Shaded: $95\%$ bootstrap interval at each $k$ ($4{,}000$ resamples of the $800$ held-out pairs and of the drawn subsets). (a, b) Against the full discovered set D, with the same scoring: D$_{\mathrm{r}}$ is ahead at every $k$, by $1.0$ point on average over $k$ under both scorings ($95\%$ CIs $[-0.1, 2.2]$ for PW, $[0.0, 2.1]$ for BT). (c, d) Against the sample-elicited set S, scored pointwise only: D$_{\mathrm{r}}$\,+\,PW is ahead at every $k$, by $3.6$ points on average ($[1.8, 5.4]$); D$_{\mathrm{r}}$\,+\,BT's advantage grows with $k$, reaching $3.5$ points at $k = 16$ ($[0.5, 6.6]$).}
  \label{fig:cmv-diff-dr}
\end{figure}

As we can see, D + PW and D$_{\mathrm{r}}$ + PW rank among the best methods despite their smaller training and feature set size, beaten only by the opaque embeddings trained on the whole training set and by \citet{tan2016winning}'s features, while being interpretable. Additionally, all four panels of Figure~\ref{fig:cmv-diff-dr} show that the reliable discovered features are more predictive, at matched feature count, than either the full discovered set or the sample-elicited set: the reliability filter of the statistical model selects the features that carry the signal.

\begin{table}[t]
\centering
\small
\setlength{\tabcolsep}{6pt}
\begin{tabular}{@{}l r r c@{}}
\toprule
Arm & Dims & Accuracy & $95\%$ CI \\
\midrule
Emb. (full) & 3{,}072 & 64.88\% & $[61.50, 68.00]$ \\
Tan & 3{,}205 & 63.75\% & $[60.38, 67.00]$ \\
D + PW & 29 & 62.63\% & $[59.38, 66.00]$ \\
D$_{\mathrm{r}}$ + PW & 16 & 62.13\% & $[58.88, 65.50]$ \\
Emb. & 3{,}072 & 61.63\% & $[58.25, 65.00]$ \\
D$_{\mathrm{r}}$ + BT & 16 & 61.50\% & $[58.13, 64.88]$ \\
D + BT & 29 & 61.38\% & $[58.00, 64.75]$ \\
BOW (full) & 11{,}348 & 61.25\% & $[57.88, 64.63]$ \\
P + PW & 16 & 61.25\% & $[58.00, 64.63]$ \\
P + BT & 16 & 61.13\% & $[57.75, 64.38]$ \\
POS (full) & 42 & 60.50\% & $[57.00, 63.75]$ \\
S + PW & 42 & 59.88\% & $[56.50, 63.25]$ \\
\#words & 1 & 59.25\% & $[55.88, 62.75]$ \\
Zero-shot & -- & 56.94\% & $[54.19, 59.88]$ \\
\bottomrule
\end{tabular}
\caption{\textbf{Held-out accuracy of each method.} D$_{\mathrm{r}}$: the reliable subset of the discovered features D. Emb.\ denotes \textsc{text-embedding-3-large} embeddings. Arms marked (full) are trained on all $3{,}411$ training pairs; all others on $400$. Accuracies for PW arms are obtained from the average of two scoring runs. Zero-shot asks the judge directly to predict the $\Delta$-winning reply, with the query issued in both presentation orders and the two predictions averaged. Chance $=50\%$; $95\%$ bootstrap percentile intervals ($4{,}000$ resamples of the $800$ held-out pairs, shared across arms).}
\label{tab:cmv-predictive-2}
\end{table}

\paragraph{Measuring effective dimensionality.}
We ask how many independent directions of variation a feature set resolves, using the participation ratio $\mathrm{PR} = (\sum_i \lambda_i)^2 / \sum_i \lambda_i^2$ of the eigenvalues $\lambda$ of the feature set's score correlation matrix over the $1{,}600$ held-out exchanges. $\mathrm{PR} \in [1, k]$, equalling $1$ when all $k$ features are one construct and $k$ when they are mutually independent. 

Three choices make the comparison meaningful. \textbf{a) Rank correlations.} Bradley--Terry scores live on an arbitrary logit scale and pointwise ratings on a coarse $0$--$10$ one, and a covariance-based participation ratio weights a feature by its variance; we therefore use the Spearman correlation matrix, which is invariant to scale and to any monotone reparameterization. 

\textbf{b) Matched feature set size.} A feature set with more features trivially has more room for independent variation, so for each $k$ we average the participation ratio over $k$-subsets of the feature set --- enumerated exactly while $\binom{K}{k} \le 3{,}000$ and sampled ($3{,}000$ draws) otherwise --- and read comparisons at matched $k$. 

\textbf{c) Measurement noise correction.}  The participation ratio counts independent directions and cannot distinguish signal from independent measurement error, so a noisier scorer is credited with extra dimensions; since a Bradley--Terry score aggregates $16$ comparisons while a pointwise score is a single coarse draw, the uncorrected comparison is biased in favor of pointwise. We therefore measure each feature set twice, independently: for the pairwise arms by refitting on disjoint halves of the $32{,}000$ comparisons (the anchor graph split in two, and each test exchange's $16$ anchor comparisons split $8$/$8$, so the halves share no data), and for the pointwise arms by a second scoring run with the features reshuffled into different bundles and order. 

Writing $C_{ij} = \mathrm{corr}(A_i, B_j)$ for the cross-measurement correlation, the independent errors cancel in expectation, so that $C_{ii}$ estimates feature $i$'s reliability and $R_{ij} = S_{ij} / \sqrt{S_{ii} S_{jj}}$ with $S = (C + C^\top)/2$ disattenuates the correlation between the latent features. The participation ratio is then computed on $R$. The correction is substantial and asymmetric for the full feature sets: it removes $15$--$20\%$ of the raw participation ratio for the pairwise arms and $25$--$28\%$ for the pointwise ones; for the reliable subset D$_{\mathrm{r}}$ it is smaller and nearly symmetric ($11\%$ under BT, $13\%$ under PW). Reliabilities are high enough for the division to be safe (median $0.83$ and $0.77$ for the D and P pointwise score sets; $0.76$ and $0.71$ for the half-data Bradley--Terry fits, i.e.\ $0.86$ and $0.83$ for the full score sets by the Spearman--Brown correction), and no disattenuated matrix carried any negative eigenvalue mass. Bands are $95\%$ percentile bootstrap over the $1{,}600$ exchanges, with the cross-measurement correlation rebuilt inside each of $200$ replicates.

\subsubsection{Discovered Features and Winning Argumentation Techniques}
\label{app:discovered-features-cmv}

\begin{figure}
    \centering
    \includegraphics[width=\linewidth]{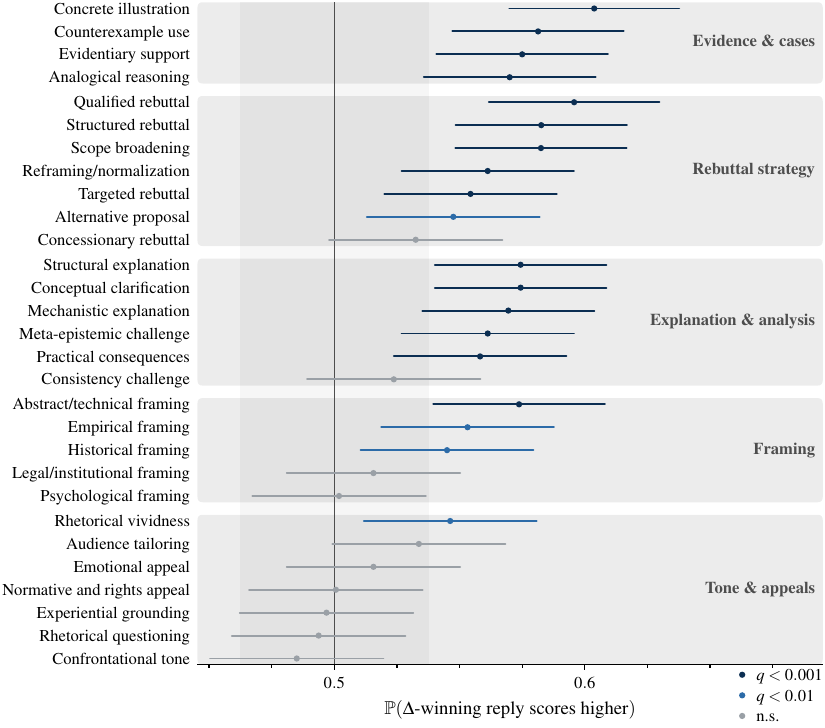}
    \caption{\textbf{Effectiveness of the discovered argumentation techniques (D + PW).} The figure is the complete version of Figure~\ref{fig:cmv-win-rates} and shows each discovered feature's \emph{effectiveness}, i.e. the fraction of the test pairs in which the reply that earned the $\Delta$ scores higher on that feature. $0.5$ is chance. Brackets give $95\%$ Wald intervals and $q$ is the Benjamini--Hochberg false-discovery rate across the $29$ tests. The discovered features are gathered by category. Exact values of the BH $q$ and each feature's description can be found in Table~\ref{tab:cmv-win-rates}.}
    \label{fig:cmv-win-rates-full}
\end{figure}

This subsection is devoted to additional results on the discovered features: Figure~\ref{fig:cmv-win-rates-full} completes Figure~\ref{fig:cmv-win-rates}, while Table~\ref{tab:cmv-win-rates} lists each discovered feature with its description, together with its effectiveness. 

\begin{table}[!htbp]
\centering
\scriptsize
\setlength{\tabcolsep}{4pt}
\renewcommand{\arraystretch}{1.05}
\begin{tabularx}{\textwidth}{@{}>{\raggedright\arraybackslash}p{0.20\textwidth} X r r@{}}
\toprule
Feature & Description & Win rate (95\% CI) & $q$ \\
\midrule
\bfseries Concrete illustration$^{*}$ & The argument makes its case through specific, vivid, or worked-through examples rather than abstraction alone. & 0.604 {\tiny$[0.570, 0.638]$} & $<$0.001 \\
\bfseries Qualified rebuttal$^{*}$ & The challenger narrows the claim, marks exceptions, or emphasizes contextual dependence instead of arguing absolutely. & 0.596 {\tiny$[0.562, 0.630]$} & $<$0.001 \\
\bfseries Structured rebuttal development$^{*}$ & The challenger builds an organized, multi-step, or multi-pronged case rather than a single bare point. & 0.583 {\tiny$[0.548, 0.617]$} & $<$0.001 \\
\bfseries Scope broadening & The exchange expands or contracts the discussion by moving between narrow cases and wider principles or issue areas. & 0.583 {\tiny$[0.548, 0.617]$} & $<$0.001 \\
\bfseries Counterexample use$^{*}$ & The challenger undermines the OP with exceptions, alternative cases, or edge-case scenarios that break the generalization. & 0.581 {\tiny$[0.547, 0.616]$} & $<$0.001 \\
\bfseries Challenger evidentiary support$^{*}$ & The challenger backs the rebuttal with external evidence, citations, factual corrections, or expert authority. & 0.575 {\tiny$[0.541, 0.609]$} & $<$0.001 \\
\bfseries Structural explanation & The challenger explains the issue through broader social, cultural, or systemic forces rather than isolated choices. & 0.574 {\tiny$[0.540, 0.609]$} & $<$0.001 \\
\bfseries Conceptual clarification$^{*}$ & The challenger disputes definitions, categories, meanings, or criteria in order to reframe the disagreement. & 0.574 {\tiny$[0.540, 0.609]$} & $<$0.001 \\
\bfseries Abstract or technical framing$^{*}$ & The exchange reasons at a more theoretical, philosophical, scientific, or specialized level rather than in everyday concrete terms. & 0.574 {\tiny$[0.539, 0.608]$} & $<$0.001 \\
\bfseries Analogical reasoning$^{*}$ & The argument persuades by comparing the case to parallel real or hypothetical cases, including extended analogies and thought experiments. & 0.570 {\tiny$[0.536, 0.604]$} & $<$0.001 \\
\bfseries Mechanistic explanation$^{*}$ & The challenger explains or critiques the causal mechanism by which the disputed outcome would arise. & 0.569 {\tiny$[0.535, 0.604]$} & $<$0.001 \\
\bfseries Meta-level epistemic challenge & The challenger contests the framing, method, evidence standards, or burden of proof rather than only the first-order claim. & 0.561 {\tiny$[0.527, 0.596]$} & $<$0.001 \\
\bfseries Reframing and normalization & The challenger tries to shift the evaluative lens by redescribing the issue, normalizing it, or offering an alternative way to view it. & 0.561 {\tiny$[0.527, 0.596]$} & $<$0.001 \\
\bfseries Practical consequences and feasibility$^{*}$ & The challenger evaluates the view in terms of real-world effects, costs, risks, tradeoffs, or implementability. & 0.558 {\tiny$[0.524, 0.593]$} & $<$0.001 \\
\bfseries Targeted rebuttal & The challenger engages the OP's specific claims, premises, or subpoints directly and often one by one. & 0.554 {\tiny$[0.520, 0.589]$} & $<$0.001 \\
\bfseries Exchange empirical framing$^{*}$ & The exchange is framed in empirical or quantitative terms, emphasizing measurement, data, or evidentiary specificity. & 0.553 {\tiny$[0.519, 0.588]$} & 0.002 \\
\bfseries Alternative proposal and guidance$^{*}$ & The challenger offers concrete substitute policies, remedies, or practical steps instead of only criticizing. & 0.547 {\tiny$[0.513, 0.582]$} & 0.006 \\
\bfseries Rhetorical vividness & The argument uses especially vivid phrasing, storytelling, or memorable imagery to make the point stick. & 0.546 {\tiny$[0.512, 0.581]$} & 0.007 \\
\bfseries Historical framing and legacy & The argument appeals to historical context, precedent, innovation, or long-run cultural legacy. & 0.545 {\tiny$[0.510, 0.580]$} & 0.003 \\
Audience tailoring & The challenger tailors the argument to particular subgroups, audiences, or the OP's specific situation. & 0.534 {\tiny$[0.499, 0.568]$} & 0.0502 \\
Concessionary rebuttal & The challenger softens resistance by explicitly granting part of the OP's point before disagreeing. & 0.532 {\tiny$[0.498, 0.567]$} & 0.070 \\
Consistency challenge$^{*}$ & The reply pressures the OP by showing that their own standard, principle, or rule yields inconsistency or self-undermining conclusions. & 0.524 {\tiny$[0.489, 0.558]$} & 0.195 \\
Legal and institutional framing & The challenger grounds the rebuttal in law, institutions, official roles, or procedural distinctions. & 0.516 {\tiny$[0.481, 0.550]$} & 0.369 \\
Emotional appeal and empathy & The reply persuades through feelings, empathy, reassurance, or existential/personal significance. & 0.516 {\tiny$[0.481, 0.550]$} & 0.412 \\
Psychological framing and diagnosis & The challenger interprets views or behavior in terms of motives, traits, psychology, or developmental state. & 0.502 {\tiny$[0.467, 0.537]$} & 0.966 \\
Normative and rights appeal$^{*}$ & The challenger argues from moral principle, fairness, rights, duty, or harm rather than only from descriptive claims. & 0.501 {\tiny$[0.466, 0.535]$} & 1.000 \\
Challenger experiential grounding$^{*}$ & The challenger or exchange relies on firsthand experience, self-disclosure, or identity-based testimony as support. & 0.497 {\tiny$[0.462, 0.532]$} & 0.907 \\
Rhetorical questioning & The challenger presses the point through pointed questions or requests for clarification rather than only direct assertion. & 0.494 {\tiny$[0.459, 0.528]$} & 0.773 \\
Confrontational tone$^{*}$ & The reply relies on combative, loaded, mocking, or emotionally aggressive rhetoric. & 0.485 {\tiny$[0.450, 0.520]$} & 0.427 \\
\bottomrule
\end{tabularx}
\caption{\textbf{Which argumentation techniques separate winners from losers (D + PW).} This table details the results shown in Figures~\ref{fig:cmv-win-rates} and~\ref{fig:cmv-win-rates-full}. It presents each discovered feature (its name and description) together with its \emph{effectiveness}, i.e. the fraction of the test pairs in which the reply that earned the $\Delta$ scores higher on that feature. $0.5$ is chance. Brackets give $95\%$ Wald intervals and $q$ is the Benjamini--Hochberg false-discovery rate across the $29$ tests; features in bold satisfy $q<0.05$. $19$ of the $29$ features separate winners, all of them positively. Features marked $^{*}$ form the reliable subset D$_{\mathrm{r}}$ ($16$ features, $\kappa \le 50$ and $\rho \geq 0.75$).}
\label{tab:cmv-win-rates}
\end{table}

\begin{table}[!htbp]
\centering
\scriptsize
\setlength{\tabcolsep}{4pt}
\renewcommand{\arraystretch}{1.05}
\begin{tabularx}{\textwidth}{@{}>{\raggedright\arraybackslash}p{0.20\textwidth} X@{}}
\toprule
Feature & Description \\
\midrule
Core Claim Relevance & It measures the degree to which the reply focuses on the central claim under dispute rather than drifting to side issues. \\
Reason Engagement & It measures the degree to which the reply directly addresses the original poster's stated reasons, premises, or assumptions. \\
View Understanding & It measures the degree to which the reply accurately represents the other person's position before challenging it. \\
Logical Coherence & It measures the degree to which the reply's conclusions follow clearly and consistently from its premises. \\
Evidence Strength & It measures the degree to which the reply supports its claims with substantial, pertinent evidence or examples. \\
Source Trustworthiness & It measures the degree to which the reply relies on credible, authoritative, and verifiable sources when offering support. \\
Concrete Specificity & It measures the degree to which the reply uses precise details or examples instead of vague generalities. \\
Counterargument Handling & It measures the degree to which the reply anticipates and answers plausible objections to its own case. \\
Nuance and Balance & It measures the degree to which the reply acknowledges complexity, uncertainty, or limits rather than overstating its case. \\
Novel Insight & It measures the degree to which the reply offers a fresh framing, distinction, or information the reader is unlikely to have already considered. \\
Value Alignment & It measures the degree to which the reply connects its argument to the other person's stated values, goals, or concerns. \\
Consequence Salience & It measures the degree to which the reply makes the practical or moral implications of the issue vivid and important. \\
Emotional Resonance & It measures the degree to which the reply evokes fitting emotions that make its point more compelling. \\
Respectful Tone & It measures the degree to which the reply treats the other person charitably and without dismissiveness or contempt. \\
Clarity of Expression & It measures the degree to which the reply is easy to follow because its wording and structure are clear. \\
Economy of Argument & It measures the degree to which the reply makes its case efficiently without unnecessary length or repetition. \\
\bottomrule
\end{tabularx}
\caption{\textbf{Prior-elicited features (P).} The $16$ features named by \textsc{GPT-5.4 (high reasoning)} for assessing the persuasiveness of a reply, with no data shown (Prompt~\ref{prm:cmv-prior-elicit}), in the order produced.}
\label{tab:cmv-features-p}
\end{table}

\begin{table}[!htbp]
\centering
\scriptsize
\setlength{\tabcolsep}{4pt}
\renewcommand{\arraystretch}{1.05}
\begin{tabularx}{\textwidth}{@{}>{\raggedright\arraybackslash}p{0.20\textwidth} X@{}}
\toprule
Feature & Description \\
\midrule
Topical responsiveness & How directly a reply addresses the poster's actual view rather than pivoting to meta-issues or adjacent topics. \\
Premise targeting & Whether a reply attacks the core assumption or inference supporting the view instead of focusing on side details. \\
Clarification demand & How much a reply advances the discussion by asking the poster to specify vague claims, terms, or examples. \\
Conceptual reframing & Whether a reply changes the frame of the debate by redefining the key concept or explaining that the dispute is really about something else. \\
Thought experiment & Whether a reply uses an invented scenario to test the implications or consistency of the poster's view. \\
Analogy use & How much a reply explains its point by mapping the issue onto a different familiar domain. \\
Consistency test & Whether a reply probes the poster's logic by applying it to parallel cases and checking whether the result is acceptable. \\
Reductio extension & Whether a reply undermines a view by extending it to an absurd or unacceptable conclusion. \\
Counterexample case & Whether a reply primarily challenges a general claim by presenting a concrete case that does not fit it. \\
Firsthand testimony & How much a reply relies on the writer's own experiences or preferences as evidence or illustration. \\
Stakeholder perspective & Whether a reply argues from direct membership in the affected group or a role that gives special practical exposure to the issue. \\
Empirical citation & How much a reply supports its claims with studies, data, or directly linked factual evidence. \\
Expert consensus appeal & Whether a reply leans on the agreement of qualified experts as a reason to update the view. \\
Legal mechanics & How much a reply explains the governing law, procedure, or institutional rules behind the issue. \\
Implementation realism & Whether a reply evaluates how workable a proposal would be once logistics, administration, and real-world constraints are considered. \\
Economic reasoning & How much a reply argues in terms of prices, incentives, resources, employment, or market effects. \\
Quantified argument & Whether a reply strengthens its case by attaching numerical magnitude or rates to the claims it makes. \\
Causal mechanism & Whether a reply explains the process by which one thing is supposed to produce another, rather than merely asserting a correlation or outcome. \\
Function reframing & Whether a reply challenges the view by arguing that the institution or behavior serves a different purpose than the poster assumes. \\
Use-case focus & Whether a reply argues from the specific practical situations in which the disputed thing works especially well or badly. \\
Institutional purpose & How much a reply justifies or criticizes something by appealing to the purpose of the larger system it belongs to. \\
Common-ground signaling & Whether a reply explicitly agrees with part of the poster's position before presenting disagreement. \\
Empathetic tone & Whether a reply validates the poster's feelings or perspective in a supportive way while disagreeing. \\
Motive attack & Whether a reply questions the poster's motives, sincerity, or openness instead of mainly addressing the substance of the claim. \\
Confrontational tone & How much a reply uses insult, scorn, or aggressive language toward the poster or target. \\
Humor deployment & Whether a reply uses jokes, playful phrasing, or irony as part of its persuasive approach. \\
Declared stance framing & Whether a reply explicitly frames itself as devil's advocate, partial agreement, or some other interpretive stance before arguing. \\
Epistemic caution & How much a reply qualifies its claims, notes uncertainty, or avoids overstating what it can prove. \\
Conditional reasoning & Whether a reply limits its claim to specific conditions or exceptions instead of treating the issue as all-or-nothing. \\
Abstraction level & Whether a reply operates mainly at a high-level philosophical or theoretical plane rather than a concrete one. \\
Explanatory correction & Whether a reply mainly persuades by supplying overlooked background information or correcting a misunderstanding. \\
Personal remedy suggestion & Whether a reply addresses the issue by recommending a practical step the poster could take in their own life. \\
Audience reframing & Whether a reply changes the evaluation of the thing in question by showing it was made for a different audience or goal than the poster assumes. \\
Scope challenge & Whether a reply argues that the poster's principle is too narrow to handle the full range of relevant cases. \\
Temporal horizon & Whether a reply evaluates the claim by emphasizing long-term effects rather than immediate ones, or vice versa. \\
Norm distinction & Whether a reply distinguishes between what is logically valid, socially useful, morally justified, or legally allowed rather than treating them as the same question. \\
Tradition appeal & Whether a reply supports or critiques a position by invoking established custom, continuity, or inherited social practice. \\
Privacy sensitivity & Whether a reply foregrounds the need to protect personal vulnerability, anonymity, or intimate information when evaluating the issue. \\
Collective agency framing & Whether a reply explains outcomes as the aggregate result of many individuals' choices rather than the intent of a single actor. \\
Resource referral & Whether a reply mainly tries to persuade by handing off a longer external resource instead of making the argument in the reply itself. \\
Form substance split & Whether a reply distinguishes between objection to labels, rhetoric, or presentation and objection to the underlying idea itself. \\
Meta-argumentation & Whether a reply persuades by analyzing the quality or structure of the argument itself rather than primarily debating the topic. \\
\bottomrule
\end{tabularx}
\caption{\textbf{Sample-elicited features (S).} The $42$ features named by \textsc{GPT-5.4 (high reasoning)} after reading a sample of $752$ exchange pairs, without their outcomes.}
\label{tab:cmv-features-s}
\end{table}

\clearpage

\subsubsection{Prompts}

\begin{promptbox}[listing options app={moredelim={[is][\color{blue!55!black}]{@@}{@@}}}]{Comparative method feature-discovery, CMV corpus.}{prm:cmv-comp}
You are helping build an interpretable framework for what drives persuasion in debate, by examining how two attempts to change someone's mind differ. The framework works by collecting many pairwise comparisons: for each pair, the ways the two exchanges differ are articulated in plain terms, and aggregated across many pairs these reveal the distinct dimensions along which persuasion-relevant material varies. There is no fixed rubric — the dimensions are recovered from comparisons like this one, not specified in advance.
Background. Each item is an online discussion in which an original poster (the "OP") states a view they currently hold and invites others to change it; a challenger then replies with an argument aimed at changing that specific view. You will be shown two such exchanges, Exchange A and Exchange B, each presented as a block that begins with the OP's view (marked [OP VIEW]) followed by the challenger's reply (marked [CHALLENGE]). Crucially, the two exchanges concern different posters holding different views on different topics.

Your task. Identify the dimensions on which the two exchanges meaningfully differ in ways that could bear on whether the OP's view gets changed. Consider the whole exchange: both the challenger's argument and the OP's post itself. Report the differences that genuinely stand out to you on reading this pair — bring up whatever you find salient, and do not work from a predetermined list. Favor real, noticeable differences over minor ones, but do not force the list to be short: report as many as actually distinguish these two exchanges.
For each dimension, first name the concrete, specific feature you actually saw — a short phrase pointing to the specific evidence. Then, grounded in that evidence, name the dimension, describe in one sentence what it captures, and indicate which exchange shows more of it.
Judge each dimension on its own terms, from its own evidence. Exchanges have uneven profiles: it is entirely normal for one to be stronger on some dimensions and the other stronger on others, so let the exchange that shows more of each dimension fall wherever the evidence for that dimension points.
Out of scope — do not treat these as dimensions of difference:
- the topic or subject matter either exchange is about, and how interesting, important, difficult, or agreeable either OP's view is: the two exchanges concern different topics by design, so topic or subject-matter difference is never a dimension here;
- which side you personally agree with, or whether you think either OP's view is correct;
- the username or any other identifier of either poster or challenger;
- length in itself
How to report. List each dimension as exactly one item — do not combine dimensions with "and", "/", "&", or a comma; split related dimensions into separate items. For the evidence field, give a short phrase naming the concrete feature you saw. For the dimension field, give a concise general noun phrase naming the dimension (the kind of short label you might use as a rubric column). For the description field, give one sentence saying what that dimension captures. Indicate which exchange shows more of it.
Output only JSON of the form: {"differences": [{"evidence": "<short phrase>", "dimension": "<concise noun phrase>", "description": "<one sentence>", "winner": "A" | "B"}, ...]}

Exchange A: {arg_a}
Exchange B: {arg_b}
Following the standard above, list the dimensions on which these two exchanges differ in ways bearing on persuasion — whether in the challenger's argument or in the OP's post — for each giving the concrete feature you saw, the dimension, a one-sentence description, and which exchange shows more of it. Compare on persuasion-relevant qualities, not topic. Respond with the JSON object only.
\end{promptbox}

\begin{promptbox}{Comparative method, CMV corpus -- feature canonicalization prompt.}{prm:cmv-taxo}
You are organizing a flat list of evaluation dimensions into a set of canonical features.

These dimensions were produced by judges comparing pairs of online exchanges — in each, an original poster (the "OP") states a view and a challenger replies with an argument aimed at changing it. On each comparison a judge named the dimensions on which two exchanges differed in ways bearing on whether the OP's view gets changed; these span both the challenger's argument and the OP's own post. Aggregated across many comparisons, the same underlying construct was named in many slightly different ways. Your job is to group the different phrasings of the SAME underlying construct together.

Below is the full list. Each entry has an INDEX in square brackets, a short NAME, a one-sentence DESCRIPTION of what it captures, and CITATIONS (how many times it was named — for your awareness only).

(*@\ph{indexed dimension list}@*)

Instructions:
- Group entries that denote the SAME underlying construct, judging by MEANING, not by surface wording. Different words for the same idea belong in one group; the same word used for different ideas belongs in different groups.
- A dimension describing the CHALLENGER's argument and a dimension describing the OP's POST are DIFFERENT constructs even when they share a word — group by the specific feature and whose it is, keeping OP-side and argument-side constructs in separate groups.
- Do NOT force everything into a small number of groups, but do NOT split one construct into its facets either. A dimension that does not clearly share a construct with any other — including a framing or appeal specific to one debate's subject matter that does not recur as a general persuasion feature — does NOT become its own feature: place all such genuinely-apart, idiosyncratic dimensions together in a single catch-all group whose canonical name is exactly "Other". Do not create many single-member groups, and do not absorb a genuinely-apart dimension into a loosely-related group to tidy up — send it to "Other".
- Do NOT merge two constructs just because they often co-occur or because persuasive exchanges tend to show both. Group only by shared MEANING, never by how exchanges might score.
- Citations are context only. Do not merge a frequently-named dimension into a rarely-named one or vice versa on the basis of counts; counts do not determine grouping.

For each group, give:
- a canonical NAME (a concise noun phrase suitable as a rubric column; use exactly "Other" for the catch-all group),
- a one-sentence DEFINITION of the construct the group captures,
- the list of members, each as its INDEX and its NAME copied verbatim from that index.

Rules on membership (these will be checked programmatically):
- Every input index must appear in exactly one group, exactly once. Place all of them; drop none; duplicate none.
- Do not invent indices that are not in the input.
- Each member's "name" must be the exact input NAME at that index — this is a cross-check, so copy it from the bracketed entry, do not paraphrase.

Output only JSON:
{"features": [{"name": "<canonical noun phrase>", "definition": "<one sentence>", "members": [{"index": <int>, "name": "<verbatim input name at that index>"}, ...]}, ...]}
\end{promptbox}

\begin{promptbox}[listing options app={moredelim={[is][\color{blue!55!black}]{@@}{@@}}}]{BT feature scoring, CMV corpus.}{prm:cmv-score-bt}
You are helping apply an interpretable framework for what drives persuasion in debate. The framework scores how two attempts to change someone's mind differ along a fixed set of dimensions, listed below.
Background. Each item is an online discussion in which an original poster (the "OP") states a view they hold and invites others to change it; a challenger then replies with an argument aimed at changing that specific view. You will be shown two such exchanges, Exchange A and Exchange B, each presented as a block that begins with the OP's view (marked [OP VIEW]) followed by the challenger's reply (marked [CHALLENGE]). The two exchanges concern different posters holding different views on different topics.
The dimensions:
[[features]]
Your task. For EACH listed dimension, in order, give a short evidence phrase (the concrete feature you saw) and then indicate which exchange shows more of it — 'A' or 'B'. Some dimensions describe the challenger's argument, others describe the OP's post — assess each exactly as its definition states. Judge each dimension on its own terms: exchanges have uneven profiles, so it is entirely normal for A to show more of some dimensions and B more of others. Even when the two are close on a dimension, pick the exchange that shows more of it.
Out of scope — do not let these drive any judgment: the topic or subject matter and how interesting, important, difficult, or agreeable either OP's view is (the two differ by design); which side you personally agree with, or whether an OP's view is correct; usernames or identifiers; length in itself.
Return one item per listed dimension, in the same order, naming the dimension exactly as given, each with a short evidence phrase and the winner ('A' or 'B').
Output only JSON of the form: {"features": [{"name": "<dimension>", "evidence": "<short phrase>", "winner": "A" | "B"}, ...]}

Exchange A: {exchange_a}
Exchange B: {exchange_b}
Following the standard above, go through each listed dimension in order: give the concrete evidence you saw and then say which exchange shows more of it. Compare on the listed dimensions, not on topic. Respond with the JSON object only.
\end{promptbox}

\begin{promptbox}[listing options app={moredelim={[is][\color{blue!55!black}]{@@}{@@}}}]{Prior feature elicitation, CMV corpus.}{prm:cmv-prior-elicit}
You are helping build a framework for assessing how persuasive an argument is.
In online debates, a person states a view they hold and invites others to change it. Others reply with arguments aimed at changing that view.
Your task. Name exactly 16 dimensions for assessing how persuasive such a reply is.
Each dimension will be used as a measurement scale: a judge will be shown two replies and asked which of the two shows more of it. So each dimension must be a matter of degree rather than a yes/no fact, and the 16 must be distinct from one another.
For each dimension give:
- "name": a short label, 2-5 words;
- "definition": one sentence saying what it measures.
Output only JSON of the form:
{"features": [{"name": "<short label>", "definition": "<one sentence>"}, ...]}

Name the 16 dimensions now, as a single JSON object. Respond with the JSON only.
\end{promptbox}

\begin{promptbox}{Sample-based feature elicitation, CMV corpus.}{prm:cmv-sample-elicit}
You design an interpretable scoring rubric and output only JSON.

You are helping build an interpretable scoring framework by examining how attempts to change someone's mind differ from one another. There is no fixed set of dimensions — they are recovered from what actually distinguishes these exchanges, not specified in advance.
In online debates, a person states a view they hold and invites others to change it. Others reply with arguments aimed at changing that view.
Below is a sample of such exchanges. Each begins with the poster's view (marked [OP VIEW]), followed by two separate replies to it, written independently by different people (marked [REPLY A] and [REPLY B]). The two replies answer the same view; they are not responses to each other, and their order carries no meaning.
=== EXCHANGES ===
(*@\ph{sample of exchanges}@*)
=== END EXCHANGES ===
Reading across them, attend to the ways the replies differ from one another — the axes along which you can see one reply showing more, less, or something different from another. Identify the dimensions on which these replies meaningfully differ as attempts to change the poster's view. Report whatever genuinely stands out to you; do not work from a predetermined list, and do not force the list to be short — report as many dimensions as actually distinguish these replies. Do not pad the list either: add a dimension only when it names something none of the others already capture.
A dimension may be something a reply shows more or less of along a range, or something a reply either shows or does not.
Every dimension must be a property of a reply, not of the poster's view or of the poster: name only dimensions on which the two replies to the same view can differ.
Keep genuinely distinct dimensions separate even when they are related or tend to co-occur — do not merge two dimensions because strong replies tend to show both. List each dimension as exactly one item — do not combine dimensions with "and", "/", "&", or a comma; split related dimensions into separate items.
For each dimension, first name the concrete, specific feature you actually saw in the replies — a short phrase pointing to the evidence. Then, grounded in that evidence, name the dimension and define in one sentence what it measures. Write the definition so a rater can apply it to a new reply without having seen this sample: describe the characteristic itself, and never refer to specific exchanges or to the reply labels.
Out of scope — do not treat these as dimensions of difference:
- the topic or subject matter, and how interesting, important, difficult or agreeable the poster's view is;
- which side you personally agree with, or whether the poster's view is correct;
- usernames or identifiers;
- length in itself.
Output only JSON:
{"features": [{"evidence": "<short phrase>", "name": "<short label, 2-5 words>", "definition": "<one sentence>"}, ...]}
\end{promptbox}

\begin{promptbox}[listing options app={moredelim={[is][\color{blue!55!black}]{@@}{@@}}}]{PW feature scoring, CMV corpus.}{prm:cmv-score-pw}
You are helping apply an interpretable framework for what drives persuasion in debate. The framework scores an attempt to change someone's mind along a fixed set of dimensions, listed below.
Background. Each item is an online discussion in which an original poster (the "OP") states a view they hold and invites others to change it; a challenger then replies with an argument aimed at changing that specific view. You will be shown one such exchange, presented as a block that begins with the OP's view (marked [OP VIEW]) followed by the challenger's reply (marked [CHALLENGE]).
The dimensions:
[[features]]
Your task. For EACH listed dimension, in order, give a short evidence phrase (the concrete feature you saw) and then rate the exchange on that dimension from 0 to 10. Some dimensions describe the challenger's argument, others describe the OP's post — assess each exactly as its definition states.
What the numbers mean. Rate each dimension relative to online arguments of this kind in general, not relative to this exchange's other dimensions: 0 means the dimension is entirely absent, 5 means it is present to a typical degree, and 10 means it is present about as strongly as such a reply ever gets. The same rating must mean the same thing on every exchange you score, because these scores are compared across exchanges you will not see together.
Judge each dimension on its own terms: exchanges have uneven profiles, so it is entirely normal for one to score high on some dimensions and low on others. Use the full range — do not cluster every dimension in the middle.
Out of scope — do not let these drive any rating: the topic or subject matter and how interesting, important, difficult, or agreeable the OP's view is; which side you personally agree with, or whether the OP's view is correct; usernames or identifiers; length in itself.
Return one item per listed dimension, in the same order, naming the dimension exactly as given, each with a short evidence phrase and an integer rating.
Output only JSON of the form: {"features": [{"name": "<dimension>", "evidence": "<short phrase>", "score": <integer 0-10>}, ...]}

Exchange: {exchange}
Following the standard above, go through each listed dimension in order: give the concrete evidence you saw and then its rating. Respond with the JSON object only.
\end{promptbox}

\end{document}